\documentclass[10pt,twocolumn,letterpaper]{article}

\usepackage[pagenumbers]{cvpr} 

\usepackage{color}
\usepackage{graphicx}
\usepackage{amsmath}
\usepackage{amssymb}
\usepackage{bm}
\usepackage{booktabs}
\usepackage{float}
\usepackage{tipa}

\usepackage[pagebackref,breaklinks,colorlinks]{hyperref}

\usepackage[capitalize]{cleveref}
\crefname{section}{Sec.}{Secs.}
\Crefname{section}{Section}{Sections}
\Crefname{table}{Table}{Tables}
\crefname{table}{Tab.}{Tabs.}

\usepackage[]{multicol}

\def\cvprPaperID{7625} 
\def\confName{CVPR}
\def\confYear{2023}

\begin{document}

\title{NDJIR: Neural Direct and Joint Inverse Rendering for Geometry, Lights, and Materials of Real Object}

\author{Kazuki Yoshiyama \qquad Takuya Narihira\\
Sony Group Corporation Research \& Development Center \\
{\tt\small \{Kazuki.Yoshiyama, Takuya.Narihira\}@sony.com}
}
\maketitle

\begin{abstract}
    The goal of inverse rendering is to decompose geometry, lights, and materials given pose multi-view images. To achieve this goal, we propose neural direct and joint inverse rendering, NDJIR. Different from prior works which relies on some approximations of the rendering equation, NDJIR directly addresses the integrals in the rendering equation and jointly decomposes geometry: signed distance function, lights: environment and implicit lights, materials: base color, roughness, specular reflectance using the powerful and flexible volume rendering framework, voxel grid feature, and Bayesian prior. Our method directly uses the physically-based rendering, so we can seamlessly export an extracted mesh with materials to DCC tools and show material conversion examples. We perform intensive experiments to show that our proposed method can decompose semantically well for real object in photogrammetric setting and what factors contribute towards accurate inverse rendering.
  \end{abstract}


\captionsetup[subfigure]{labelformat=empty}

\section{Introduction}
\label{sec:introduction}

Inverse rendering -- given multi-view posed images, we decompose geometry, lights, materials -- is a long standing problem in computer vision and graphics. In the era of deep learning, it mainly started in 2D image space \cite{DBLP:conf/3dim/AsselinLL20,DBLP:journals/tog/GaoLDP0019,DBLP:conf/cvpr/LiSRSC20,DBLP:conf/cvpr/BossJKLK20,DBLP:conf/eccv/SangC20,DBLP:conf/eccv/SangC20}. After the great success of neural renderings of implicit representation \cite{DBLP:conf/nips/YarivKMGABL20,DBLP:conf/eccv/MildenhallSTBRN20}, directly working in 3D space is prominent \cite{DBLP:conf/iccv/BossBJBLL21,DBLP:journals/tog/ZhangSDDFB21,DBLP:conf/cvpr/MunkbergCHES0GF22,DBLP:conf/nips/ZhangYTR21,DBLP:conf/cvpr/ZhangLWBS21,DBLP:journals/tog/KuangOCHAT22,DBLP:journals/corr/abs-2206-03380,DBLP:conf/nips/BossJBLBL21,DBLP:conf/cvpr/ZhangSHFJZ22}. 

However, modeling the rendering equation \cite{DBLP:conf/siggraph/Kajiya86} directly requires sampling lights, which might be prohibitive in memory footprint and computationally expensive when we integrate with the volume rendering \cite{DBLP:conf/cvpr/Martin-BruallaR21}. To tackle that problem, there are some approximations proposed: Spherical Gaussian (SG)s \cite{DBLP:conf/iccv/BossBJBLL21,DBLP:conf/cvpr/ZhangLWBS21,DBLP:conf/cvpr/ZhangSHFJZ22}, pre-integrated light \cite{DBLP:conf/nips/BossJBLBL21}, split-sum \cite{DBLP:conf/cvpr/MunkbergCHES0GF22}, and co-located light \cite{DBLP:journals/corr/abs-2008-03824,DBLP:conf/cvpr/ZhangLLS22}. Multi-stage training \cite{DBLP:conf/cvpr/ZhangLWBS21,DBLP:conf/cvpr/ZhangSHFJZ22,DBLP:journals/tog/KuangOCHAT22,DBLP:conf/cvpr/ZhangLLS22} is another form to address the problem. Different direction of approach is using surface rendering \cite{DBLP:conf/cvpr/MunkbergCHES0GF22,DBLP:journals/corr/abs-2206-03380}, but such methods require accurate object mask. None of these approaches do not directly and jointly optimize for the integral of the rendering equation under the powerful and flexible volume rendering. Additionally, prior works are mainly evaluated on synthetic data, which limits practical applicability.

\textbf{Contribution:} In this paper, we propose neural direct and joint inverse rendering, NDJIR [\textipa{\'en(d)\textdyoghlig i\textrevepsilon :r}] which light model is tailored to real object in photogrammetric setting. NDJIR directly addresses the integral of the rendering equation and jointly solves the inverse rendering problem with priors. Our key insights are four folds: 1) an indicator function multiplied to the integrand of the specular term allows for simple joint training, 2) directly addressing the integral with importance sampling mitigates degeneration of material networks, 3) introducing Bayesian priors reduces degeneration more, and 4) dense voxel grid feature reinforces more plausible results. We also conducted intensive studies using real datasets. Code is available in https://github.com/sony/NDJIR.

\section{Related work}
\label{sec:related_work}

\textbf{Geometry extraction:} There are several 3D geometry representations, e.g., mesh, signed distance field (SDF), and volume density. Correspondingly, differentiable renderers were proposed \cite{DBLP:conf/eccv/LoperB14,DBLP:conf/iccv/Liu0LL19,DBLP:journals/tog/LaineHKSLA20,DBLP:conf/nips/ChenLGSLJF19,DBLP:conf/cvpr/KatoUH18,DBLP:conf/cvpr/NiemeyerMOG20,DBLP:conf/cvpr/LiuZPSPC20,DBLP:conf/nips/LinWL20,DBLP:conf/nips/YarivKMGABL20,DBLP:journals/tog/LombardiSSSLS19,DBLP:conf/eccv/MildenhallSTBRN20} in a decade in order to extract such a representation. Notably, NeuS \cite{DBLP:conf/nips/WangLLTKW21} and VolSDF \cite{DBLP:conf/nips/YarivGKL21} integrates SDF representation with volume rendering \cite{DBLP:conf/eccv/MildenhallSTBRN20}, SDF is implicit representation and $0$-level set corresponds to surface, so we can easily extract mesh once SDF is acquired, and volume rendering \cite{DBLP:conf/eccv/MildenhallSTBRN20} mitigates the need of corresponding accurate object masks. As a result, we can extract a mesh given multi-view posed images of an object. Based on the works \cite{DBLP:conf/nips/WangLLTKW21,DBLP:conf/eccv/MildenhallSTBRN20}, we jointly decompose geometry, lights, and materials. 

\textbf{Materials and lights estimation:} Neural-based materials and lights estimation started in 2D image space \cite{DBLP:conf/3dim/AsselinLL20,DBLP:journals/tog/GaoLDP0019,DBLP:conf/cvpr/LiSRSC20,DBLP:conf/cvpr/BossJKLK20,DBLP:conf/eccv/SangC20,DBLP:conf/eccv/SangC20}. Soon after the success of differentiable rendering of implicit representation \cite{DBLP:conf/nips/YarivKMGABL20,DBLP:conf/eccv/MildenhallSTBRN20}; materials and lights estimation is mainly based on neural rendering \cite{DBLP:conf/iccv/BossBJBLL21,DBLP:journals/tog/ZhangSDDFB21,DBLP:conf/cvpr/MunkbergCHES0GF22,DBLP:conf/nips/ZhangYTR21,DBLP:conf/cvpr/ZhangLWBS21,DBLP:journals/tog/KuangOCHAT22,DBLP:journals/corr/abs-2206-03380,DBLP:conf/nips/BossJBLBL21,DBLP:conf/cvpr/ZhangSHFJZ22}. Spherical Gaussian (SG)s are often used in \cite{DBLP:conf/iccv/BossBJBLL21,DBLP:conf/cvpr/ZhangLWBS21,DBLP:conf/cvpr/ZhangSHFJZ22} for illumination, but it can only represent low to middle frequency and specular materials. Evaluating physically-based rendering (PBR) directly is considered costly, so there are other approximations to PBR: pre-integrated lighting \cite{DBLP:conf/nips/BossJBLBL21}, split-sum \cite{DBLP:conf/cvpr/MunkbergCHES0GF22}, and co-located light \cite{DBLP:journals/corr/abs-2008-03824,DBLP:conf/cvpr/ZhangLLS22}. Recently, Hasselgren \textit{et al}. \cite{DBLP:journals/corr/abs-2206-03380} directly optimizes physically-based rendering using Monte Carlo integration and denoiser on top of rendered image. However, applying denoiser requires full resolution image, which might be bounded by image resolution. Their work relies on DMTet \cite{DBLP:conf/nips/ShenGYLF21} so uses the surface rendering on extracted mesh at each iteration of training, which reduces memory footprint compared to using the volume rendering and may allow for full resolution rendering. Nonetheless, this type of approaches sacrifices level of details of geometry and needs accurate foreground object mask. Contrarily, our approach utilizes the powerfully volume rendering to directly evaluate PBR by circumventing illumination evaluation in 3D space but in pixel space and introducing an indicator function, being able to produce high-poly mesh and high frequent materials.

\textbf{Acceleration structure:} One of shortcomings of the volume rendering is slowness. Prior works to mitigate slow rendering in runtime \cite{DBLP:conf/iccv/HedmanSMBD21,DBLP:conf/iccv/GarbinK0SV21,DBLP:conf/nips/LiuGLCT20,DBLP:conf/iccv/YuLT0NK21,DBLP:conf/cvpr/WuLBWF22} are based on voxel grid structure, which influences following works for faster training \cite{DBLP:conf/cvpr/0004SC22,DBLP:conf/cvpr/WuLBWF22,DBLP:conf/cvpr/Fridovich-KeilY22,DBLP:conf/siggraph/TakikawaET0MJF22}. Density is directly optimized on voxel grid \cite{DBLP:conf/cvpr/0004SC22,DBLP:conf/cvpr/Fridovich-KeilY22,DBLP:conf/siggraph/Karnewar0WM22} to remove evaluations of MLP. TensorF \cite{DBLP:journals/corr/abs-2203-09517} factorizes voxel grid to vector-matrix low-rank representation to speed up look-up from voxel grid representation. Instant NGP \cite{DBLP:journals/tog/MullerESK22} uses hash table on voxel grid to further acceleration. These works aims for novel view synthesis, an acquired underlying geometry is not smooth so can not easily be exported to existing DCC tools as 3D asset. Thus, we use the simple dense voxel grid feature.

\section{Method}
\label{sec:method}

In this section, we formulate the novel forward rendering tailored to inverse rendering of real object. Our proposed method is based on two important prior works: NeuS \cite{DBLP:conf/nips/WangLLTKW21} and Physically-based rendering with Cook-Torrance specular BRDF (Bidirectional Reflectance Distribution Function) model \cite{DBLP:journals/tog/CookT82}. These models are briefly described in \cref{subsec:background}. In \cref{subsec:joint_inverse_rendering}, we introduce light model specific to photogrammetry setting and integrate the light model to PBR in computationally efficient and trainable way. Even if we can train inverse rendering system, we may not be able to obtain meaningful materials since we only have ground truth color as supervision, there are plenty of solutions for materials to be estimated. Therefore, we pose priors on materials in \cref{subsec:loss_and_prior}. At the end of this section \cref{subsec:implementation}, we elaborate implementation of NDJIR. Note we assume object is opaque and dielectric, and lights are white. Furthermore, we only denote the dimension of sample points along a ray for simplicity, and several input arguments to a function in equations are excluded for visibility in \cref{subsec:background,subsec:joint_inverse_rendering}. For rigorous inputs and outputs, see \cref{subsec:network}. More detailed equations and network diagram can be found in the supplementary material.

\subsection{Background}
\label{subsec:background}

\textbf{Volume rendering with SDF:} Given sample points $\{\bm{x}_i\}_{i=0}^{N}$ along a ray $\bm{r}$, NeuS \cite{DBLP:conf/nips/WangLLTKW21} formulates the volume rendering with SDF $s$ to render a color $\hat{C}$ as follows, 
\begin{equation}
    \label{eq:volume_rendering_with_sdf}
    \begin{split}
        \hat{C}(\bm{r}) = \sum_{i=1}^{N}{T_i \alpha_{i} \bm{c}_i}, 
        \hfill T_{i} = \prod_{j=1}^{i-1}{(1 - \alpha_{j})}, \\
        \alpha_{j} = \max \left(\frac{\Phi(s_{j}, \sigma) - \Phi(s_{j+1}, \sigma)}{\Phi(s_{j}, \sigma)}, 0 \right), 
    \end{split}
\end{equation}
where $\bm{c}$ is radiance, $\alpha$ is opacity using adjacent SDFs, $T$ is accumulated transmittance, $\Phi$ is the cumulative distribution function of logistic distribution with the globally trainable scale $\sigma$.

\textbf{Physically-based rendering:} Rendering equation \cite{DBLP:conf/siggraph/Kajiya86} with Cook-Torrance specular BRDF model \cite{DBLP:journals/tog/CookT82} is the principle of modern graphics engines e.g., \cite{WinNT,Blender,unrealengine}. Given a surface point $\hat{\bm{x}}$, normals $\hat{\bm{n}}$, corresponding spatially-varying materials: base color $\hat{\bm{c}}_b$, roughness $\hat{\alpha}_{r}$, and specular reflectance $\bm{\hat{f}}_{0}$, viewing direction $\bm{v}$, and light direction $\bm{l}$ over hemisphere $\Omega_{+}$ aligned with $\hat{\bm{n}}$, the rendered color $\hat{C}(\bm{r}) $ is modeled as 
\begin{equation}
    \label{eq:physically_based_rendering}
    \begin{split}
        & \underbrace{\frac{\hat{\bm{c}}_b}{\pi} \int_{\Omega_{+}} L(\hat{\bm{x}}, \bm{l}) (\hat{\bm{n}} \cdot \bm{l}) d\bm{l}}_{diffuse\ term} \\
        + & \underbrace{\int_{\Omega_{+}} \frac{D(\bm{v}, \bm{l}, \hat{\bm{n}}, \hat{\alpha}_{r}) G(\bm{v}, \bm{l}, \hat{\bm{n}}, \hat{\alpha}_{r}) F(\bm{v}, \bm{l}, \hat{\bm{f}}_{0})}{4 (\hat{\bm{n}} \cdot \bm{v}) (\hat{\bm{n}} \cdot \bm{l})} L(\hat{\bm{x}}, \bm{l}) (\hat{\bm{n}} \cdot \bm{l}) d\bm{l}}_{specular\ term}.
    \end{split}
\end{equation}
$L$ is incoming light intensity, $D$ is normal distribution function, $G$ is masking-shadowing function, and $F$ is Fresnel function. With arbitrary two vectors $\bm{a}$ and $\bm{b}$, $(\bm{a} \cdot \bm{b})$ is saturated dot product $\max(\bm{a}^{T}\bm{b}, \epsilon_{\textrm{dot}})$. We use Filament BRDF model \cite{WinNT} for $D$, $G$, and $F$. For space, explicit models of $D$, $G$, and $F$ are excluded but in the supplementary.

\subsection{Joint inverse rendering}
\label{subsec:joint_inverse_rendering}

NDJIR models the incoming light intensity attenuated by the dot product $L(\bm{x}, \bm{l})(\bm{n}, \bm{l})$ for the diffuse and specular term, respectively:
\begin{equation}
    \label{eq:ndjir_diffuse_light_model}
    L_{P}(\bm{x}, \bm{v}, \bm{n})
        \left(L_{V}(\bm{x}, \bm{l}) L_{E}(\bm{l}) (\bm{n} \cdot \bm{l}) + L_{I}(\bm{x})\right), 
\end{equation}
\begin{equation}
    \label{eq:ndjir_specular_light_model}
    L_{P}(\bm{x}, \bm{v}, \bm{n})
        L_{V}(\bm{x}, \bm{l}) L_{E}(\bm{l}) (\bm{n} \cdot \bm{l}), 
\end{equation}
where $L_{V}$ is soft visibility, $L_{E}$ is intensity of environment map, $L_{I}$ is intensity of implicit illumination, and $L_{P}$ is photogrammetric light intensity.

Environment map $L_{E}$ is same as in the classical PBR to query light intensity directionally, and the implicit illumination $L_{I}$ resembles the photon map \cite{DBLP:conf/rt/Jensen96} to query light intensity in space, both of which together captures direct and indirect illumination well. Implicit illumination $L_{I}$ is one of keys to successful training; otherwise, training fails to bad local minima. Notable light model in NDJIR is photogrammetric light $L_{P}$ tailored to photogrammetry setting where a photographer dynamically becomes masks of lights and casts shadows, which is not modeled by classical PBR using a typical pinhole camera model. Photogrammetric light $L_{P}$ is also a importance factor for joint inverse rendering.

Plugging both \cref{eq:ndjir_diffuse_light_model} and \cref{eq:ndjir_specular_light_model} to \cref{eq:physically_based_rendering}, $L_{P}$ is factored out, but still we have the integrals. If we integrate the volume rendering \cref{eq:volume_rendering_with_sdf} with the physically-based rendering \cref{eq:physically_based_rendering} naively, we must sample incident lights per batch, ray, and, point during training, which prohibitively occupies memory footprint and is computationally expensive. Therefore, for computing the specular term in \cref{eq:physically_based_rendering}, we apply the volume rendering \cref{eq:volume_rendering_with_sdf} to spatially-varying quantities: roughness $\alpha_{r}$, specular reflectance $\bm{f}_{0}$, analytically computed normals $\bm{n}$, photogrammetric light $L_{P}$, and implicit illumination $L_{I}$, one exception is computing $L_{V}$, we apply the volume rendering \cref{eq:volume_rendering_with_sdf} to $\bm{x}$. In other words, we first reduce point dimension by volume rendering and compute the specular term in pixel space. Then, we perform incident light sampling to compute the integrals in \cref{eq:physically_based_rendering}. This idea is similar to \cite{DBLP:conf/iccv/BossBJBLL21}; however, we can directly optimize thanks to the indicator function of \cref{eq:ndjir_specular_term} and do not use any approximations. For computing diffuse component in \cref{eq:physically_based_rendering}, we first entangle $\bm{c}_b$ and $L_{p}$ and utilize the same way as in computing the specular term for the integral then multiply them. Finally, using shorthand notation $\hat{\cdot}$ as the volume rendering \cref{eq:volume_rendering_with_sdf} to each quantity $(\cdot)$ other than $\bm{c}$, the rendered color $\hat{C}(\bm{r})$ of NDJIR is the summation of the diffuse and specular terms formulated as 
\begin{equation}
    \label{eq:ndjir_diffuse_term}
    \frac{\widehat{L_{p}\bm{c}_{b}}}{\pi} 
        \int_{\Omega_{+}} 
            \left(L_{V}(\hat{\bm{x}}, \bm{l}) L_{E}(\bm{l}) (\hat{\bm{n}} \cdot \bm{l}) + \widehat{L_{I}(\bm{x})}\right) d\bm{l},
\end{equation}
\begin{equation}
    \label{eq:ndjir_specular_term}
    \begin{split}
        \widehat{L_{p}} \int_{\Omega_{+}} & \frac{D(\bm{v}, \bm{l}, \hat{\bm{n}}, \hat{\alpha}_{r}) G(\bm{v}, \bm{l}, \hat{\bm{n}}, \hat{\alpha}_{r}) F(\bm{v}, \bm{l}, \hat{\bm{f}}_{0})}{4 (\hat{\bm{n}} \cdot \bm{v}) (\hat{\bm{n}} \cdot \bm{l})} \\
            & L_{V}(\hat{\bm{x}}, \bm{l}) L_{E}(\bm{l}) (\hat{\bm{n}} \cdot \bm{l}) 
                I(\bm{v}, \bm{l}, \hat{\bm{n}}) d\bm{l}, 
    \end{split}
\end{equation}
where $I(\bm{v}, \bm{l}, \hat{\bm{n}})$ is the indicator function of $1$ if $\left( (\hat{\bm{n}} \cdot \bm{l}) \ge \epsilon_{I} \right) \land \left( (\hat{\bm{n}} \cdot \bm{v}) \ge \epsilon_{I} \right) \land \left( (\hat{\bm{n}} \cdot \frac{\bm{v} + \bm{l}}{||\bm{v} + \bm{l}||_{2}}) \ge \epsilon_{I} \right)$, otherwise $0$.

Entanglement of $L_{p}$ and $\bm{c}_{b}$ is important; otherwise, training does not successfully start. Indicator function is also necessary for joint-training. As geometry and corresponding normals are changing during training, dot products of numerator in \cref{eq:ndjir_specular_term} can be negative. Negative values are omitted by the saturation of dot product, but we also have several dot products in denominator of the specular term, so does $DGF$, such that saturated dot products increase value of the specular term, which results in really large gradients in backward pass. To prevent such gradient explosion, we introduce the indicator function. In that sense, the indicator function can be viewed as hard visibility factor specific to the specular term in joint inverse rendering.

We are mixing the volume rendering and physically based rendering; however, once training starts, the trainable scale $\sigma$ of \cref{eq:volume_rendering_with_sdf} is going up, geometry is gradually being formed, then the specular term is considered in training, and rendering is going towards surface rendering. Therefore, NDJIR approaches to $L_{P}$ $\times$ PBR computed on surface in the end of training.

\subsection{Network}
\label{subsec:network}

All geometry, lights, and materials are represented by naive multi-layer perceptron (MLP); geometric network is denoted by $\bm{f}_{G}(\bm{x}) \in \mathbb{R}^{D_{G} + 1}$, base color network $\bm{c}_{b}(\bm{x}, \bm{F}_{G}) \in \mathbb{R}^{3}$, roughness network $\alpha_{r}(\bm{x}, \bm{F}_{G}, \bm{n}) \in \mathbb{R}$, specular reflectance network $\bm{f}_{0}(\bm{x}, \bm{F}_{G}, \bm{n}) \in \mathbb{R}^{3}$, environment light network $L_{E}(\bm{l}) \in \mathbb{R}$, soft visibility network $L_{V}(\bm{x}, \bm{l}, \bm{F}_{G}, \bm{n}) \in \mathbb{R}$, implicit illumination network $L_{I}(\bm{x}, \bm{F}_{G}, \bm{n}) \in \mathbb{R}$, and photogrammetric light network $L_{P}(\bm{x}, \bm{v}, \bm{F}_{G}, \bm{n}, d^{-1}) \in \mathbb{R}$. $\bm{F}_{G} \in \mathbb{R}^{D_{G}}$ is geometric feature as one of the outputs of $\bm{f}_{G}$ in addition to SDF $s \in \mathbb{R}$. We add the inverse squared distance $d^{-2}$ as input of $L_{p}$, which improves geometric reconstruction empirically.

Choice of inputs is based on our preliminary experiments where normals $\bm{n}$ captures light distribution spatially. For base color network, we do not expect that light is baked into base color, thus we exclude normals from inputs. Light networks use a single channel output because of white light assumption to reduce ambiguity. For specular reflectance, we use $3$ channels in order to increase degree of freedom.

Positional encoding $\bm{\gamma}(\cdot, \ell)$ \cite{DBLP:conf/eccv/MildenhallSTBRN20} is applied to the geometric network as $\bm{\gamma}(\bm{x}, \ell_{G})$, the environment light network as $\bm{\gamma}(\bm{l}, \ell_{E})$, the soft visibility network as $\bm{\gamma}(\bm{l}, \ell_{V})$, and the photogrammetric light network as $\bm{\gamma}(\bm{v}, \ell_{P})$, where $\ell$ is the number of frequencies. Results are concatenated to the original input.

Dense voxel grid feature of linear interpolation with grid size $G$ is used but only for $\bm{x}$ in the geometric network. We tried several variants: the one with varying interpolations (cosine and lanzcos), triplane and triline grid feature similar to tensor factorization \cite{DBLP:journals/corr/abs-2203-09517}, and voxel hash grid feature \cite{DBLP:journals/tog/MullerESK22}. With varying interpolations achieve similar performance both quantitatively and qualitatively. Triplane and triline grid feature extracts smoother geometry but sometimes bad separation between foreground object and background, depending on scene; also, semantic separation of materials are worse than using dense voxel. With voxel hash feature \cite{DBLP:journals/tog/MullerESK22} can not be trained, we conjecture this is due to the simultaneous use of Eikonal regularization \cite{DBLP:conf/icml/GroppYHAL20}.

\subsection{Loss and prior}
\label{subsec:loss_and_prior}

Our loss functions are composed of RGB color loss $\mathcal{L}_{RGB}$, Eikonal regularization $\mathcal{L}_{E}$ \cite{DBLP:conf/icml/GroppYHAL20}, total variation loss $\mathcal{L}_{TV}$ \cite{DBLP:conf/cvpr/Fridovich-KeilY22}, base color prior $\mathcal{L}_{\bm{c}_{b}}$, roughness prior $\mathcal{L}_{\alpha_{r}}$, and specular prior $\mathcal{L}_{\bm{f}_{0}}$. These losses are computed as follows.

RGB color loss $\mathcal{L}_{RGB}$ is defined as $L_{1}$ loss of the residual between render color $\hat{C}(\bm{r})$ and ground truth color $C(\bm{r})$: 
\begin{equation}
    \label{eq:color_loss}
    \mathcal{L}_{RGB} = ||\hat{C}(\bm{r}) - C(\bm{r})||_{1}.
\end{equation}

Eikonal regularization $\mathcal{L}_{E}$ is the constraint SDF meets:
\begin{equation}
    \label{eq:eikoanl_regularization}
    \mathcal{L}_{E} = (||s||_2 - 1)^{2}.
\end{equation}

Total variation loss $\mathcal{L}_{TV}$ of voxel grid feature is computed as 
\begin{equation}
    \label{eq:tv_loss}
    \mathcal{L}_{TV} = \sqrt{
        \Delta \bm{f}_{i+1, j, k}^{2} 
        + \Delta \bm{f}_{i, j+1, k}^{2} 
        + \Delta \bm{f}_{i, j, k+1}^{2} }
\end{equation}
where $\Delta \bm{f}_{i+1, j, k}$ is the difference of adjacent features along $x$ axis, same for $y$ and $z$ axes, $\bm{f}_{i, j, k}$ is positionally-embedded trainable feature corresponding to discretized query point.

Base color prior $\mathcal{L}_{\bm{c}_{b}}$ is defined spatially between query point and near point perturbed by $\bm{\epsilon}_{\bm{c}_{b}}$, 
\begin{equation}
    \label{eq:base_color_prior}
    \mathcal{L}_{\bm{c}_{b}} = ||\bm{c}_{b}(\bm{x}) - \bm{c}_{b}(\bm{x} + \bm{\epsilon}_{\bm{c}_{b}})||_{1}.
\end{equation}

Even if we can train inverse rendering system jointly, we found roughness and specular reflectance tend to degenerate and outputs either unary ($0$ or $1$) or binary ($0$ and $1$) value(s). This can be mitigated by using smaller MLPs and increasing the number of incident lights to some extent. Even though, roughness and specular reflectance networks would degenerate, so we introduce priors on both networks.

For specular reflectance, we know that specular reflectance of common object is $0.04$ \cite{WinNT} as prior knowledge, but we do not know for roughness, so we simply set a certain value $r_{p}$ as prior. Because using a deterministic value is too strong, we pose Laplacian distribution as prior distribution under Bayesian deep learning framework \cite{DBLP:conf/nips/KendallG17}.
\begin{eqnarray}
    \label{eq:roughness_specular_priors}
    \mathcal{L}_{\alpha_{r}} &=& \frac{||\alpha_{r} - r_{p}||_1}{\sigma_{\alpha_{r}}} + \log(\sigma_{\alpha_{r}}), \\
    \mathcal{L}_{\bm{f}_{0}} &=& \frac{||\bm{f}_{0} - 0.04||_1}{\sigma_{\bm{f}_{0}}} + \log(\bm{\sigma}_{\bm{f}_{0}}), 
\end{eqnarray}
where the roughness network is forked at the last layer of MLP and outputs $\sigma_{\alpha_{r}}$ additionally and same for $\bm{\sigma}_{\bm{f}_{0}}$ of the specular reflectance network.

Our final objective to be minimized with corresponding weights, AABB-ray intersection mask $M_{A}$, and denominators $N$ and $M$ is
\begin{equation}
    \begin{split}
        \frac{\overline{\mathcal{L}}_{RGB}}{N}
        + \lambda_{E} \frac{\overline{\mathcal{L}_{E} M_{A}}}{M}
        + \lambda_{TV} \frac{\overline{\mathcal{L}_{TV} M_{A}}}{M} \\
        + \lambda_{\bm{c}_{b}} \frac{\overline{\mathcal{L}_{\bm{c}_{b}} M_{A}}}{M}
        + \lambda_{\alpha_{r}} \frac{\overline{\mathcal{L}_{\alpha_{r}} M_{A}}}{M} 
        + \lambda_{\bm{f}_{0}} \frac{\overline{\mathcal{L}_{\bm{f}_{0}} M_{A}}}{M},
    \end{split}
\end{equation}
where $\overline{\cdot}$ is the summation of all dimensions.

\subsection{Implementation}
\label{subsec:implementation}

Our scene is normalized in the same way as \cite{DBLP:conf/nips/YarivKMGABL20} such that an object is approximately inside a unit-sphere located on the origin. We put on the origin the voxel grid which $\min$ and $\max$ coordinates are $(-1, -1, -1)$ and $(1, 1, 1)$, respectively. Grid size is $512$, and feature size are $4$. During training, we sample $4$ camera locations and cast $512$ rays at each iteration; if a ray hits the axis-aligned bounding box (AABB) of the voxel grid, we sample $64$ points between near and far hits, then upsample $4$ times $16$ points same as \cite{DBLP:conf/nips/WangLLTKW21}. For background modeling, we use NeRF++ \cite{DBLP:journals/corr/abs-2010-07492} and sample $32$ points per ray staring from the far hit of AABB. If a ray does not hit AABB, we sample points only for background modeling from the location of the distance $||\textrm{camera location}||_{2} - 1$ along that ray.

Voxel grid feature is implemented in CUDA \cite{cuda} and supports double-backward \cite{155328}. This is needed; otherwise, extracted geometry is locally jagged. For total variation loss, the asymmetric backward is used in \cite{DBLP:conf/cvpr/Fridovich-KeilY22}, but we found the symmetric backward mitigates the baked light on base color $\bm{c}_{b}$, so the symmetric backward is used. When using the voxel grid feature, the extracted mesh is rugged, but applying simple average filter dramatically mitigates and produces smoother mesh. Application of average filters is $2$ times as default.

$\bm{f}_{G}$ has $8$ layers and one skip connection at the middle as in \cite{DBLP:conf/nips/WangLLTKW21} with $D_{G}=256$. $\bm{c}_{b}$ and $L_{p}$ are $4$ layers of $256$ feature size. The other MLPs have $4$ layers of $128$ feature size. We use $\textrm{softplus}(\beta=100)$ \cite{DBLP:conf/ijcnn/ZhengYLLL15} as activation function, and output is ranged in $[-1, 1]$ by the sigmoid function in all MLPs except for $L_{E}$, $\sigma_{\alpha_{r}}$, and $\sigma_{\bm{f_{0}}}$ which use $\textrm{softplus}(\beta=1)$ as output activation function. As in \cite{WinNT}, we remap roughness and specular reflectance networks such that actual roughness is $\alpha_{r} \leftarrow \alpha_{r}^{2}$ and specular reflectance is $\bm{f}_{0} \leftarrow 0.16 \times \bm{f}_{0}^{2}$. To prevent $0$ division, we clip as $\max(\alpha_{r}, 0.089)$. $\ell_{G} = 6$, $\ell_{E} = \ell_{V} = 6$, and $\ell_{P} = 4$ are set in positional encodings. All networks are initialized using Glorot initialization \cite{DBLP:journals/jmlr/GlorotB10} except for $\bm{f}_{G}$ where the geometric initialization \cite{DBLP:conf/cvpr/AtzmonL20} is used. Voxel grid features are directly initialized by sampling from $\mathcal{N}(0, (10^{-3})^{2})$. 

When computing the integrals, we use Monte Carlo integration. The number of incident lights is $128$ per pixel. In \cref{eq:ndjir_diffuse_term}, the uniform sampling is used and the importance sampling in consideration of roughness $\alpha_{r}$ in \cref{eq:ndjir_specular_term} same as \cite{WinNT}. As suggested in \cite{DBLP:journals/tog/ZeltnerSGJ21}, we do not propagate gradients in the sampling process.

$\epsilon_{\bm{n}} = 10^{-16}$ is added to volume rendered normals $\hat{\bm{n}}$, then the result is unit-vector normalized. Even small $\epsilon_{\bm{n}}$ produces bias, but this is necessary to prevent $0$-division especially at the beginning of training. $\bm{\epsilon}_{\bm{c}_b}$ is sampled from Normal distribution $\mathcal{N}(0, (\sqrt{3}\frac{2}{G})^{2})$. We set $\epsilon_{\textrm{dot}} = \epsilon_{I} = 10^{-8}$, $\lambda_{E} = 10^{-1}$, $\lambda_{TV} = 10^{-1}$, $\lambda_{\bm{c}_{b}} = 10^{-1}$, $\lambda_{\alpha_{r}} = 10^{-5}$, $\lambda_{\bm{f}_{0}} = 10^{-3}$, and $r_{p} = 0.5$ as default. $N$ is multiplication of the batch size of $4$ and sampled pixels of $512$, and $M$ is multiplication of the number of rays hitting AABB and sampled points of $128$.

For training, we use Adam optimizer \cite{DBLP:journals/corr/KingmaB14} with $10^{-3}$ weight decay and $5 \times 10^{-4}$ learning rate, and the warmup \cite{DBLP:journals/corr/GoyalDGNWKTJH17} is applied for $1.5$ \% of the total epoch $1500$. After the warmup period, the learning rate cosine-decays towards $5 \times 10^{-6}$. Training takes $2.8$ and $3.7$ hours for 49 and 64 images of DTU MVS dataset \cite{DBLP:conf/cvpr/JensenDVTA14}, respectively using A100 GPU.

To extract mesh, we use MarchingCubs algorithm with grid size of $512$. For texture baking, we use Blender \cite{Blender}, first we use Smart UV unwrap to get uv-coordinates which is then applied to baking vertex attributes to texture maps in Cycles renderer. To physically-based render extracted meshes with decomposed materials, we utilize Open3D \cite{Zhou2018} and set the sun light direction above object(s).

\section{Experiment}
\label{sec:experiment}

\textbf{Dataset:} We use DTU MVS dataset \cite{DBLP:conf/cvpr/JensenDVTA14} curated by \cite{DBLP:conf/nips/YarivKMGABL20}. Dataset contains $15$ scenes where real object(s) is captured with a industrial robot. $16$ LEDs are located above object(s). 49 and 64 images are given with camera poses.

\textbf{Evaluation:} For inverse rendering of real object, it is hard to evaluate quantitatively in material space, so we compare our method in image space using PSNR and SSIM. Geometrically, we evaluate decomposition quantity with reference given by DTU MVS dataset. In the other cases, we show qualitative comparisons.

\textbf{Baseline:} We use NeuS \cite{DBLP:conf/nips/WangLLTKW21} as quantitative baseline. For qualitative comparison, we implement a method bearing the similar spirit of split-sum \cite{DBLP:conf/cvpr/MunkbergCHES0GF22} and pre-integrated light \cite{DBLP:conf/nips/BossJBLBL21} under our framework. The derivation and more results are found in the supplementary material.

\subsection{Primary result}
\label{subsec:primary_result}

\begin{figure*}[tbp]
    \centering
    \rotatebox[origin=b]{90}{scan24}\quad
    \begin{subfigure}[h]{0.14\paperwidth}
        \caption{normals}
        \includegraphics[width=\textwidth]{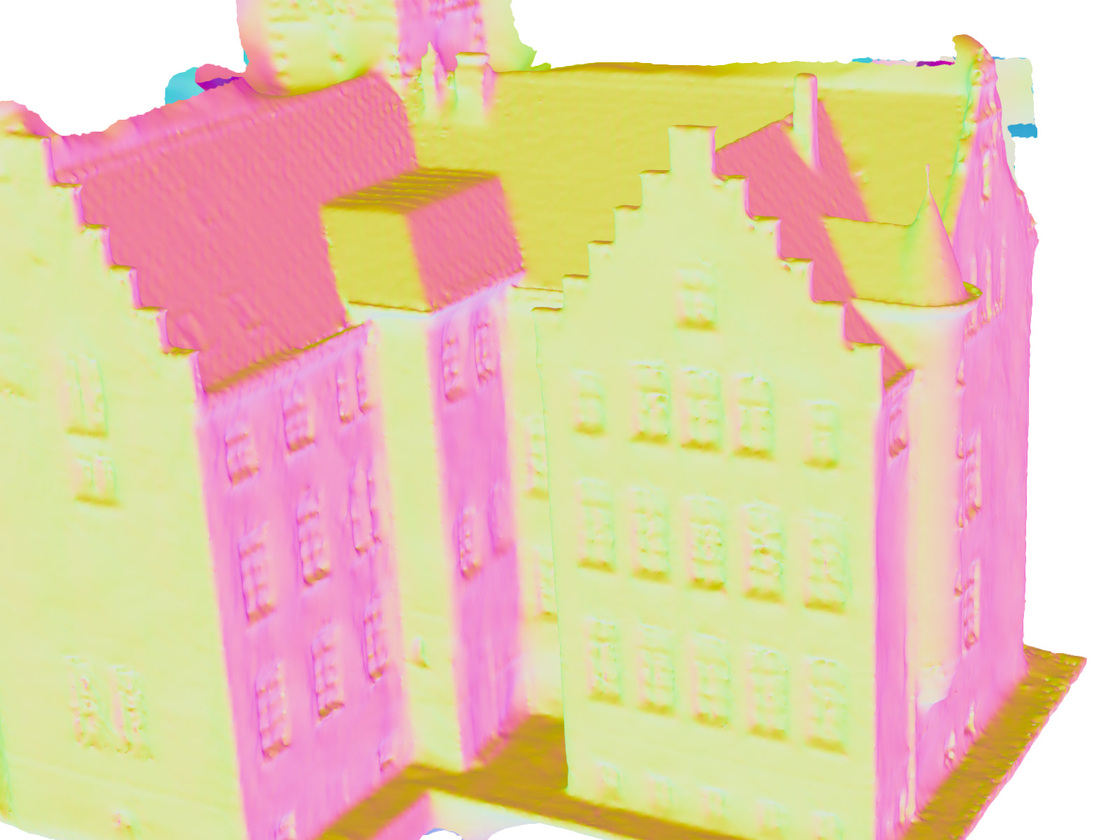}
    \end{subfigure}
    \hspace{1pt}
    \begin{subfigure}[h]{0.14\paperwidth}
        \caption{base color}
        \includegraphics[width=\textwidth]{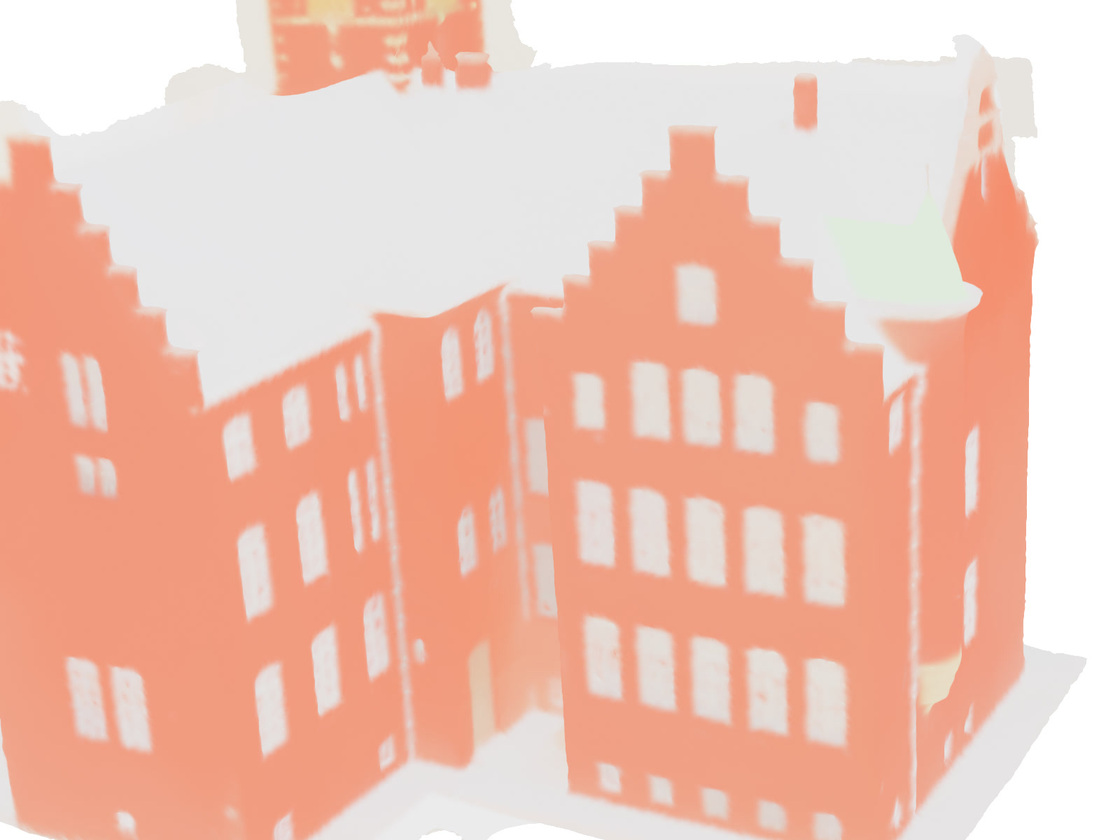}
    \end{subfigure}
    \hspace{1pt}
    \begin{subfigure}[h]{0.14\paperwidth}
        \caption{roughness}
        \includegraphics[width=\textwidth]{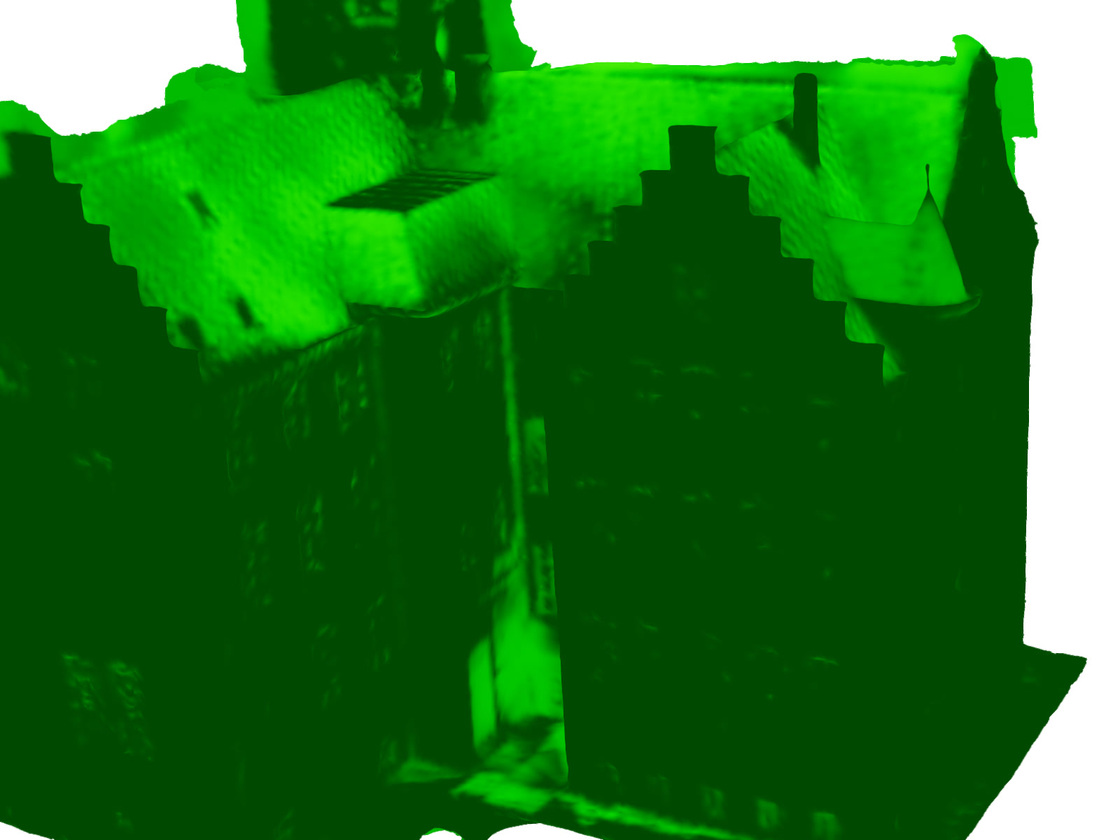}
    \end{subfigure}
    \hspace{1pt}
    \begin{subfigure}[h]{0.14\paperwidth}
        \caption{specular reflectance}
        \includegraphics[width=\textwidth]{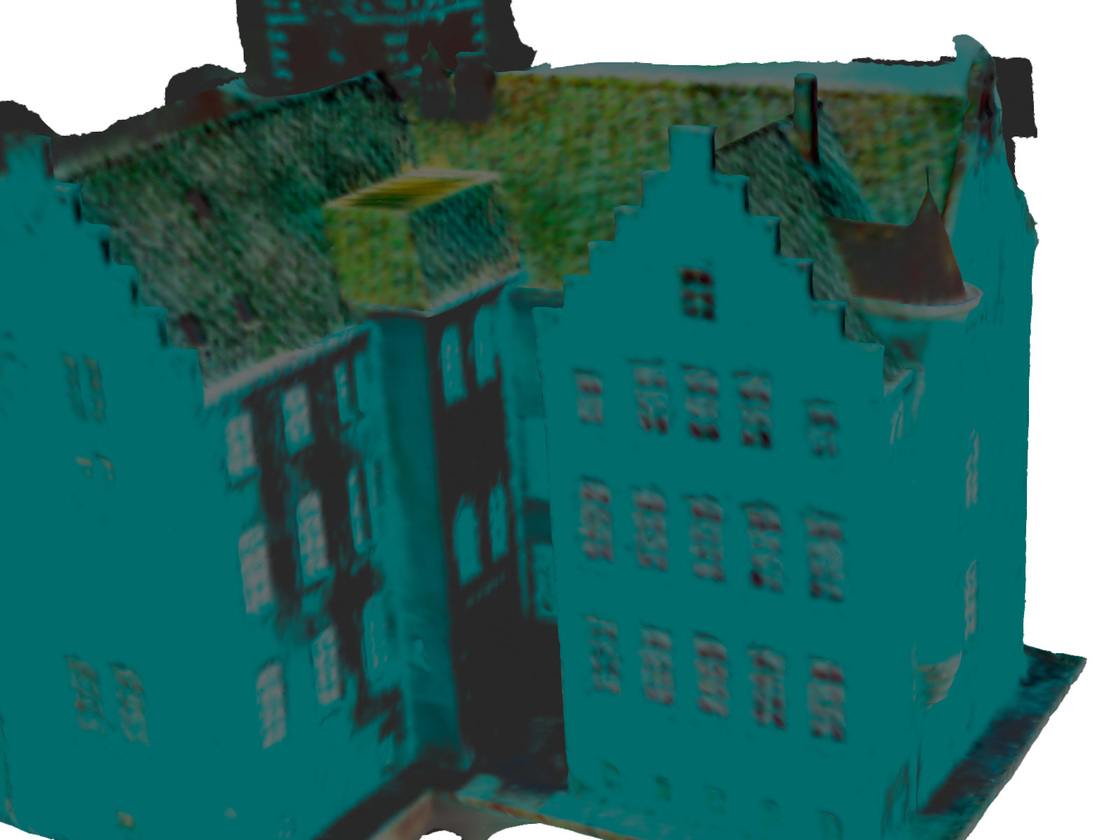}
    \end{subfigure}
    \hspace{1pt}
    \begin{subfigure}[h]{0.14\paperwidth}
        \caption{implicit illumination}
        \includegraphics[width=\textwidth]{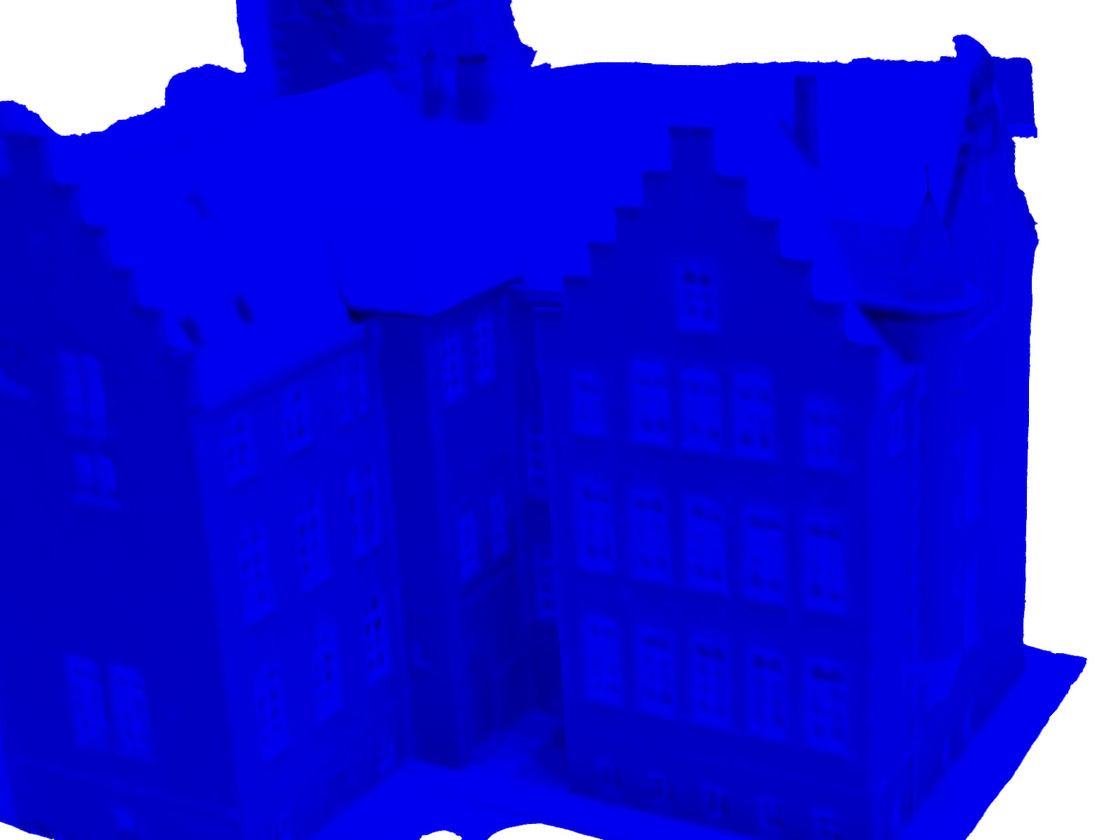}
    \end{subfigure}

    \smallskip
    \rotatebox[origin=b]{90}{scan65}\quad
    \begin{subfigure}[h]{0.14\paperwidth}
        \includegraphics[width=\textwidth]{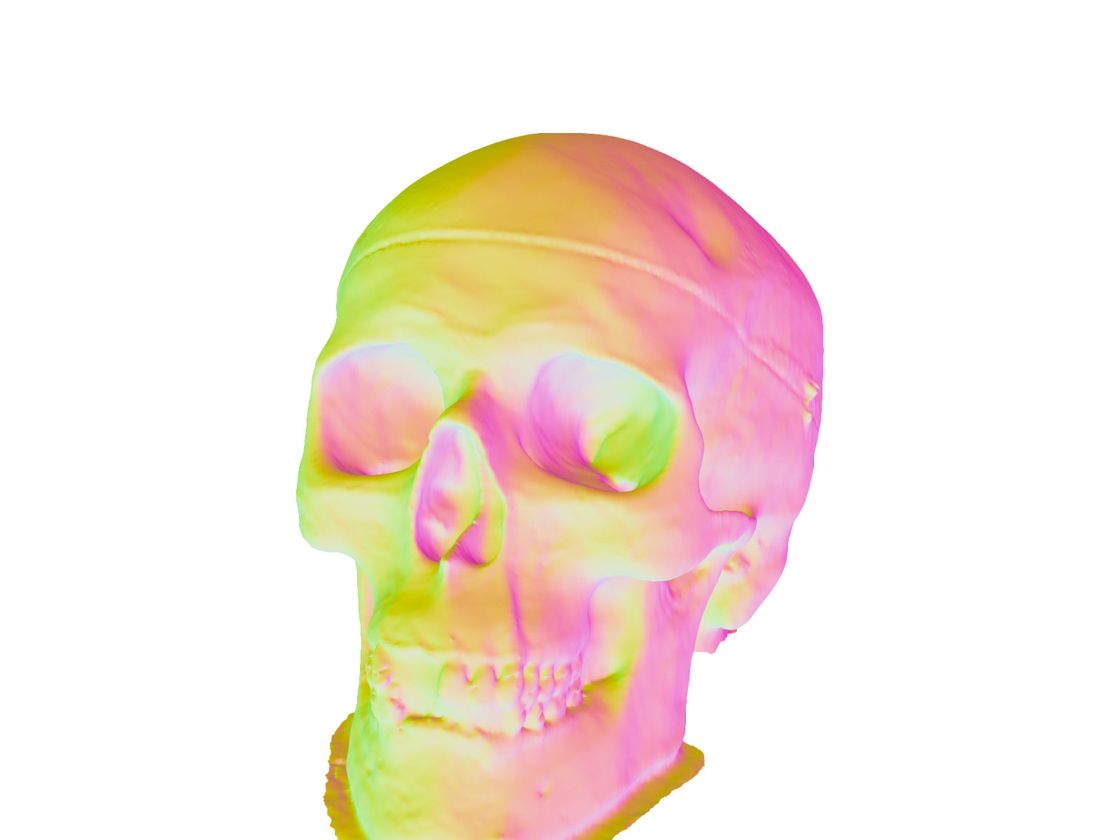}
    \end{subfigure}
    \hspace{1pt}
    \begin{subfigure}[h]{0.14\paperwidth}
        \includegraphics[width=\textwidth]{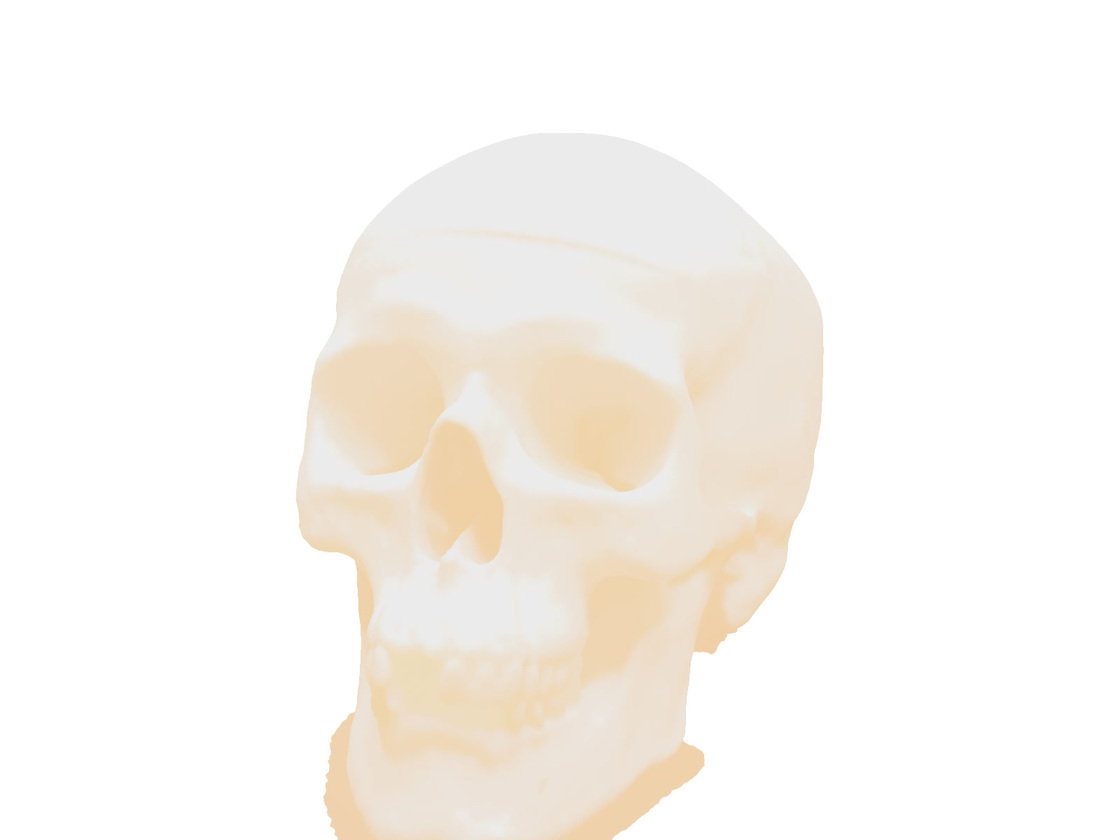}
    \end{subfigure}
    \hspace{1pt}
    \begin{subfigure}[h]{0.14\paperwidth}
        \includegraphics[width=\textwidth]{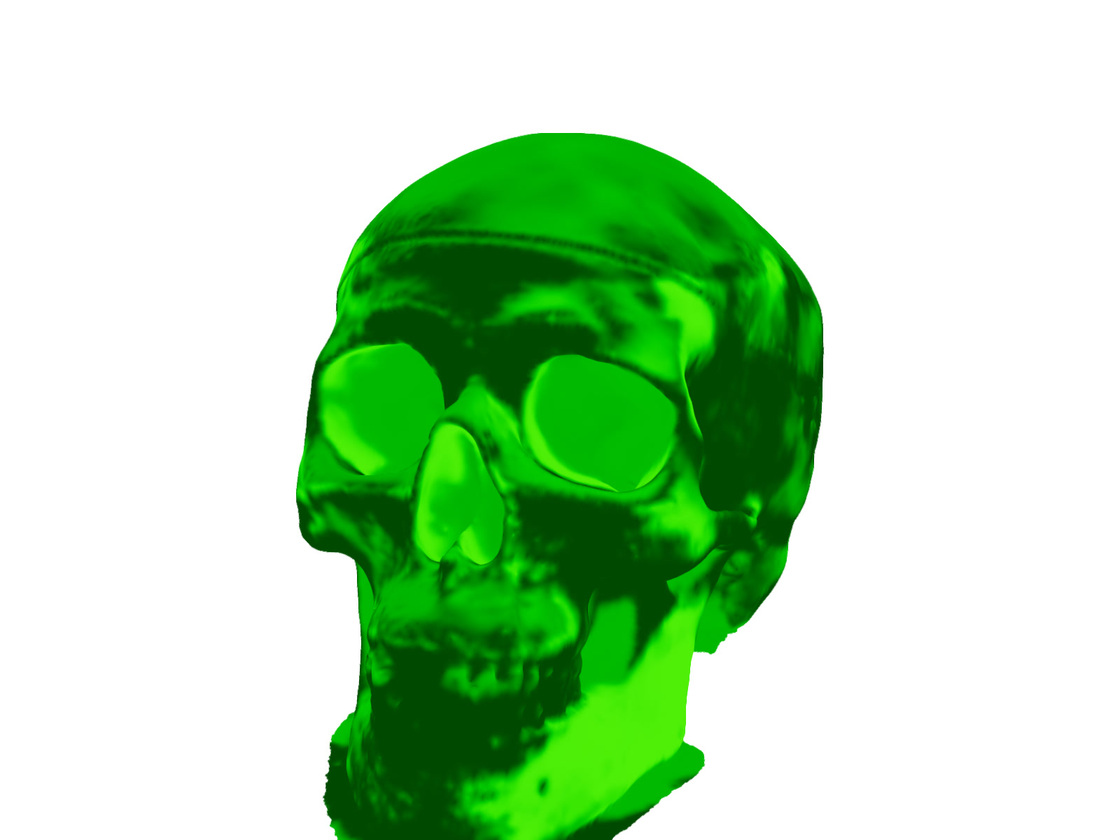}
    \end{subfigure}
    \hspace{1pt}
    \begin{subfigure}[h]{0.14\paperwidth}
        \includegraphics[width=\textwidth]{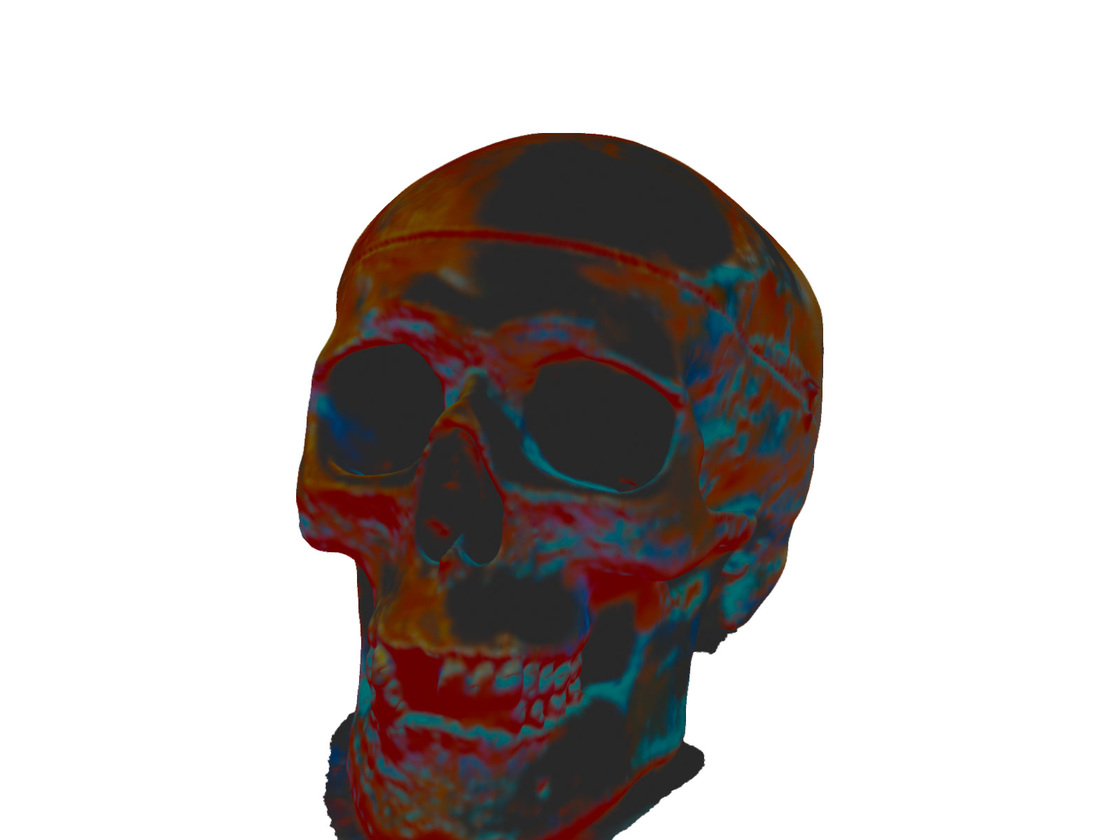}
    \end{subfigure}
    \hspace{1pt}
    \begin{subfigure}[h]{0.14\paperwidth}
        \includegraphics[width=\textwidth]{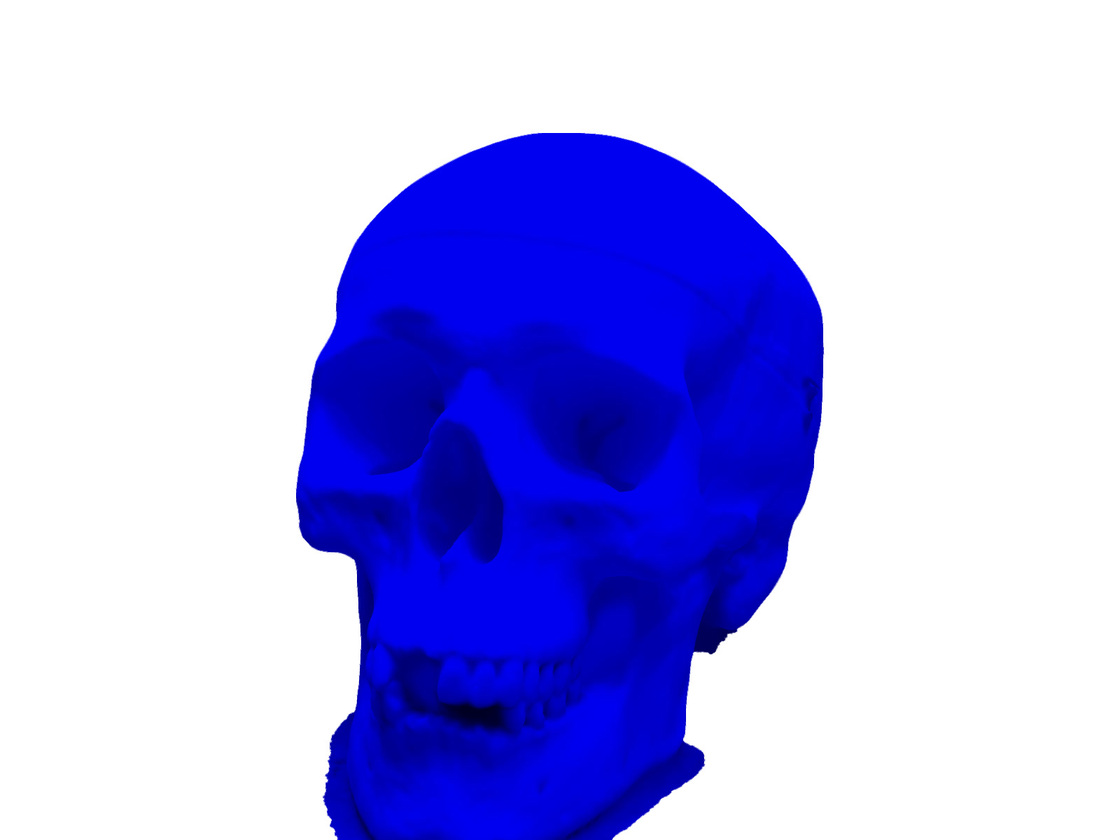}
    \end{subfigure}

    \smallskip
    \rotatebox[origin=b]{90}{scan69}\quad
    \begin{subfigure}[h]{0.14\paperwidth}
        \includegraphics[width=\textwidth]{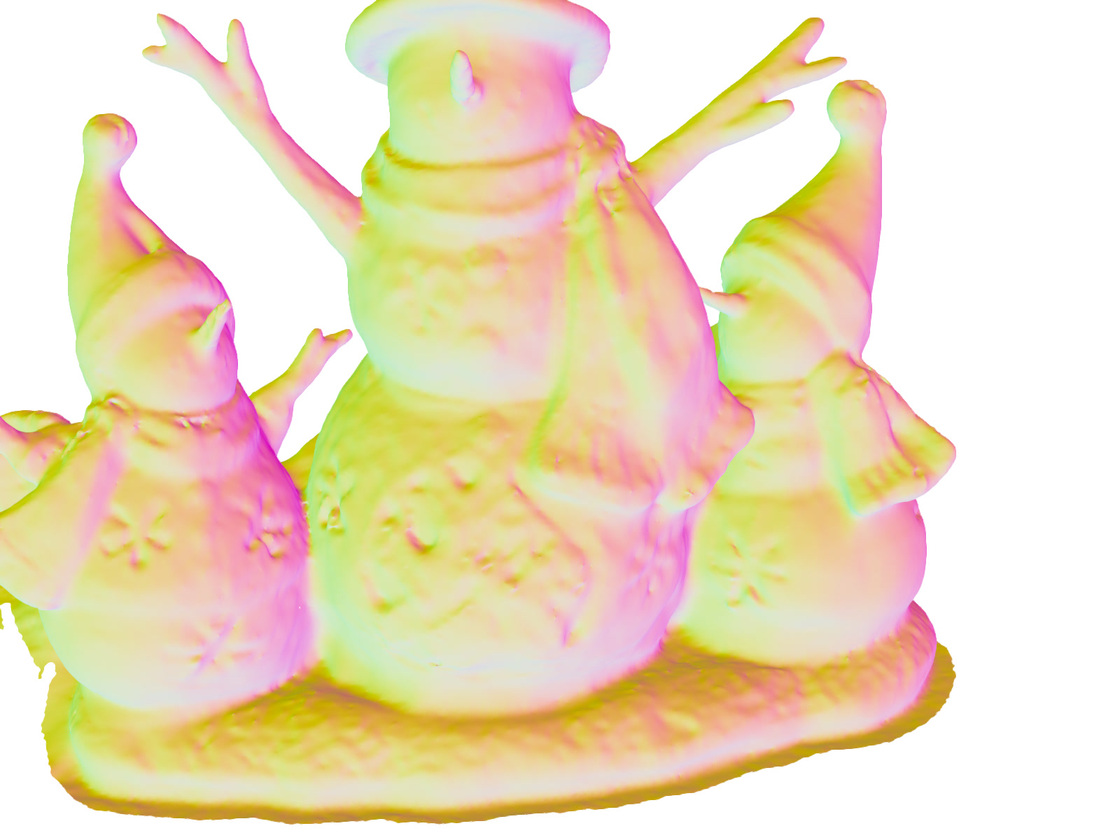}
    \end{subfigure}
    \hspace{1pt}
    \begin{subfigure}[h]{0.14\paperwidth}
        \includegraphics[width=\textwidth]{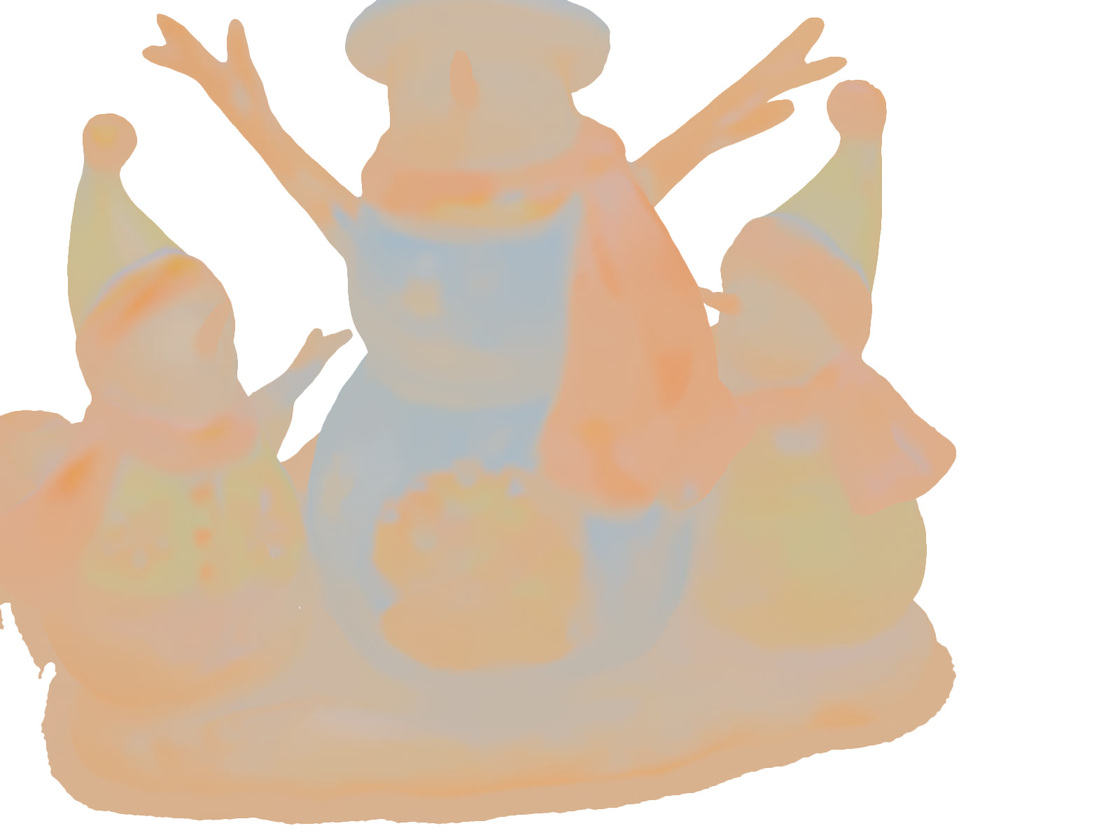}
    \end{subfigure}
    \hspace{1pt}
    \begin{subfigure}[h]{0.14\paperwidth}
        \includegraphics[width=\textwidth]{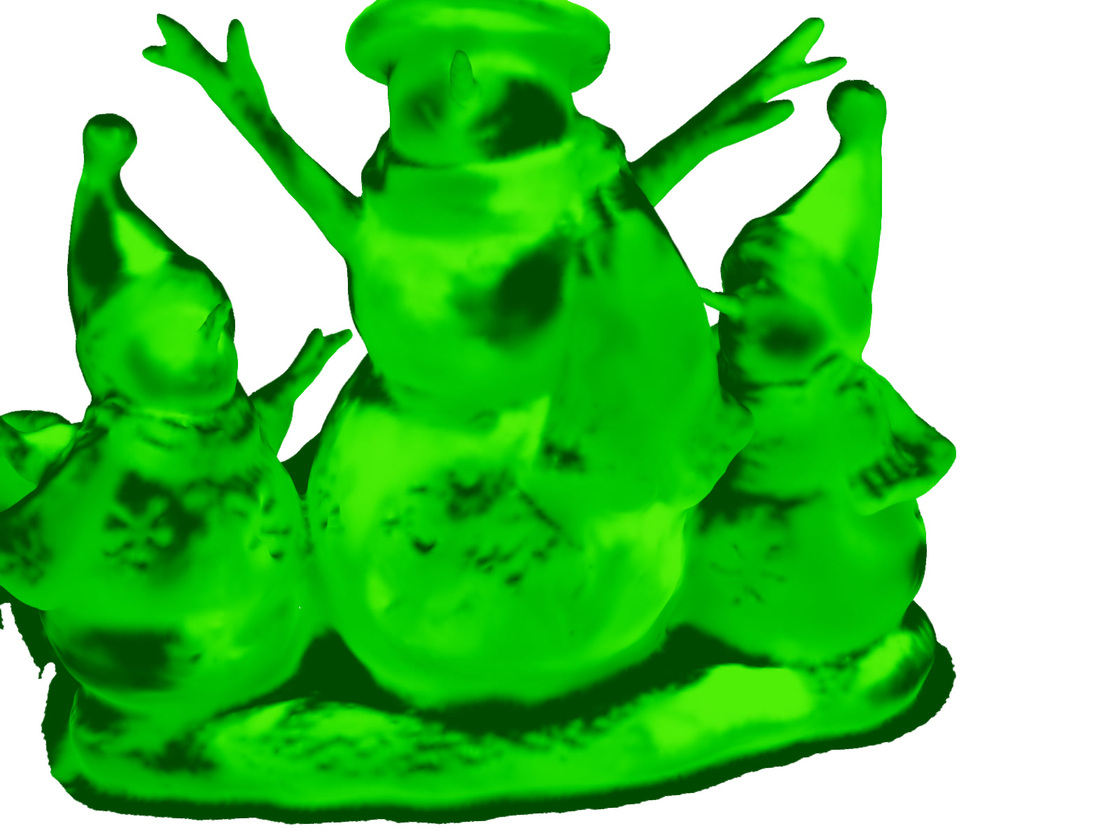}
    \end{subfigure}
    \hspace{1pt}
    \begin{subfigure}[h]{0.14\paperwidth}
        \includegraphics[width=\textwidth]{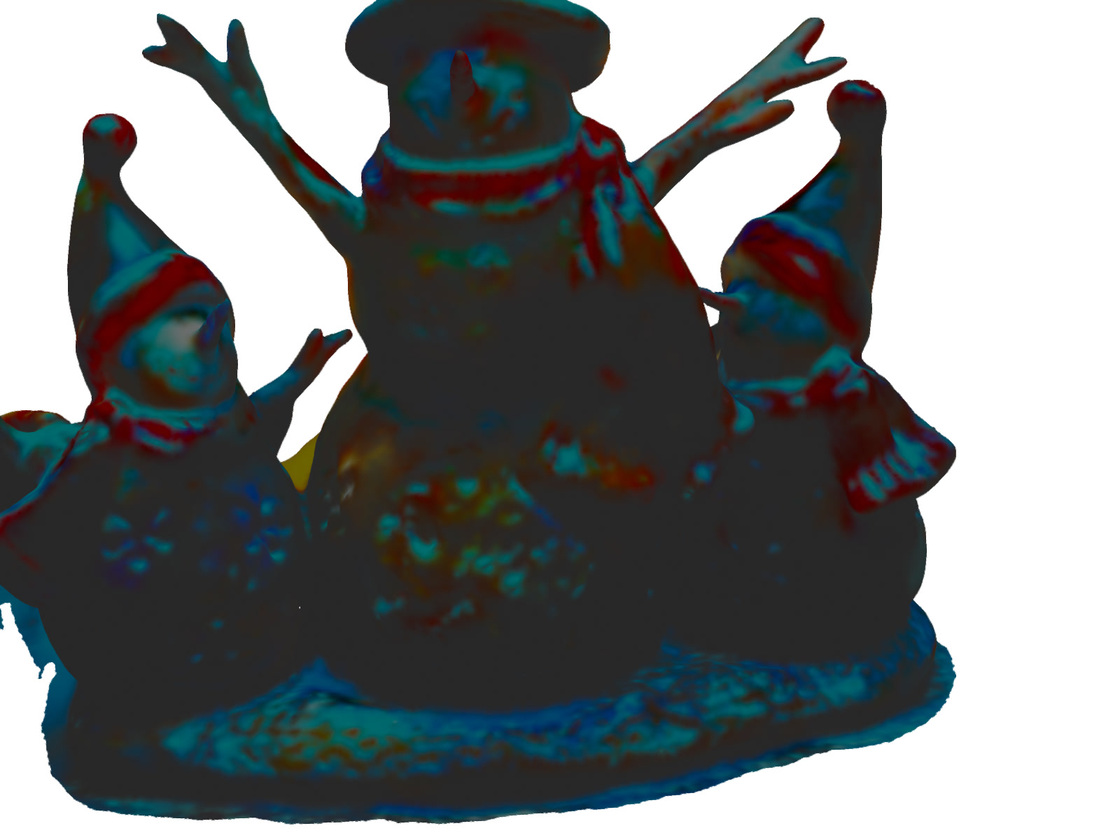}
    \end{subfigure}
    \hspace{1pt}
    \begin{subfigure}[h]{0.14\paperwidth}
        \includegraphics[width=\textwidth]{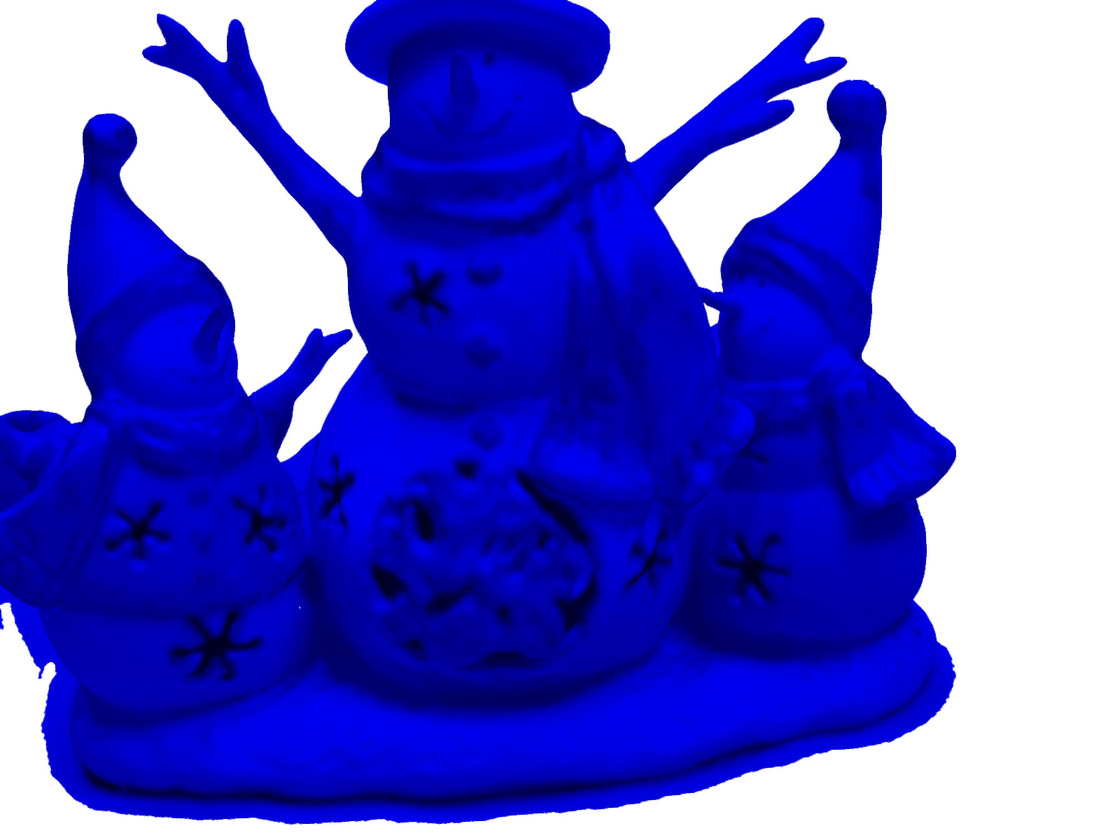}
    \end{subfigure}

    \smallskip
    \rotatebox[origin=b]{90}{scan118}\quad
    \begin{subfigure}[h]{0.14\paperwidth}
        \includegraphics[width=\textwidth]{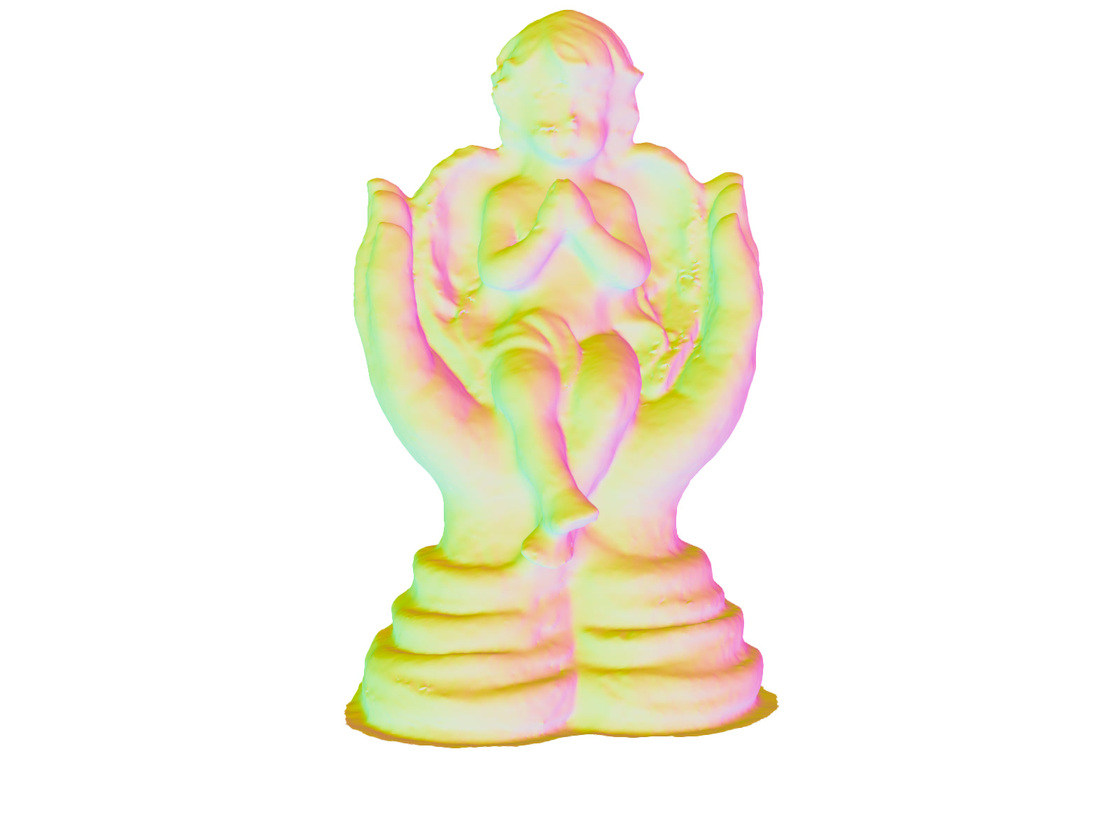}
    \end{subfigure}
    \hspace{1pt}
    \begin{subfigure}[h]{0.14\paperwidth}
        \includegraphics[width=\textwidth]{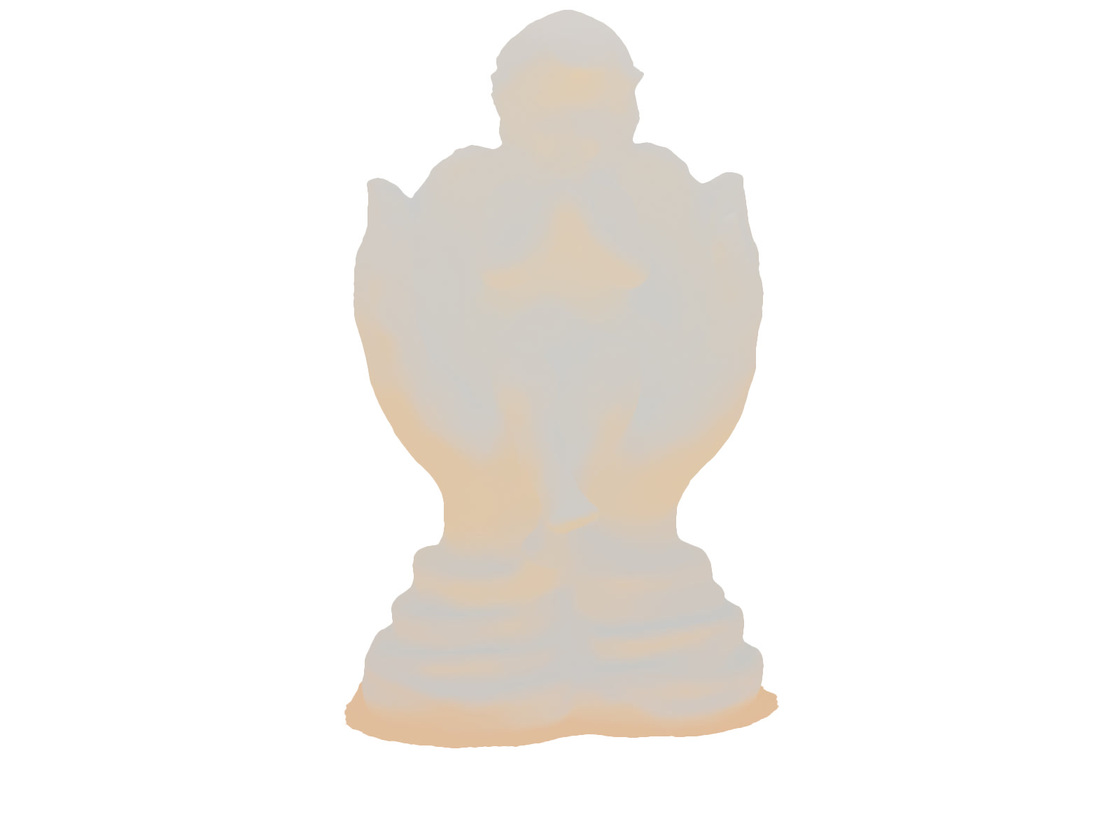}
    \end{subfigure}
    \hspace{1pt}
    \begin{subfigure}[h]{0.14\paperwidth}
        \includegraphics[width=\textwidth]{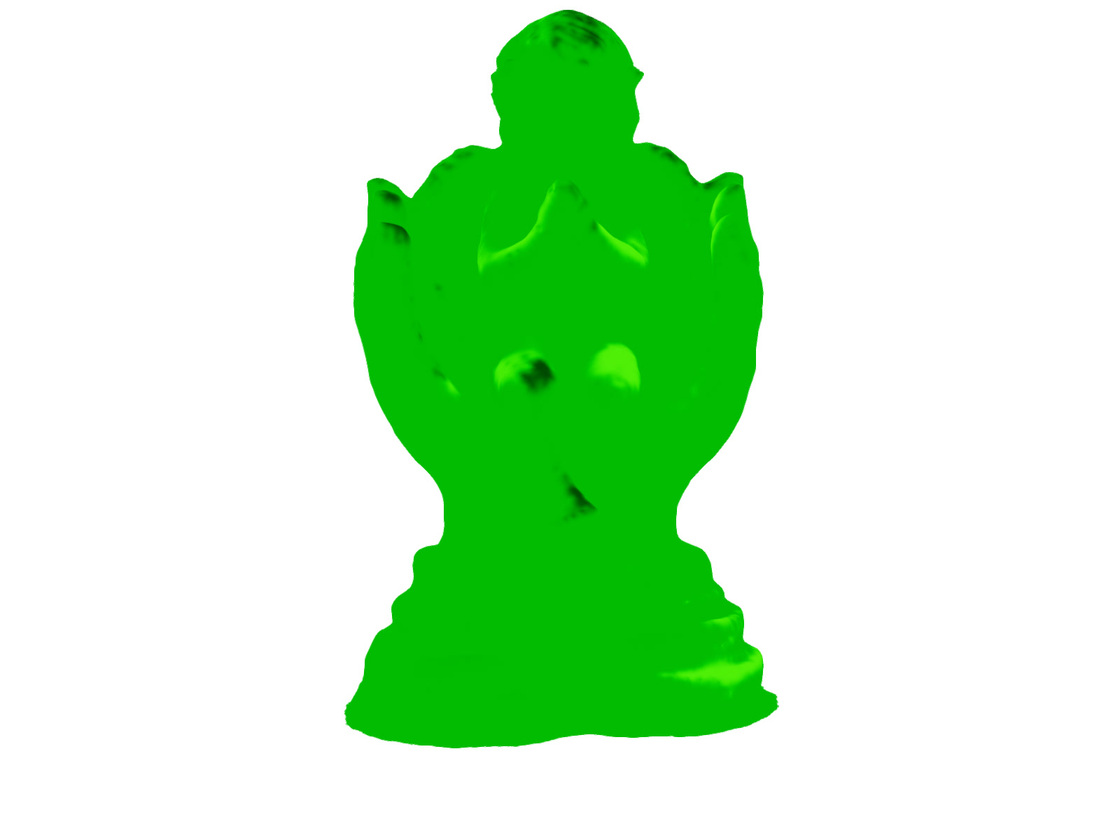}
    \end{subfigure}
    \hspace{1pt}
    \begin{subfigure}[h]{0.14\paperwidth}
        \includegraphics[width=\textwidth]{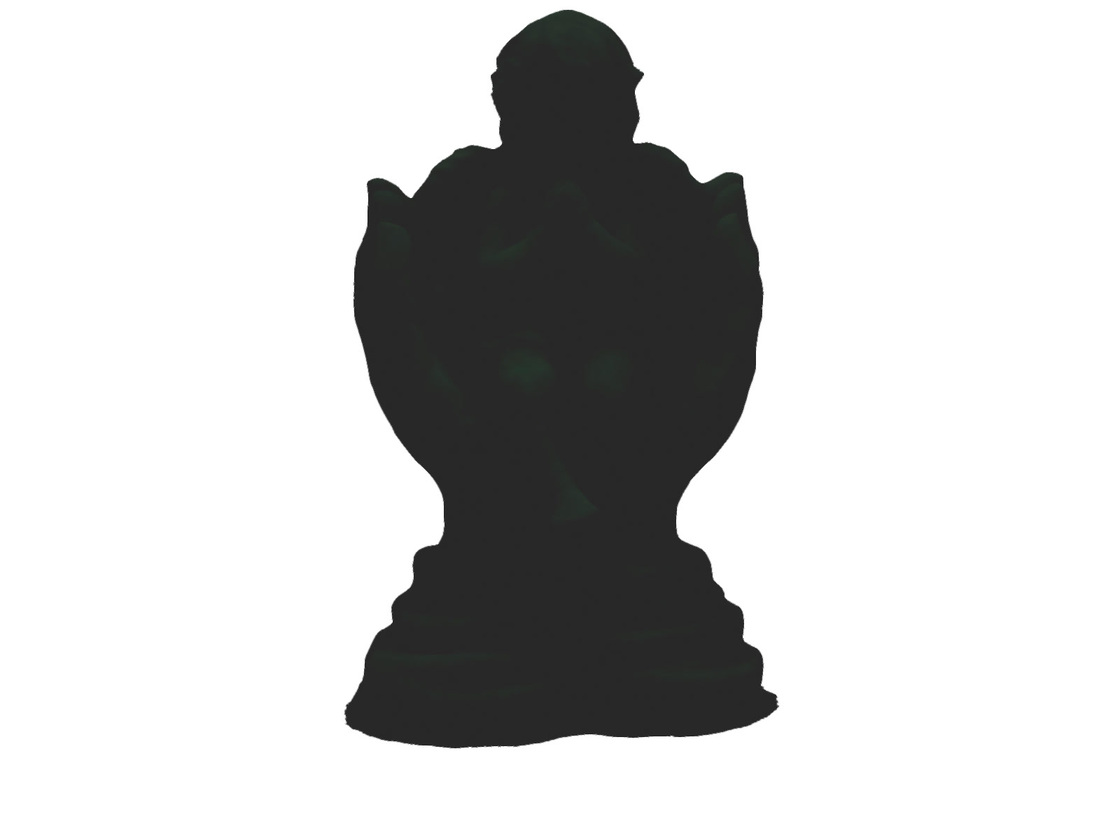}
    \end{subfigure}
    \hspace{1pt}
    \begin{subfigure}[h]{0.14\paperwidth}
        \includegraphics[width=\textwidth]{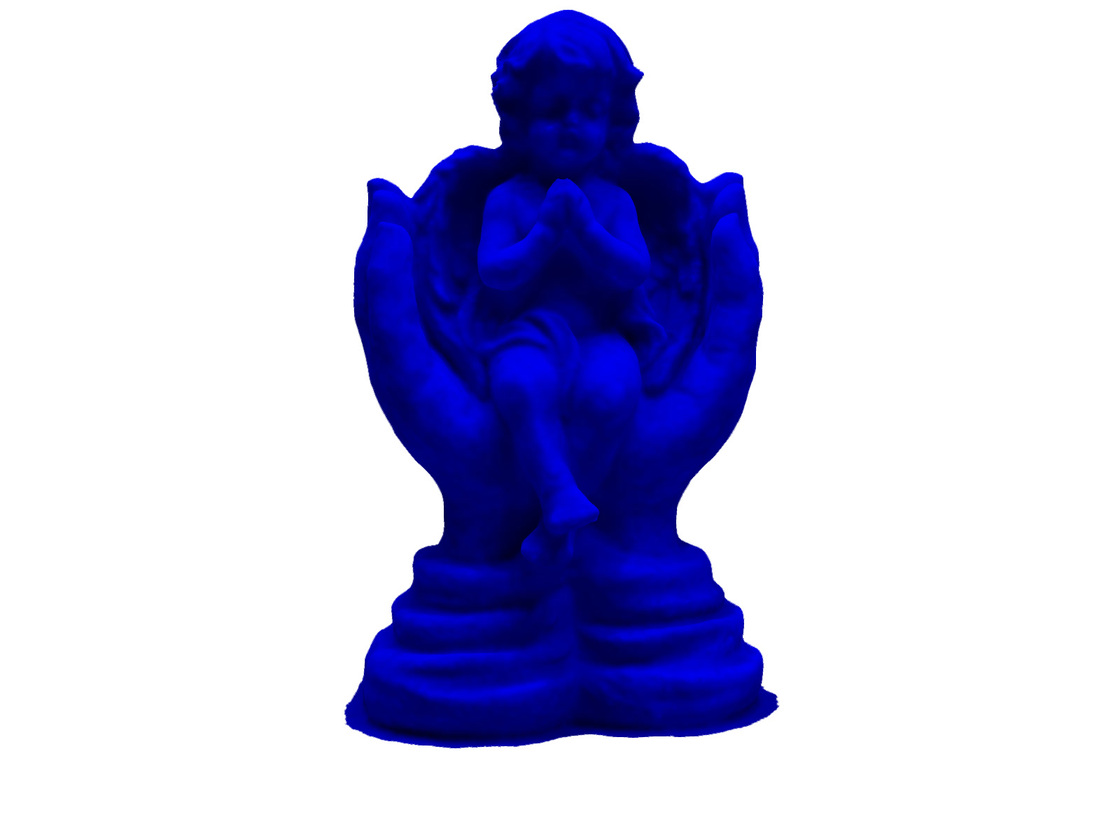}
    \end{subfigure}

    \smallskip
    \rotatebox[origin=b]{90}{scan24}\quad
    \begin{subfigure}[h]{0.17\paperwidth}
        \caption{PBR (default)}
        \includegraphics[width=\textwidth]{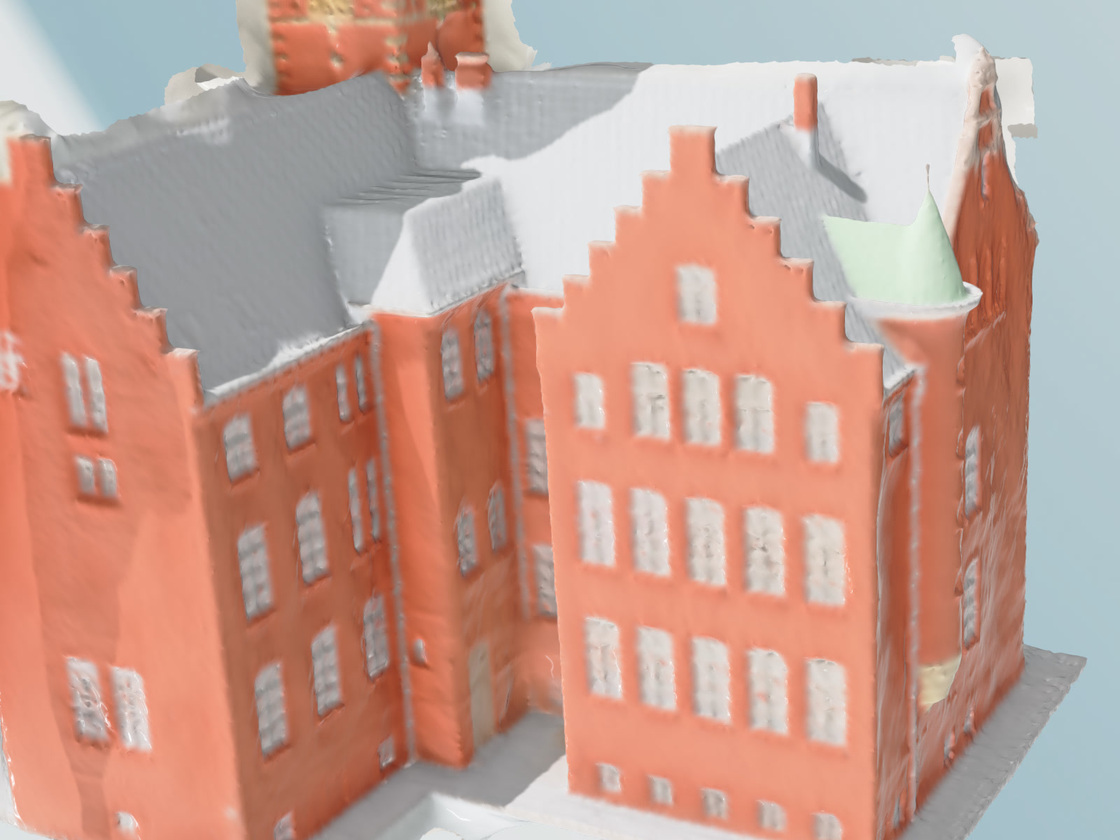}
    \end{subfigure}
    \hspace{1pt}
    \begin{subfigure}[h]{0.17\paperwidth}
        \caption{PBR (pillars)}
        \includegraphics[width=\textwidth]{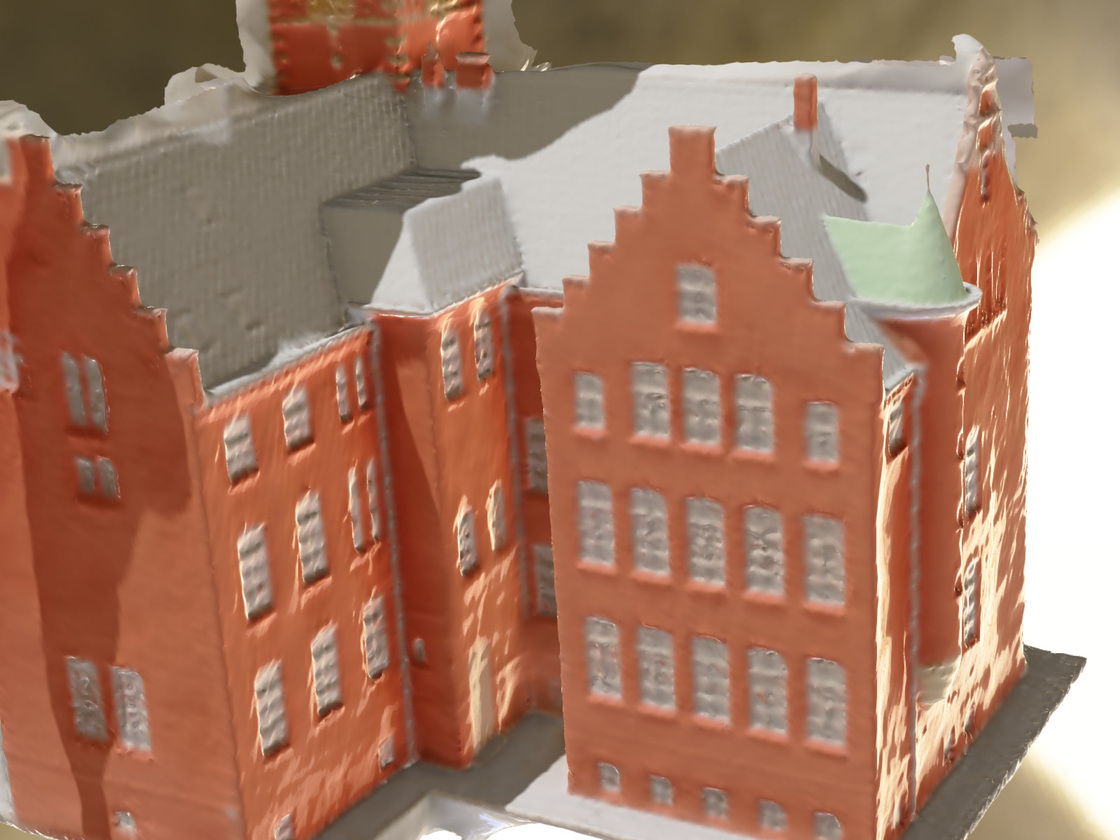}
    \end{subfigure}
    \hspace{1pt}
    \begin{subfigure}[h]{0.17\paperwidth}
        \caption{Neural rendering}
        \includegraphics[width=\textwidth]{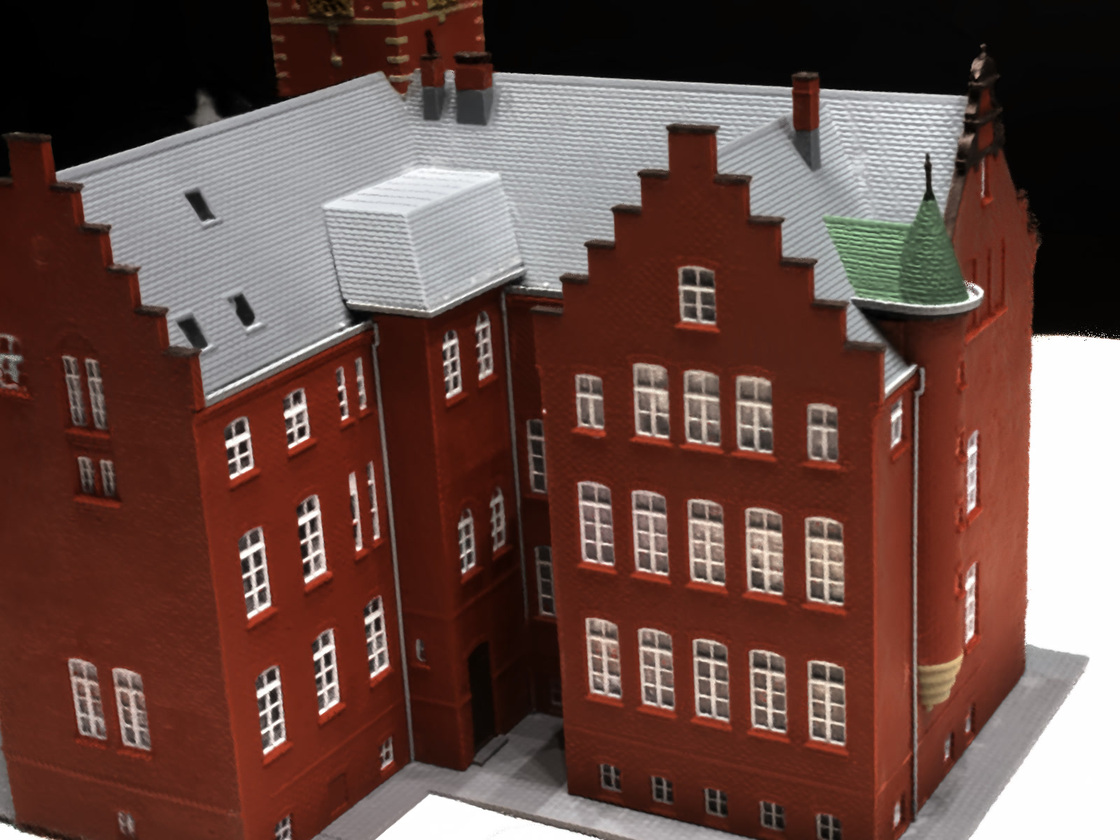}
    \end{subfigure}
    \hspace{1pt}
    \begin{subfigure}[h]{0.17\paperwidth}
        \caption{Groundtruth}
        \includegraphics[width=\textwidth]{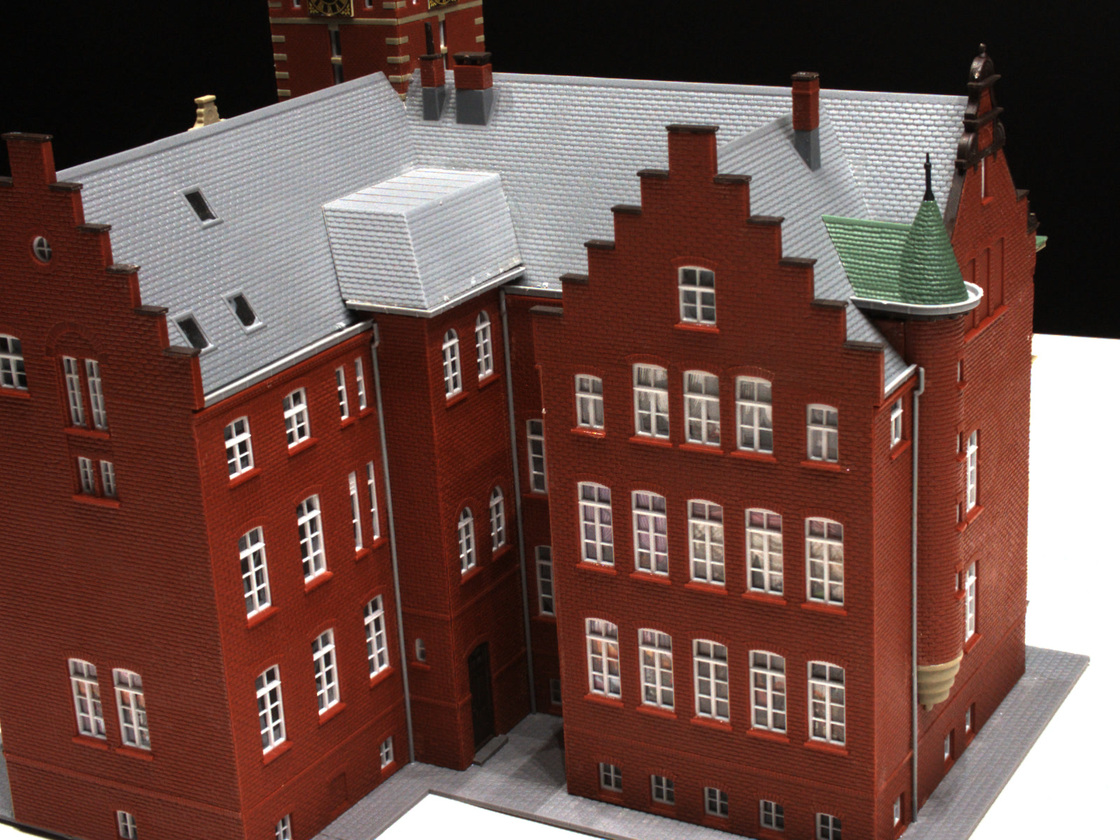}
    \end{subfigure}

    \smallskip
    \rotatebox[origin=b]{90}{scan65}\quad
    \begin{subfigure}[h]{0.17\paperwidth}
        \includegraphics[width=\textwidth]{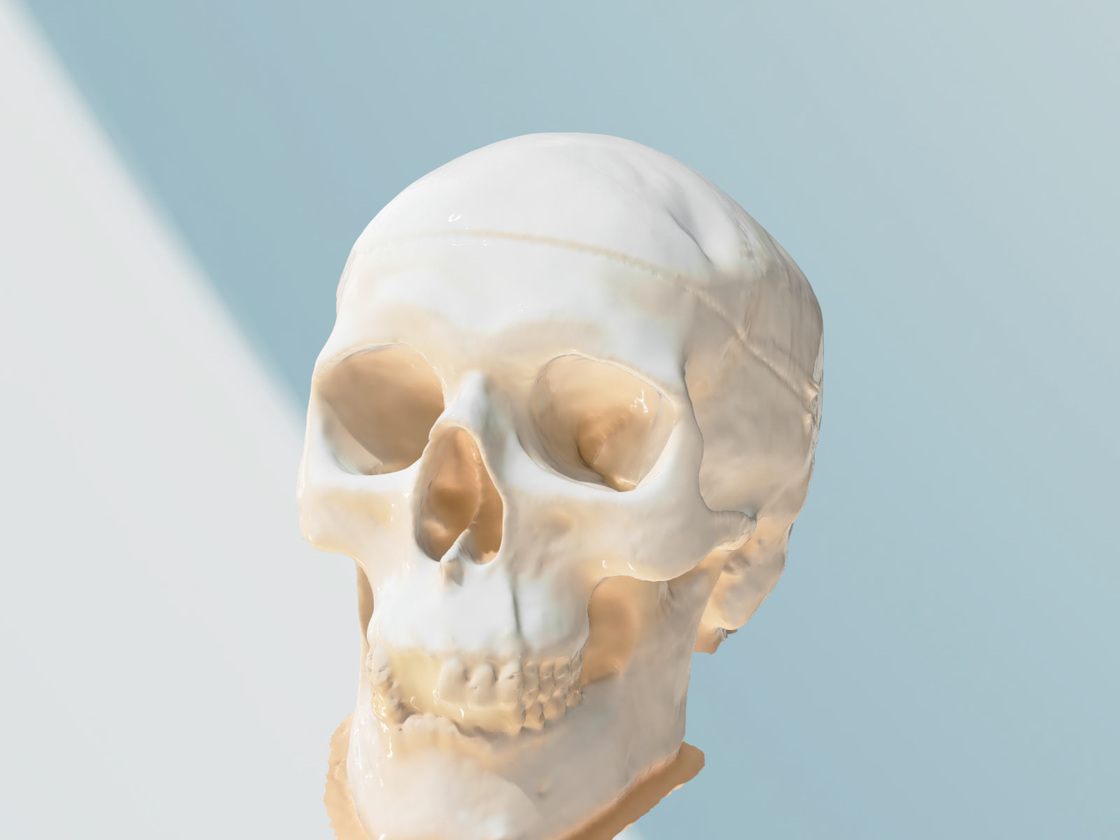}
    \end{subfigure}
    \hspace{1pt}
    \begin{subfigure}[h]{0.17\paperwidth}
        \includegraphics[width=\textwidth]{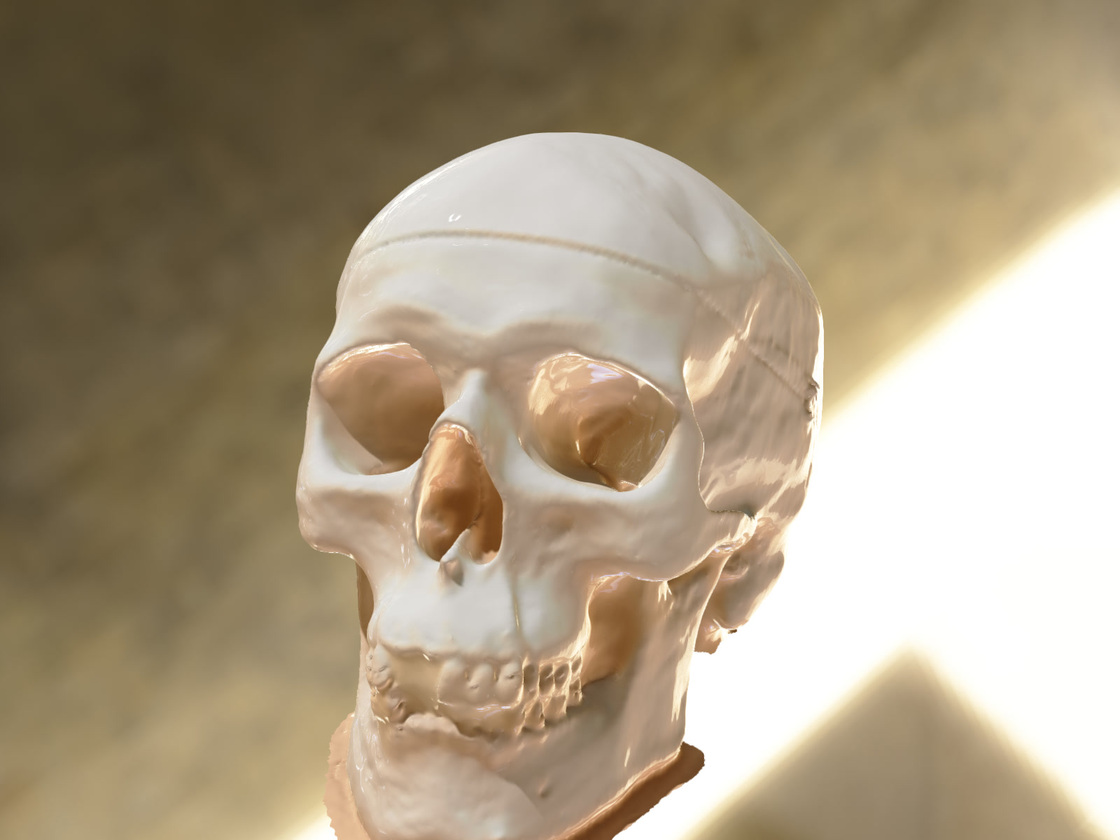}
    \end{subfigure}
    \hspace{1pt}
    \begin{subfigure}[h]{0.17\paperwidth}
        \includegraphics[width=\textwidth]{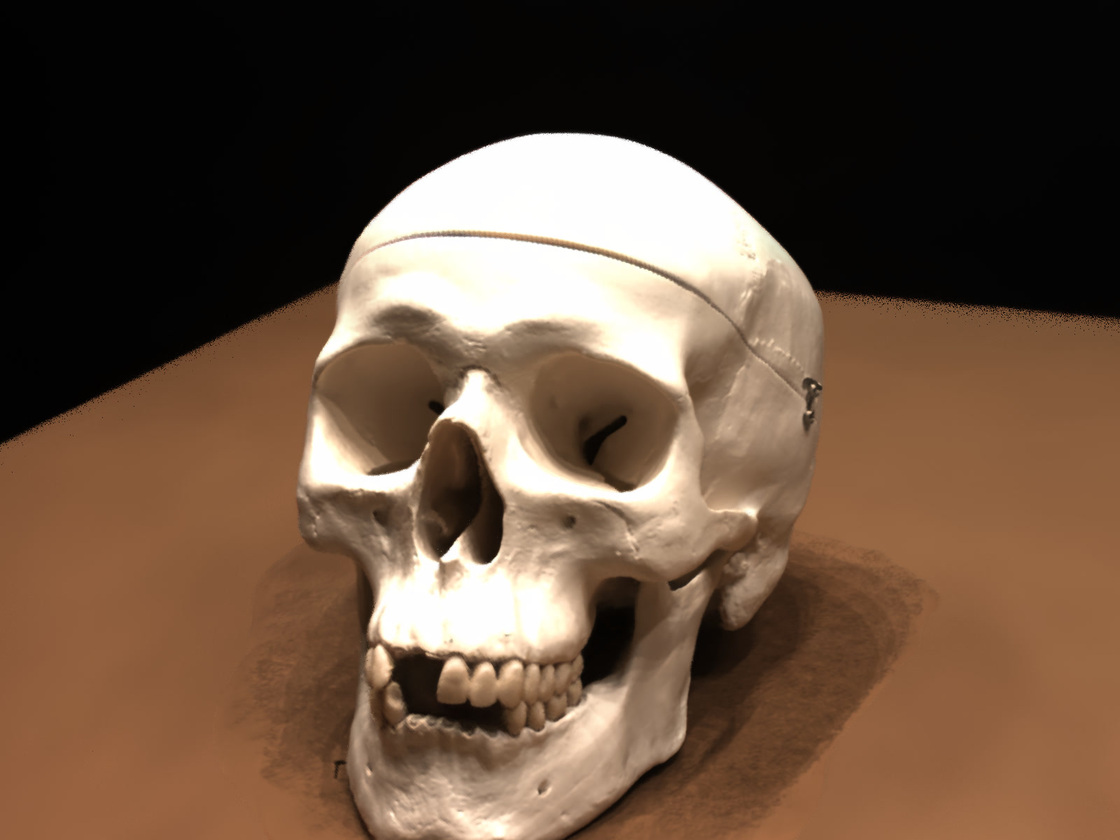}
    \end{subfigure}
    \hspace{1pt}
    \begin{subfigure}[h]{0.17\paperwidth}
        \includegraphics[width=\textwidth]{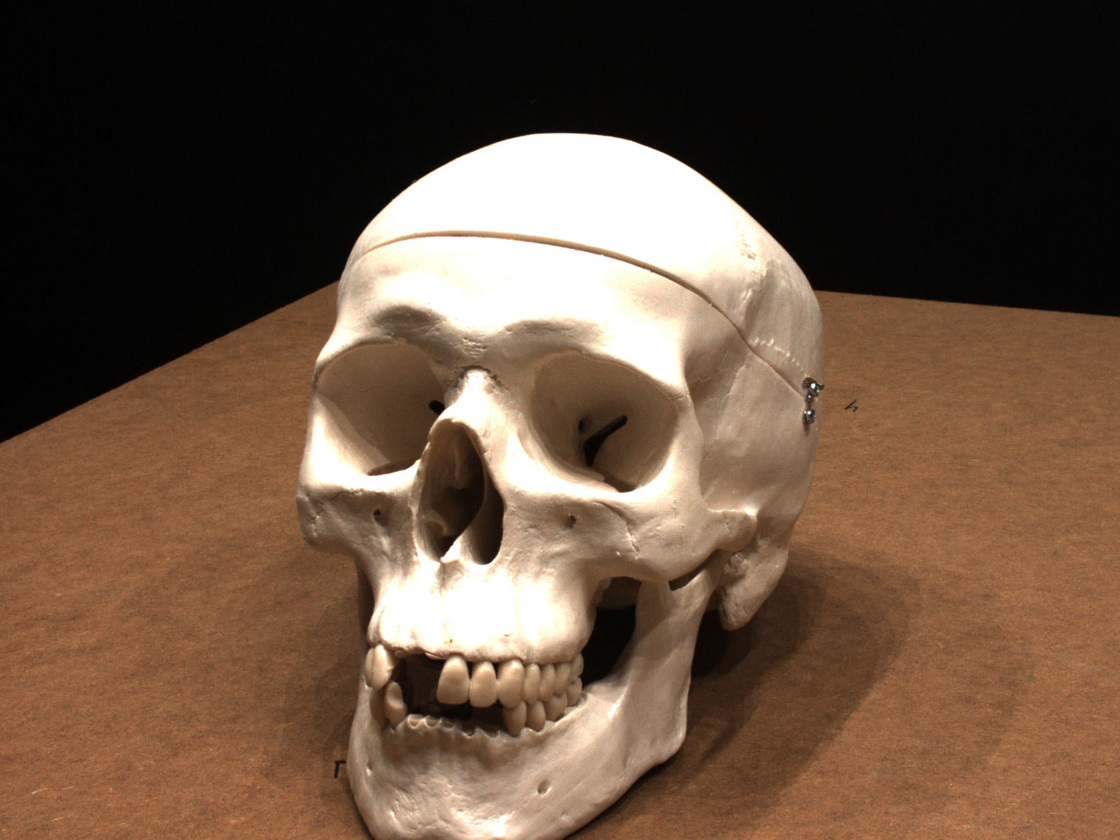}
    \end{subfigure}

    \smallskip
    \rotatebox[origin=b]{90}{scan69}\quad
    \begin{subfigure}[h]{0.17\paperwidth}
        \includegraphics[width=\textwidth]{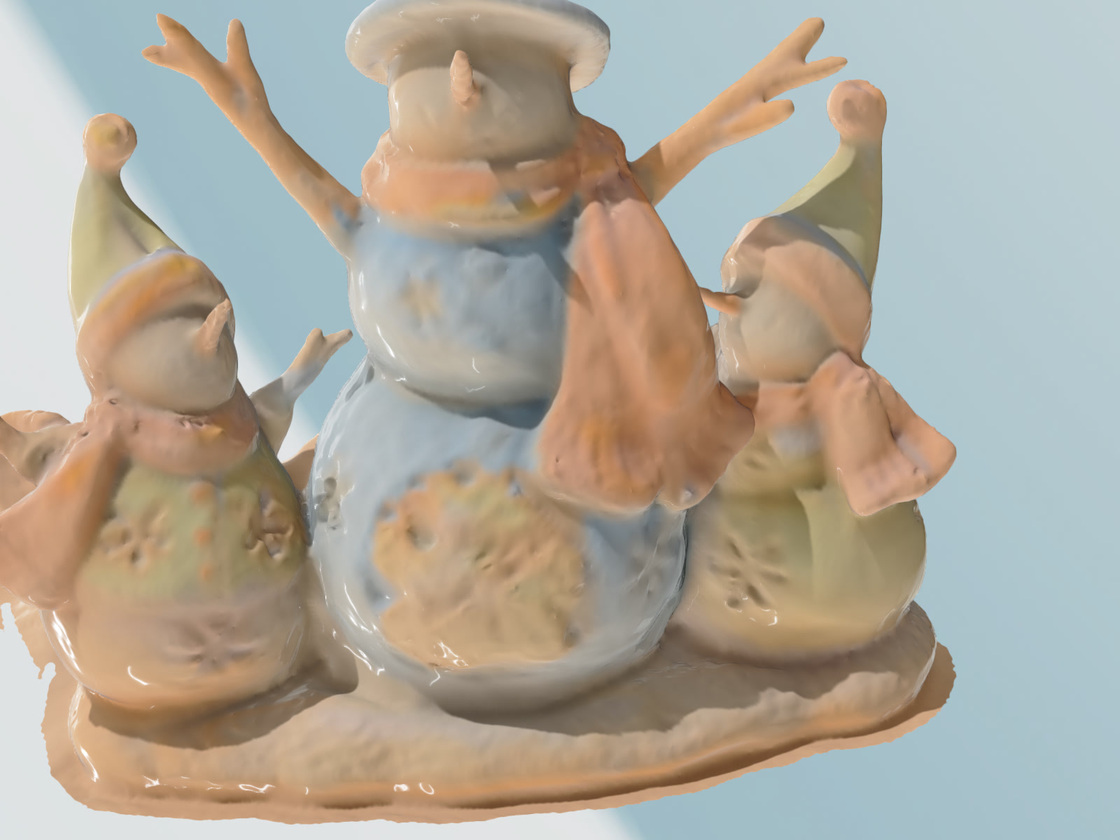}
    \end{subfigure}
    \hspace{1pt}
    \begin{subfigure}[h]{0.17\paperwidth}
        \includegraphics[width=\textwidth]{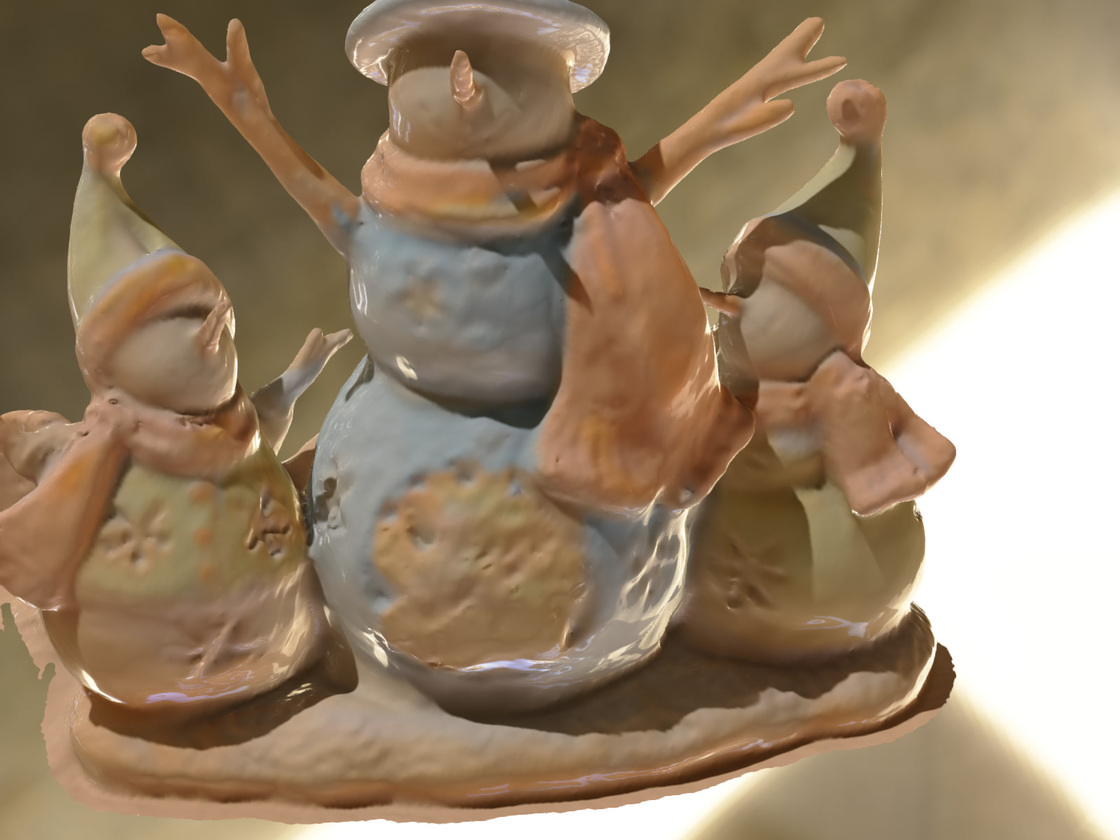}
    \end{subfigure}
    \hspace{1pt}
    \begin{subfigure}[h]{0.17\paperwidth}
        \includegraphics[width=\textwidth]{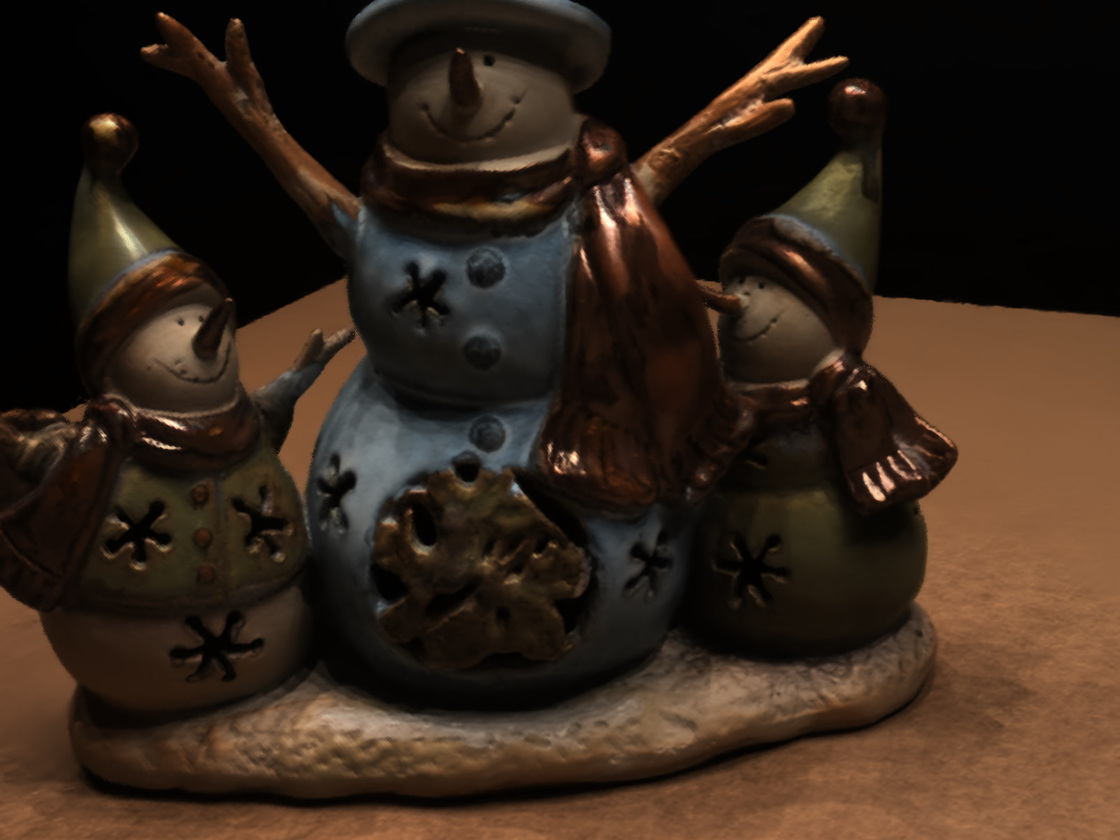}
    \end{subfigure}
    \hspace{1pt}
    \begin{subfigure}[h]{0.17\paperwidth}
        \includegraphics[width=\textwidth]{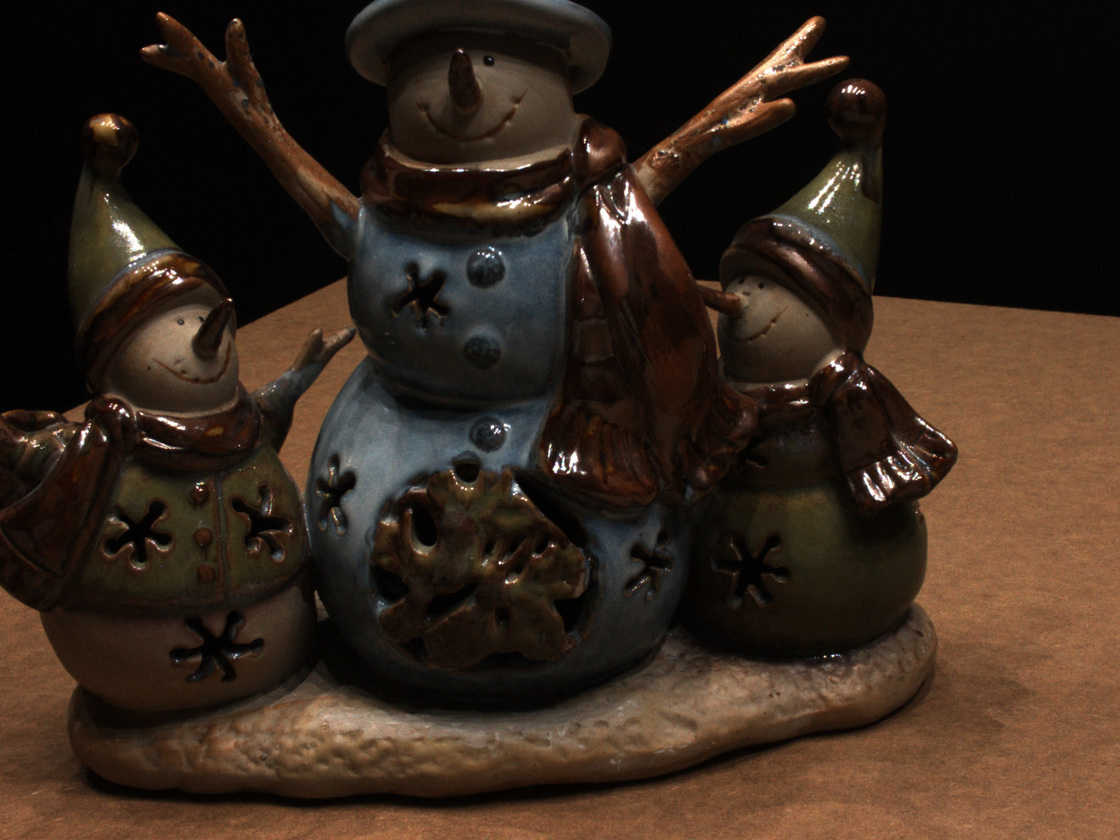}
    \end{subfigure}

    \smallskip
    \rotatebox[origin=b]{90}{scan118}\quad
    \begin{subfigure}[h]{0.17\paperwidth}
        \includegraphics[width=\textwidth]{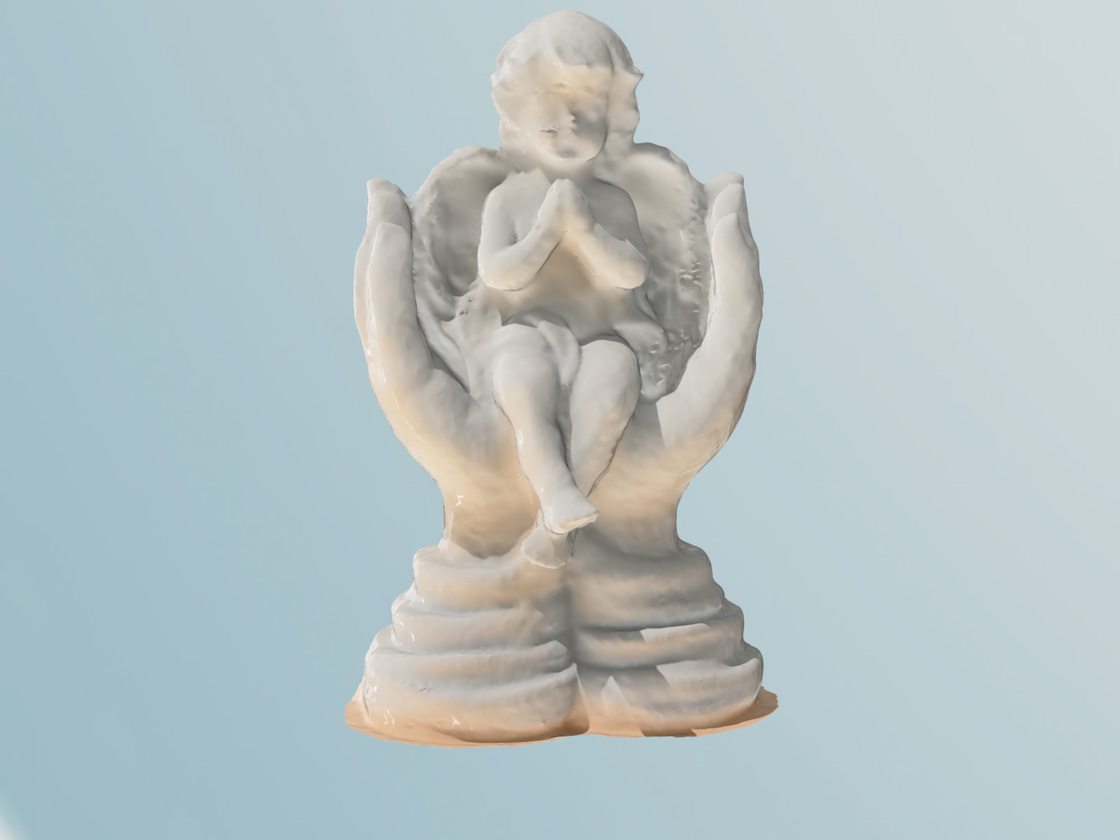}
    \end{subfigure}
    \hspace{1pt}
    \begin{subfigure}[h]{0.17\paperwidth}
        \includegraphics[width=\textwidth]{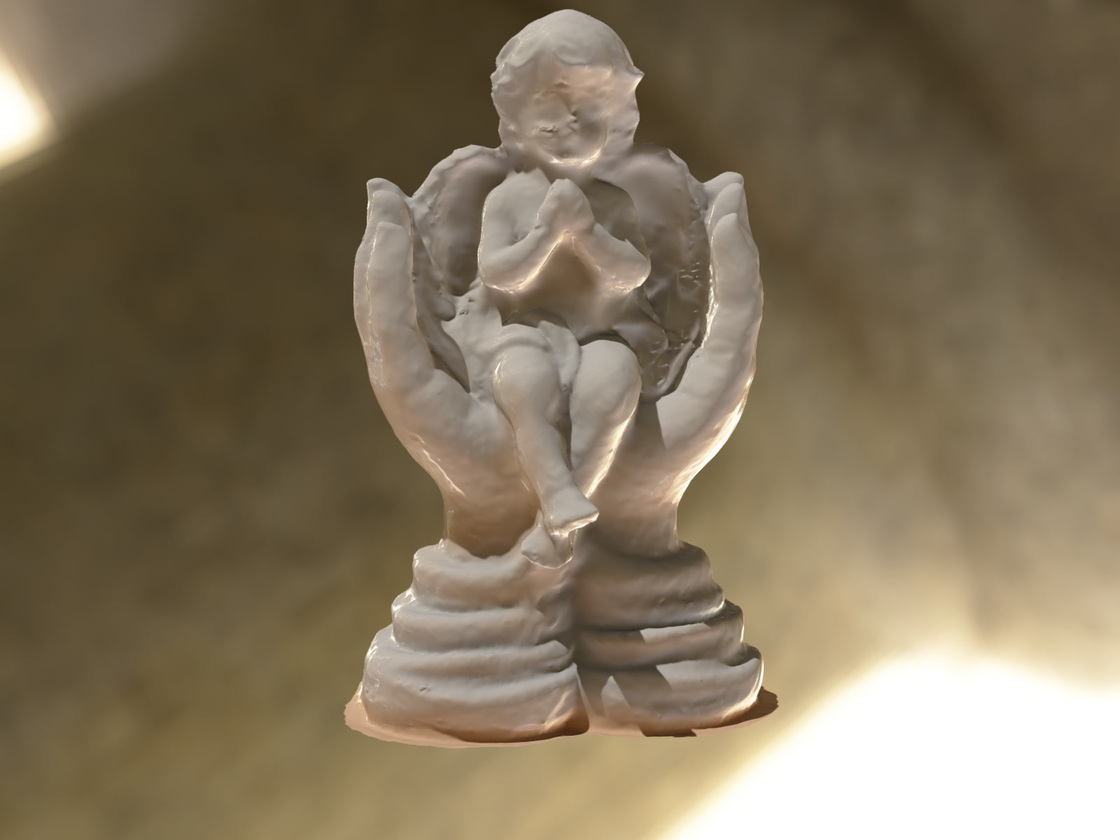}
    \end{subfigure}
    \hspace{1pt}
    \begin{subfigure}[h]{0.17\paperwidth}
        \includegraphics[width=\textwidth]{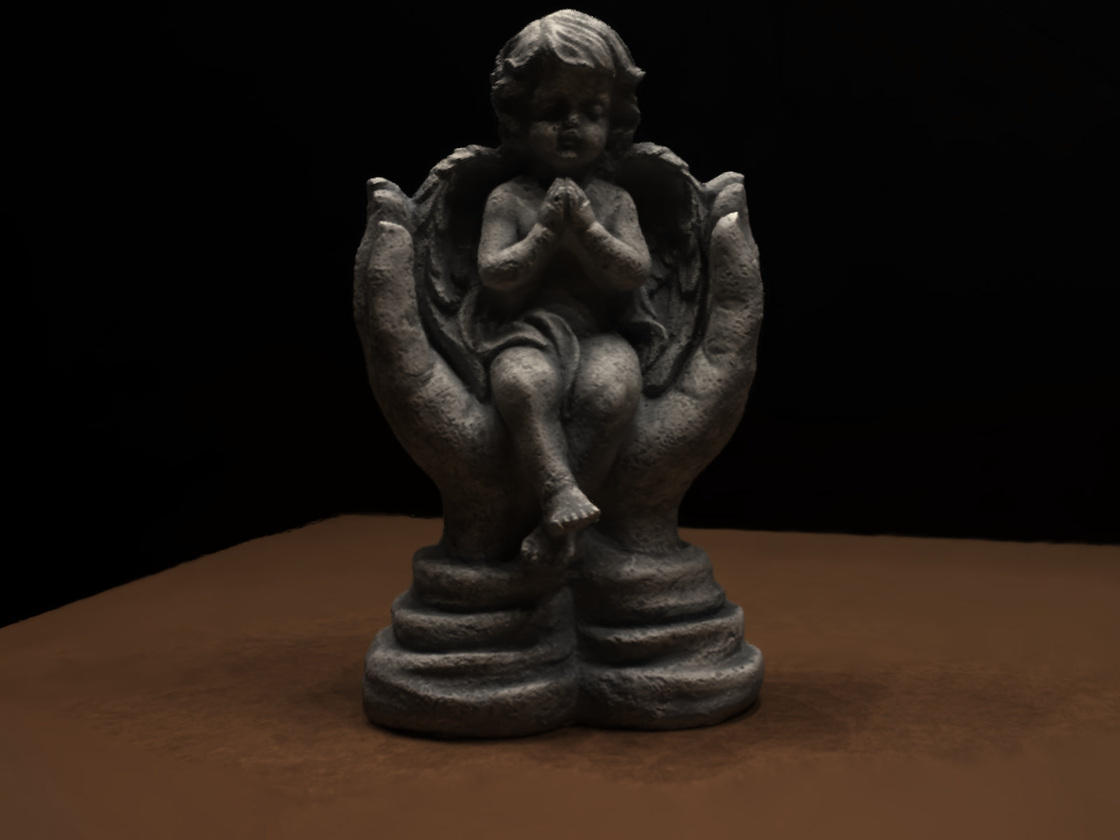}
    \end{subfigure}
    \hspace{1pt}
    \begin{subfigure}[h]{0.17\paperwidth}
        \includegraphics[width=\textwidth]{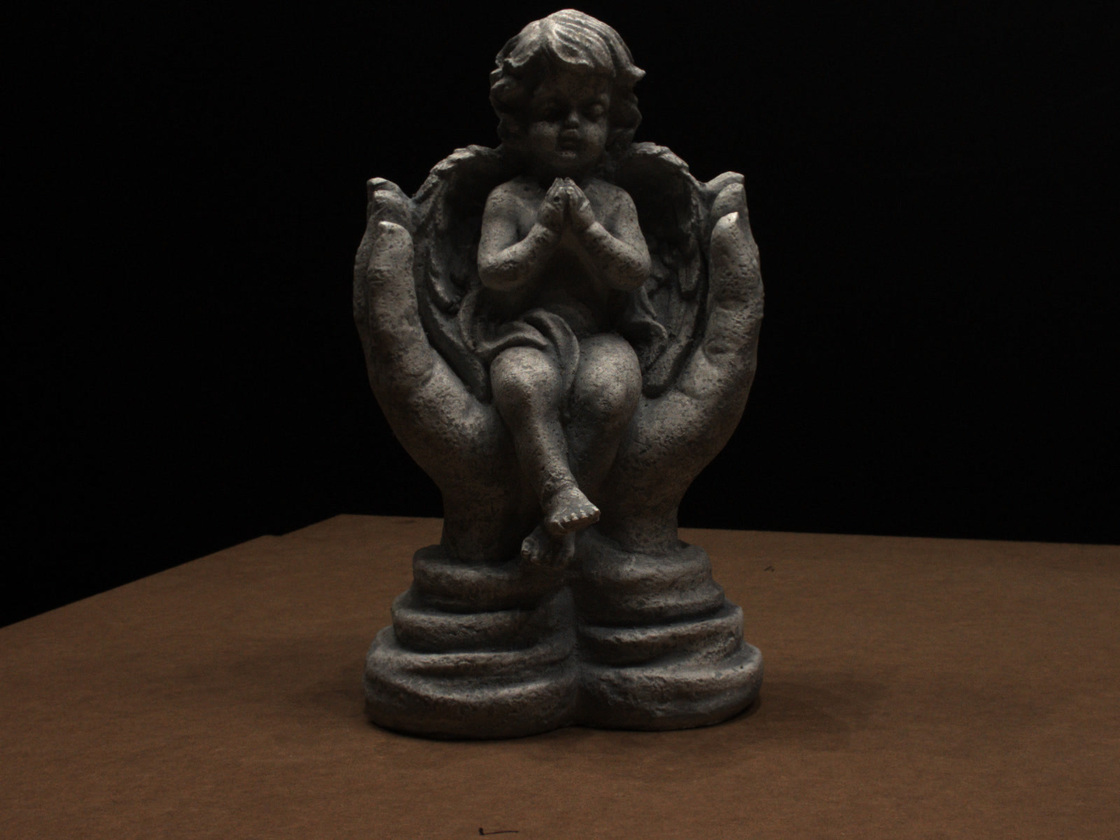}
    \end{subfigure}

    \caption{\textbf{Decomposed materials and rendered images.} ($\cdot$) is a given environment map in Open3D.}
    \label{fig:main_results}
\end{figure*}

\textbf{Decomposed geometry and materials:} \cref{fig:main_results} (top) shows decomposed geometry and materials. Geometries are well extracted. Base colors are flat and do not contain high light intensity and specularity. Roughness and specular reflectance reflect light intensity, specularity, and semantic object color in the scene. Higher specularity means lower roughness and higher specular reflectance. Parts of similar colors are segmented with close values of roughness and specular reflectance, yet light frequencies are considered. Implicit illumination certainly captures light distribution spatially for each scene.

\textbf{Rendered images:} \cref{fig:main_results} (bottom) illustrates the rendered images. Neurally-rendered images looks real, and PBR images are also towards photorealistic.

\begin{table}[tbp]
    \footnotesize
    \centering
    \caption{\textbf{Qualitative evaluations}.}
    \label{tab:chamfer_psnr_ssim}
    \begin{tabular}{c|c|c|c|c|c|c}
        ~       & \multicolumn{2}{c}{Chamfer $\downarrow$} & \multicolumn{2}{|c}{PSNR $\uparrow$} & \multicolumn{2}{|c}{SSIM $\uparrow$}                                                    \\
        \hline \hline
        Scan ID & NeuS                                     & NDJIR                                & NeuS                                 & NDJIR          & NeuS           & NDJIR          \\
        \hline \hline
        24      & 1.00                                     & \textbf{0.70}                        & 23.98                                & \textbf{28.10} & 0.732          & \textbf{0.820} \\
        37      & 1.37                                     & \textbf{1.19}                        & 22.79                                & \textbf{23.67} & 0.778          & \textbf{0.803} \\
        40      & 0.93                                     & \textbf{0.64}                        & 25.21                                & \textbf{27.51} & 0.722          & \textbf{0.761} \\
        55      & \textbf{0.43}                            & 0.49                                 & 26.03                                & \textbf{28.61} & 0.739          & \textbf{0.783} \\
        63      & \textbf{1.10}                            & 1.53                                 & 28.32                                & \textbf{30.53} & \textbf{0.915} & 0.890          \\
        65      & \textbf{0.65}                            & 0.83                                 & 29.80                                & \textbf{31.39} & 0.809          & \textbf{0.839} \\
        69      & \textbf{0.57}                            & 0.78                                 & 27.45                                & \textbf{28.98} & 0.818          & \textbf{0.847} \\
        83      & 1.48                                     & \textbf{1.04}                        & 28.89                                & \textbf{31.90} & 0.831          & \textbf{0.842} \\
        97      & \textbf{1.09}                            & 1.21                                 & 26.03                                & \textbf{29.15} & 0.812          & \textbf{0.819} \\
        105     & 0.83                                     & \textbf{0.79}                        & 28.93                                & \textbf{31.71} & 0.815          & \textbf{0.836} \\
        106     & \textbf{0.52}                            & \textbf{0.52}                        & 32.47                                & \textbf{33.30} & 0.866          & \textbf{0.877} \\
        110     & \textbf{1.20}                            & 2.09                                 & 30.78                                & \textbf{31.81} & 0.863          & \textbf{0.868} \\
        114     & \textbf{0.35}                            & \textbf{0.35}                        & 29.37                                & \textbf{30.34} & \textbf{0.847} & 0.842          \\
        118     & \textbf{0.49}                            & 0.64                                 & \textbf{34.23}                       & 33.29          & \textbf{0.878} & 0.870          \\
        122     & \textbf{0.54}                            & 0.63                                 & \textbf{33.95}                       & 27.55          & \textbf{0.878} & 0.869
    \end{tabular}
\end{table}
\begin{figure}[tbp]
    \centering
    \captionsetup[subfigure]{font=scriptsize}
    \rotatebox[origin=b]{90}{normals}\quad
    \begin{subfigure}[h]{0.16\paperwidth}
        \caption{NDJIR}
        \includegraphics[width=\textwidth]{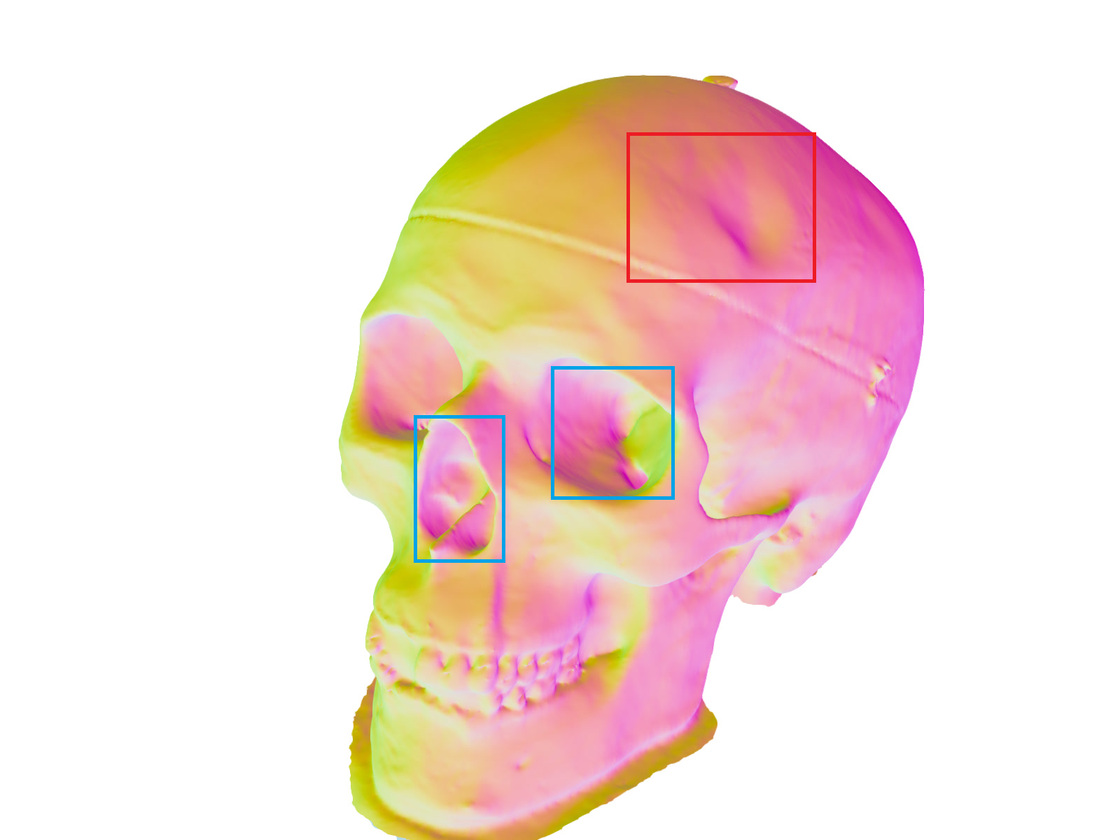}
    \end{subfigure}
    \hspace{1pt}
    \begin{subfigure}[h]{0.16\paperwidth}
        \caption{NDJIR w/ split-sum \cite{DBLP:conf/cvpr/MunkbergCHES0GF22} + PIL \cite{DBLP:conf/nips/BossJBLBL21}}
        \includegraphics[width=\textwidth]{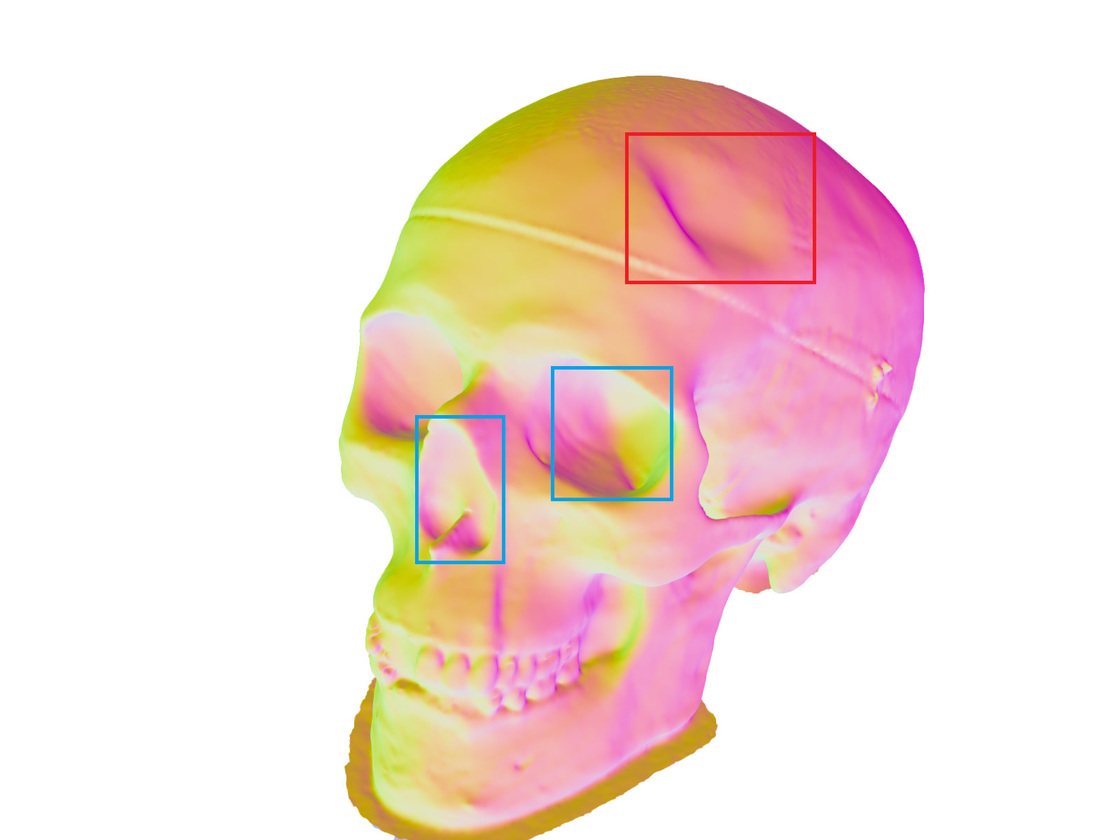}
    \end{subfigure}

    \smallskip
    \rotatebox[origin=b]{90}{roughness}\quad
    \begin{subfigure}[h]{0.16\paperwidth}
        \includegraphics[width=\textwidth]{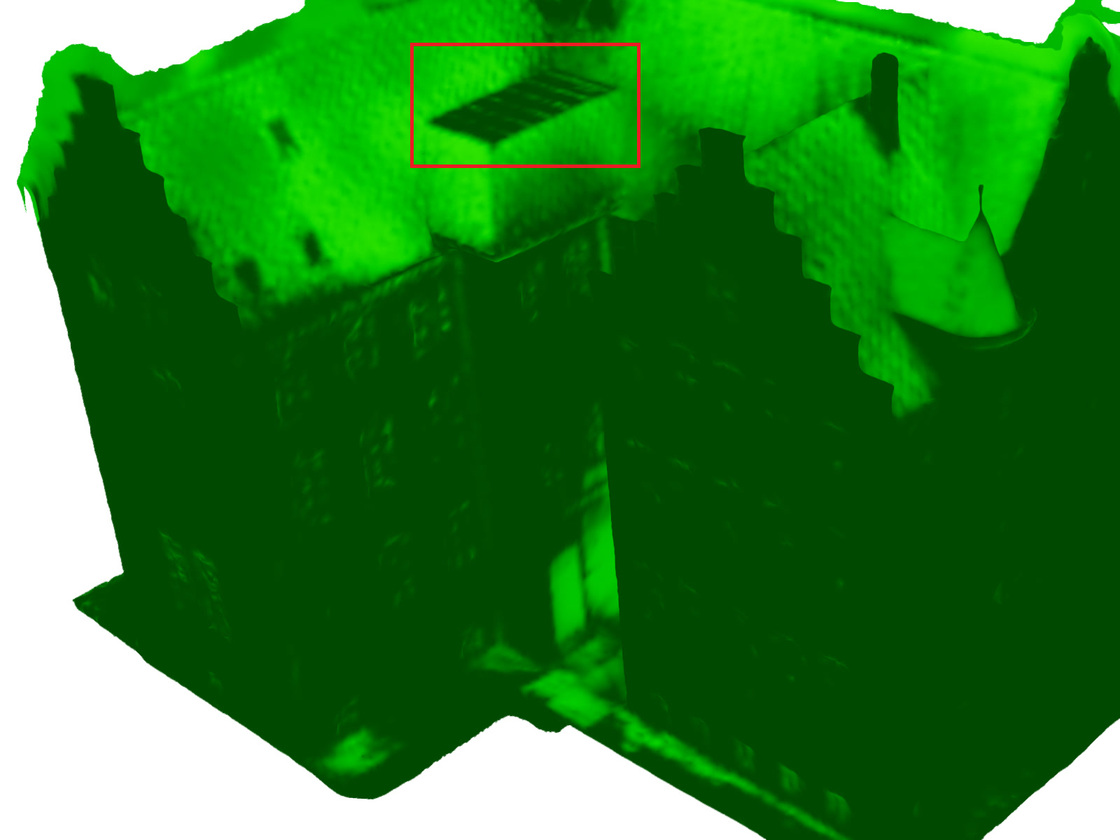}
    \end{subfigure}
    \hspace{1pt}
    \begin{subfigure}[h]{0.16\paperwidth}
        \includegraphics[width=\textwidth]{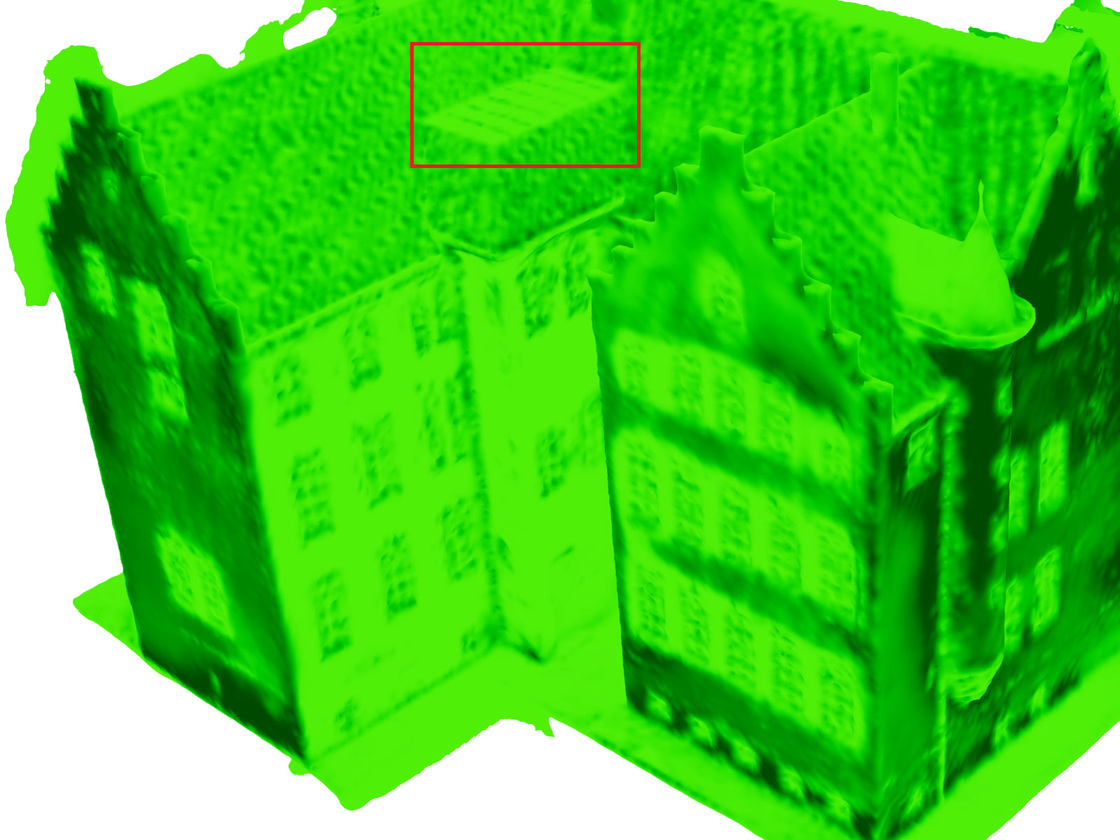}
    \end{subfigure}

    \caption{\textbf{Comparison with split-sum and pre-integrated light} using scan65 (top) and scan24 (bottom). Red boxes indicate high light intensity region, and blue boxes represent deep concave parts.}
    \label{fig:comparison_with_split-sum_and_pre-integrated_light}
\end{figure}

\textbf{Comparison to baseline:} \cref{tab:chamfer_psnr_ssim} shows quantitative comparison. NDJIR achieves competitive performance in all metrics even as it decomposes all the components: geometry, lights, and materials. In \cref{fig:comparison_with_split-sum_and_pre-integrated_light}, with the approximation, geometry is not properly reconstructed on higher intensity or deep concave region. Also, the roughness does not reflects geometric smoothness in part of the roof and same material property of the building wall.

\subsection{Analysis and ablation study}
\label{subsec:analysis_and_ablation_study}

\begin{figure}[tbp]
    \centering
    \captionsetup[subfigure]{font=scriptsize}
    
    \begin{subfigure}[h]{0.18\paperwidth}
        \includegraphics[width=\textwidth]{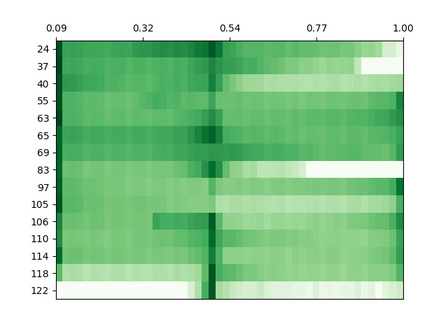}
        \caption{NDJIR}
    \end{subfigure}
    \begin{subfigure}[h]{0.18\paperwidth}
        \includegraphics[width=\textwidth]{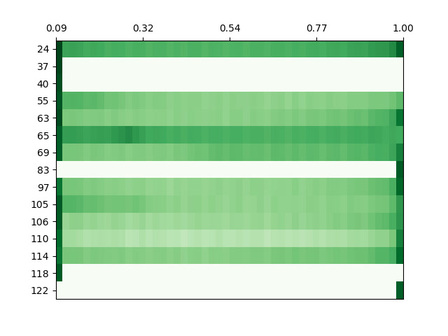}
        \caption{w/o priors ($128$)}
    \end{subfigure}

    \smallskip
    \begin{subfigure}[h]{0.18\paperwidth}
        \includegraphics[width=\textwidth]{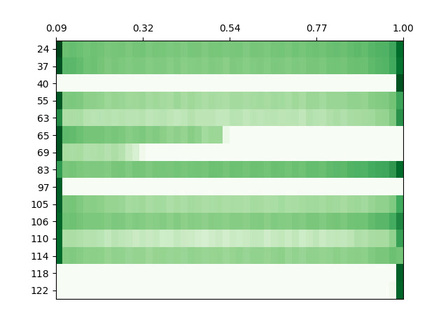}
        \caption{w/o priors ($32$)}
    \end{subfigure}
    \begin{subfigure}[h]{0.18\paperwidth}
        \includegraphics[width=\textwidth]{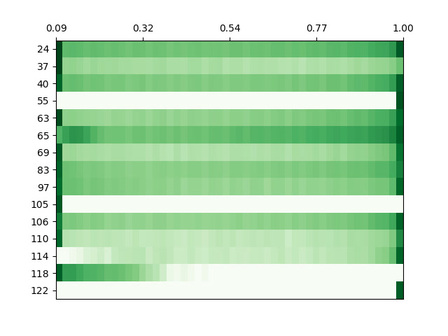}
        \caption{w/o priors ($8$)}
    \end{subfigure}

    \caption{\textbf{Roughness distributions} with varying number of light samples per pixel ($N$) over all scenes of DTU MVS dataset.}
    \label{fig:roughness_distributions}
\end{figure}
\begin{figure}[tbp]
    \centering
    \captionsetup[subfigure]{font=scriptsize}

    \begin{subfigure}[h]{0.18\paperwidth}
        \includegraphics[width=\textwidth]{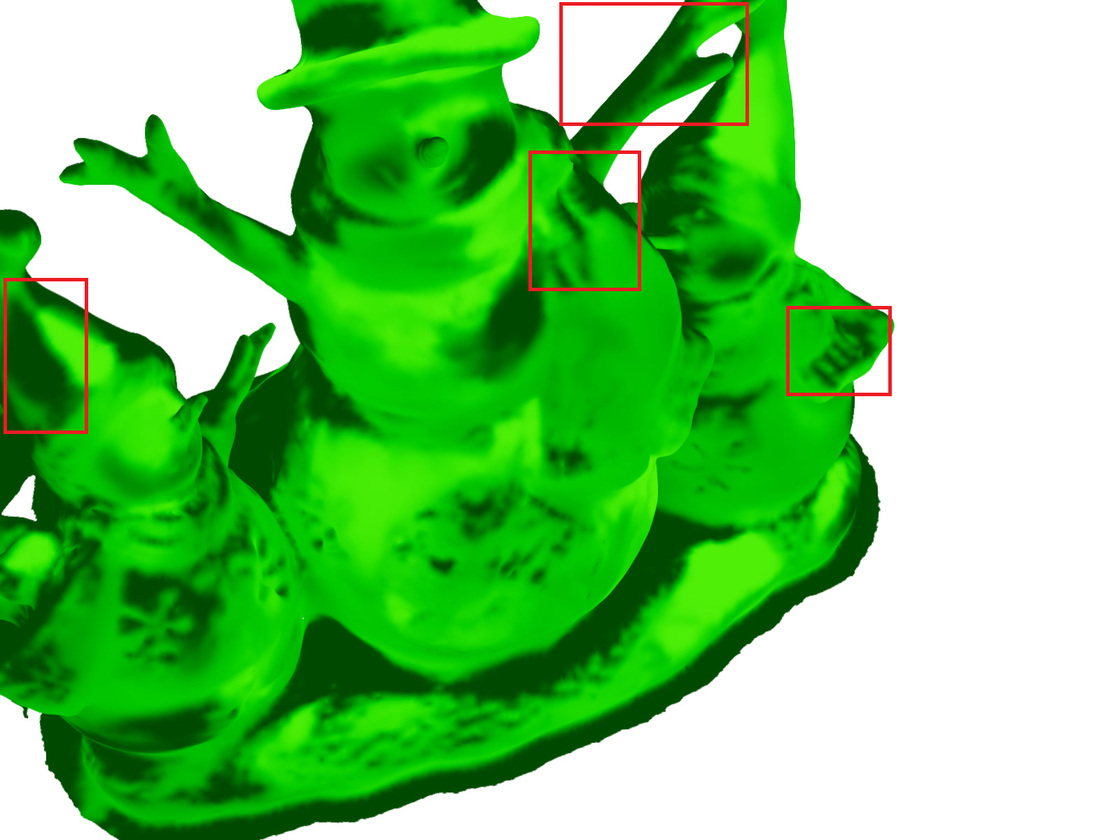}
        \caption{NDJIR}
    \end{subfigure}
    \hspace{1pt}
    \begin{subfigure}[h]{0.18\paperwidth}
        \includegraphics[width=\textwidth]{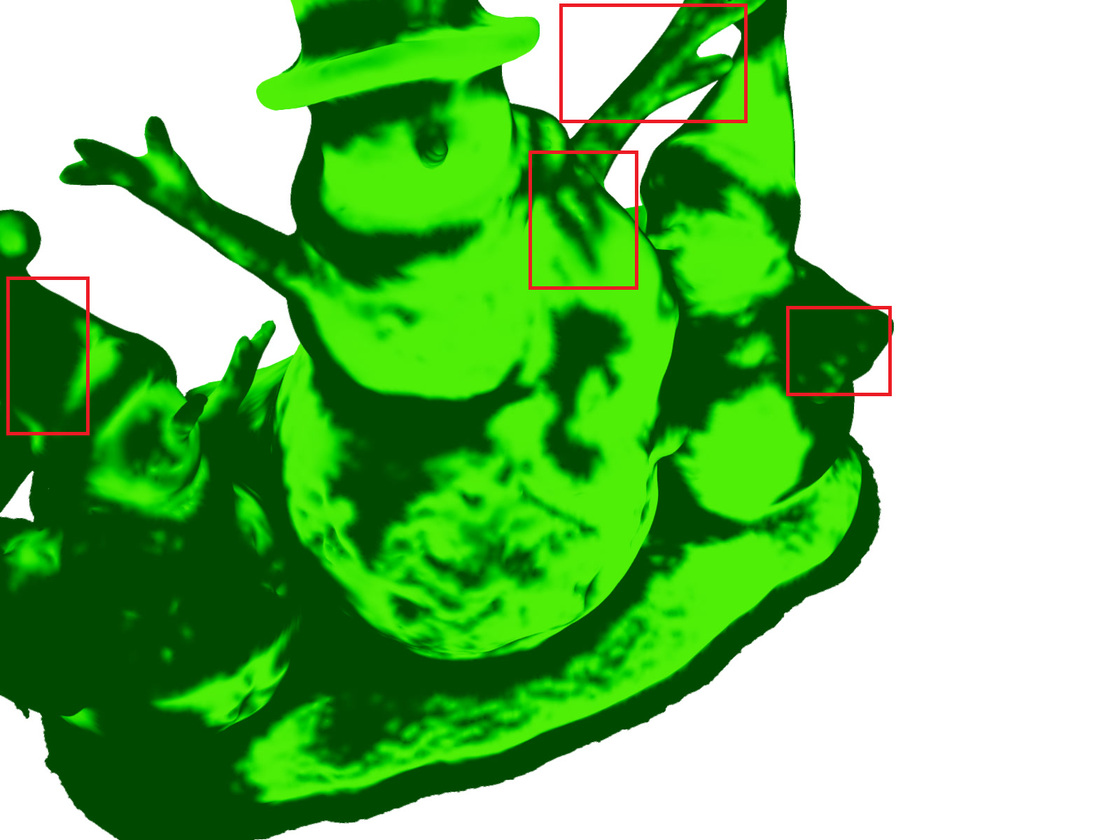}
        \caption{w/o priors}
    \end{subfigure}

    \smallskip
    \begin{subfigure}[h]{0.18\paperwidth}
        \includegraphics[width=\textwidth]{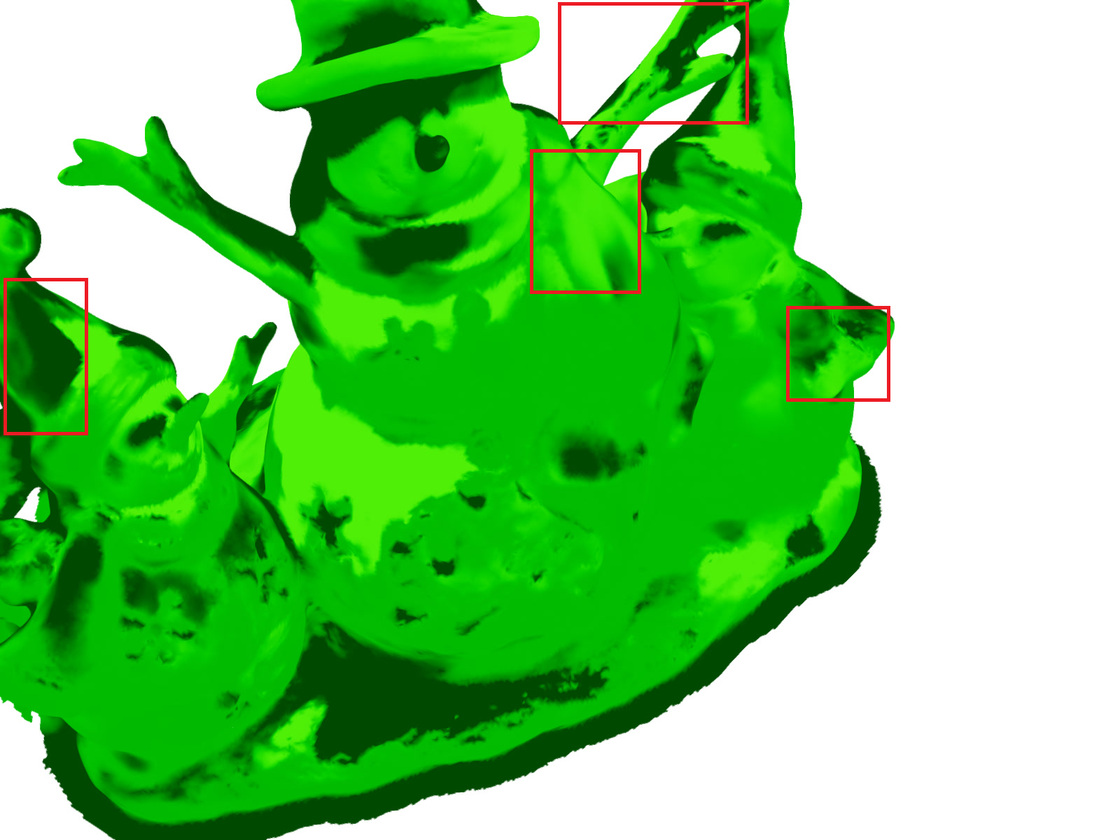}
        \caption{w/ triplane $\times$ triline}
    \end{subfigure}
    \hspace{1pt}
    \begin{subfigure}[h]{0.18\paperwidth}
        \includegraphics[width=\textwidth]{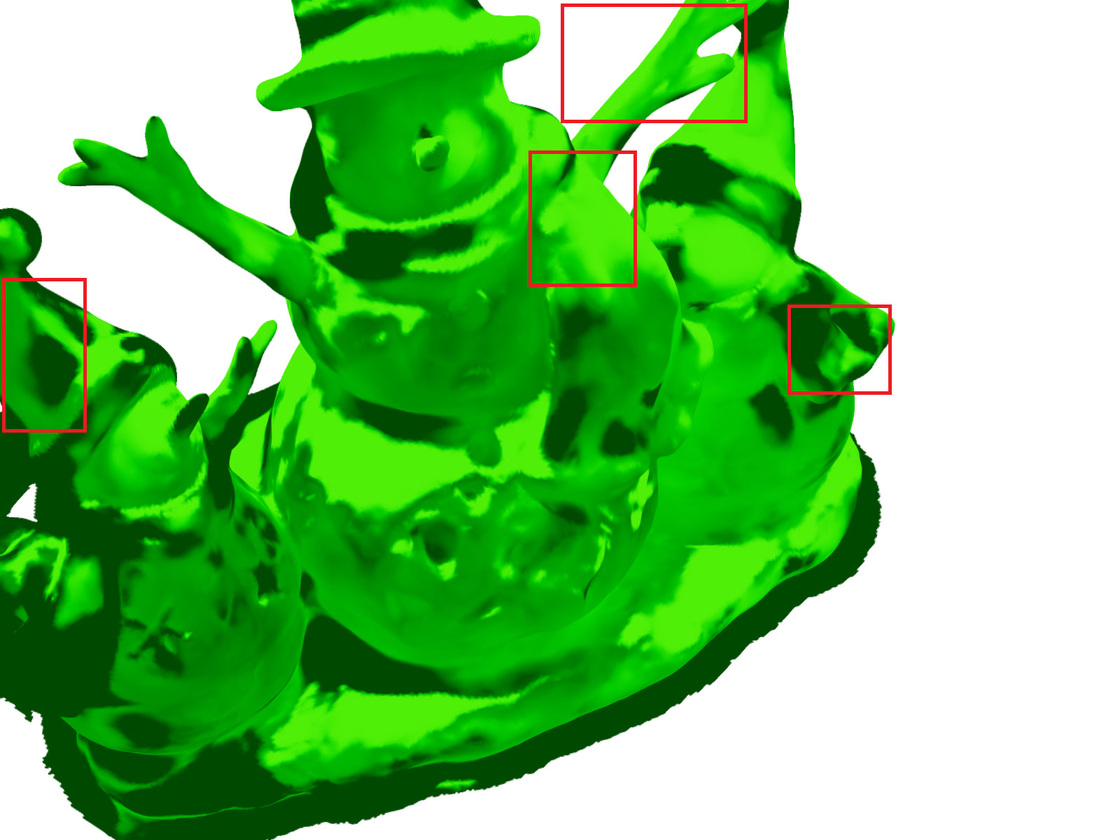}
        \caption{w/o voxel}
    \end{subfigure}

    \caption{\textbf{Comparison of spatially-varying roughness} with different configurations (scan69). Red boxes mean high specularity regions.}
    \label{fig:comparison_of_spatially-varying_roughness}
\end{figure}

\textbf{Distribution of roughness:} \cref{fig:roughness_distributions} illustrates how effective the increasing number of light samples per pixel (spp) and Bayesian prior are. When we use lower spps, roughness networks seem degenerated, outputting unary ($0$ or $1$) or binary ($0$ and $1$) value(s). However, once we increase spps, especially to $128$, it is mitigated for some scenes, but we still have degeneration. Using Bayesian prior more reduces degeneration, and the peak of the distribution is around $0.5$. \cref{fig:comparison_of_spatially-varying_roughness} shows spatially-varying roughnesses. Without Bayesian prior, the values are binarily distributed. Using triplane $\times$ triline acceleration structure or without voxel grid feature bears lower spatially-varying property and can not capture high specularity.

\begin{figure}[tbp]
    \centering
    \captionsetup[subfigure]{font=scriptsize}

    \begin{subfigure}[h]{0.12\paperwidth}
        \includegraphics[width=\textwidth]{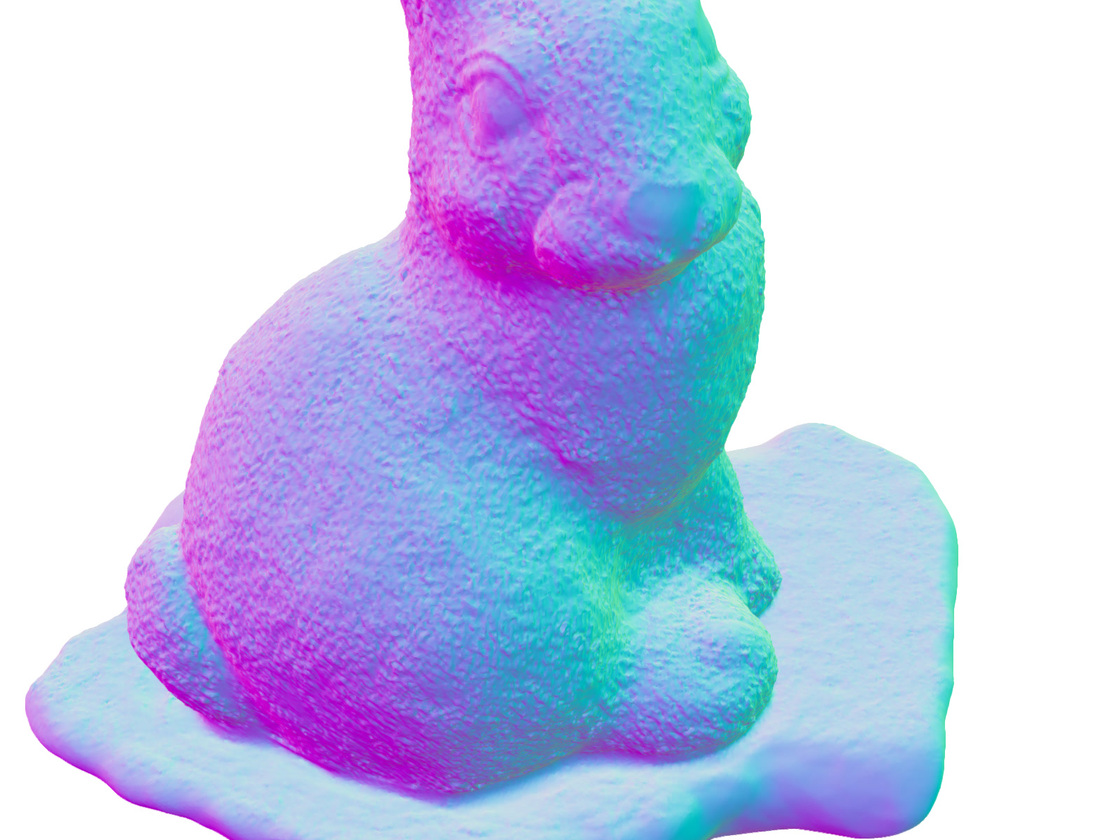}
        \caption{NDJIR ($0 \times$)}
    \end{subfigure}
    \begin{subfigure}[h]{0.12\paperwidth}
        \includegraphics[width=\textwidth]{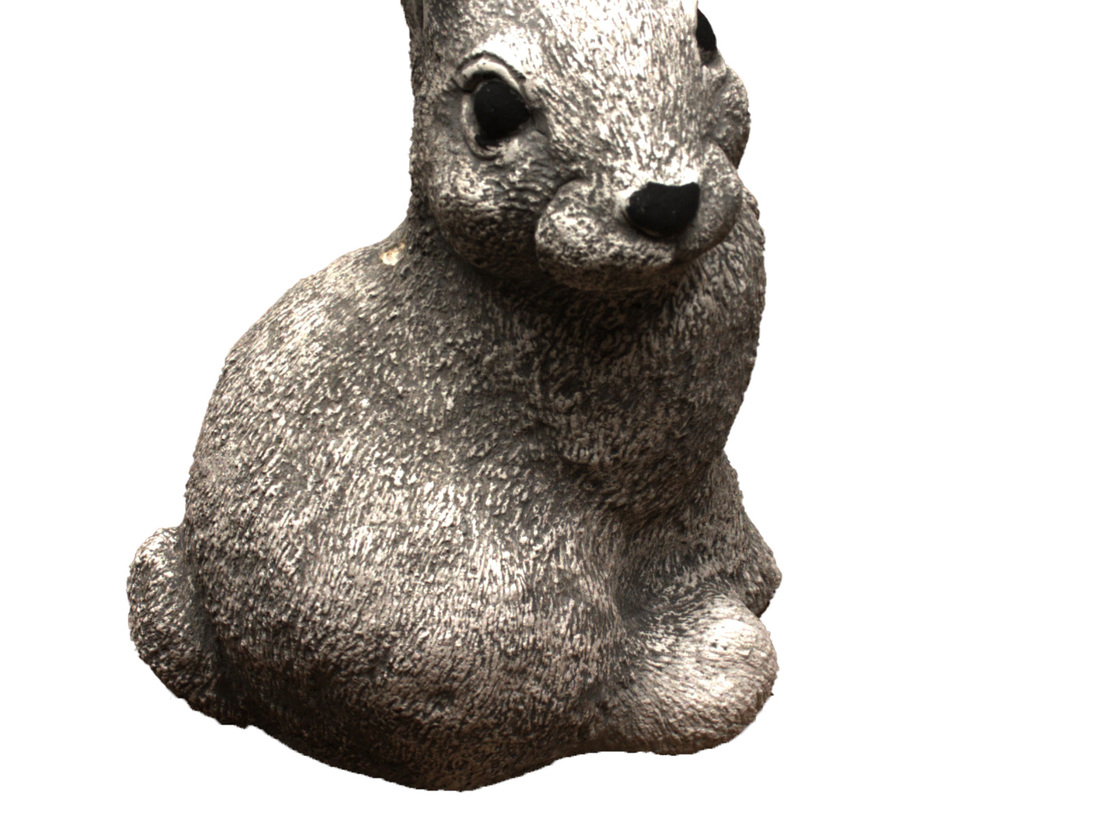}
        \caption{Groundtruth}
    \end{subfigure}
    \begin{subfigure}[h]{0.12\paperwidth}
        \includegraphics[width=\textwidth]{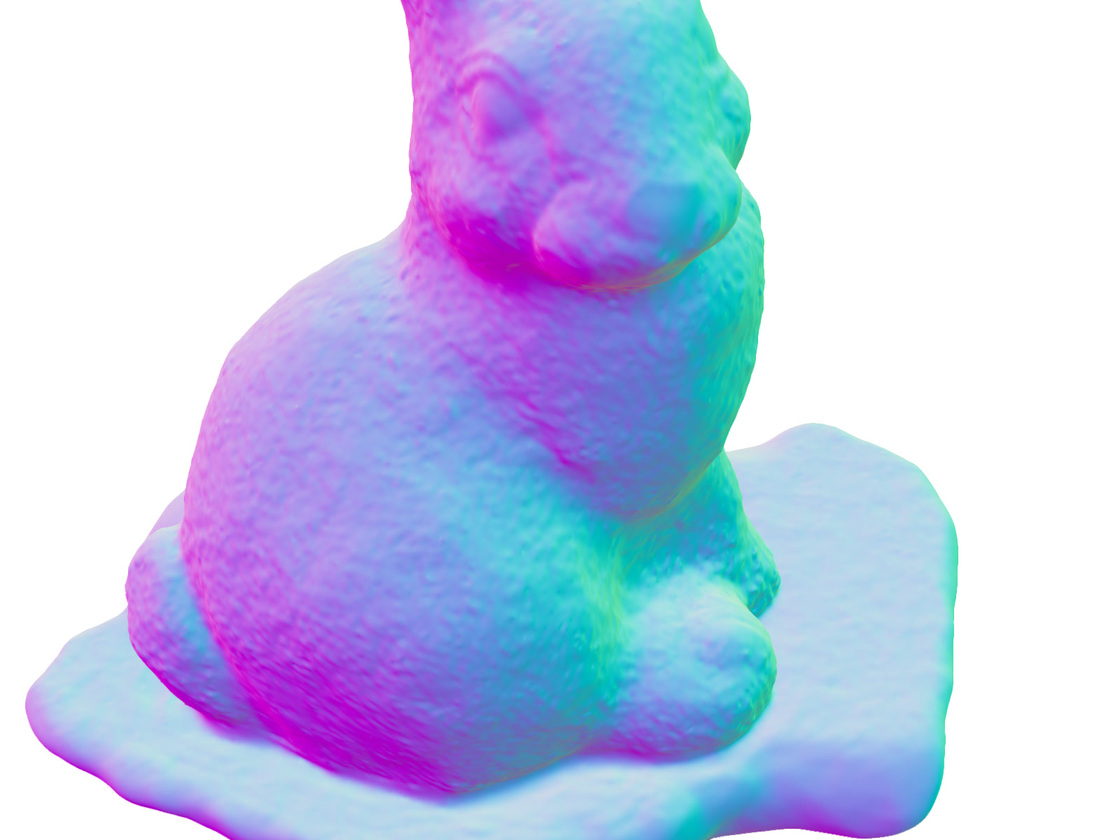}
        \caption{NDJIR ($2 \times$)}
    \end{subfigure}

    \smallskip
    \begin{subfigure}[h]{0.12\paperwidth}
        \includegraphics[width=\textwidth]{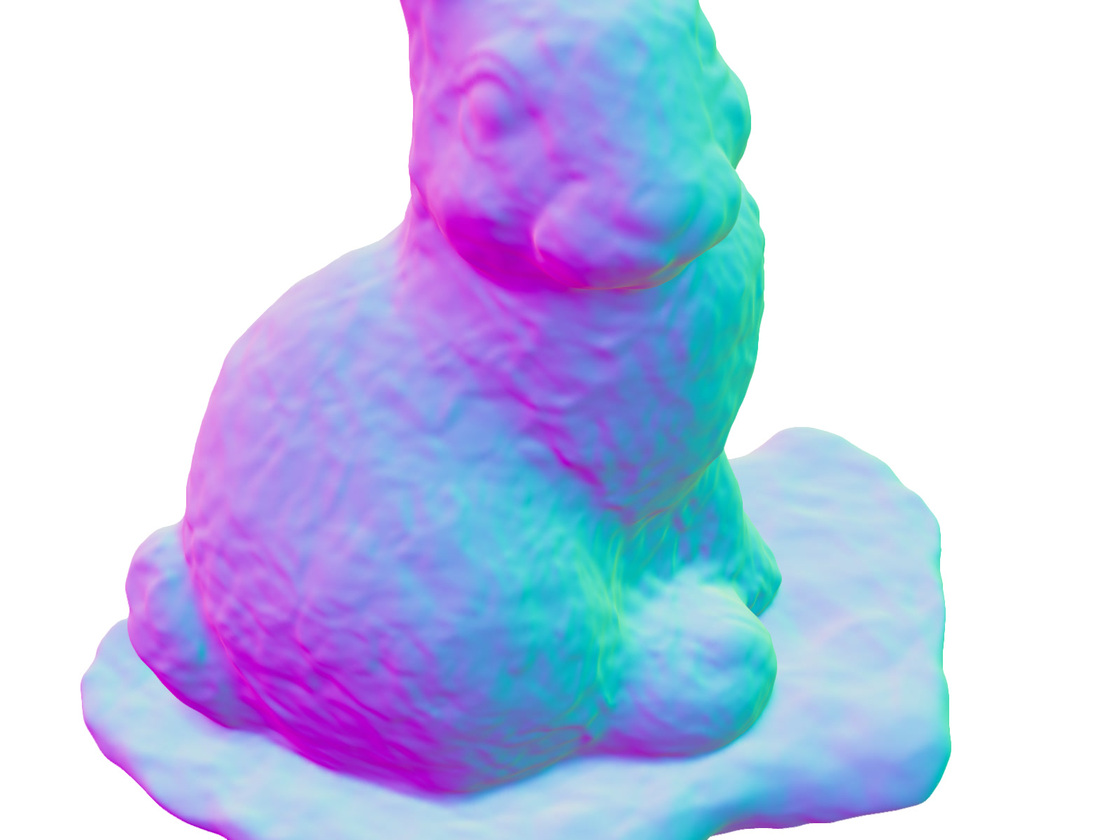}
        \caption{w/o voxel ($0 \times$)}
    \end{subfigure}
    \begin{subfigure}[h]{0.12\paperwidth}
        \includegraphics[width=\textwidth]{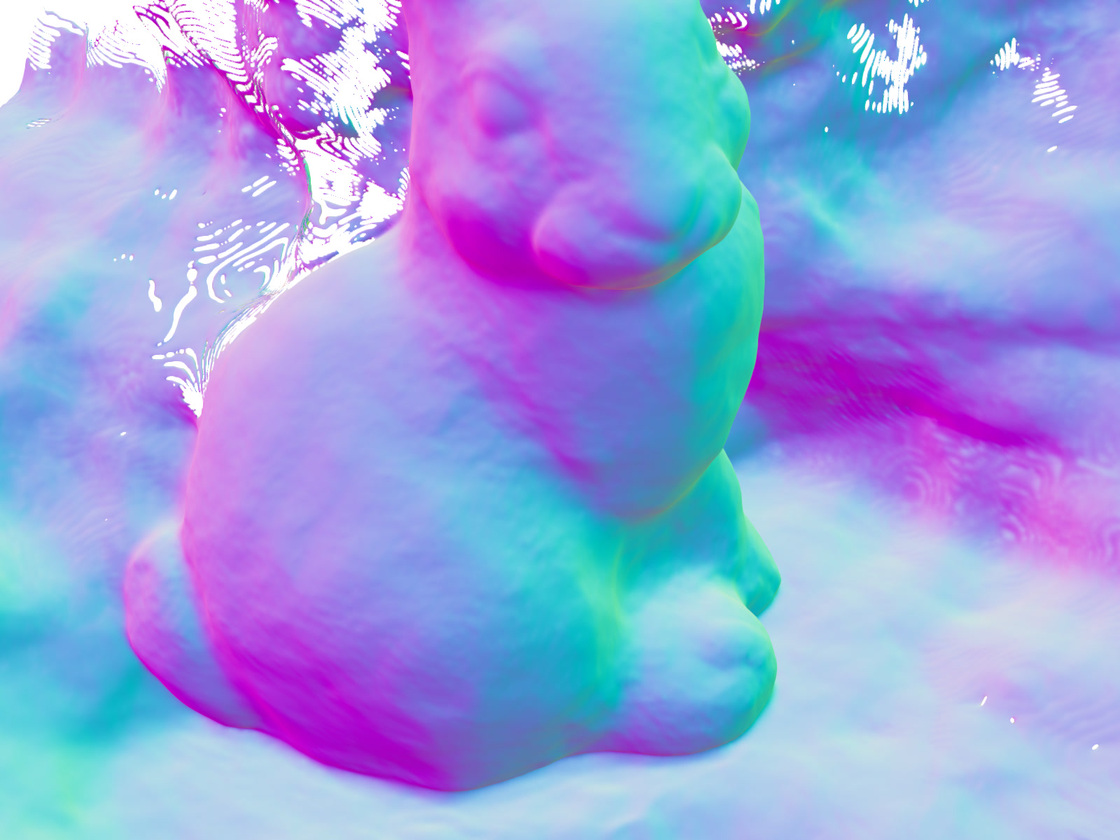}
        \caption{w/ triplane $\times$ triline ($0 \times$)}
    \end{subfigure}
    \begin{subfigure}[h]{0.12\paperwidth}
        \includegraphics[width=\textwidth]{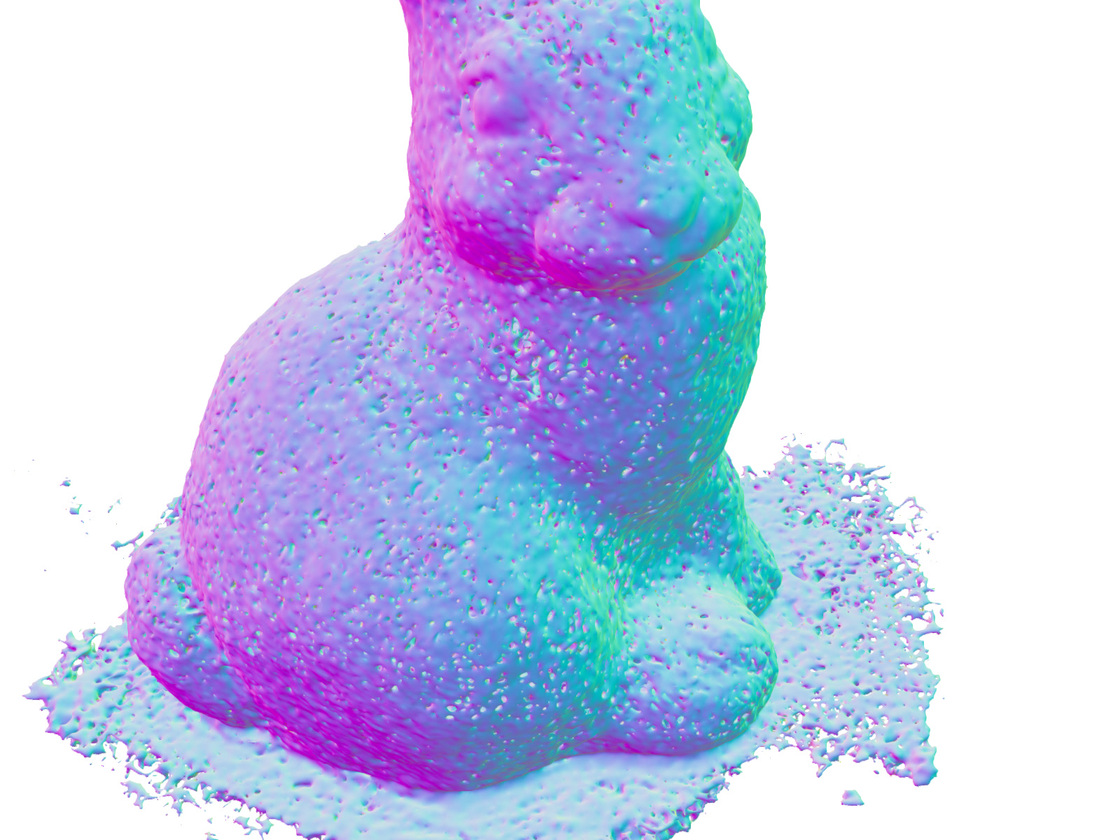}
        \caption{w/ STE ($2 \times$)}
    \end{subfigure}
    \caption{\textbf{Comparison of geometric details} with different acceleration structures and average filter ($N \times$) as post-processing (scan55).}
    \label{fig:comparison_of_geometric_details}
\end{figure}

\textbf{Geometric effect of acceleration structure:} In addition to material decomposition, the dense voxel grid feature influences geometry \cref{fig:comparison_of_geometric_details}. Voxel grid features captures geometric details but often produces some tiny geometric artifacts. However, it can be reduced and/or controlled by a few simple average filters as post-processing. Straight through estimator (STE) \cite{DBLP:conf/nips/CourbariauxBD15} is one way to compute gradients of interpolation, but STE produces lots of non-erasable jagged artifacts, meaning normals of the voxel grid feature are important. Without acceleration structure cannot capture geometric details with large batch and smaller iteration training, also with triplane $\times$ triline acceleration structure cannot separate foreground and background well, depending on scene.

\begin{figure}[tbp]
    \captionsetup[subfigure]{justification=centering,font=scriptsize}
    \centering    

    \begin{subfigure}[h]{0.12\paperwidth}
        \includegraphics[width=\textwidth]{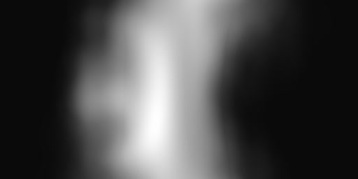}
        \caption{$PE(\bm{l}, 4)$ \\ $(0.69, 17.53)$}
    \end{subfigure}
    \hspace{1pt}
    \begin{subfigure}[h]{0.12\paperwidth}
        \includegraphics[width=\textwidth]{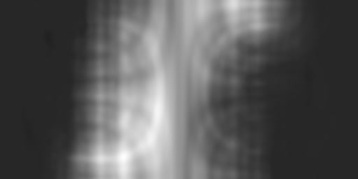}
        \caption{$PE(\bm{l}, 6)$ \\ $(0.54, 4.23)$}
    \end{subfigure}
    \hspace{1pt}
    \begin{subfigure}[h]{0.12\paperwidth}
        \includegraphics[width=\textwidth]{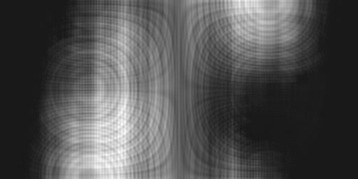}
        \caption{$PE(\bm{l}, 8)$ \\ $(0.74, 7.63)$}
    \end{subfigure}
    
    \caption{\textbf{Environment lights} with different number of frequencies in positional encoding (scan55). Values below $PE$ are the actual (min, max) values. For visibility, values are normalized such that the maximum is $255$.}
    \label{fig:environment_lights}
\end{figure}

\textbf{Decomposed environment light:} Environment light network represents directional light distribution as in \cref{fig:environment_lights}, it captures light intensity only in an upper hemisphere like the scene setting of DTU MVS dataset. When increasing frequencies of the positional encoding, the network more represents light interference of LEDs.

\begin{figure*}[tbp]
    \centering
    \captionsetup[subfigure]{font=scriptsize}
    \rotatebox[origin=b]{90}{Unlit}\quad
    \begin{subfigure}[h]{0.14\paperwidth}
        \caption{$c=0$}
        \includegraphics[width=\textwidth]{assets/raw_images/NDJIR/default_epoch1500__groups_gcb50379_dataset_DTU_scan69/model_01499_512grid_trimmed_base_color_mesh00_filtered02_defaultUnlit_default/32.png.jpg}
    \end{subfigure}
    \hspace{1pt}
    \begin{subfigure}[h]{0.14\paperwidth}
        \caption{$c=0.25$}
        \includegraphics[width=\textwidth]{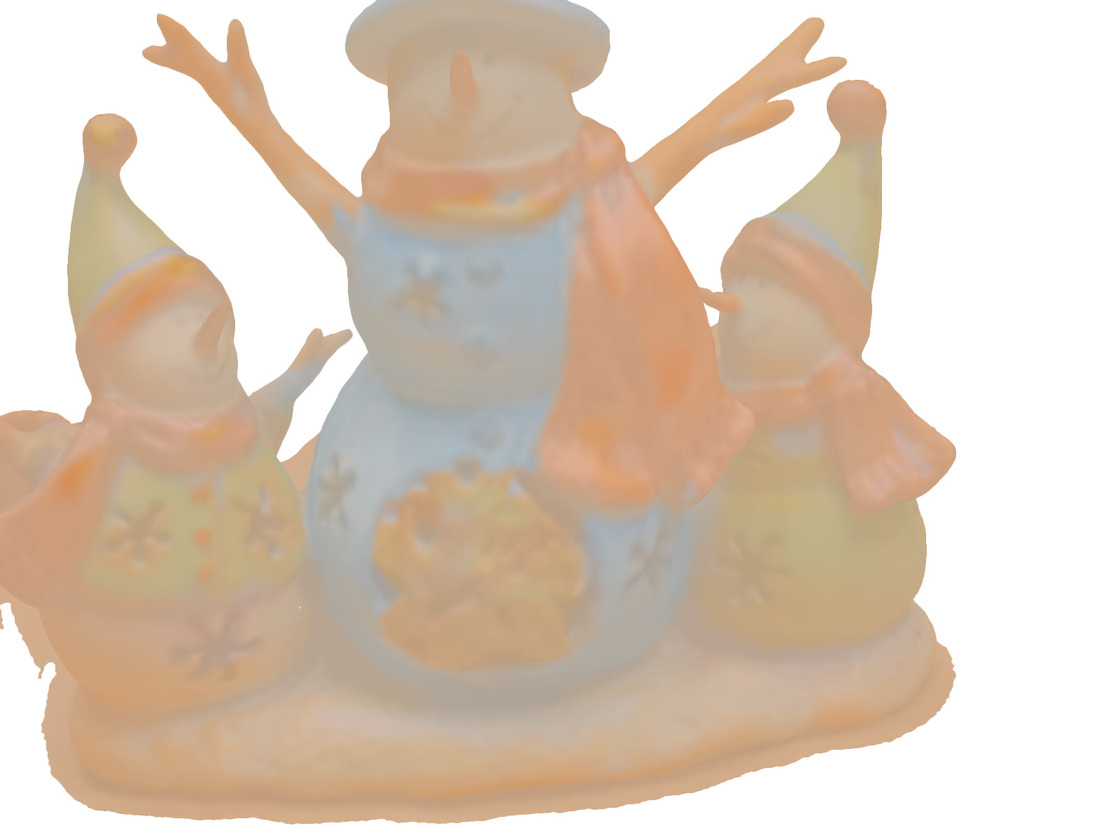}
    \end{subfigure}
    \hspace{1pt}
    \begin{subfigure}[h]{0.14\paperwidth}
        \caption{$c=0.5$}
        \includegraphics[width=\textwidth]{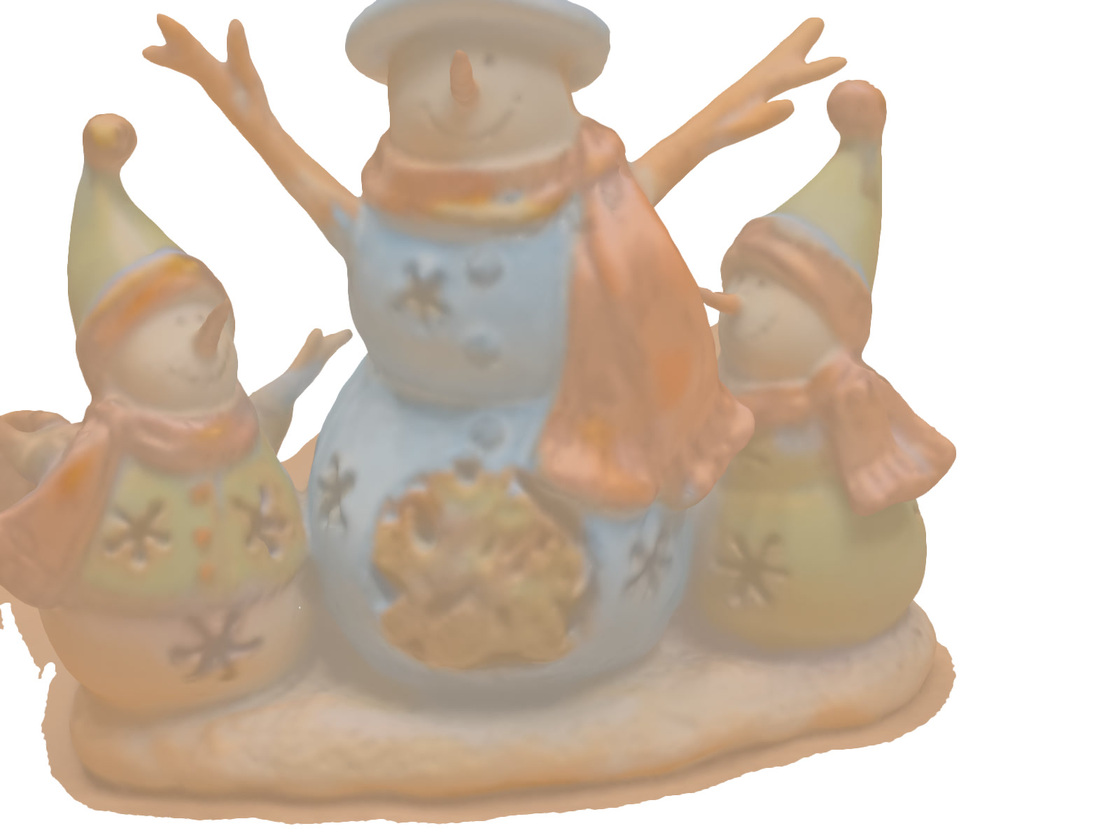}
    \end{subfigure}
    \hspace{1pt}
    \begin{subfigure}[h]{0.14\paperwidth}
        \caption{$c=0.75$}
        \includegraphics[width=\textwidth]{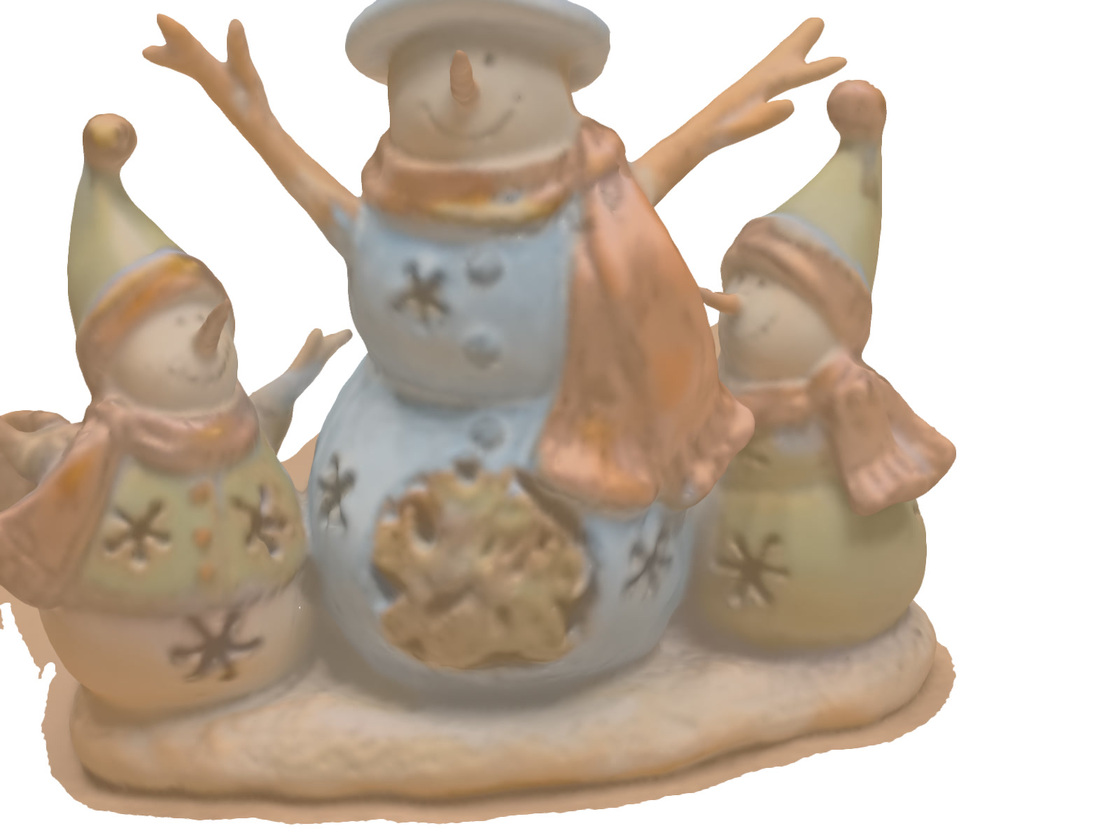}
    \end{subfigure}
    \hspace{1pt}
    \begin{subfigure}[h]{0.14\paperwidth}
        \caption{$c=1$}
        \includegraphics[width=\textwidth]{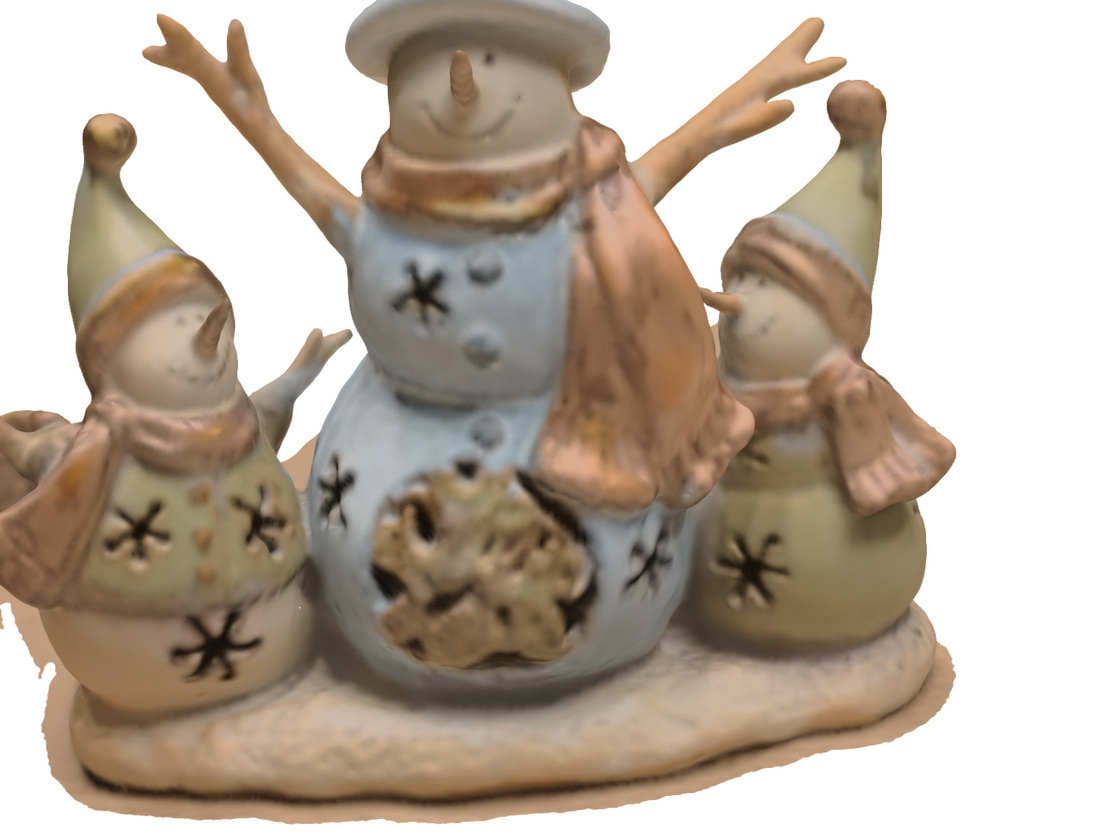}
    \end{subfigure}

    \smallskip
    \rotatebox[origin=b]{90}{PBR-lit}\quad
    \begin{subfigure}[h]{0.14\paperwidth}
        \includegraphics[width=\textwidth]{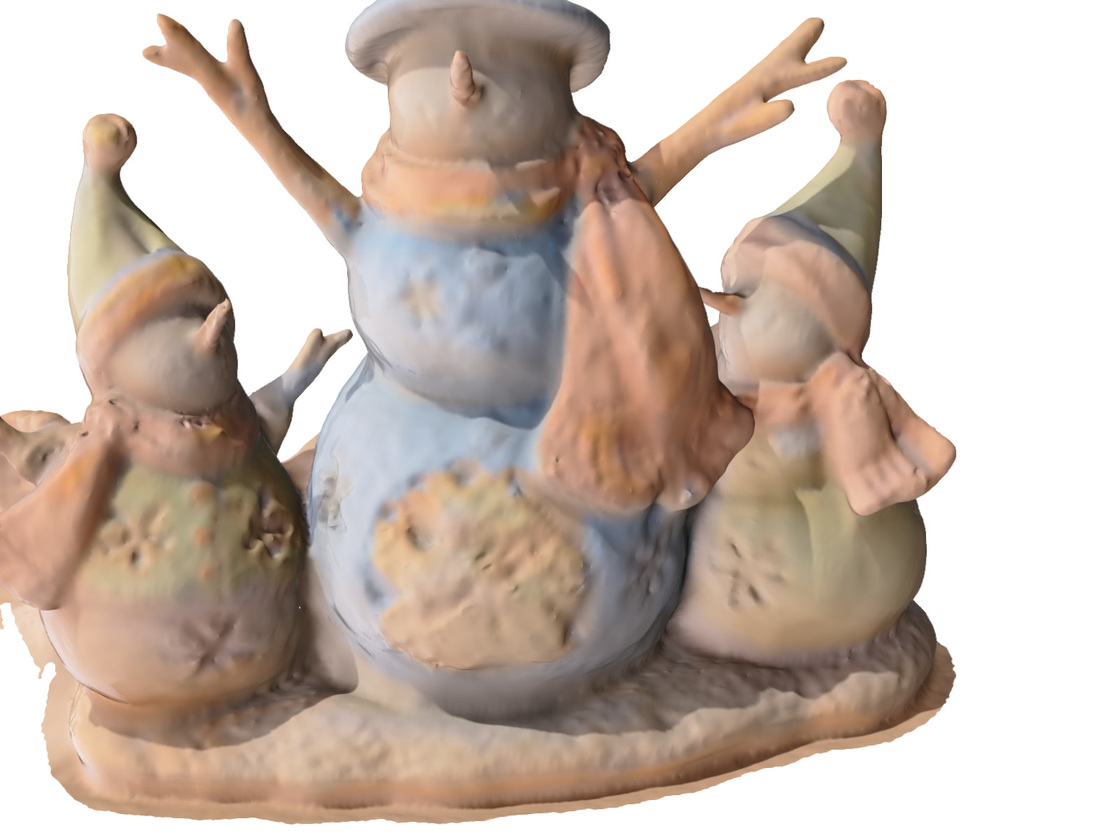}
    \end{subfigure}
    \hspace{1pt}
    \begin{subfigure}[h]{0.14\paperwidth}
        \includegraphics[width=\textwidth]{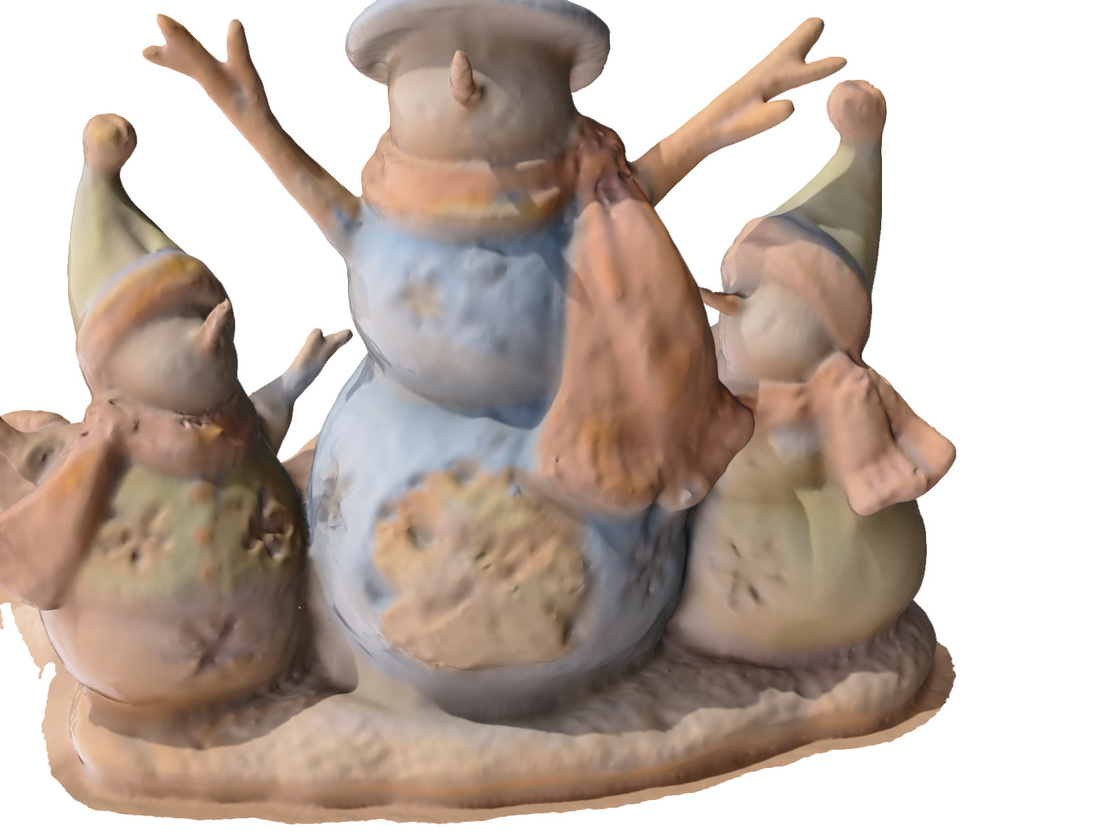}
    \end{subfigure}
    \hspace{1pt}
    \begin{subfigure}[h]{0.14\paperwidth}
        \includegraphics[width=\textwidth]{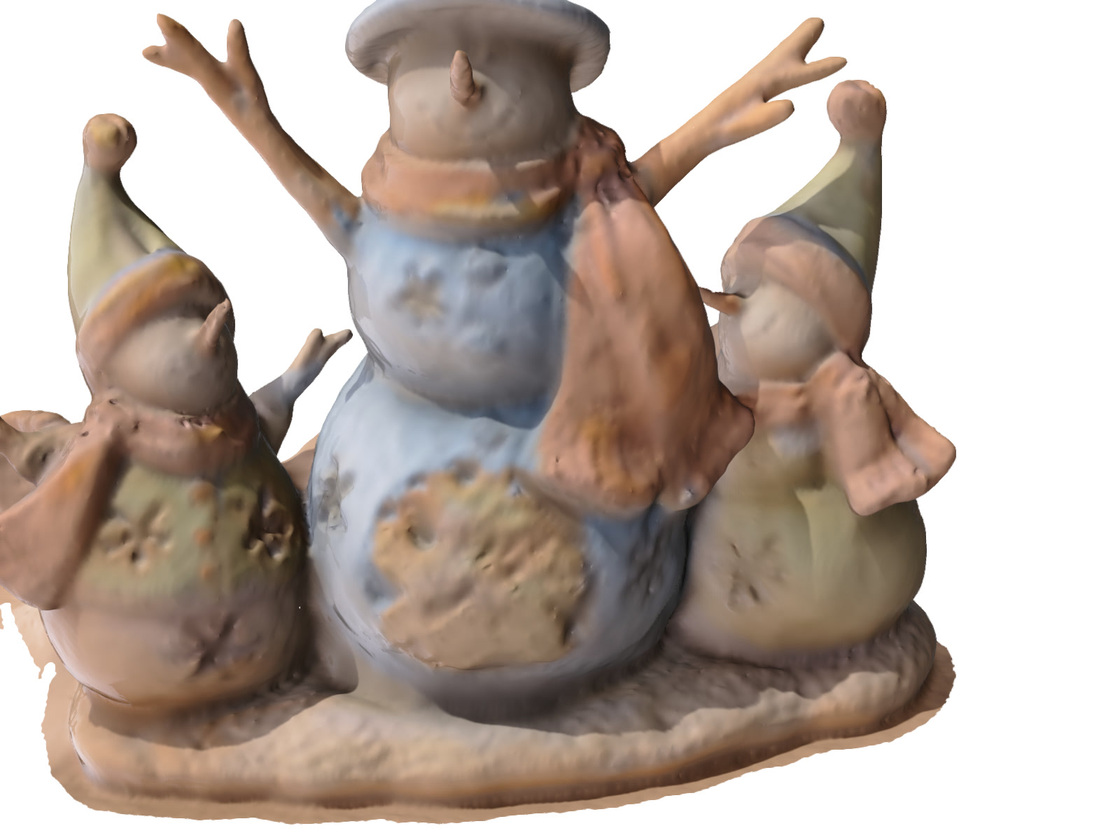}
    \end{subfigure}
    \hspace{1pt}
    \begin{subfigure}[h]{0.14\paperwidth}
        \includegraphics[width=\textwidth]{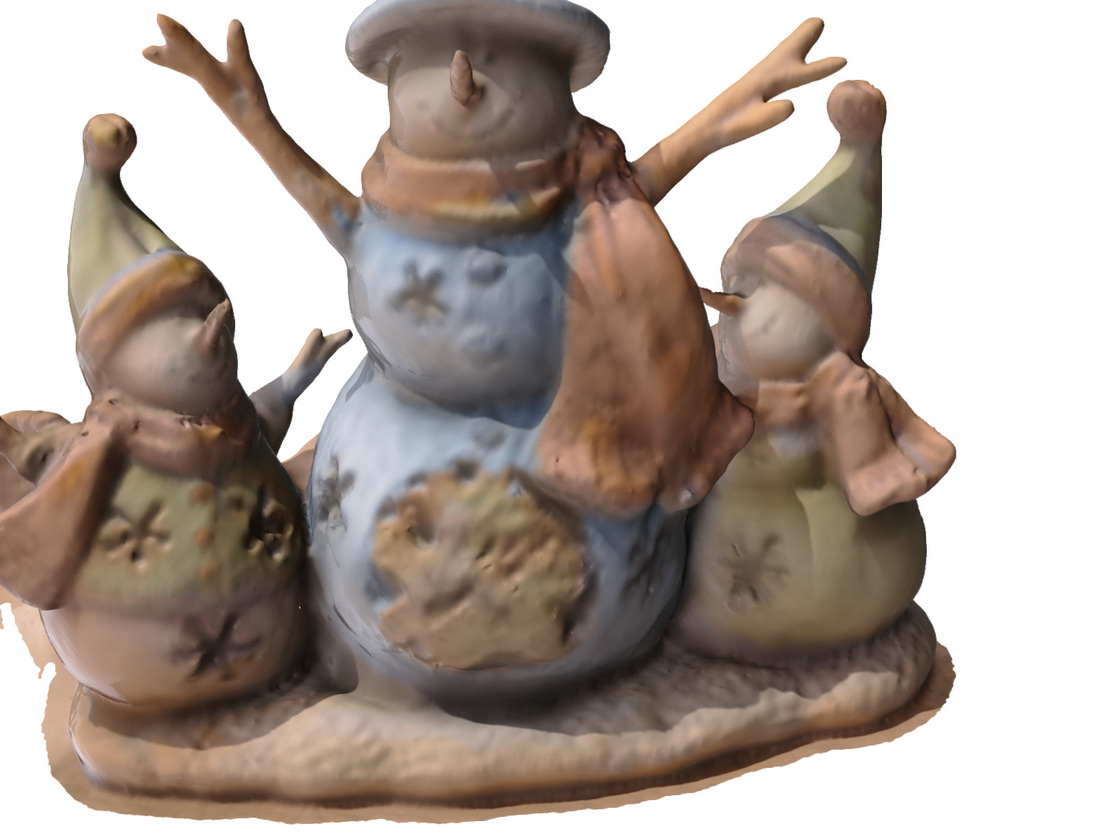}
    \end{subfigure}
    \hspace{1pt}
    \begin{subfigure}[h]{0.14\paperwidth}
        \includegraphics[width=\textwidth]{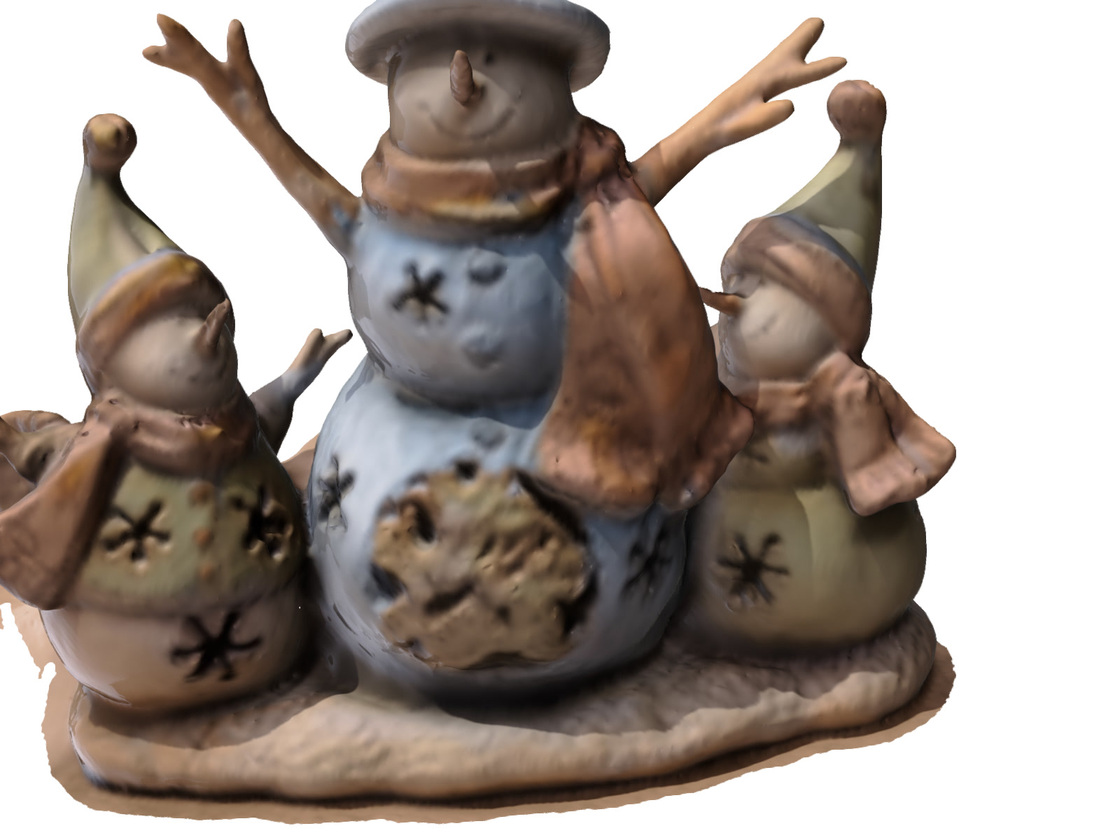}
    \end{subfigure}

    \caption{\textbf{Rebaking implicit illumination to base color} (scan69) with    interpolation coefficient $c$.
    }
    \label{fig:rebaking_implicit_illumination_to_base_color}
\end{figure*}

\begin{figure}[tbp]
    \centering
    \captionsetup[subfigure]{font=scriptsize}
    \begin{subfigure}[h]{0.18\paperwidth}
        \caption{importance}
        \includegraphics[width=\textwidth]{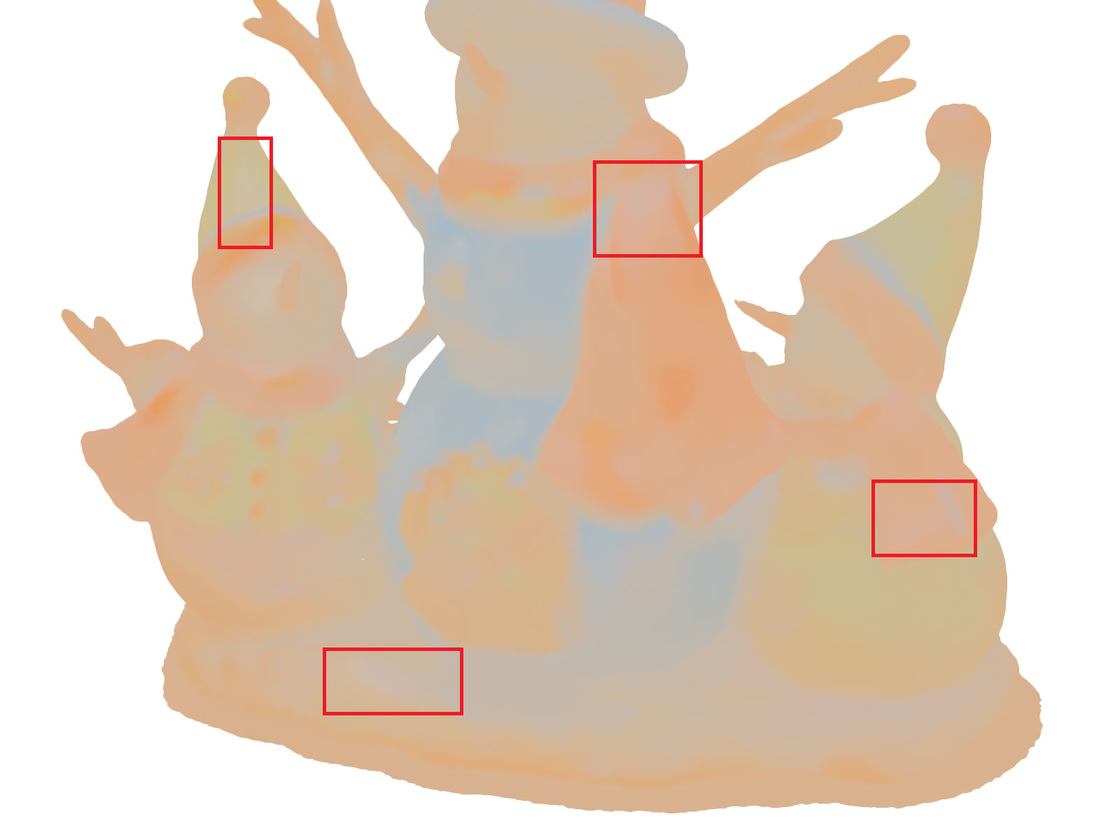}
    \end{subfigure}
    \hspace{1pt}
    \begin{subfigure}[h]{0.18\paperwidth}
        \caption{uniform}
        \includegraphics[width=\textwidth]{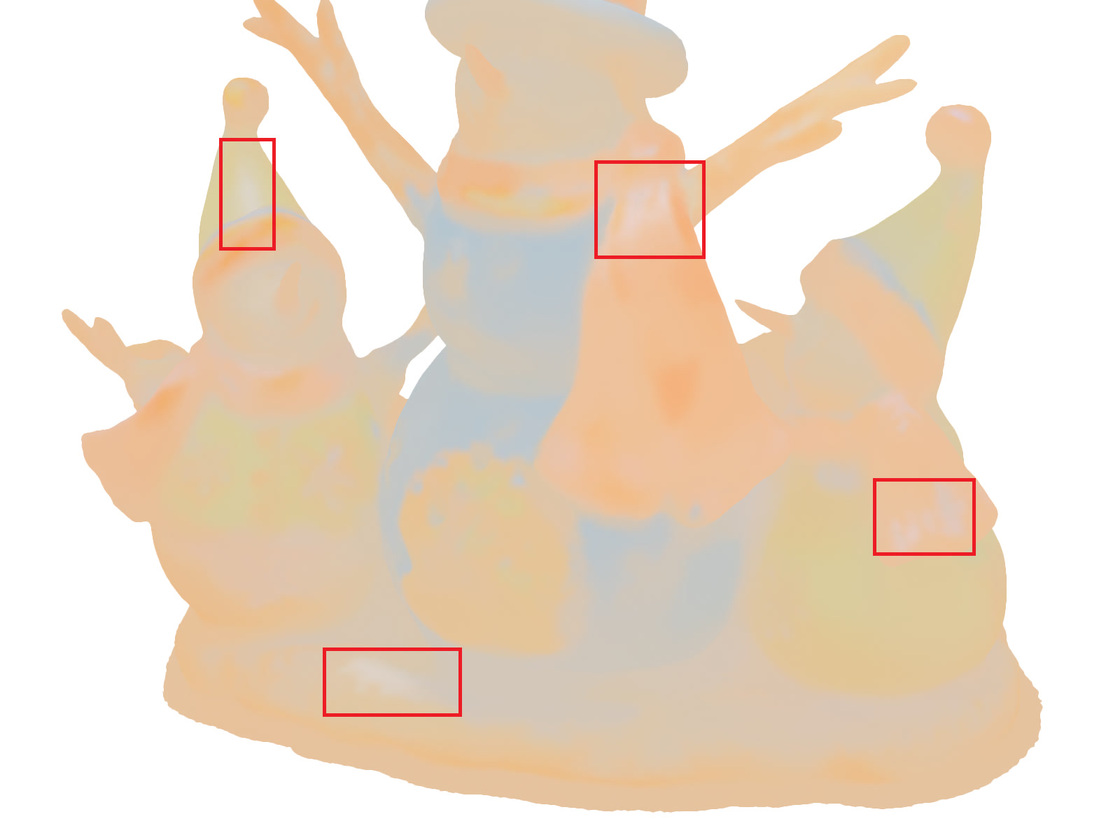}
    \end{subfigure}

    \smallskip
    \begin{subfigure}[h]{0.18\paperwidth}
        \includegraphics[width=\textwidth]{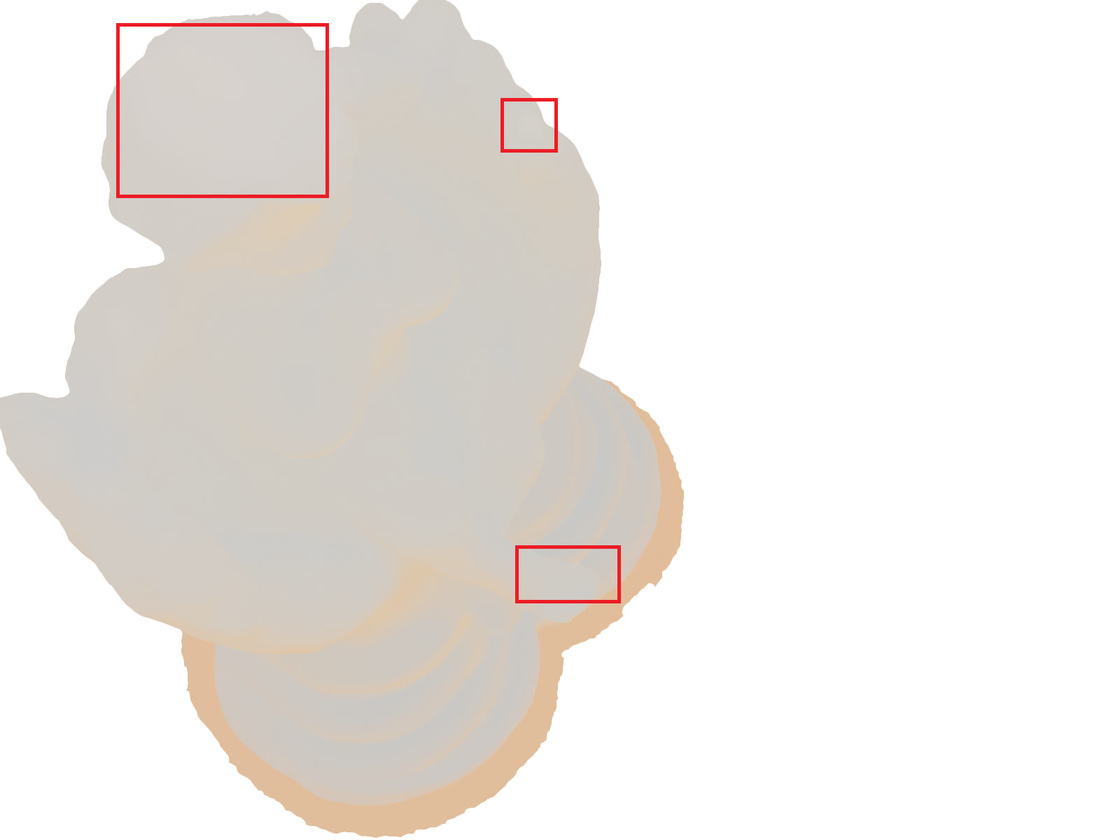}
    \end{subfigure}
    \hspace{1pt}
    \begin{subfigure}[h]{0.18\paperwidth}
        \includegraphics[width=\textwidth]{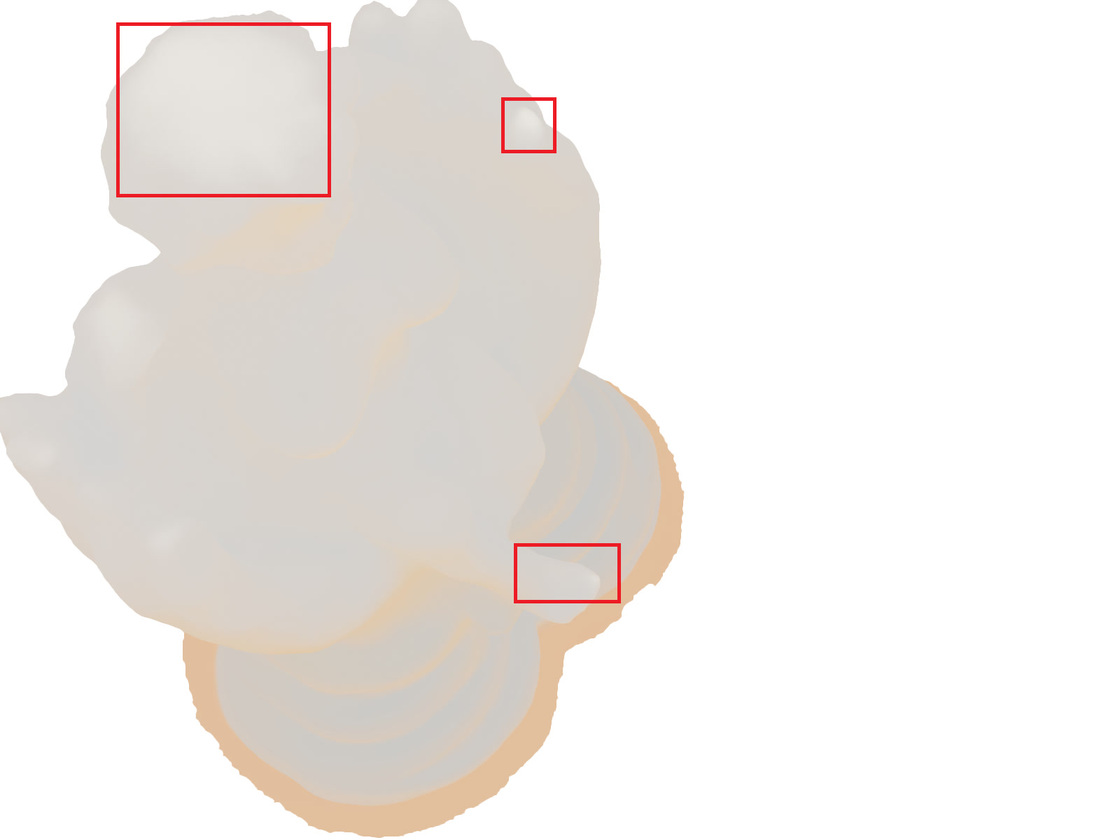}
    \end{subfigure}

    \caption{\textbf{Comparison of base color} with different sampling methods in Monte Carlo integration. Red boxes indicate high light intensity or specularity regions.}
    \label{fig:different_samplings}
\end{figure}

\textbf{Sampling method:} Importance sampling of the specular term \cref{eq:ndjir_specular_term} is a key to model specularity precisely. As in \cref{fig:different_samplings}, using uniform sampling still bakes high light intensity and specularity on base color; importance sampling mitigates strong light baking to more extent.

\subsection{Material conversion}
\label{subsec:material_conversion}

\begin{figure}[tbp]
    \captionsetup[subfigure]{justification=centering,font=scriptsize}
    \centering

    \rotatebox[origin=b]{90}{scan37}\quad
    \begin{subfigure}[h]{0.11\paperwidth}
        \caption{PBR w/ metallic}
        \includegraphics[width=\textwidth]{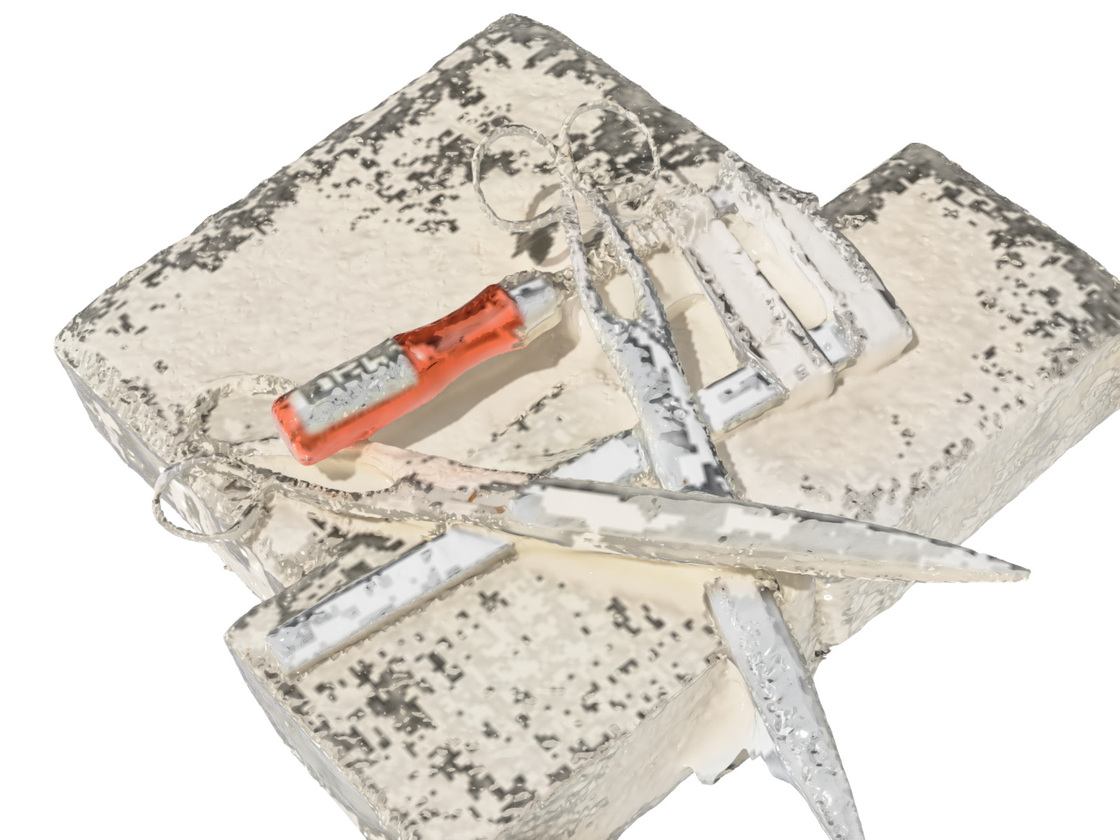}
    \end{subfigure}
    \hspace{1pt}
    \begin{subfigure}[h]{0.11\paperwidth}
        \caption{Groundtruth}
        \includegraphics[width=\textwidth]{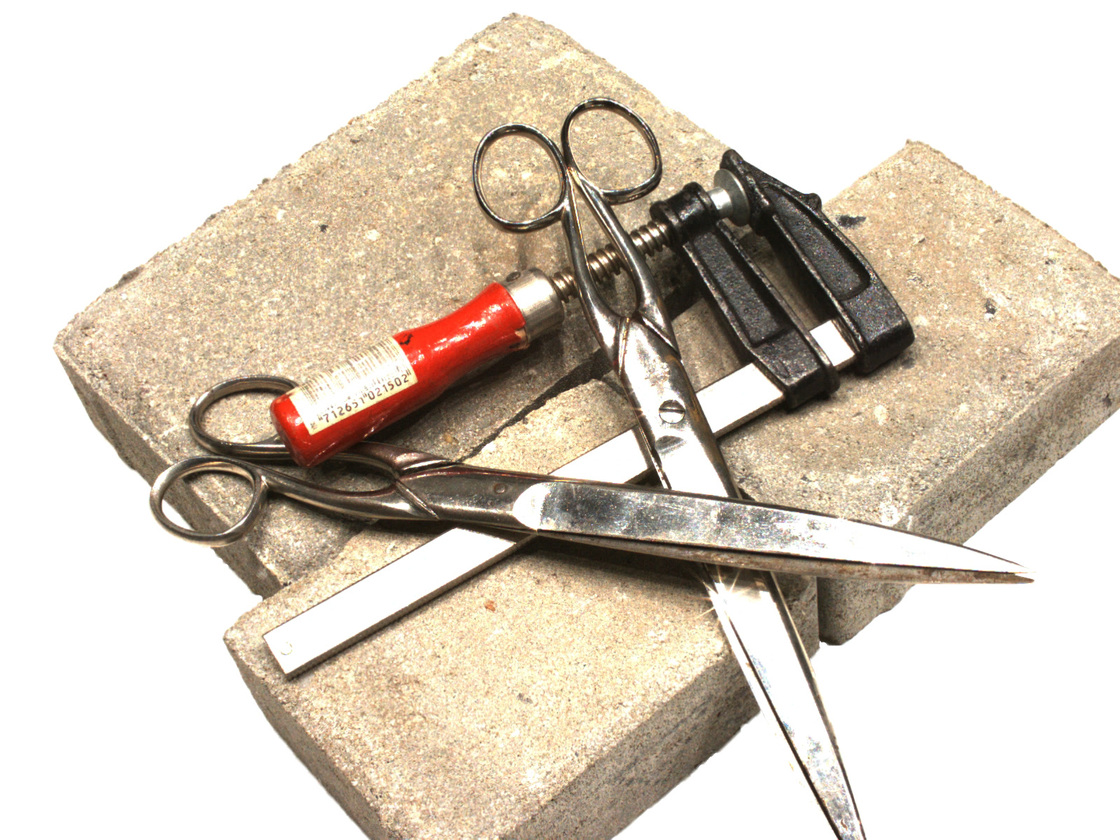}
    \end{subfigure}
    \hspace{1pt}
    \begin{subfigure}[h]{0.11\paperwidth}
        \caption{PBR}
        \includegraphics[width=\textwidth]{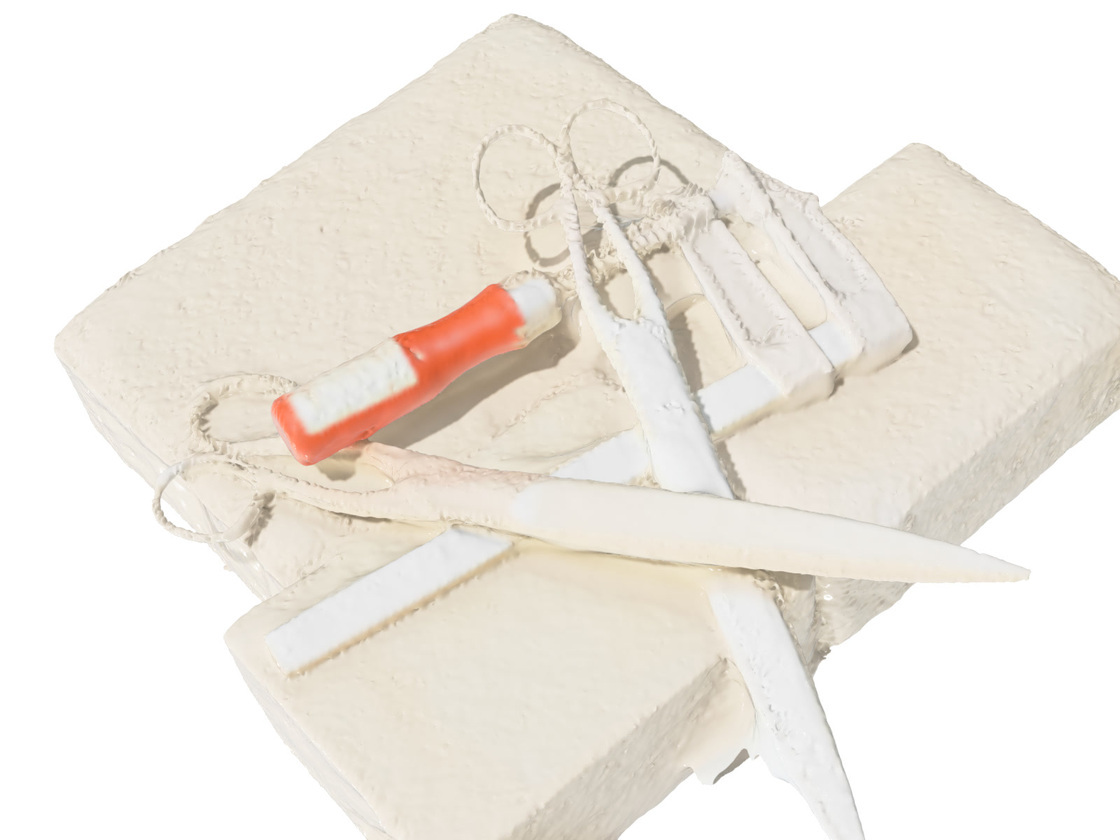}
    \end{subfigure}

    \smallskip
    \rotatebox[origin=b]{90}{scan97}\quad
    \begin{subfigure}[h]{0.11\paperwidth}
        \includegraphics[width=\textwidth]{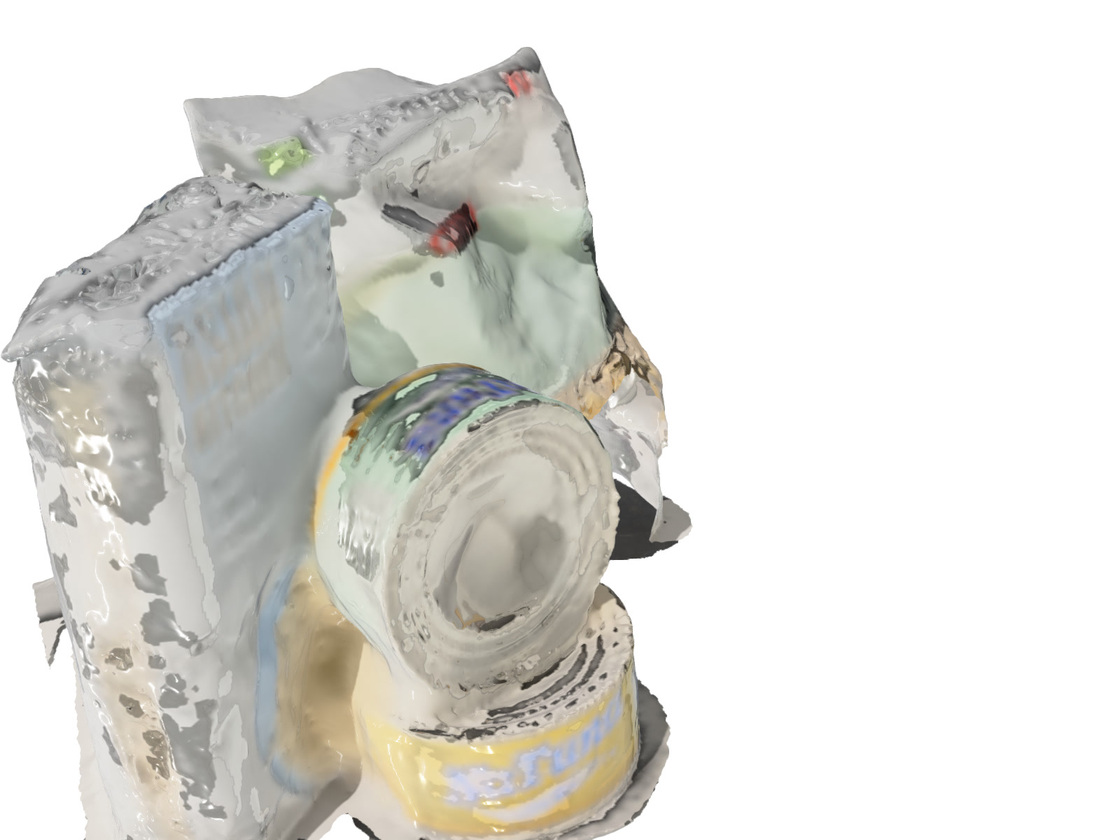}
    \end{subfigure}
    \hspace{1pt}
    \begin{subfigure}[h]{0.11\paperwidth}
        \includegraphics[width=\textwidth]{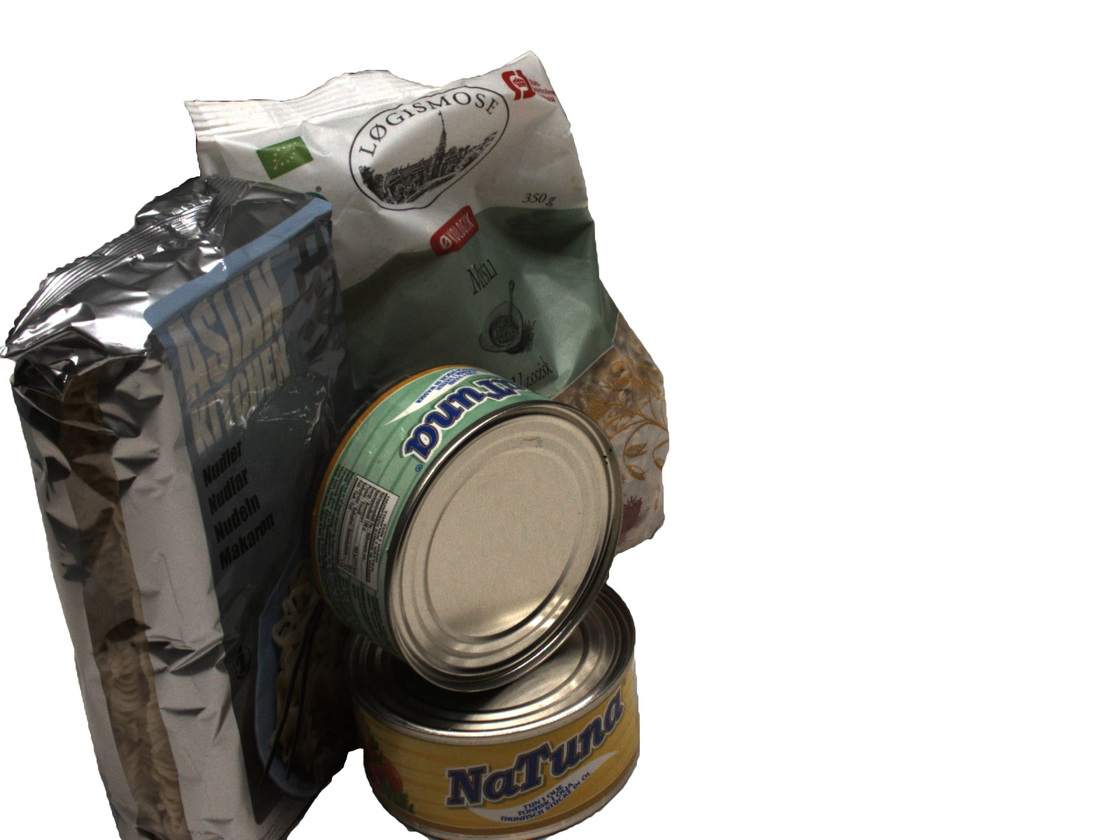}
    \end{subfigure}
    \hspace{1pt}
    \begin{subfigure}[h]{0.11\paperwidth}
        \includegraphics[width=\textwidth]{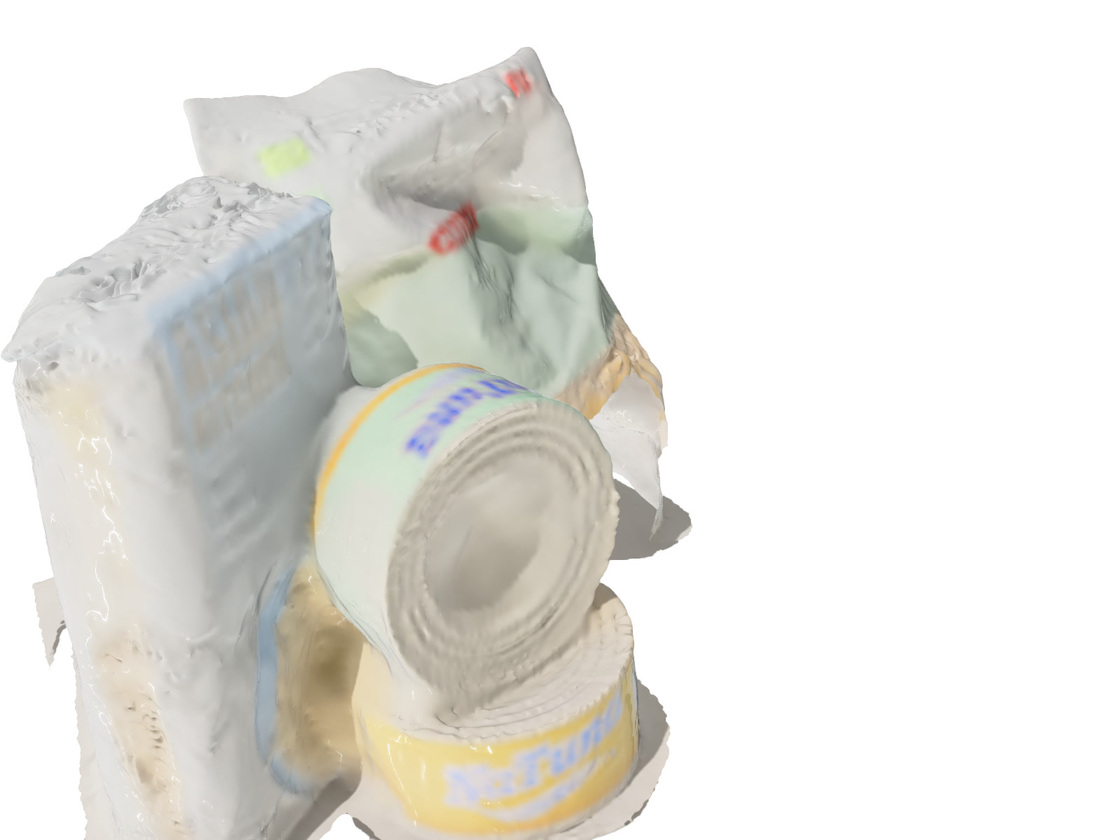}
    \end{subfigure}

    \caption{\textbf{Metallic conversion} of metallic object.}
    \label{fig:metallic_conversion}
\end{figure}

\textbf{Metallic conversion:} Metalness is not modeled in our method, but once specular reflectance is obtained, we can simulate metalness according to the strength of specular reflectance to some extent. \cref{fig:metallic_conversion} are the results of reflectance-aware metallic conversion. Metal-like objects exhibit more metallic than non-metal PBR.

\textbf{Rebaking of implicit illumination:} We sometimes want to bake light distribution of a captured scene to base color, e.g., 3D printing. \cref{fig:rebaking_implicit_illumination_to_base_color} shows the interpolation results between base color and base color $\times$ implicit illumination. As seen, when increasing the interpolation coefficient, we can more re-bake light distribution to base color as post-processing.

\section{Discussion and conclusion}
\label{sec:discussion_and_conclusion}

\textbf{Discussion:} Here, we mainly discuss limitations of our method. NDJIR sometimes underestimates or overestimates e.g., in \cref{fig:main_results}, the left cheek and chest of scan69 and the left jaw of scan65. Increasing the number of light samples and introducing Bayesian prior prevents roughness and specular networks from degenerating, but we still have extreme values as in \cref{fig:roughness_distributions}. We believe in that more reasonable priors can mitigate degeneration and extract more plausible results. Dense voxel grid feature produces tiny noisy artifacts and/or non-smooth geometry even if we apply average filters as in \cref{fig:metallic_conversion}, e.g., scissors . Importance sampling of \cref{eq:ndjir_specular_term}, modeling implicit illumination as in \cref{fig:rebaking_implicit_illumination_to_base_color}, and strong regularization of \cref{eq:base_color_prior} can make base color flatter in \cref{fig:different_samplings} at the cost of removal of small colors, e.g., dot eyes and thin mouth of scan69. We do not model metalness, so metallic conversion in \cref{fig:metallic_conversion} is object-agnostic.

\textbf{Conclusion:} We propose NDJIR, neural direct and joint inverse rendering for geometry, lights, and materials of real object tailored to photogrammetry setting. Our proposed method is direct and joint optimization. NDJIR achieves better performances with the prior work and semantically well decomposition of real object.

\clearpage


\appendix

\setcounter{equation}{0}

\makeatletter
\renewcommand \thesection{S\@arabic\c@section}
\renewcommand \thetable{S\@arabic\c@table}
\renewcommand \thefigure{S\@arabic\c@figure}
\renewcommand \theequation{S\@arabic\c@equation}
\makeatother

\begin{table}[h]
    \centering
    \caption{\textbf{Notation table}.}
    \footnotesize
    \label{tab:}
    \begin{tabular}{l|l}
        Notation                                                                   & Description                            \\
        \hline \hline
        $\bm{x} \in \mathbb{R}^{3}$                                                & query point                            \\
        $s \in \mathbb{R}$                                                         & signed distance                        \\
        $\alpha \in \mathbb{R}$                                                    & opacity                                \\
        $T \in \mathbb{R}$                                                         & accumulated transmittance              \\
        $\bm{F}_{G} \in \mathbb{R}^{D_{G}}$                                        & geometric feature                      \\
        $\bm{n} \in \mathbb{R}^{3}$                                                & normals                                \\
        $\bm{l} \in \mathbb{R}^{3}$                                                & light direction                        \\
        $\bm{v} \in \mathbb{R}^{3}$                                                & viewing direction                      \\
        $d^{-2} \in \mathbb{R}$                                                    & inverse squared distance               \\
        $\bm{c}_{b} \in \mathbb{R}^{3}$                                            & base color                             \\
        $\alpha_{r} \in \mathbb{R}$                                                & roughness                              \\
        $\bm{f}_{0} \in \mathbb{R}^{3}$                                            & specular reflectance                   \\
        $L_{E} \in \mathbb{R}$                                                     & environment light intensity            \\
        $L_{V} \in \mathbb{R}$                                                     & soft visibility                        \\
        $L_{I} \in \mathbb{R}$                                                     & implicit illumination                  \\
        $L_{P} \in \mathbb{R}$                                                     & photogrammetric light intensity        \\
        $\hat{\bm{C}} \in \mathbb{R}^{3}$                                          & rendered color                         \\
        $D \in \mathbb{R}$                                                         & normal distribution function           \\
        $G \in \mathbb{R}$                                                         & masking-shadowing function             \\
        $F \in \mathbb{R}^{3}$                                                     & Fresnel function                       \\
        $\bm{h} \in \mathbb{R}^{3}$                                                & half vector                            \\
        $I \in \mathbb{R}$                                                         & indicator function                     \\
        $\Omega_{+}$                                                               & upper-hemisphere domain                \\
        $(\bm{a} \cdot \bm{b}) \in \mathbb{R}$                                     & saturated dot-product                  \\
        $\epsilon_{dot} \in \mathbb{R}$                                            & small threshold for dot-product        \\
        $\epsilon_{I} \in \mathbb{R}$                                              & small threshold for indicator function \\
        $\hat{\bm{a}}$                                                             & volume rendering, or on surface        \\
        $\bm{c} \in \mathbb{R}^{3}$                                                & camera location                        \\
        $t_{near} \in \mathbb{R}$                                                  & start distance of point sampling       \\
        $t_{far} \in \mathbb{R}$                                                   & end distance of point sampling         \\
        $\bm{G} \in \mathbb{R}^{3}$                                                & grid sizes                             \\
        $\bm{F}_{V} \in \mathbb{R}^{G_{x} \times G_{y} \times G_{z} \times D_{V}}$ & voxel grid feature                     \\
        $w \in \mathbb{R}$                                                         & interpolation coefficient              \\
        $\bm{f}_{i,j,j} \in \mathbb{R}^{D_{V}}$                                    & $i, j, k$-th voxel grid feature        \\
        $\bar{\bm{f}} \in \mathbb{R}^{D_{V}}$                                      & interpolated voxel grid feature        \\
        $\bm{m}$                                                                   & mins of AABB                           \\
        $\bm{M}$                                                                   & maxes of AABB                          \\
        $\bm{s}$                                                                   & scales of AABB                         \\
    \end{tabular}
\end{table}

\section{Network diagram}
\label{sec:network_diagram}

\begin{figure*}[t]
    \centering
    \includegraphics[width=\textwidth]{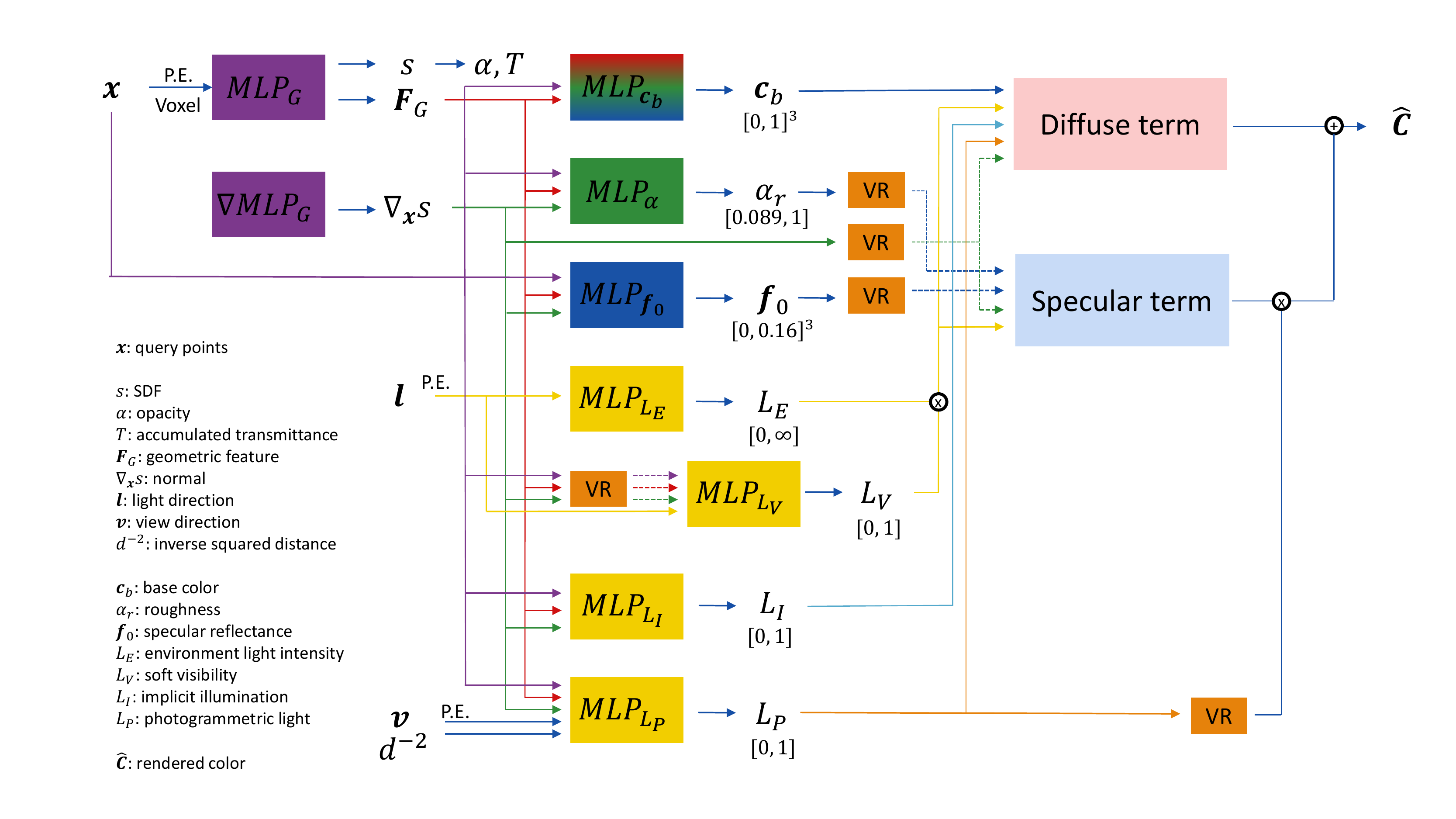}
    \caption{\textbf{Network diagram.} P.E. means the positional encoding. Voxel indicates the look-up result of the dense voxel grid feature. VR is the abbreviation of the volume rendering, and the dotted lines mean the result of VR. Ranges below some notations are the output ranges. $\oplus$ and $\otimes$ is element-wise addition and multiplication, respectively. Note $\alpha$ and $T$ go to all VRs, but for visibility, such lines are omitted. Also, $\bm{l}$ is the result of the uniform or importance samplings so depends on $\alpha_{r}$ and $\hat{\bm{n}}$, but its line is omitted for visibility.}
    \label{fig:network_diagram}
\end{figure*}

\cref{fig:network_diagram} shows the network diagram for visual understanding of inputs and outputs.

\section{Scene layout and point sampling}
\label{sec:scene_layout_and_point_sampling}

\begin{figure}[tbp]
    \captionsetup[subfigure]{justification=centering,font=scriptsize}
    \centering
    \includegraphics[scale=0.45]{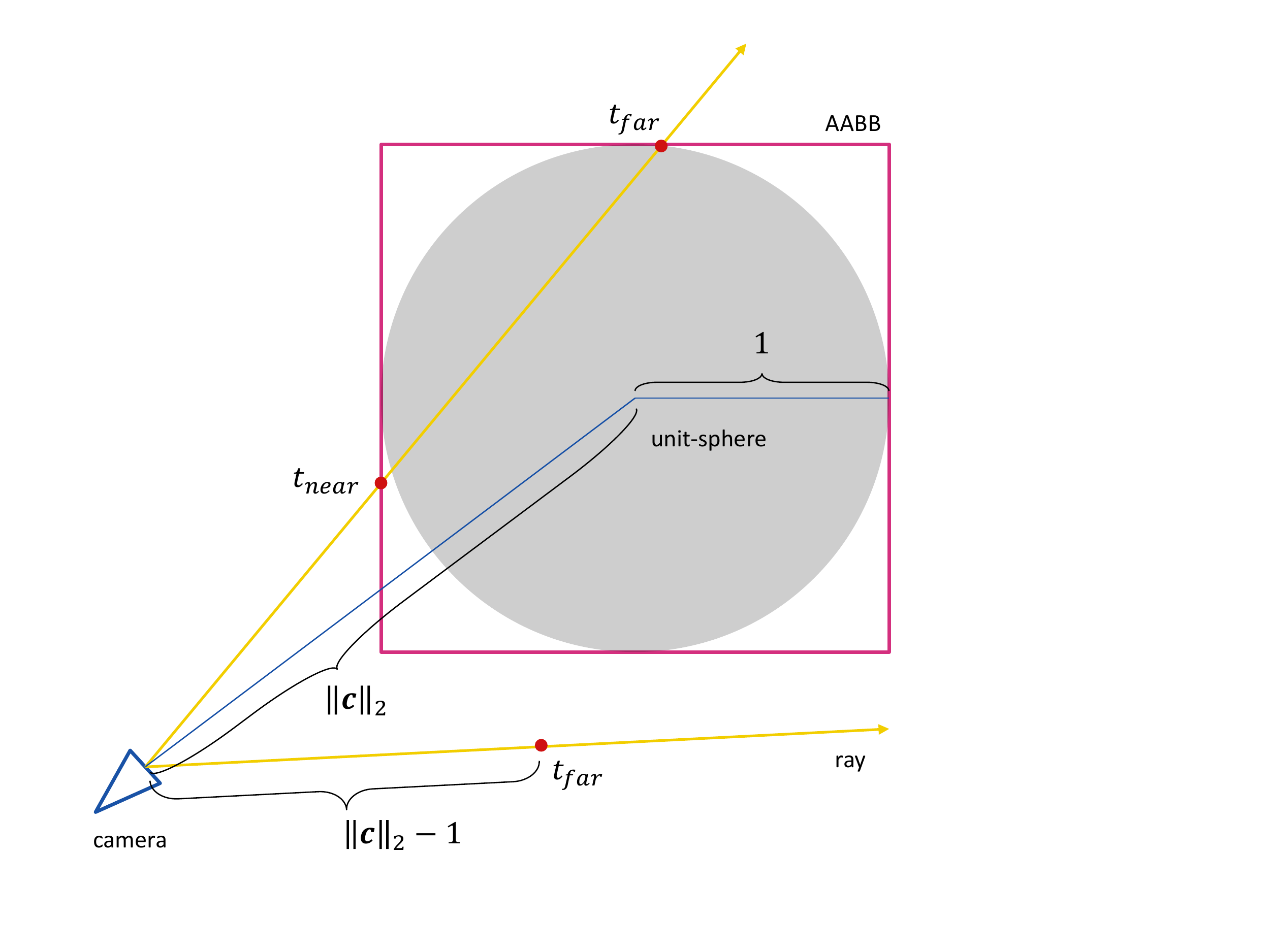}
    \caption{\textbf{Scene layout.} For visibility, we draw the corresponding 2D case.}
    \label{fig:scene_layout}
\end{figure}

\cref{fig:scene_layout} shows the scene layout. With the preprocessing \cite{DBLP:conf/nips/YarivKMGABL20}, we assume that an object of interest is located on the origin and inside the unit-sphere. If a ray hits the axis-aligned bounding box (AABB) with $\min=(-1, -1, -1)$ and $\max=(1, 1, 1)$, we can denote the intersection points as $(t_{near}, t_{far})$ and samples points between them. From $t_{far}$, we also sample points for background modeling as in NeRF++ \cite{DBLP:journals/corr/abs-2010-07492}. If a ray does not hit AABB, then we use $||\bm{c}||_2 - 1$ as $t_{far}$ and apply NeRF++ only.

\section{Monte Carlo integration}
\label{sec:monte_carlo_integration}

We will in detail describe how to compute the diffuse and specular term which contain the integrals. Our implementation is almost same as in \cite{WinNT}. However, to make the paper self-contained, we elaborate Monte Carlo integration.

First of all, the rendering equation \cite{DBLP:conf/siggraph/Kajiya86} with Cook-Torrance specular BRDF model \cite{DBLP:journals/tog/CookT82} is given:
\begin{equation}
    \label{eq:supp_physically_based_rendering}
    \begin{split}
        & \underbrace{\frac{\hat{\bm{c}}_b}{\pi} \int_{\Omega_{+}} L(\hat{\bm{x}}, \bm{l}) (\hat{\bm{n}} \cdot \bm{l}) d\bm{l}}_{diffuse\ term} \\
        + & \underbrace{\int_{\Omega_{+}} \frac{D(\bm{v}, \bm{l}, \hat{\bm{n}}, \hat{\alpha}_{r}) G(\bm{v}, \bm{l}, \hat{\bm{n}}, \hat{\alpha}_{r}) F(\bm{v}, \bm{l}, \hat{\bm{f}}_{0})}{4 (\hat{\bm{n}} \cdot \bm{v}) (\hat{\bm{n}} \cdot \bm{l})} L(\hat{\bm{x}}, \bm{l}) (\hat{\bm{n}} \cdot \bm{l}) d\bm{l}}_{specular\ term},
    \end{split}
\end{equation}
where $D$, $G$, and $F$ are defined as in \cite{WinNT}, 
\begin{equation}
    \begin{split}
        D &= \frac{\hat{\alpha}_{r}^{2}}{\pi \left( (\hat{\bm{n}} \cdot \bm{h})^{2} (\hat{\alpha}_{r}^{2} - 1) + 1\right)^{2}}, \\
        G &= \frac{2(\hat{\bm{n}} \cdot \bm{l})}{(\hat{\bm{n}} \cdot \bm{l}) + \sqrt{\hat{\alpha}_{r}^{2} + (1 - \hat{\alpha}_{r}^{2})(\hat{\bm{n}} \cdot \bm{l})^{2}}} \\
        & \times \frac{2(\hat{\bm{n}} \cdot \bm{v})}{(\hat{\bm{n}} \cdot \bm{v}) + \sqrt{\hat{\alpha}_{r}^{2} + (1 - \hat{\alpha}_{r}^{2})(\hat{\bm{n}} \cdot \bm{v})^{2}}}, \\
        F &= \bm{f}_{0} + (1 - \bm{f}_{0}) (1 - (\bm{v} \cdot \bm{h}))^{5}.
    \end{split}
\end{equation}

In case of Filament model \cite{WinNT}, we can simplify the shadowing-masking function $G$ in consideration of the denominator of \cref{eq:supp_physically_based_rendering} and without the height correction of microfacet:
\begin{equation}
    \begin{split}
        V :=& \frac{G}{4 (\hat{\bm{n}} \cdot \bm{v}) (\hat{\bm{n}} \cdot \bm{l})} \\
        =& \frac{1}{(\hat{\bm{n}} \cdot \bm{l}) + \sqrt{\hat{\alpha}_{r}^{2} + (1 - \hat{\alpha}_{r}^{2})(\hat{\bm{n}} \cdot \bm{l})^{2}}} \\
        & \times \frac{1}{(\hat{\bm{n}} \cdot \bm{v}) + \sqrt{\hat{\alpha}_{r}^{2} + (1 - \hat{\alpha}_{r}^{2})(\hat{\bm{n}} \cdot \bm{v})^{2}}}.
    \end{split}
\end{equation}
Thus, the simplified rendering equation is
\begin{equation}
    \label{eq:supp_simplified_physically_based_rendering}
    \begin{split}
        & \underbrace{\frac{\hat{\bm{c}}_b}{\pi} \int_{\Omega_{+}} L(\hat{\bm{x}}, \bm{l}) (\hat{\bm{n}} \cdot \bm{l}) d\bm{l}}_{diffuse\ term} \\
        + & \underbrace{\int_{\Omega_{+}} D(\bm{v}, \bm{l}, \hat{\bm{n}}, \hat{\alpha}_{r}) V(\bm{v}, \bm{l}, \hat{\bm{n}}, \hat{\alpha}_{r}) F(\bm{v}, \bm{l}, \hat{\bm{f}}_{0}) L(\hat{\bm{x}}, \bm{l}) (\hat{\bm{n}} \cdot \bm{l}) d\bm{l}}_{specular\ term}.
    \end{split}
\end{equation}

Accordingly, the diffuse and specular terms of NDJIR forward rendering model become
\begin{equation}
    \label{eq:supp_simplified_ndjir_diffuse_term}
    \frac{\widehat{L_{p}\bm{c}_{b}}}{\pi} 
        \int_{\Omega_{+}} 
            \left(L_{V}(\hat{\bm{x}}, \bm{l}) L_{E}(\bm{l}) (\hat{\bm{n}} \cdot \bm{l}) + \widehat{L_{I}(\bm{x})}\right) d\bm{l},
\end{equation}
\begin{equation}
    \label{eq:supp_simplified_ndjir_specular_term}
    \begin{split}
        \widehat{L_{p}} \int_{\Omega_{+}} & D(\bm{v}, \bm{l}, \hat{\bm{n}}, \hat{\alpha}_{r}) V(\bm{v}, \bm{l}, \hat{\bm{n}}, \hat{\alpha}_{r}) F(\bm{v}, \bm{l}, \hat{\bm{f}}_{0}) \\
            & L_{V}(\hat{\bm{x}}, \bm{l}) L_{E}(\bm{l}) (\hat{\bm{n}} \cdot \bm{l}) 
                I(\bm{v}, \bm{l}, \hat{\bm{n}}) d\bm{l}, 
    \end{split}
\end{equation}

For computing the specular term \cref{eq:supp_simplified_ndjir_specular_term}, we use Monte Carlo integration with importance sampling:
\begin{equation}
    \label{eq:supp_specular_term_monte_carlo_integration}
    \begin{split}
        \frac{\widehat{L_{p}}}{N_{L}} \sum_{i}^{N_{L}} & \frac{D(\bm{v}, \bm{l}_{i}, \hat{\bm{n}}, \hat{\alpha}_{r}) V(\bm{v}, \bm{l}_{i}, \hat{\bm{n}}, \hat{\alpha}_{r}) F(\bm{v}, \bm{l}_{i}, \hat{\bm{f}}_{0})}{p(\bm{v}, \bm{l}_{i}, \hat{\bm{n}}, \hat{\alpha}_{r})} \\
            & L_{V}(\hat{\bm{x}}, \bm{l}_{i}) L_{E}(\bm{l}_{i}) (\hat{\bm{n}} \cdot \bm{l}_{i}) 
                I(\bm{v}, \bm{l}_{i}, \hat{\bm{n}}), 
    \end{split}
\end{equation}
where $N_{L}$ is the number of light samples per pixel, and $p$ is the probabilistic distribution function (PDF). We use $D(\bm{v}, \bm{l}_{i}, \hat{\bm{n}}, \hat{\alpha}_{r}) (\hat{\bm{n}} \cdot \bm{h})$ for the PDF but consider the reflection of viewing direction $\bm{v}$ around half vector $\bm{h}$, so the transformed PDF is 
\begin{equation}
    \label{eq:supp_transformed_pdf}
    p = D(\bm{v}, \bm{l}_{i}, \hat{\bm{n}}, \hat{\alpha}_{r}) (\hat{\bm{n}} \cdot \bm{h}) \frac{1}{4 (\bm{v} \cdot \bm{h})}.
\end{equation}
Putting \cref{eq:supp_transformed_pdf} into \cref{eq:supp_specular_term_monte_carlo_integration} leads to
\begin{equation}
    \label{eq:supp_specular_term_monte_carlo_integration_without_pdf}
    \begin{split}
        \frac{\widehat{L_{p}}}{N_{L}} \sum_{i}^{N_{L}} & V(\bm{v}, \bm{l}_{i}, \hat{\bm{n}}, \hat{\alpha}_{r}) F(\bm{v}, \bm{l}_{i}, \hat{\bm{f}}_{0})\frac{4(\bm{v} \cdot \bm{h})}{(\hat{\bm{n}} \cdot \bm{h})} \\
            & L_{V}(\hat{\bm{x}}, \bm{l}_{i}) L_{E}(\bm{l}_{i}) (\hat{\bm{n}} \cdot \bm{l}_{i}) 
                I(\bm{v}, \bm{l}_{i}, \hat{\bm{n}}).
    \end{split}
\end{equation}

Now, we have the discretized form for the specular term \cref{eq:supp_specular_term_monte_carlo_integration_without_pdf}, the final piece is the way to sample $\bm{l}$ according to the PDF. Using the solid angle representation, we can define the PDF as
\begin{equation}
    \label{eq:supp_pdf}
    p(\theta, \phi) = \frac{\hat{\alpha}_{r}^{2}}{\pi (\cos^{2}\theta (\hat{\alpha}_{r}^{2} - 1) + 1)^{2}} \cos\theta \sin\theta, 
\end{equation}
where $\theta$ and $\phi$ is azimuthal and polar angle, respectively.
Applying the inverse transform method, we can sample $\bm{l}$ using
\begin{equation}
    \label{eq:supp_importance_sampling}
    \begin{split}
        \phi &= 2 \pi \zeta_{\phi}, \\
        \theta &= \cos^{-1} \sqrt{\frac{1 - \zeta_{\theta}}{(\hat{\alpha}_{r}^{2} - 1)\zeta_{\theta} + 1}}, \\
        \bm{l} &= \left( \cos\phi\sin\theta, \sin\phi\sin\theta, \cos\theta \right),
    \end{split}
\end{equation}
where $\zeta_{\phi}$ and $\zeta_{\theta}$ are sampled from the uniform distribution of the range $[0, 1)$. Now, $\bm{l}$ is defined in the upper-hemisphere on the world coordinate system, so we transform it to the local coordinate system where the normals $\hat{\bm{n}}$ is the up-vector: $\bm{l} \leftarrow T(\bm{n}) \bm{l}$. As noted in the main script, we do not backpropagate gradients in the sampling process as suggested in \cite{DBLP:journals/tog/ZeltnerSGJ21}.

Since the range of azimuthal angle is as twice as polar angle, we sample two times larger samples regarding $\theta$ than $\phi$ then use all possible combinations of $\theta$ and $\phi$ (same as the batch-wise \texttt{meshgrid} operation). For example, when we samples $8$ points for the polar angle, the number of light samples per pixel is $8 \times 16 = 128$.

In summary, for computing the specular term, we sample light $\bm{l}$ according to \cref{eq:supp_importance_sampling}, then transform it as $\bm{l} \leftarrow T(\bm{n}) \bm{l}$, finally apply it to \cref{eq:supp_specular_term_monte_carlo_integration_without_pdf}.

Similarly, the diffuse term \cref{eq:supp_simplified_ndjir_diffuse_term} can be computed by Monte Carlo integration but with uniform sampling:
\begin{equation}
    \label{eq:supp_diffuse_term_monte_carlo_integration}
    \frac{\widehat{L_{p}\bm{c}_{b}}}{N_{L}}
        \sum_{i=1}^{N_{L}} L_{V}(\hat{\bm{x}}, \bm{l}_{i}) L_{E}(\bm{l}_{i}) (\hat{\bm{n}} \cdot \bm{l}_{i}) + \widehat{L_{p}\bm{c}_{b}} \widehat{L_{I}(\bm{x})}.
\end{equation}
Note that the normalization factor $\pi$ is canceled out, and $\theta$ of \cref{eq:supp_importance_sampling} becomes
\begin{equation}
    \label{eq:supp_uniform_sampling}
    \theta = \cos^{-1} (1 - \zeta_{\theta}).
\end{equation}

\section{Derivation of model with split-sum and PIL}
\label{sec:derivation_of_model_with_split-sum_and_pil}

As baseline, we implemented the model which bears the similar spirit of split-sum \cite{DBLP:conf/cvpr/MunkbergCHES0GF22} and pre-integrated light (PIL) \cite{DBLP:conf/nips/BossJBLBL21}.

We split the specular term of \cref{eq:supp_physically_based_rendering} into two factors:
\begin{equation}
    \label{eq:supp_split-sum}
    \int_{\Omega_{+}} L(\hat{\bm{x}}, \bm{l}) (\hat{\bm{n}} \cdot \bm{l}) d\bm{l}
    \int_{\Omega_{+}} \frac{DGF}{4 (\hat{\bm{n}} \cdot \bm{v}) (\hat{\bm{n}} \cdot \bm{l})} (\hat{\bm{n}} \cdot \bm{l}) d\bm{l}.
\end{equation}

Then, representing the light integral factor divided by $\pi$ using pre-integrated neural network $\tilde{L}(\hat{\bm{x}}, \hat{\bm{n}}, \bm{v}, d^{-2})$, the rendering equation \cref{eq:supp_physically_based_rendering} becomes 
\begin{equation}
    \label{eq:supp_split-sum_and_pil}
    \tilde{L} \hat{\bm{c}}_b 
    + (\pi \tilde{L}) \int_{\Omega_{+}} \frac{DGF}{4 (\hat{\bm{n}} \cdot \bm{v}) (\hat{\bm{n}} \cdot \bm{l})} (\hat{\bm{n}} \cdot \bm{l}) d\bm{l}.
\end{equation}

Similarly to NDJIR forward rendering model, we entangle the two factors in the diffuse term and apply volume rendering to quantities, then the forward rendering model of split-sum + PIL approximation is
\begin{equation}
    \label{eq:supp_split-sum_and_pil_forward_rendering_model}
    \widehat{\tilde{L} \bm{c}_b}
    + (\pi \hat{\tilde{L}}) \int_{\Omega_{+}} \frac{DGF}{4 (\hat{\bm{n}} \cdot \bm{v}) (\hat{\bm{n}} \cdot \bm{l})} (\hat{\bm{n}} \cdot \bm{l}) d\bm{l}.
\end{equation}

\section{Double backward of voxel grid feature}
\label{sec:double_backward_of_voxel_grid_feature}

Naive implementation of the voxel grid feature is to use function composition each of which supports \texttt{auto-grad} as implemented in modern neural network libraries \cite{paszke2017automatic,hayakawa2021neural,jax2018github}. However, such implementation constructs a long and complex computation graph which can not be optimized automatically, resulting in computational overhead in speed and large memory footprint. Thus, we implement specific CUDA \cite{cuda} kernels to support double-backward \cite{155328}.

\subsection{Forward function}
\label{subsec:forward_function}

Given query points $\{\bm{x}_{b}\} = \{ (x_{b}, y_{b}, z_{b}) \}$, voxel grid feature $\bm{F}_{V}$ with grid size $\bm{G}$ and bounded in min $\bm{m}$ and max $\bm{M}$, we first transform query point to the discrete coordinate system:
\begin{equation}
    \label{eq:supp_transform_to_discrete_coordinate}
    \bar{\bm{x}}_{b} = \frac{\bm{x}_{b} - \bm{m}}{\bm{M} - \bm{m}} \bm{G} = \bm{s} (\bm{x}_{b} - \bm{m}).
\end{equation}
Then, the voxel grid feature outputs
\begin{equation}
    \label{eq:supp_forward_function}
    \bar{\bm{f}}(\bar{\bm{x}}_{b}) = \sum_{i, j, k} w_{i, j, k}(\bar{\bm{x}}_{b}) \bm{f}_{i, j, k}(\bar{\bm{x}}_{b}),
\end{equation}
where $w_{i, j, k}(\bar{\bm{x}}_{b})$ is the interpolation coefficient, and $\bm{f}_{i, j, k} (\bar{\bm{x}}_{b})$ is the voxel grid feature queried by $\bar{\bm{x}}_{b}$ at $i, j, k$-th location of the discrete coordinate system.

\subsection{Grad function}
\label{subsec:grad_function}

For gradient function, we denote as $g^{(n)}$ gradient operator on $n$-th order gradient graph. Here, we have two inputs: query point $\bm{x}_{b}$ and voxel grid feature $\bm{f}_{i,j,k}$. Correspondingly, there are two grad functions.

\textbf{Gradient w.r.t. query point:}
\begin{equation}
    \label{eq:supp_grad_wrt_query}
    \begin{split}
        g^{(1)}x_{b} = \sum_{d} g^{(1)}\bar{f}^{d}(\bar{\bm{x}}_{b}) \sum_{i, j, k} s_{x} \frac{\partial w_{i, j, k}(\bar{\bm{x}}_{b})}{\partial \bar{x}_{b}} f_{i, j, k}^{d}(\bar{\bm{x}}_{b})\\
        g^{(1)}y_{b} = \sum_{d} g^{(1)}\bar{f}^{d}(\bar{\bm{x}}_{b}) \sum_{i, j, k} s_{y} \frac{\partial w_{i, j, k}(\bar{\bm{x}}_{b})}{\partial \bar{y}_{b}} f_{i, j, k}^{d}(\bar{\bm{x}}_{b})\\
        g^{(1)}z_{b} = \sum_{d} g^{(1)}\bar{f}^{d}(\bar{\bm{x}}_{b}) \sum_{i, j, k} s_{z} \frac{\partial w_{i, j, k}(\bar{\bm{x}}_{b})}{\partial \bar{z}_{b}} f_{i, j, k}^{d}(\bar{\bm{x}}_{b}).
    \end{split}
\end{equation}
Note that $s_{x}$, $s_{y}$, and $s_{z}$ come from the chain rule of the backpropagation, $\frac{\partial w_{i, j, k}(\bar{\bm{x}}_{b})}{\partial \bm{x}_{b}} = \frac{\partial \bar{\bm{x}}_{b}}{\partial \bm{x}_{b}} \frac{\partial w_{i, j, k}(\bar{\bm{x}}_{b})}{\partial \bar{\bm{x}}_{b}}$.

\textbf{Gradient w.r.t. voxel grid feature:}
\begin{equation}
    \label{eq:supp_grad_wrt_feature}
    g^{(1)}\bm{f}_{i, j, k}(\bar{\bm{x}}_{b}) \mathrel{+}= g^{(1)} \bar{\bm{f}}(\bar{\bm{x}}_{b}) w_{i, j, k}(\bar{\bm{x}}_{b})
\end{equation}

We can write a specific form regarding the derivative of $w_{i, j, k}(\bar{\bm{x}}_{b})$, but it depends on implementation of interpolation. Further, if we write a specific form, notations become more complicated. Thus, this form is enough to proceed. Note that in case of Straight Through Estimator (STE) \cite{DBLP:conf/nips/CourbariauxBD15}, $\frac{\partial w_{i, j, k}(\bar{\bm{x}})}{\partial \bar{x}} = 0$ and same for $y$ and $z$, meaning the normals of the voxel grid feature would be ignored.

\subsection{Grad grad function}
\label{subsec:grad_grad_function}

For gradients of gradients, there are five cases. However, we are only interested in Eikonal regularization \cite{DBLP:conf/icml/GroppYHAL20} thus solely consider two cases.

\textbf{Gradient of} \cref{eq:supp_grad_wrt_query} \textbf{w.r.t. grad output:}
\begin{equation}
    \label{eq:supp_grad_grad_wrt_gradoutput}
    \begin{split}
        g^{(2)}g^{(1)} \bar{f}^{d}(\bar{\bm{x}}_{b}) 
        & = g^{(2)}g^{(1)} x_{b}
        \sum_{i, j, k} s_{x} \frac{\partial w_{i, j, k}(\bar{\bm{x}}_{b})}{\partial \bar{x}_{b}} f_{i, j, k}^{d}(\bar{\bm{x}}_{b})\\
        & + g^{(2)}g^{(1)} y_{b}
        \sum_{i, j, k} s_{y} \frac{\partial w_{i, j, k}(\bar{\bm{x}}_{b})}{\partial \bar{y}_{b}} f_{i, j, k}^{d}(\bar{\bm{x}}_{b})\\
        & + g^{(2)}g^{(1)} z_{b}
        \sum_{i, j, k} s_{z} \frac{\partial w_{i, j, k}(\bar{\bm{x}}_{b})}{\partial \bar{z}_{b}} f_{i, j, k}^{d}(\bar{\bm{x}}_{b})
    \end{split}
\end{equation}
\textbf{Gradient of} \cref{eq:supp_grad_wrt_query} \textbf{w.r.t. voxel grid feature:}
\begin{equation}
    \label{eq:supp_grad_grad_wrt_feature}
    \begin{split}
        g^{(2)} f_{i, j, k}^{d}(\bar{\bm{x}}_{b})
        & \mathrel{+}= g^{(2)}g^{(1)}x_{b}
        g^{(1)}\bar{f}^{d}(\bar{\bm{x}}_{b}) s_{x} \frac{\partial w_{i, j, k}(\bar{\bm{x}}_{b})}{\partial \bar{x}_{b}} \\
        & + g^{(2)}g^{(1)}y_{b}
        g^{(1)}\bar{f}^{d}(\bar{\bm{x}}_{b}) s_{y} \frac{\partial w_{i, j, k}(\bar{\bm{x}}_{b})}{\partial \bar{y}_{b}} \\
        & + g^{(2)}g^{(1)}z_{b}
        g^{(1)}\bar{f}^{d}(\bar{\bm{x}}_{b}) s_{z} \frac{\partial w_{i, j, k}(\bar{\bm{x}}_{b})}{\partial \bar{z}_{b}}.
    \end{split}
\end{equation}

In CUDA implementation, we do not optimize codes, simply parallelize both $b$ and $d$ dimensions, and use the \texttt{atomic\_add} operation for the respective summations. Note that $\bm{f}_{i, j, k} (\bar{\bm{x}}_{b})$ is the result of look-up, we do not (cannot) backpropagate w.r.t. $\bm{x}$.

\section{More experimental results}
\label{sec:more_experimental_results}

\subsection{More primary result}
\label{subsec:more_primary_result}

\begin{figure*}[tbp]
    \centering
    \rotatebox[origin=b]{90}{scan24}\quad
    \begin{subfigure}[h]{0.14\paperwidth}
        \caption{normals}
        \includegraphics[width=\textwidth]{assets/raw_images/NDJIR/default_epoch1500__groups_gcb50379_dataset_DTU_scan24/model_01499_512grid_trimmed_base_color_mesh00_filtered02_normals_default/32.png.jpg}
    \end{subfigure}
    \hspace{1pt}
    \begin{subfigure}[h]{0.14\paperwidth}
        \caption{base color}
        \includegraphics[width=\textwidth]{assets/raw_images/NDJIR/default_epoch1500__groups_gcb50379_dataset_DTU_scan24/model_01499_512grid_trimmed_base_color_mesh00_filtered02_defaultUnlit_default/32.png.jpg}
    \end{subfigure}
    \hspace{1pt}
    \begin{subfigure}[h]{0.14\paperwidth}
        \caption{roughness}
        \includegraphics[width=\textwidth]{assets/raw_images/NDJIR/default_epoch1500__groups_gcb50379_dataset_DTU_scan24/model_01499_512grid_trimmed_roughness_mesh00_filtered02_defaultUnlit_default/32.png.jpg}
    \end{subfigure}
    \hspace{1pt}
    \begin{subfigure}[h]{0.14\paperwidth}
        \caption{specular reflectance}
        \includegraphics[width=\textwidth]{assets/raw_images/NDJIR/default_epoch1500__groups_gcb50379_dataset_DTU_scan24/model_01499_512grid_trimmed_specular_reflectance_mesh00_filtered02_defaultUnlit_default/32.png.jpg}
    \end{subfigure}
    \hspace{1pt}
    \begin{subfigure}[h]{0.14\paperwidth}
        \caption{implicit illumination}
        \includegraphics[width=\textwidth]{assets/raw_images/NDJIR/default_epoch1500__groups_gcb50379_dataset_DTU_scan24/model_01499_512grid_trimmed_implicit_illumination_mesh00_filtered02_defaultUnlit_default/32.png.jpg}
    \end{subfigure}

    \smallskip
    \rotatebox[origin=b]{90}{scan37}\quad
    \begin{subfigure}[h]{0.14\paperwidth}
        \includegraphics[width=\textwidth]{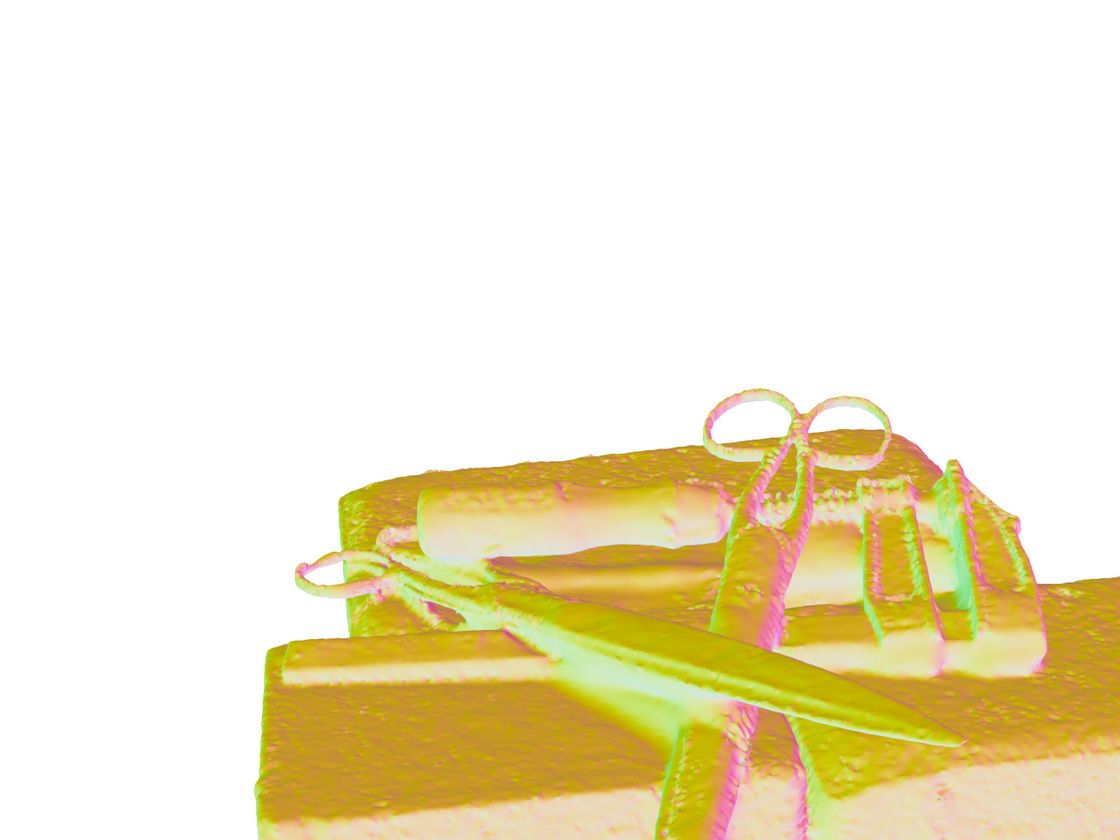}
    \end{subfigure}
    \hspace{1pt}
    \begin{subfigure}[h]{0.14\paperwidth}
        \includegraphics[width=\textwidth]{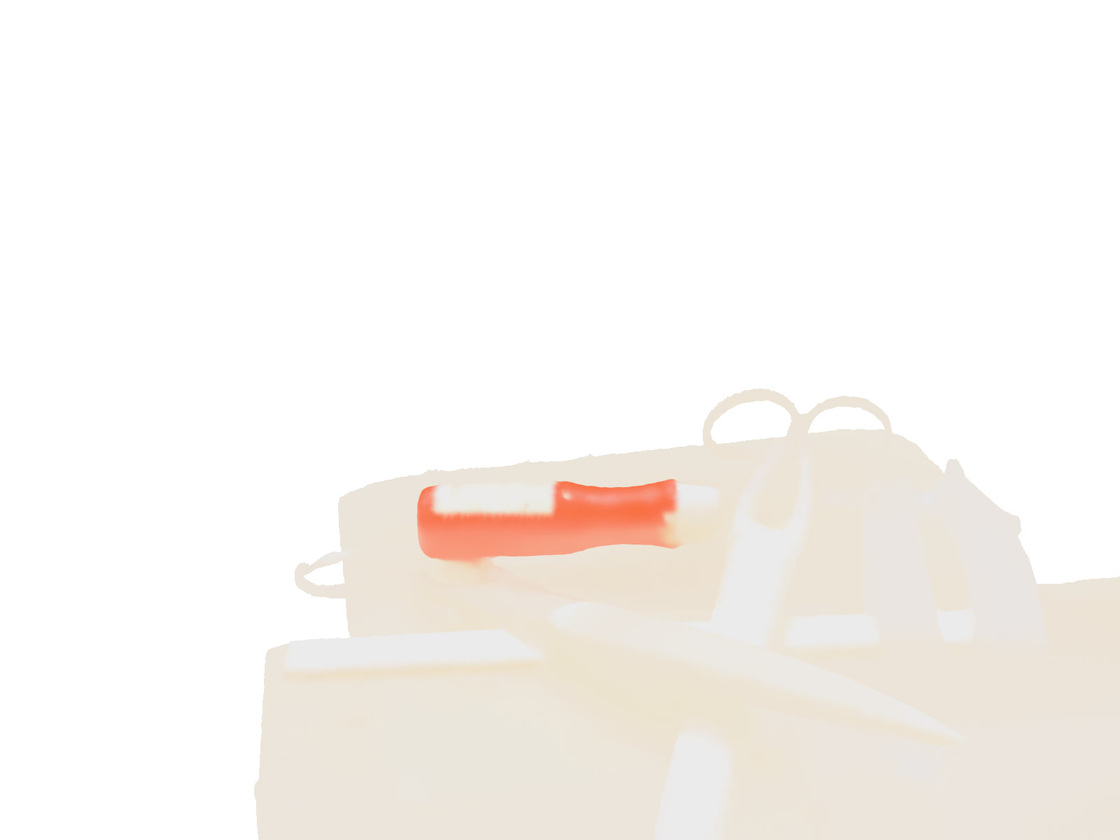}
    \end{subfigure}
    \hspace{1pt}
    \begin{subfigure}[h]{0.14\paperwidth}
        \includegraphics[width=\textwidth]{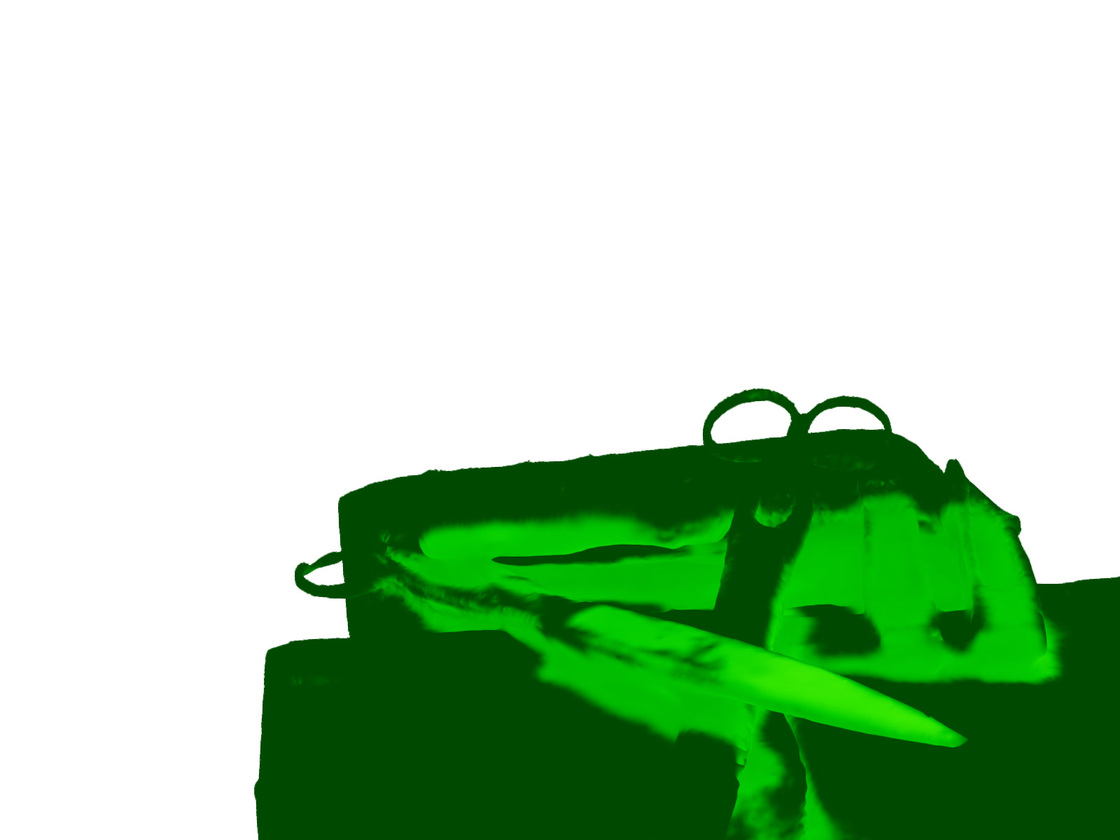}
    \end{subfigure}
    \hspace{1pt}
    \begin{subfigure}[h]{0.14\paperwidth}
        \includegraphics[width=\textwidth]{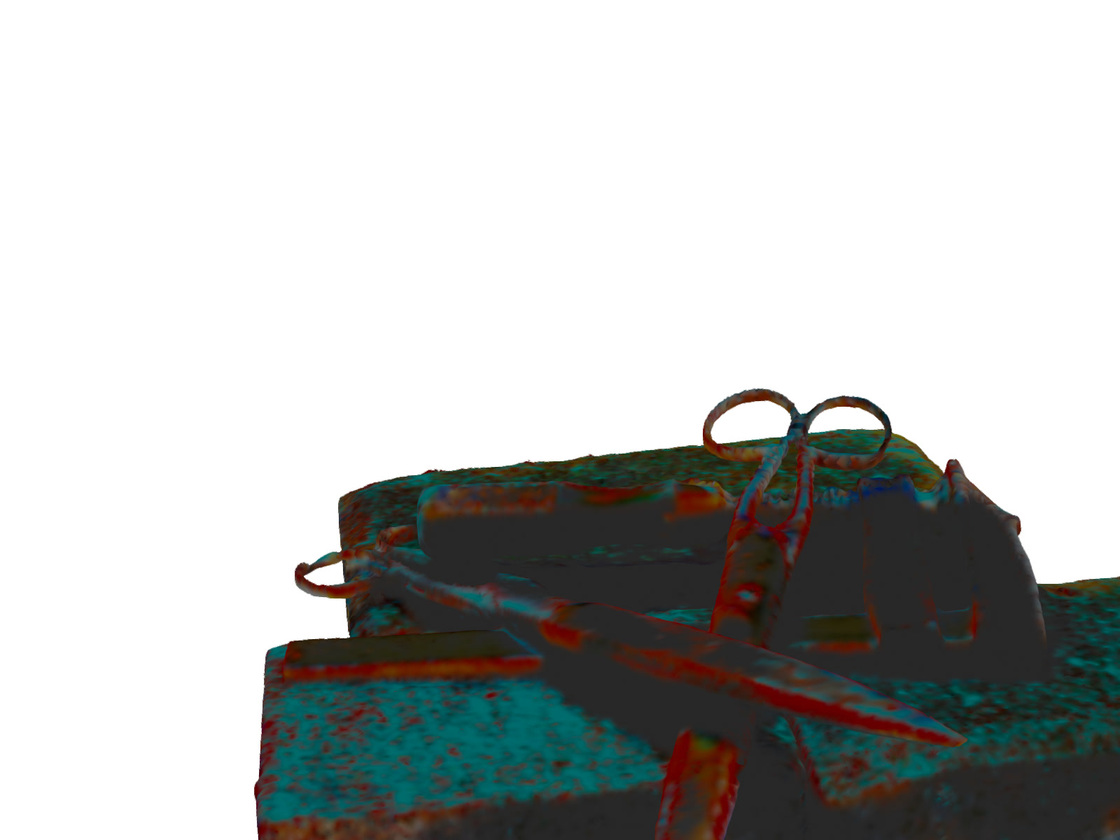}
    \end{subfigure}
    \hspace{1pt}
    \begin{subfigure}[h]{0.14\paperwidth}
        \includegraphics[width=\textwidth]{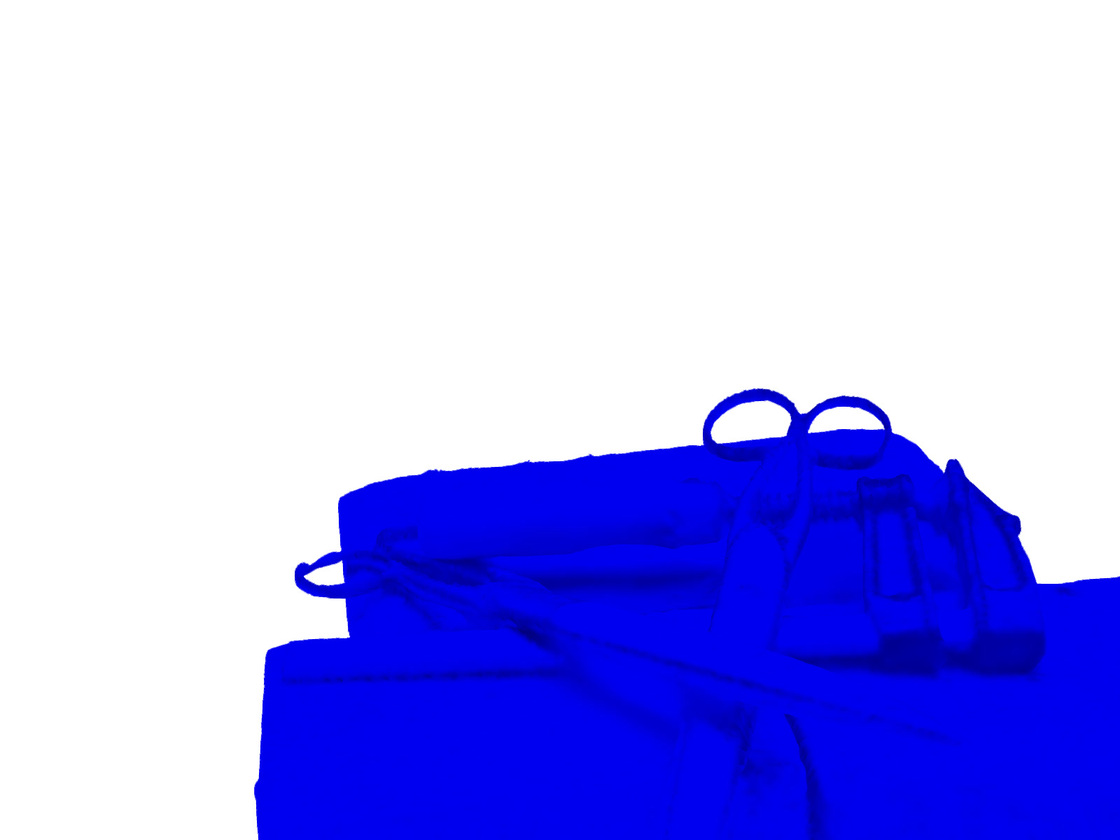}
    \end{subfigure}

    \smallskip
    \rotatebox[origin=b]{90}{scan40}\quad
    \begin{subfigure}[h]{0.14\paperwidth}
        \includegraphics[width=\textwidth]{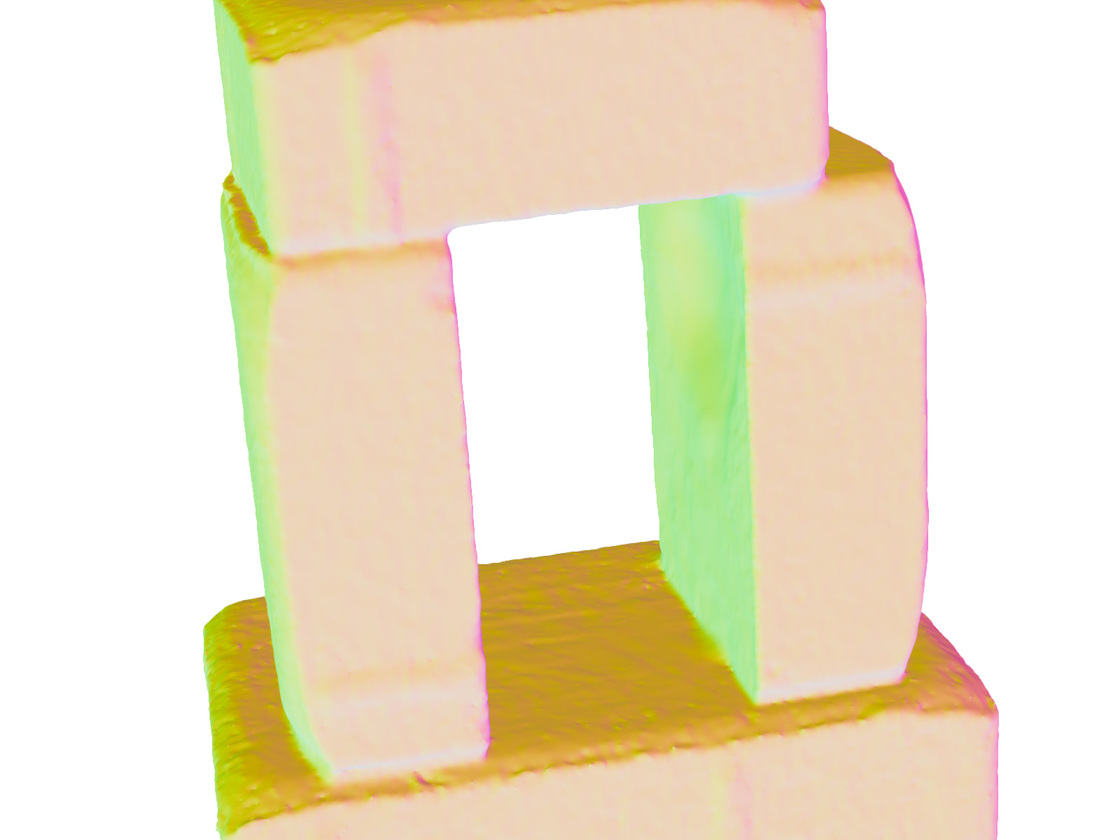}
    \end{subfigure}
    \hspace{1pt}
    \begin{subfigure}[h]{0.14\paperwidth}
        \includegraphics[width=\textwidth]{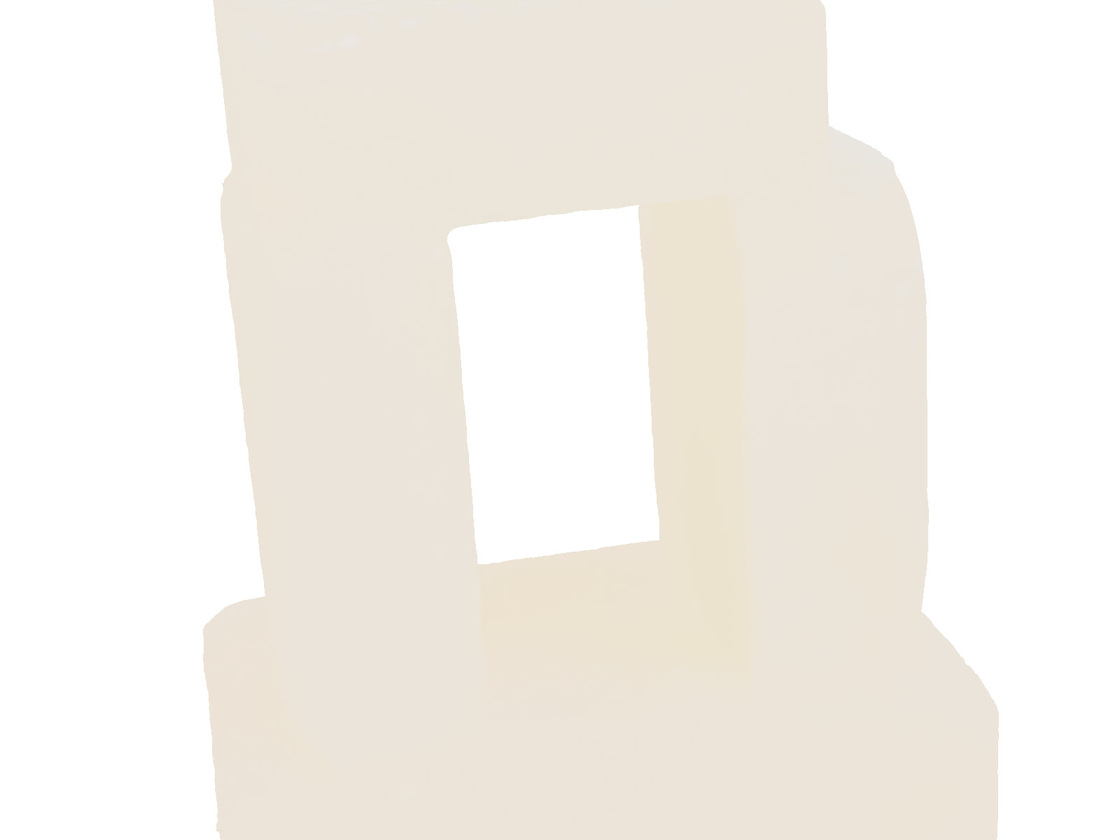}
    \end{subfigure}
    \hspace{1pt}
    \begin{subfigure}[h]{0.14\paperwidth}
        \includegraphics[width=\textwidth]{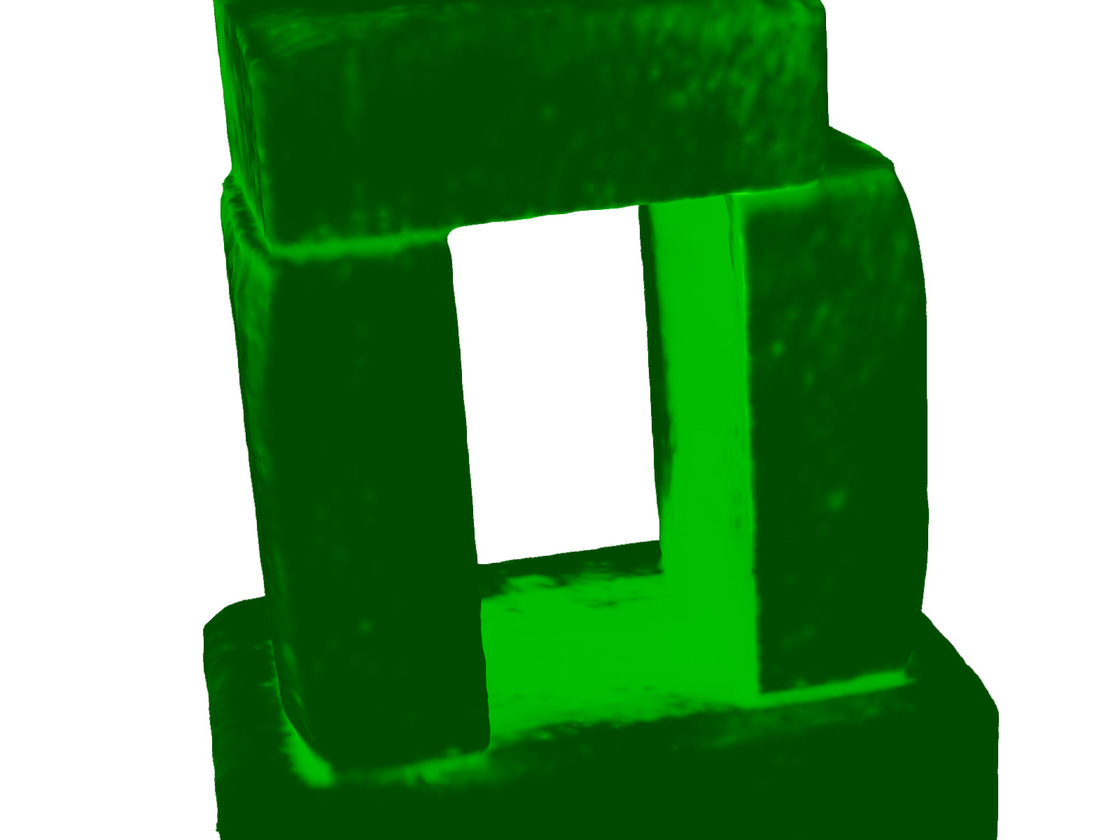}
    \end{subfigure}
    \hspace{1pt}
    \begin{subfigure}[h]{0.14\paperwidth}
        \includegraphics[width=\textwidth]{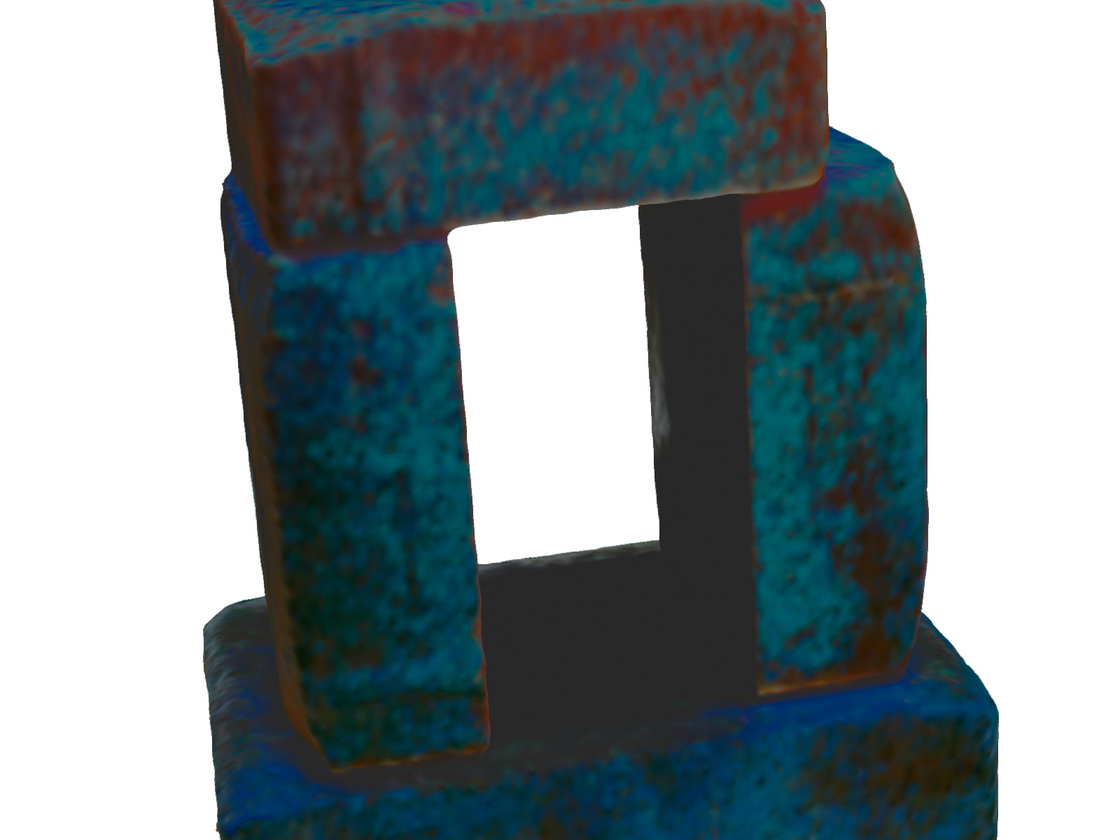}
    \end{subfigure}
    \hspace{1pt}
    \begin{subfigure}[h]{0.14\paperwidth}
        \includegraphics[width=\textwidth]{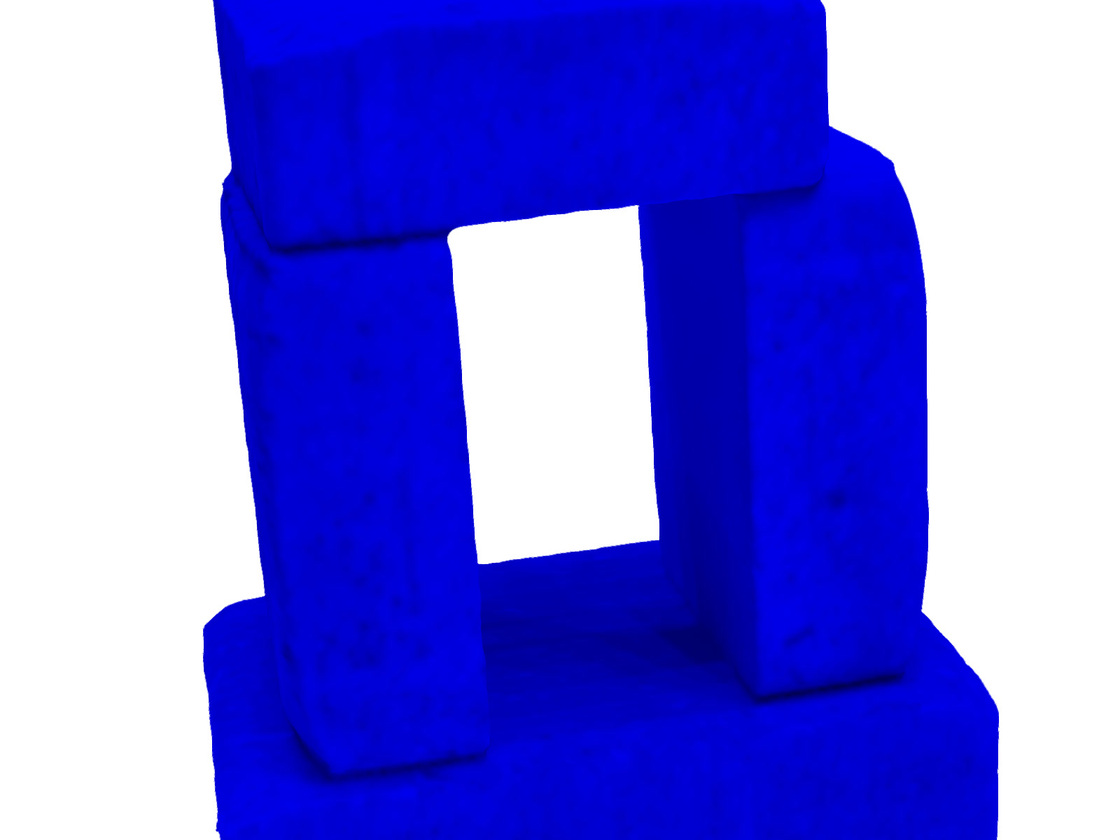}
    \end{subfigure}

    \smallskip
    \rotatebox[origin=b]{90}{scan55}\quad
    \begin{subfigure}[h]{0.14\paperwidth}
        \includegraphics[width=\textwidth]{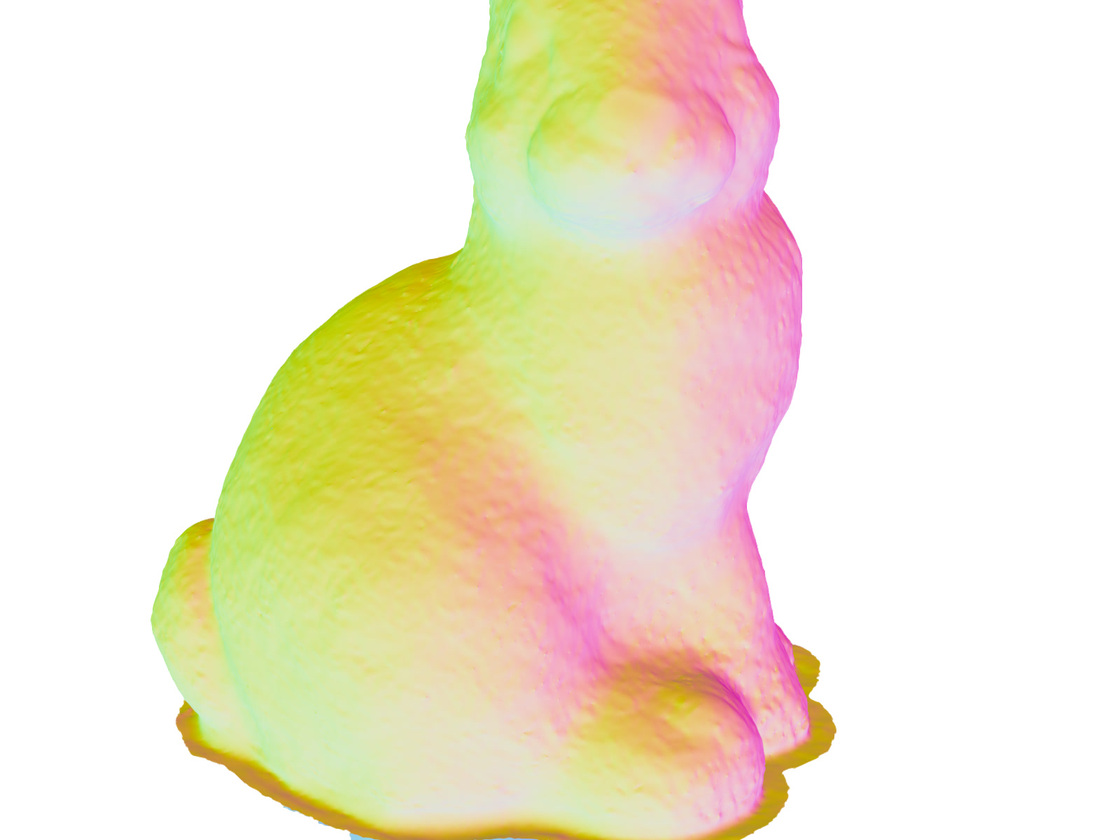}
    \end{subfigure}
    \hspace{1pt}
    \begin{subfigure}[h]{0.14\paperwidth}
        \includegraphics[width=\textwidth]{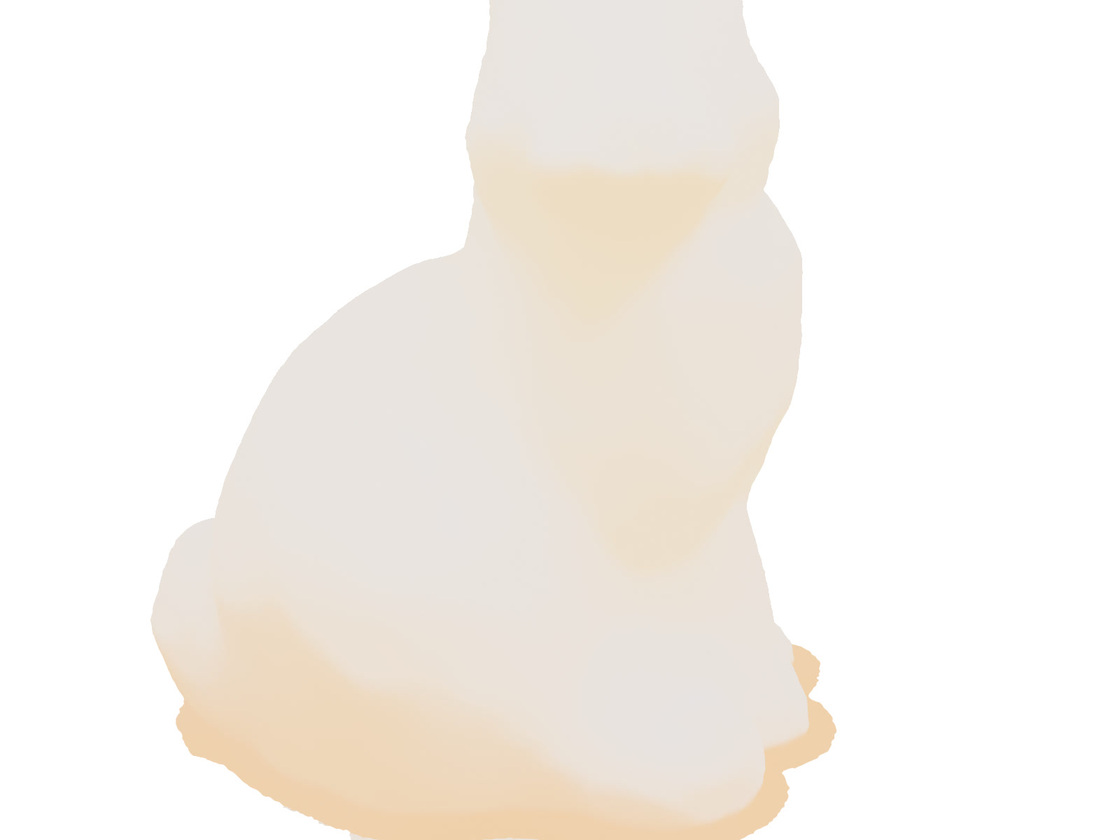}
    \end{subfigure}
    \hspace{1pt}
    \begin{subfigure}[h]{0.14\paperwidth}
        \includegraphics[width=\textwidth]{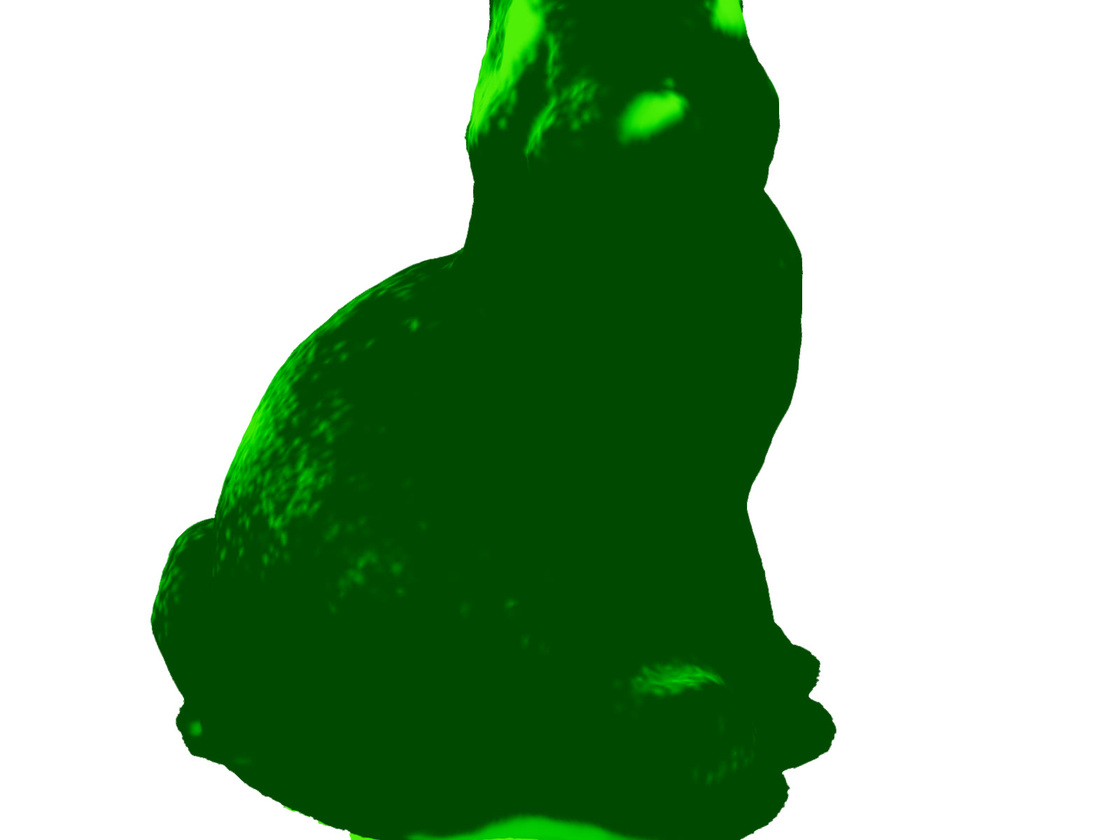}
    \end{subfigure}
    \hspace{1pt}
    \begin{subfigure}[h]{0.14\paperwidth}
        \includegraphics[width=\textwidth]{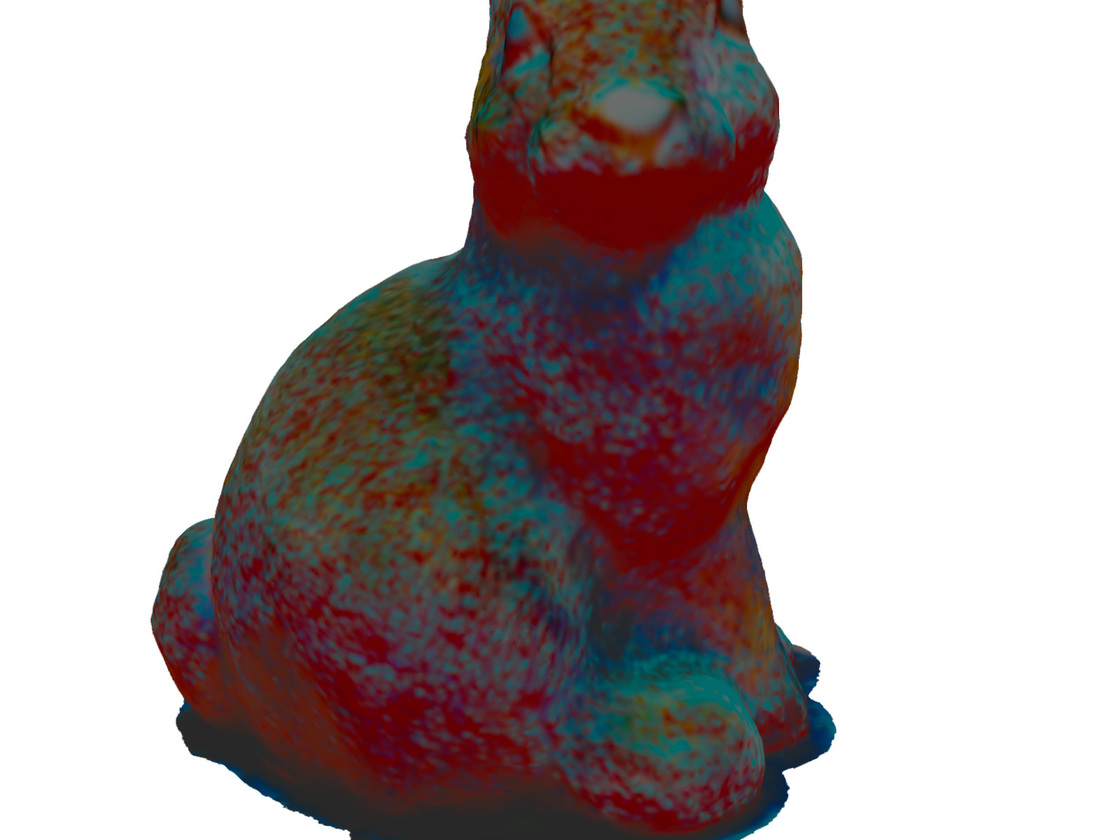}
    \end{subfigure}
    \hspace{1pt}
    \begin{subfigure}[h]{0.14\paperwidth}
        \includegraphics[width=\textwidth]{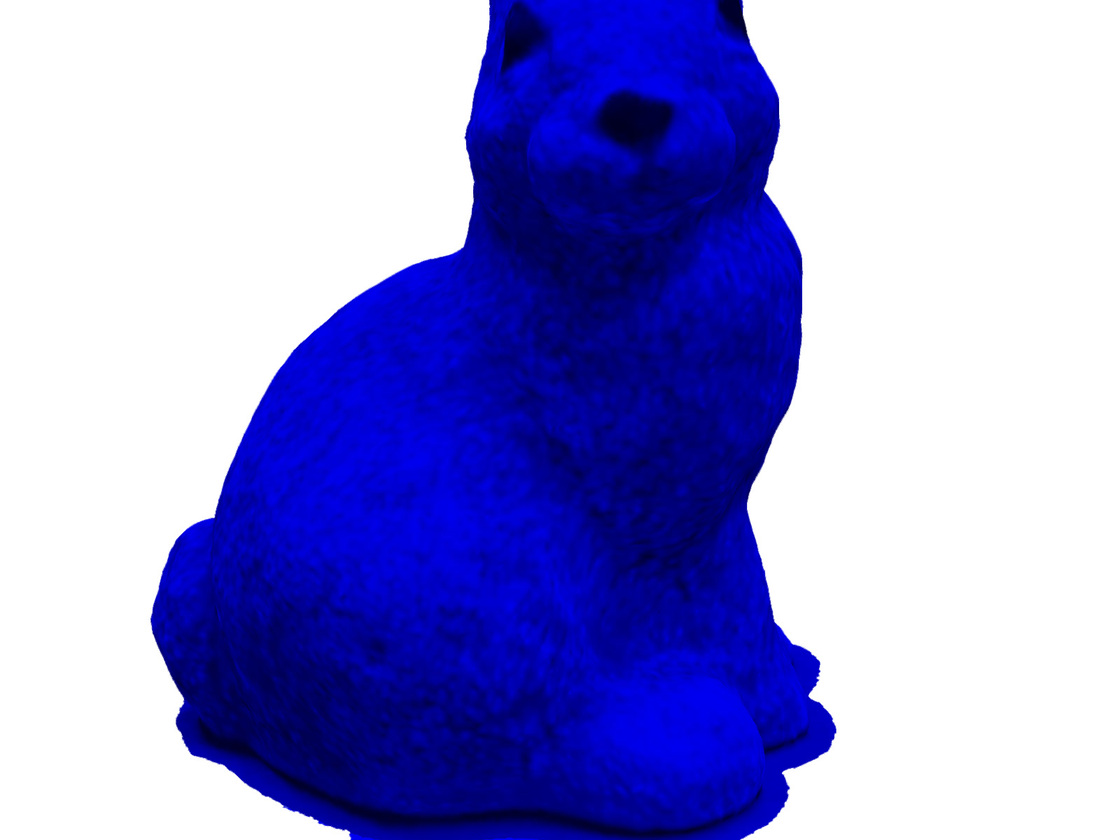}
    \end{subfigure}

    \smallskip
    \rotatebox[origin=b]{90}{scan24}\quad
    \begin{subfigure}[h]{0.17\paperwidth}
        \caption{PBR (default)}
        \includegraphics[width=\textwidth]{assets/raw_images/NDJIR/default_epoch1500__groups_gcb50379_dataset_DTU_scan24/model_01499_512grid_trimmed_base_color_mesh00_filtered02_defaultLit_default_32.png.jpg}
    \end{subfigure}
    \hspace{1pt}
    \begin{subfigure}[h]{0.17\paperwidth}
        \caption{PBR (pillars)}
        \includegraphics[width=\textwidth]{assets/raw_images/NDJIR/default_epoch1500__groups_gcb50379_dataset_DTU_scan24/model_01499_512grid_trimmed_base_color_mesh00_filtered02_defaultLit_pillars_32.png.jpg}
    \end{subfigure}
    \hspace{1pt}
    \begin{subfigure}[h]{0.17\paperwidth}
        \caption{Neural rendering}
        \includegraphics[width=\textwidth]{assets/raw_images/NDJIR/default_epoch1500__groups_gcb50379_dataset_DTU_scan24/Eval-rendered-image-1600x1200/000032-000.png.jpg}
    \end{subfigure}
    \hspace{1pt}
    \begin{subfigure}[h]{0.17\paperwidth}
        \caption{Groundtruth}
        \includegraphics[width=\textwidth]{assets/DTU/scan24/image/000032.png.jpg}
    \end{subfigure}

    \smallskip
    \rotatebox[origin=b]{90}{scan37}\quad
    \begin{subfigure}[h]{0.17\paperwidth}
        \includegraphics[width=\textwidth]{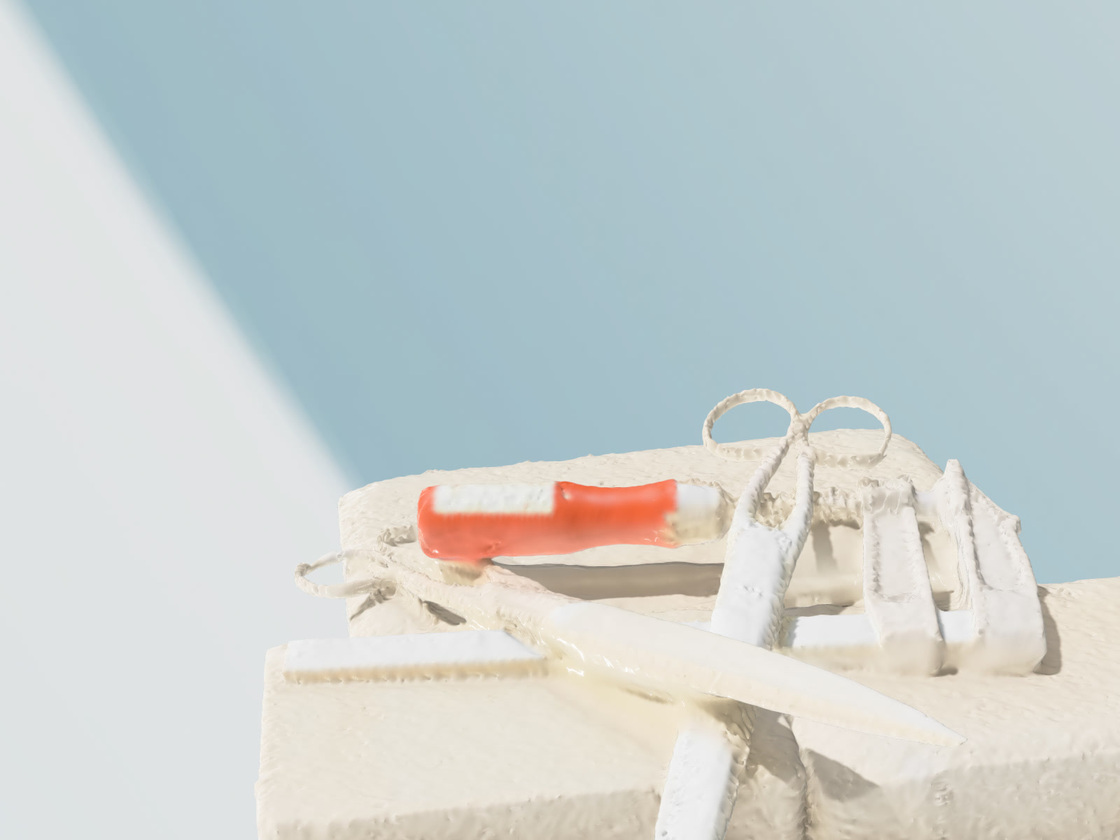}
    \end{subfigure}
    \hspace{1pt}
    \begin{subfigure}[h]{0.17\paperwidth}
        \includegraphics[width=\textwidth]{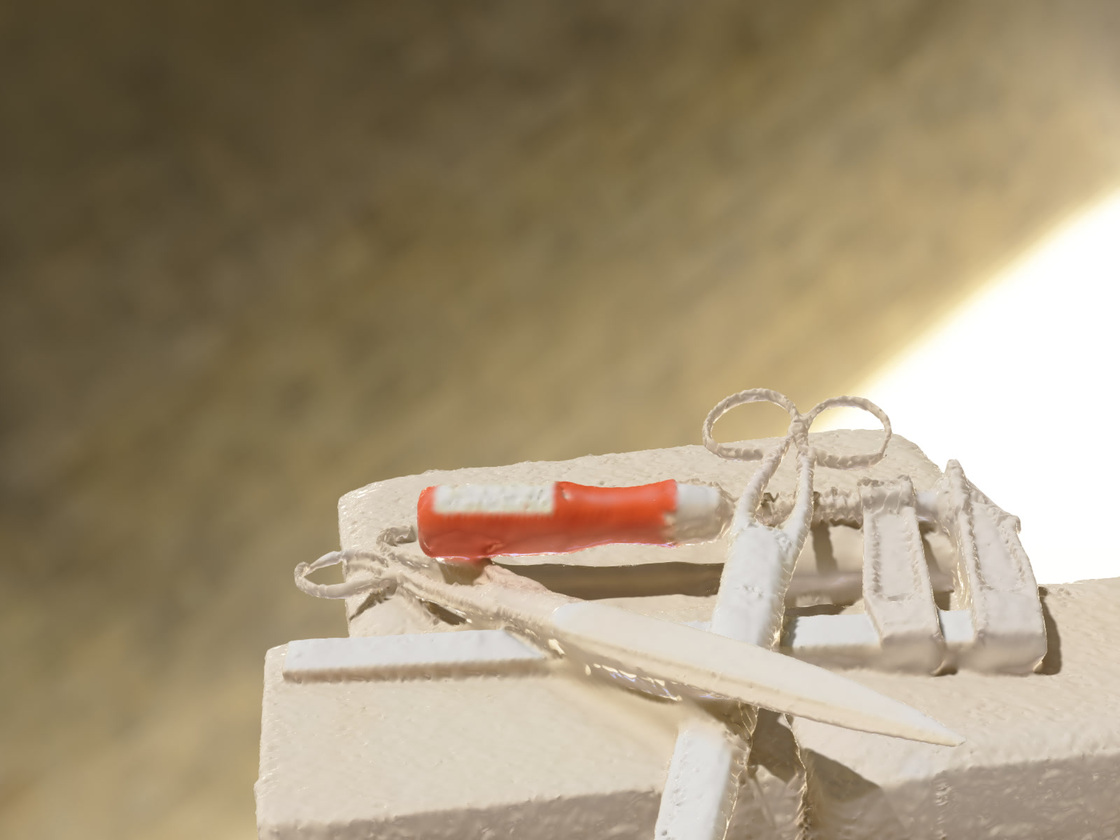}
    \end{subfigure}
    \hspace{1pt}
    \begin{subfigure}[h]{0.17\paperwidth}
        \includegraphics[width=\textwidth]{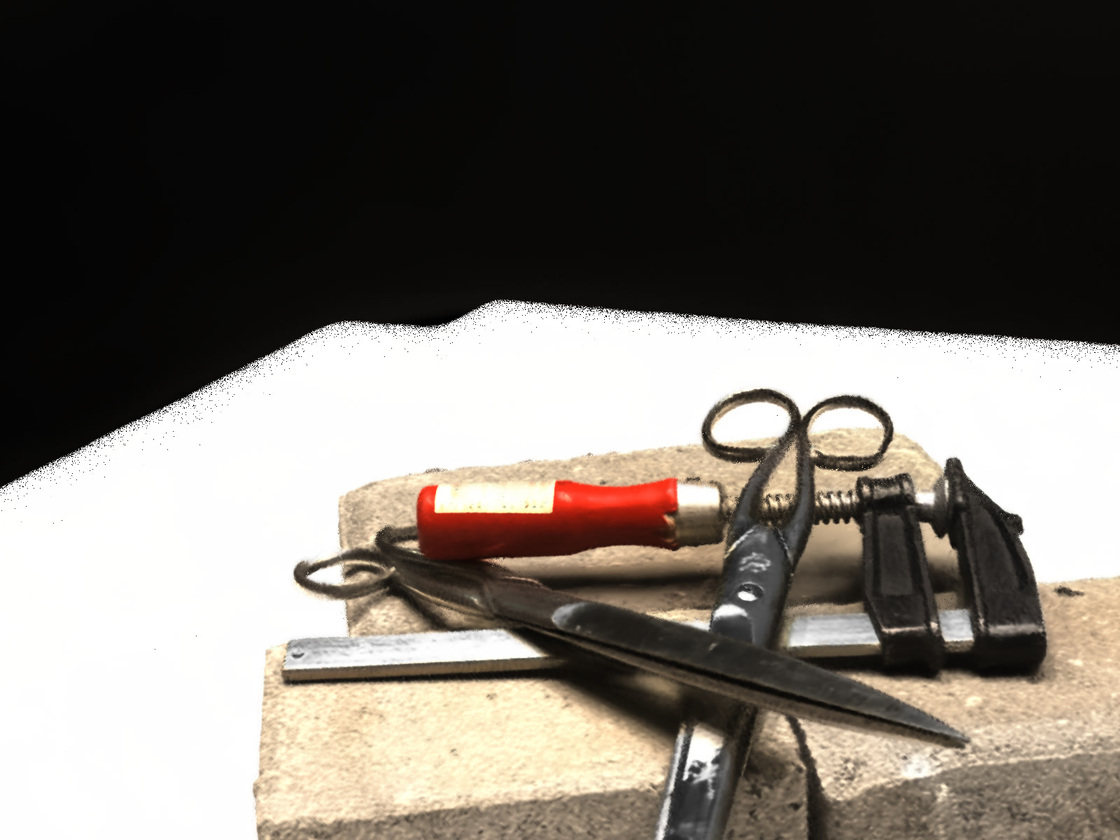}
    \end{subfigure}
    \hspace{1pt}
    \begin{subfigure}[h]{0.17\paperwidth}
        \includegraphics[width=\textwidth]{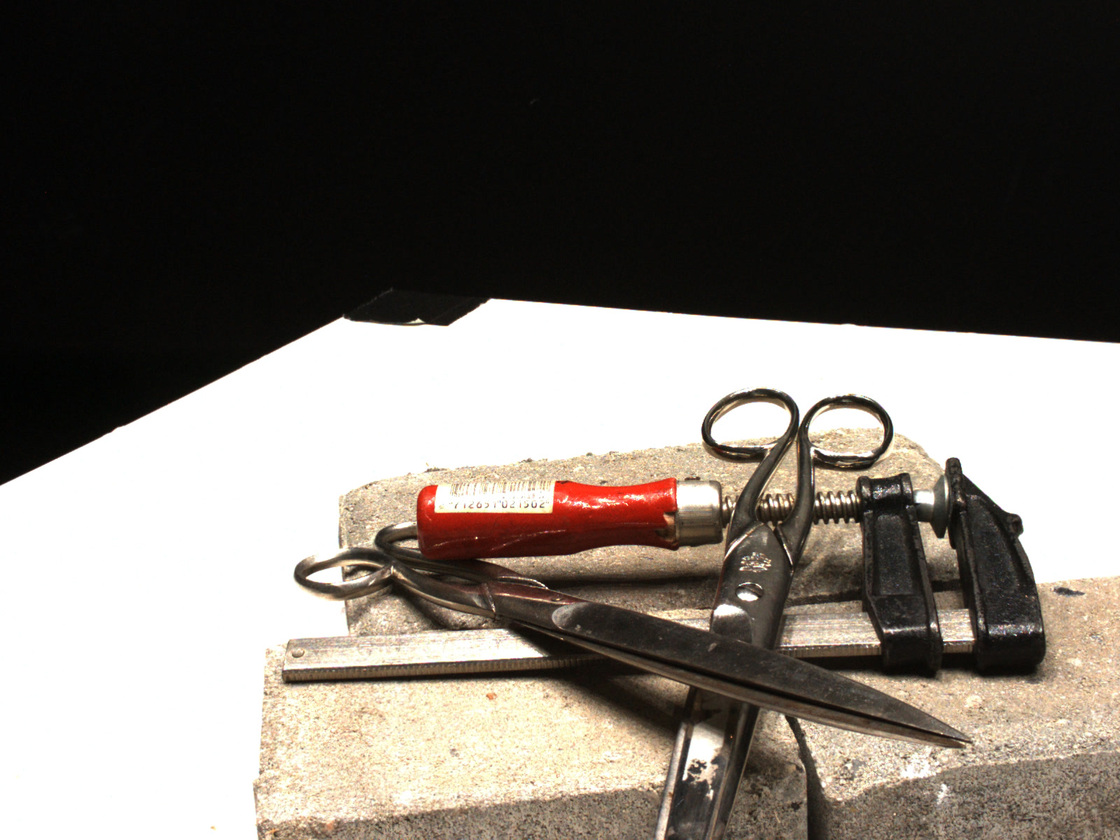}
    \end{subfigure}

    \smallskip
    \rotatebox[origin=b]{90}{scan40}\quad
    \begin{subfigure}[h]{0.17\paperwidth}
        \includegraphics[width=\textwidth]{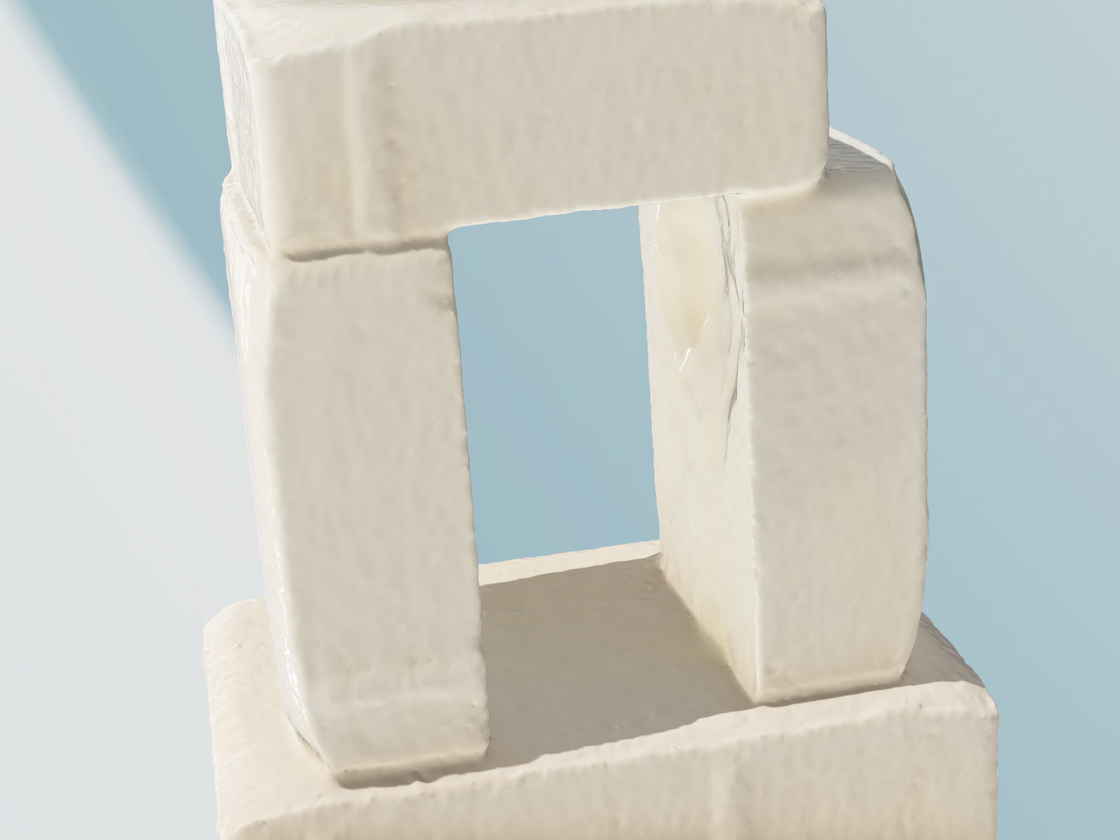}
    \end{subfigure}
    \hspace{1pt}
    \begin{subfigure}[h]{0.17\paperwidth}
        \includegraphics[width=\textwidth]{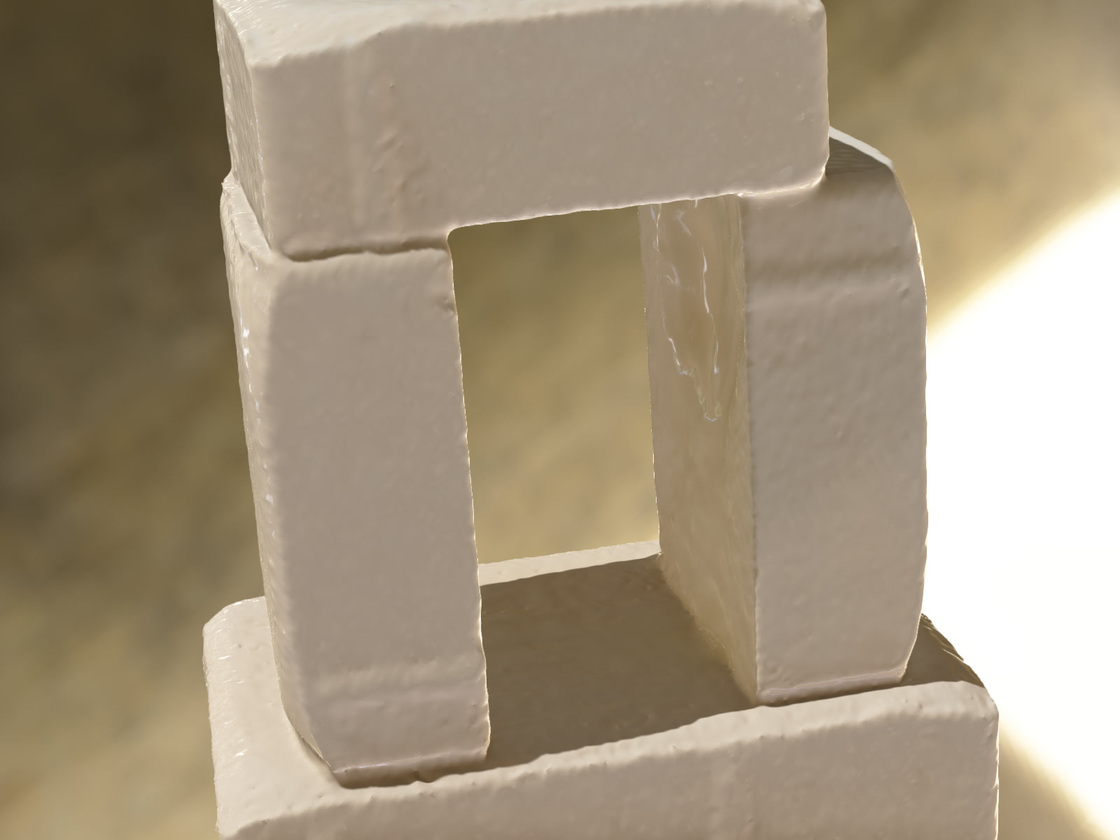}
    \end{subfigure}
    \hspace{1pt}
    \begin{subfigure}[h]{0.17\paperwidth}
        \includegraphics[width=\textwidth]{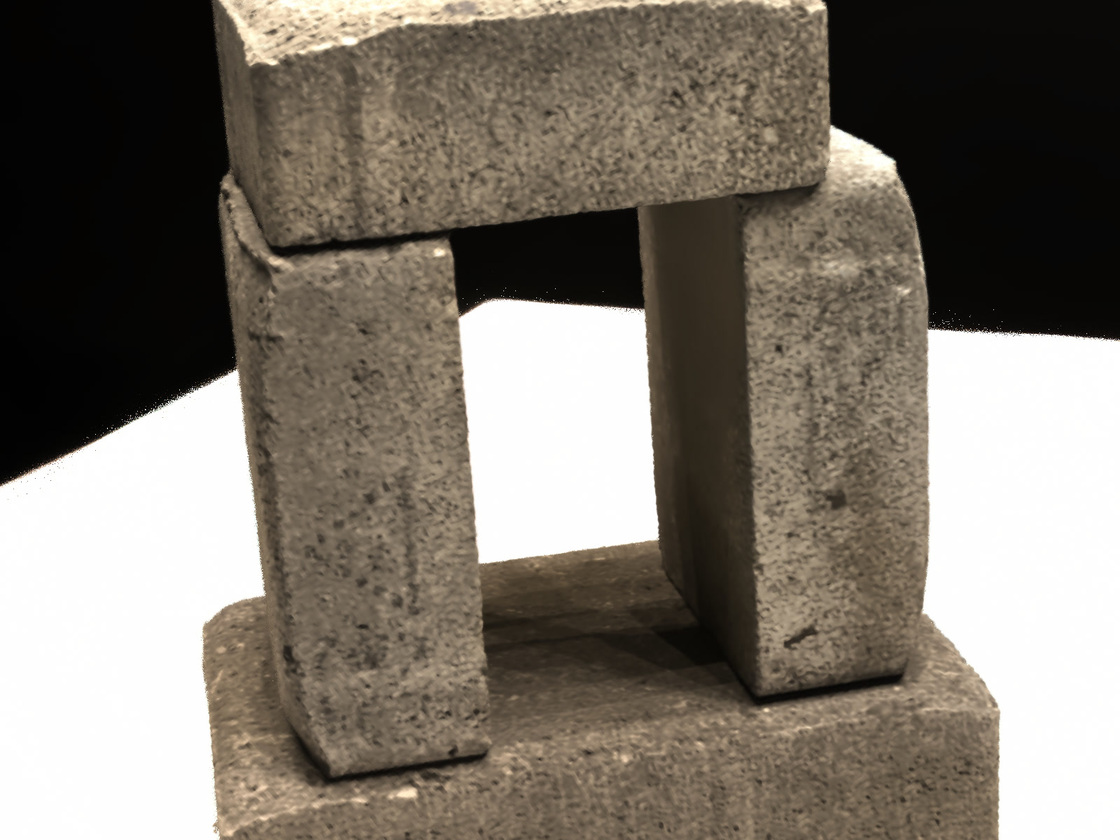}
    \end{subfigure}
    \hspace{1pt}
    \begin{subfigure}[h]{0.17\paperwidth}
        \includegraphics[width=\textwidth]{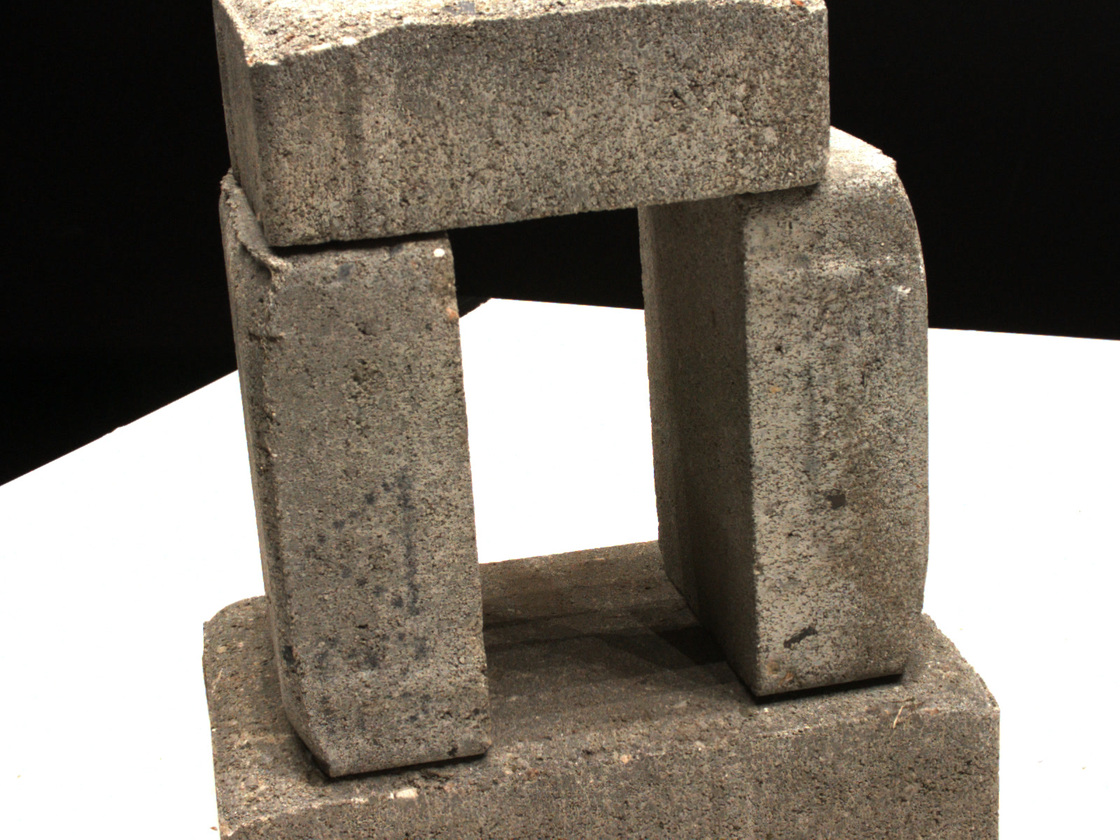}
    \end{subfigure}

    \smallskip
    \rotatebox[origin=b]{90}{scan55}\quad
    \begin{subfigure}[h]{0.17\paperwidth}
        \includegraphics[width=\textwidth]{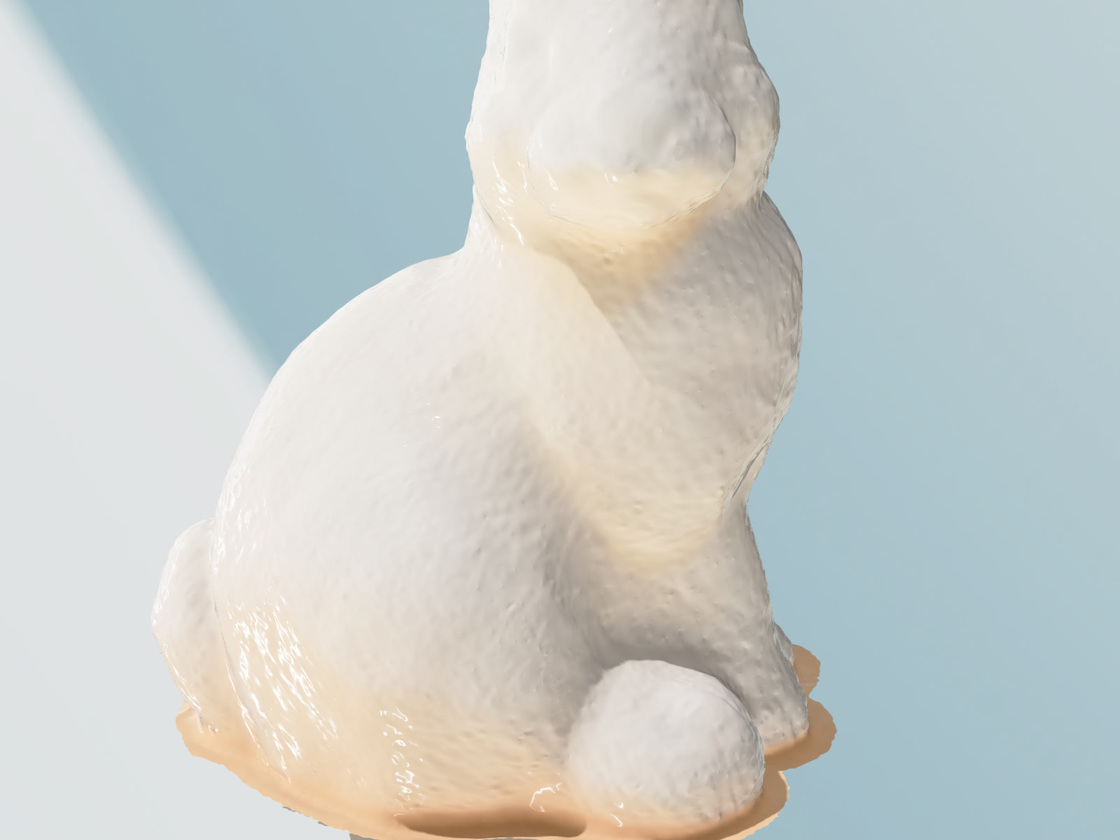}
    \end{subfigure}
    \hspace{1pt}
    \begin{subfigure}[h]{0.17\paperwidth}
        \includegraphics[width=\textwidth]{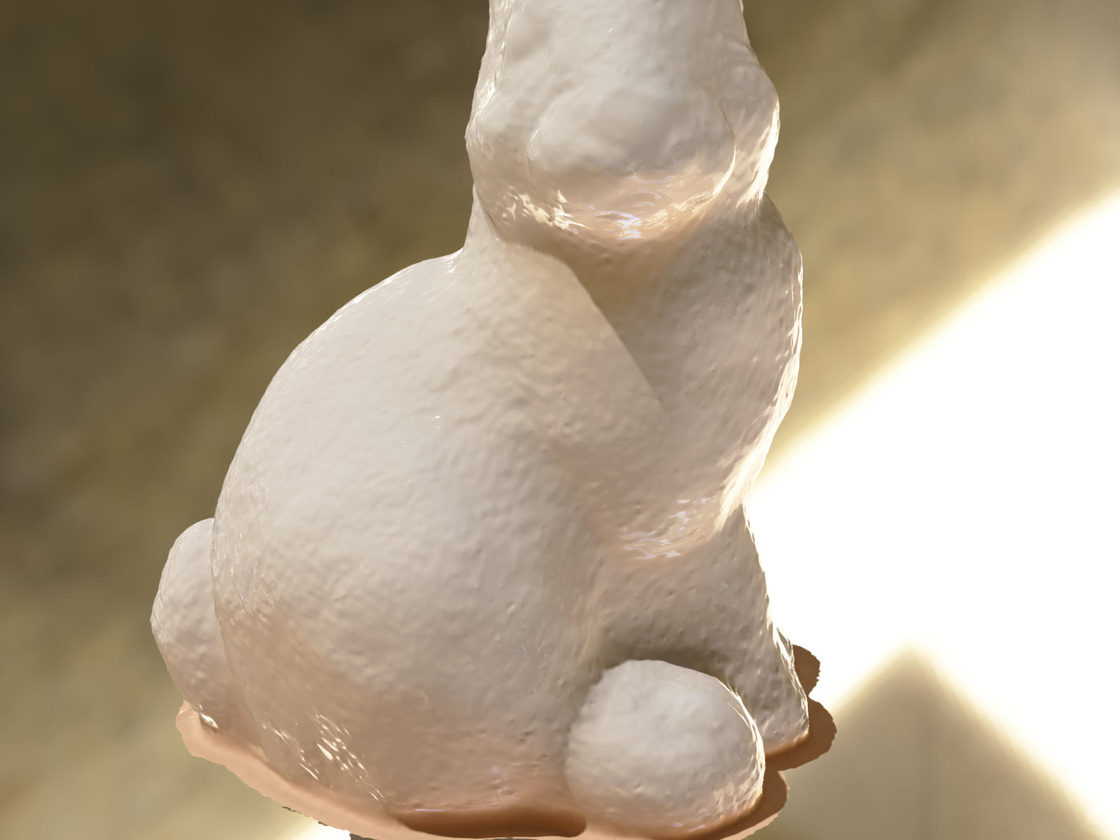}
    \end{subfigure}
    \hspace{1pt}
    \begin{subfigure}[h]{0.17\paperwidth}
        \includegraphics[width=\textwidth]{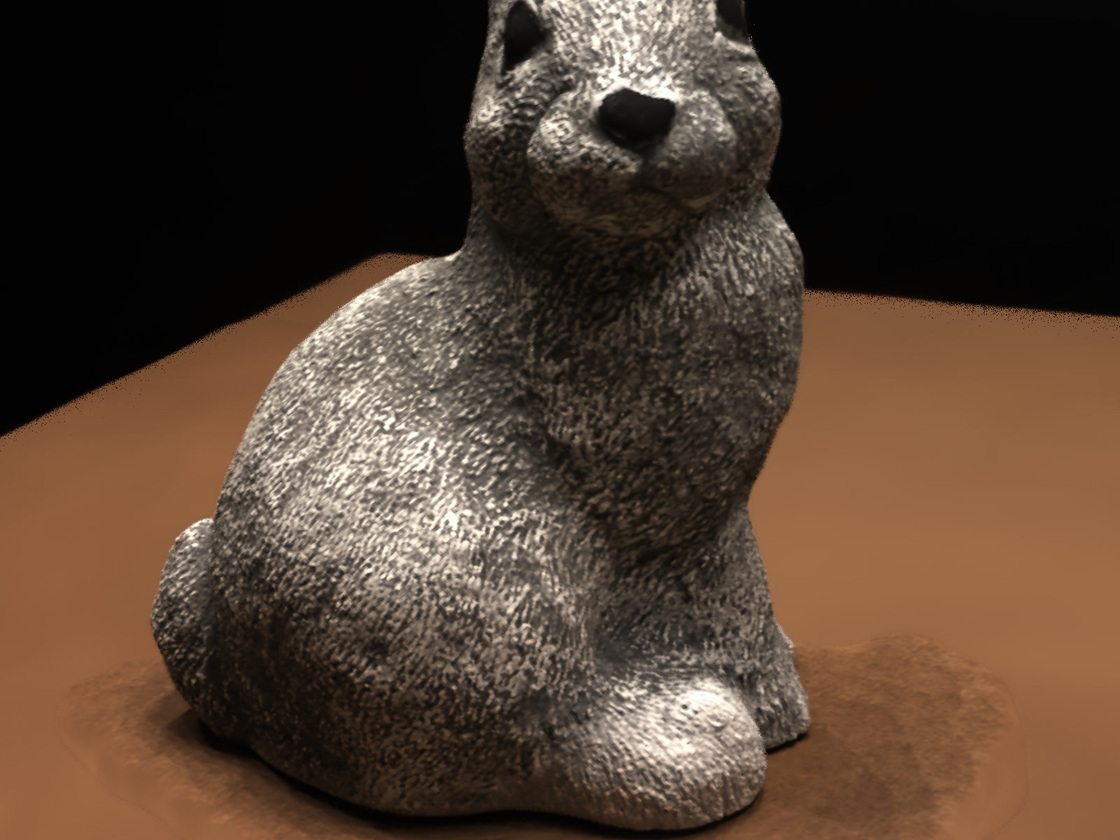}
    \end{subfigure}
    \hspace{1pt}
    \begin{subfigure}[h]{0.17\paperwidth}
        \includegraphics[width=\textwidth]{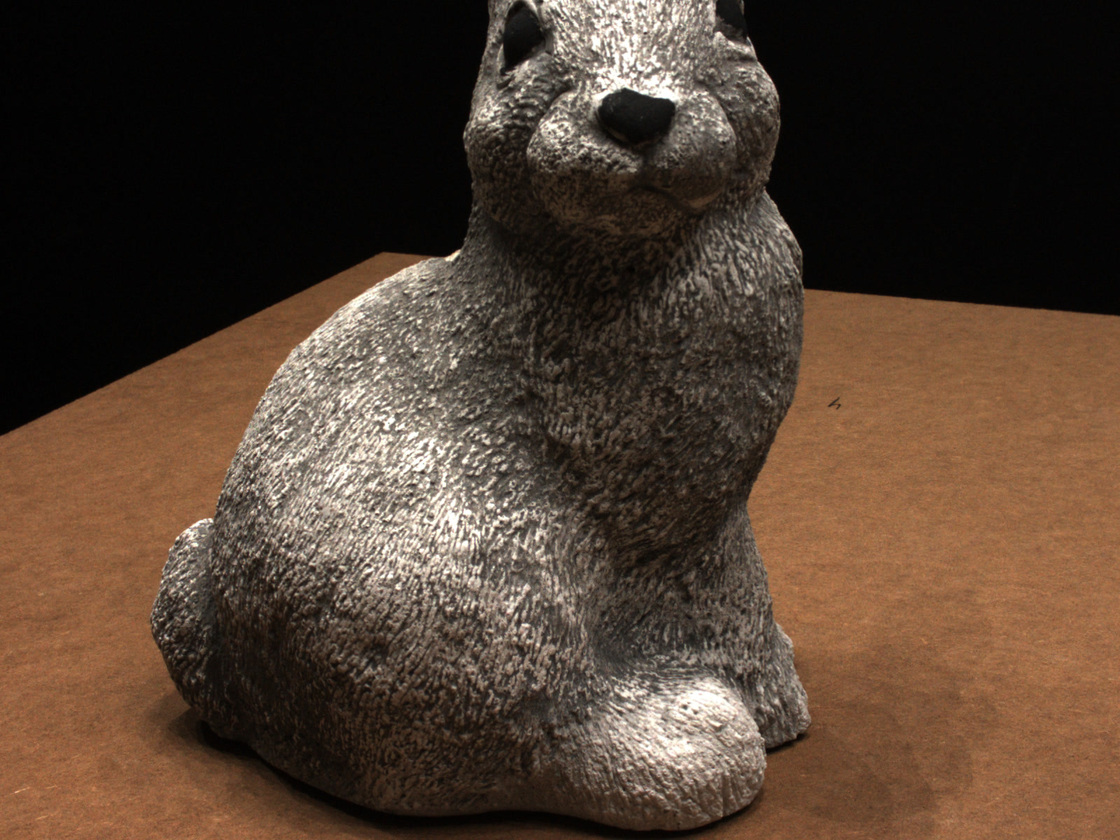}
    \end{subfigure}

    \caption{\textbf{Decomposed materials and rendered images.} ($\cdot$) is a given environment map in Open3D.}
    \label{fig:main_results_01}
\end{figure*}
\begin{figure*}[tbp]
    \centering
    \rotatebox[origin=b]{90}{scan63}\quad
    \begin{subfigure}[h]{0.14\paperwidth}
        \caption{normals}
        \includegraphics[width=\textwidth]{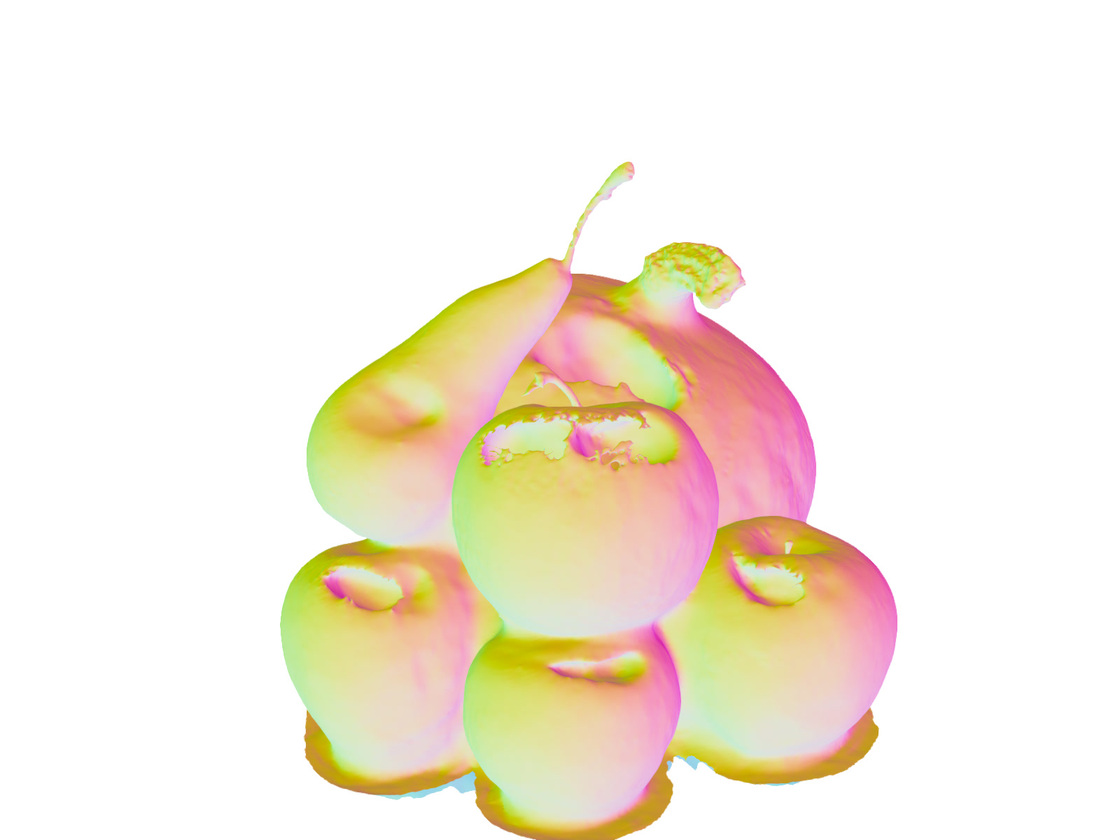}
    \end{subfigure}
    \hspace{1pt}
    \begin{subfigure}[h]{0.14\paperwidth}
        \caption{base color}
        \includegraphics[width=\textwidth]{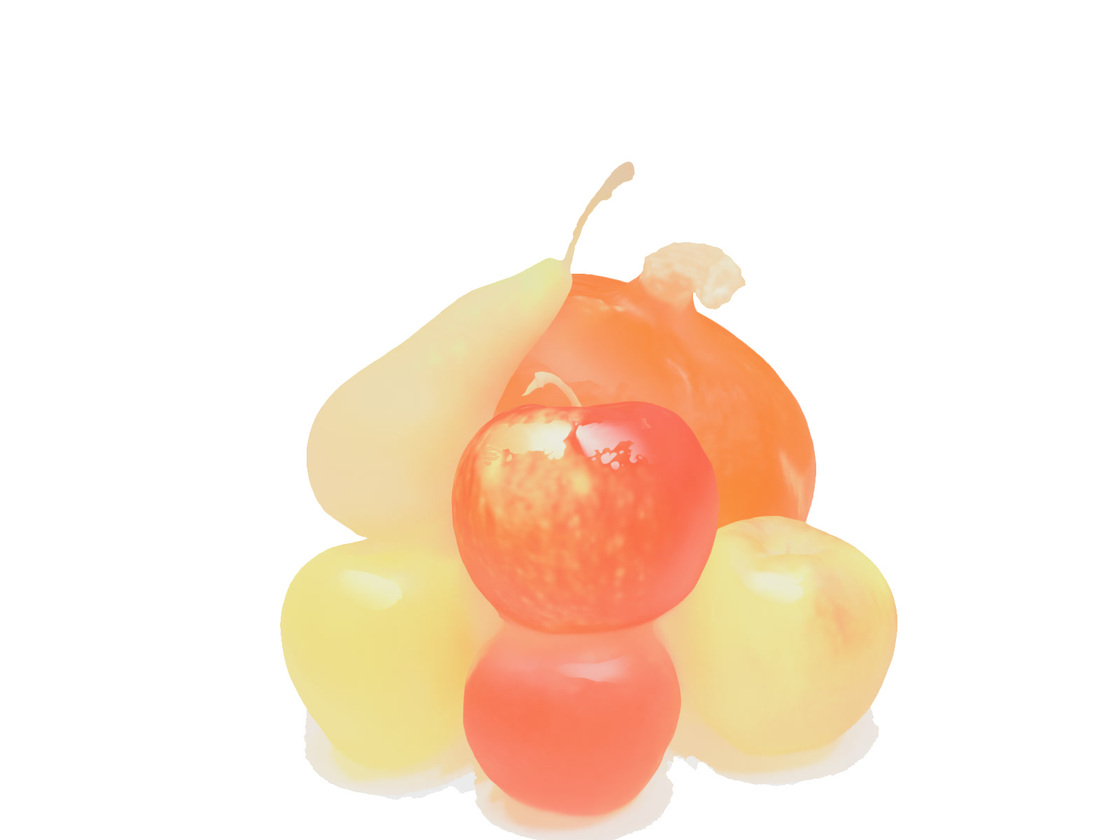}
    \end{subfigure}
    \hspace{1pt}
    \begin{subfigure}[h]{0.14\paperwidth}
        \caption{roughness}
        \includegraphics[width=\textwidth]{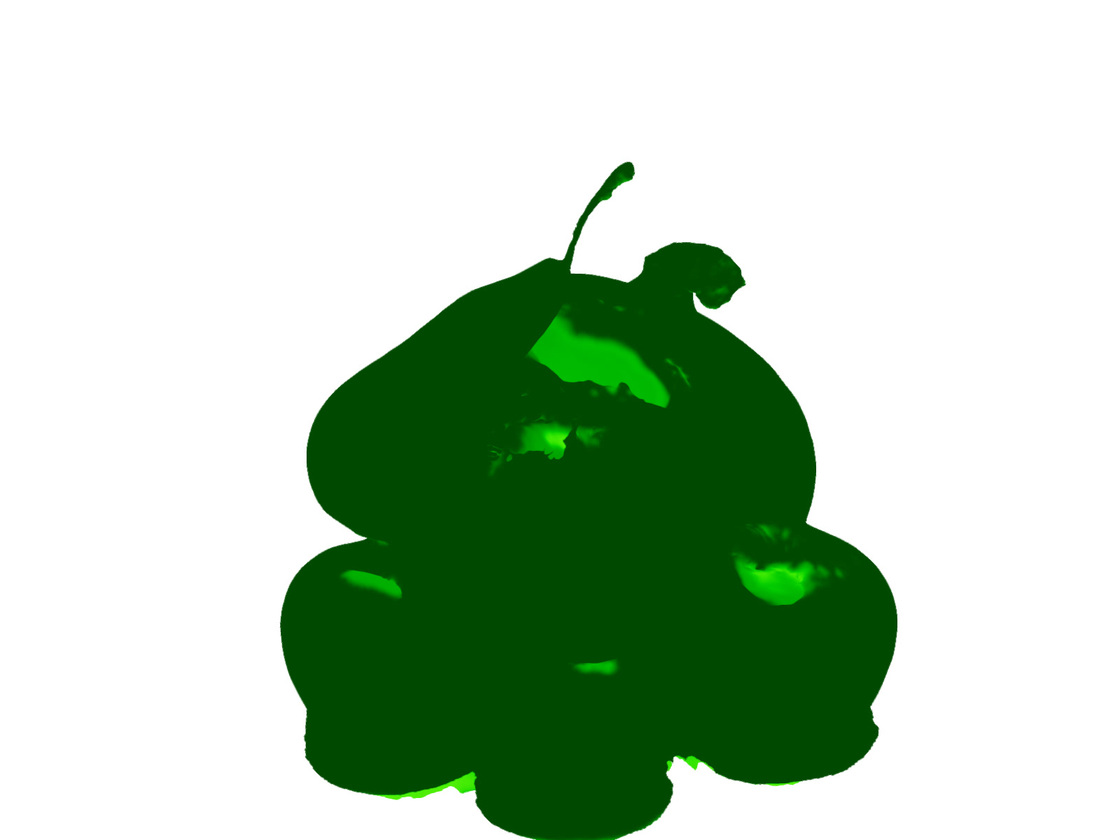}
    \end{subfigure}
    \hspace{1pt}
    \begin{subfigure}[h]{0.14\paperwidth}
        \caption{specular reflectance}
        \includegraphics[width=\textwidth]{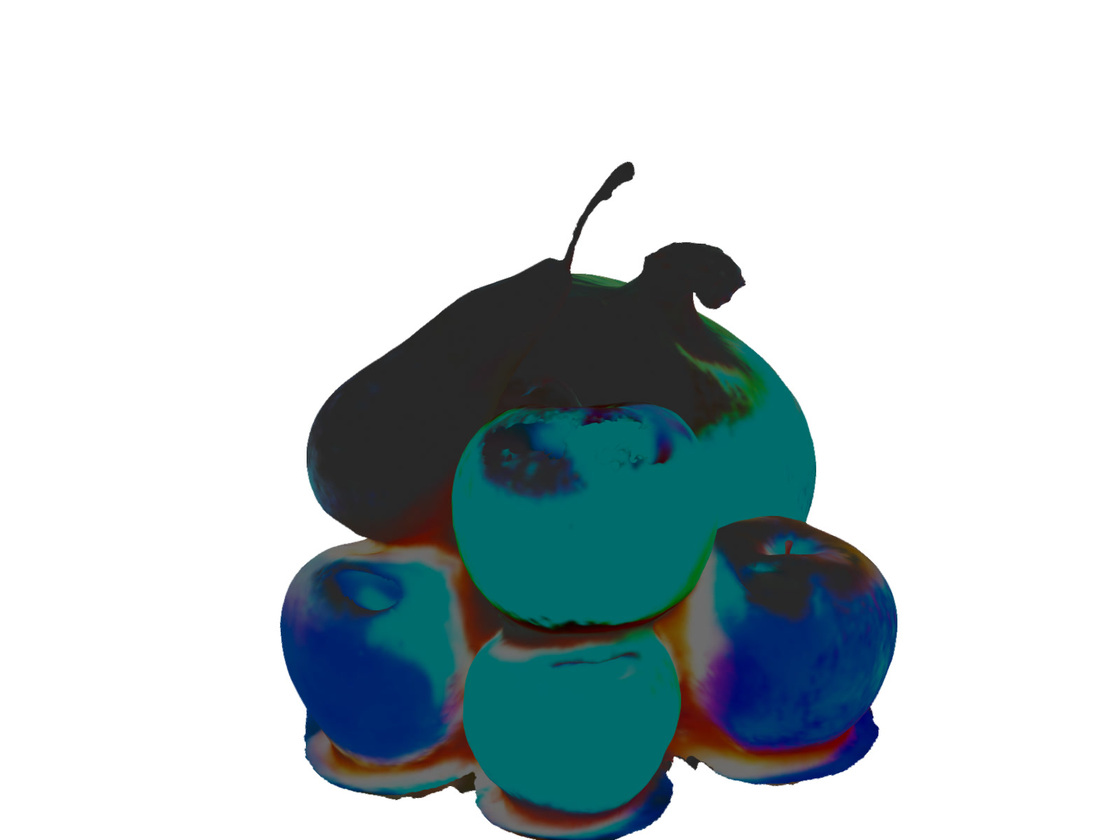}
    \end{subfigure}
    \hspace{1pt}
    \begin{subfigure}[h]{0.14\paperwidth}
        \caption{implicit illumination}
        \includegraphics[width=\textwidth]{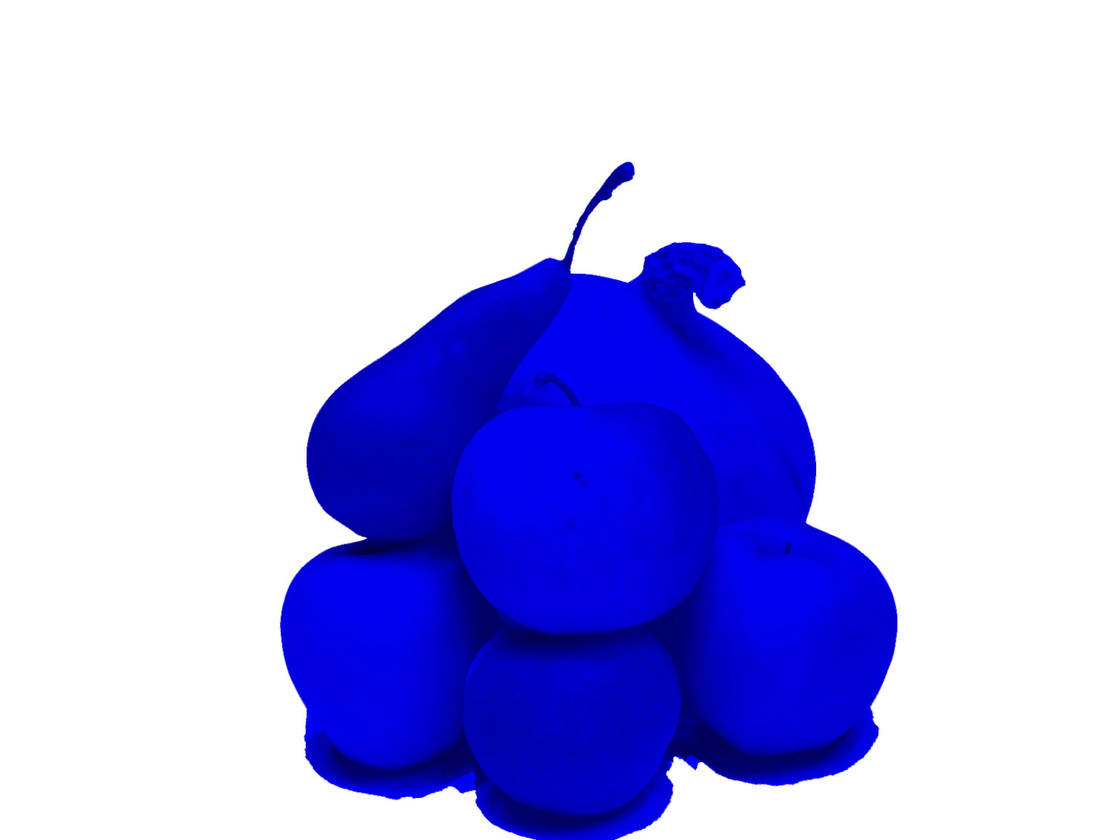}
    \end{subfigure}

    \smallskip
    \rotatebox[origin=b]{90}{scan65}\quad
    \begin{subfigure}[h]{0.14\paperwidth}
        \includegraphics[width=\textwidth]{assets/raw_images/NDJIR/default_epoch1500__groups_gcb50379_dataset_DTU_scan65/model_01499_512grid_trimmed_base_color_mesh00_filtered02_normals_default/32.png.jpg}
    \end{subfigure}
    \hspace{1pt}
    \begin{subfigure}[h]{0.14\paperwidth}
        \includegraphics[width=\textwidth]{assets/raw_images/NDJIR/default_epoch1500__groups_gcb50379_dataset_DTU_scan65/model_01499_512grid_trimmed_base_color_mesh00_filtered02_defaultUnlit_default/32.png.jpg}
    \end{subfigure}
    \hspace{1pt}
    \begin{subfigure}[h]{0.14\paperwidth}
        \includegraphics[width=\textwidth]{assets/raw_images/NDJIR/default_epoch1500__groups_gcb50379_dataset_DTU_scan65/model_01499_512grid_trimmed_roughness_mesh00_filtered02_defaultUnlit_default/32.png.jpg}
    \end{subfigure}
    \hspace{1pt}
    \begin{subfigure}[h]{0.14\paperwidth}
        \includegraphics[width=\textwidth]{assets/raw_images/NDJIR/default_epoch1500__groups_gcb50379_dataset_DTU_scan65/model_01499_512grid_trimmed_specular_reflectance_mesh00_filtered02_defaultUnlit_default/32.png.jpg}
    \end{subfigure}
    \hspace{1pt}
    \begin{subfigure}[h]{0.14\paperwidth}
        \includegraphics[width=\textwidth]{assets/raw_images/NDJIR/default_epoch1500__groups_gcb50379_dataset_DTU_scan65/model_01499_512grid_trimmed_implicit_illumination_mesh00_filtered02_defaultUnlit_default/32.png.jpg}
    \end{subfigure}

    \smallskip
    \rotatebox[origin=b]{90}{scan69}\quad
    \begin{subfigure}[h]{0.14\paperwidth}
        \includegraphics[width=\textwidth]{assets/raw_images/NDJIR/default_epoch1500__groups_gcb50379_dataset_DTU_scan69/model_01499_512grid_trimmed_base_color_mesh00_filtered02_normals_default/32.png.jpg}
    \end{subfigure}
    \hspace{1pt}
    \begin{subfigure}[h]{0.14\paperwidth}
        \includegraphics[width=\textwidth]{assets/raw_images/NDJIR/default_epoch1500__groups_gcb50379_dataset_DTU_scan69/model_01499_512grid_trimmed_base_color_mesh00_filtered02_defaultUnlit_default/32.png.jpg}
    \end{subfigure}
    \hspace{1pt}
    \begin{subfigure}[h]{0.14\paperwidth}
        \includegraphics[width=\textwidth]{assets/raw_images/NDJIR/default_epoch1500__groups_gcb50379_dataset_DTU_scan69/model_01499_512grid_trimmed_roughness_mesh00_filtered02_defaultUnlit_default/32.png.jpg}
    \end{subfigure}
    \hspace{1pt}
    \begin{subfigure}[h]{0.14\paperwidth}
        \includegraphics[width=\textwidth]{assets/raw_images/NDJIR/default_epoch1500__groups_gcb50379_dataset_DTU_scan69/model_01499_512grid_trimmed_specular_reflectance_mesh00_filtered02_defaultUnlit_default/32.png.jpg}
    \end{subfigure}
    \hspace{1pt}
    \begin{subfigure}[h]{0.14\paperwidth}
        \includegraphics[width=\textwidth]{assets/raw_images/NDJIR/default_epoch1500__groups_gcb50379_dataset_DTU_scan69/model_01499_512grid_trimmed_implicit_illumination_mesh00_filtered02_defaultUnlit_default/32.png.jpg}
    \end{subfigure}

    \smallskip
    \rotatebox[origin=b]{90}{scan83}\quad
    \begin{subfigure}[h]{0.14\paperwidth}
        \includegraphics[width=\textwidth]{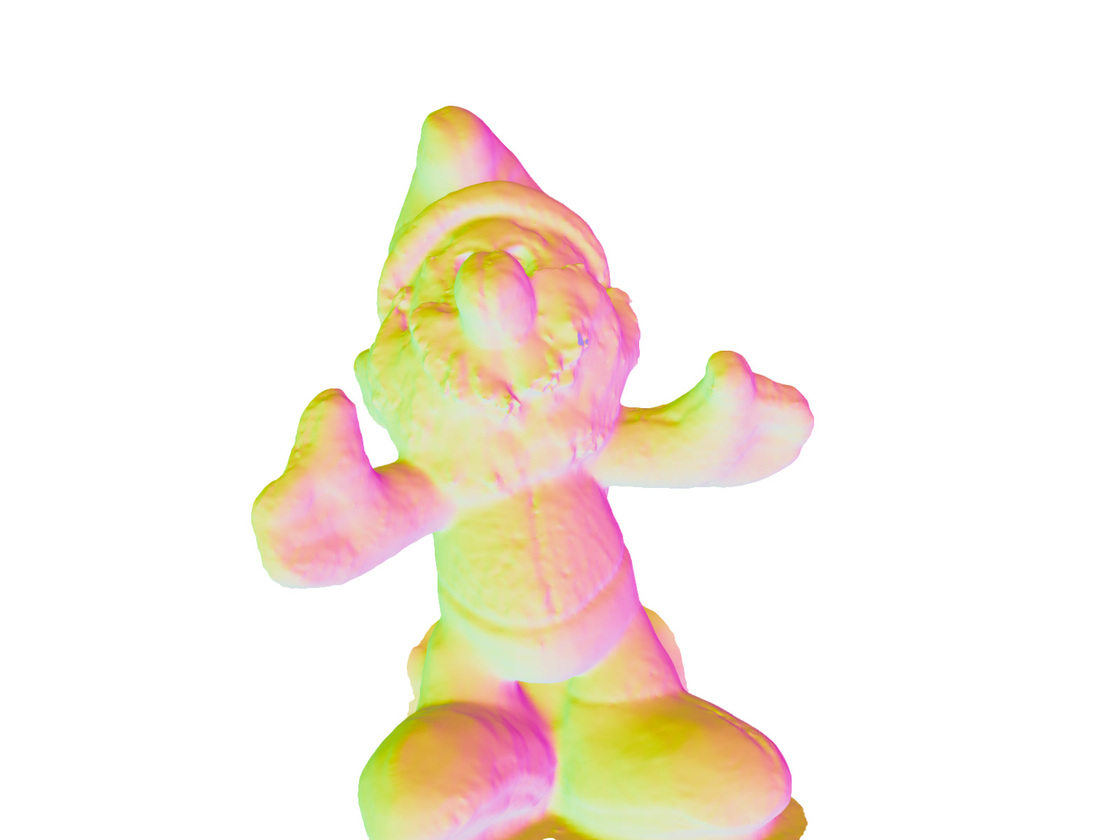}
    \end{subfigure}
    \hspace{1pt}
    \begin{subfigure}[h]{0.14\paperwidth}
        \includegraphics[width=\textwidth]{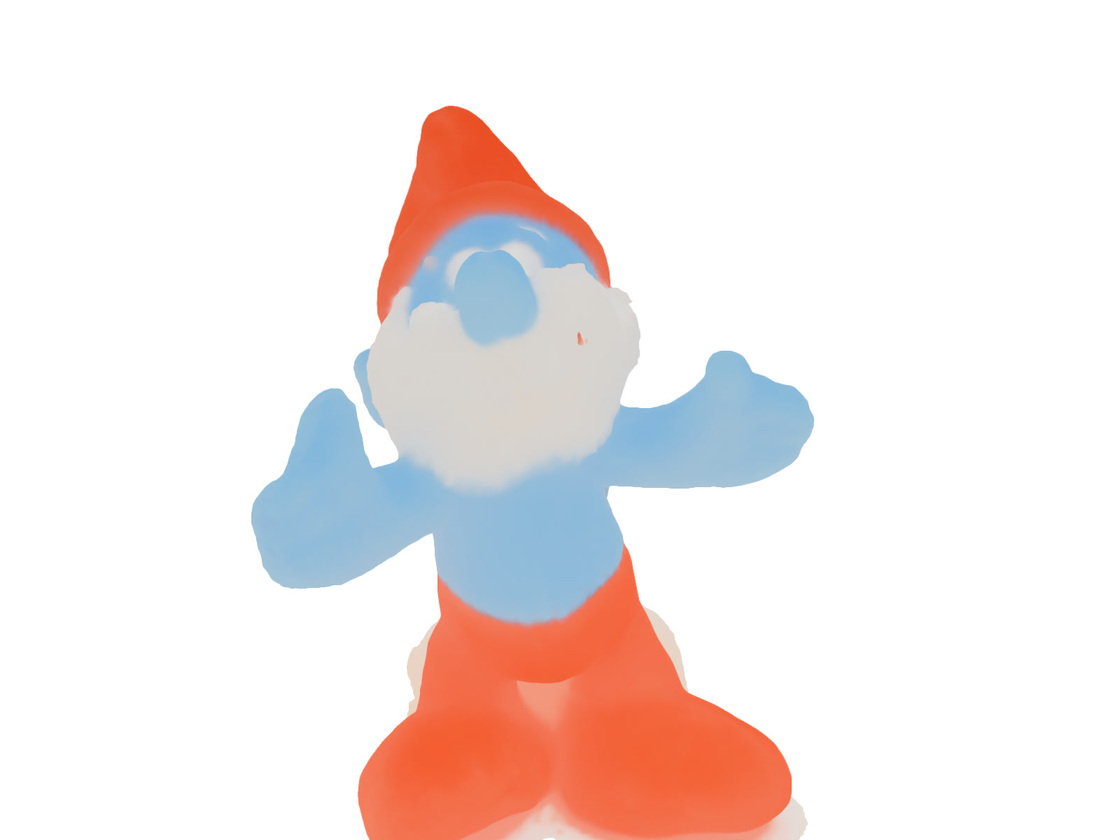}
    \end{subfigure}
    \hspace{1pt}
    \begin{subfigure}[h]{0.14\paperwidth}
        \includegraphics[width=\textwidth]{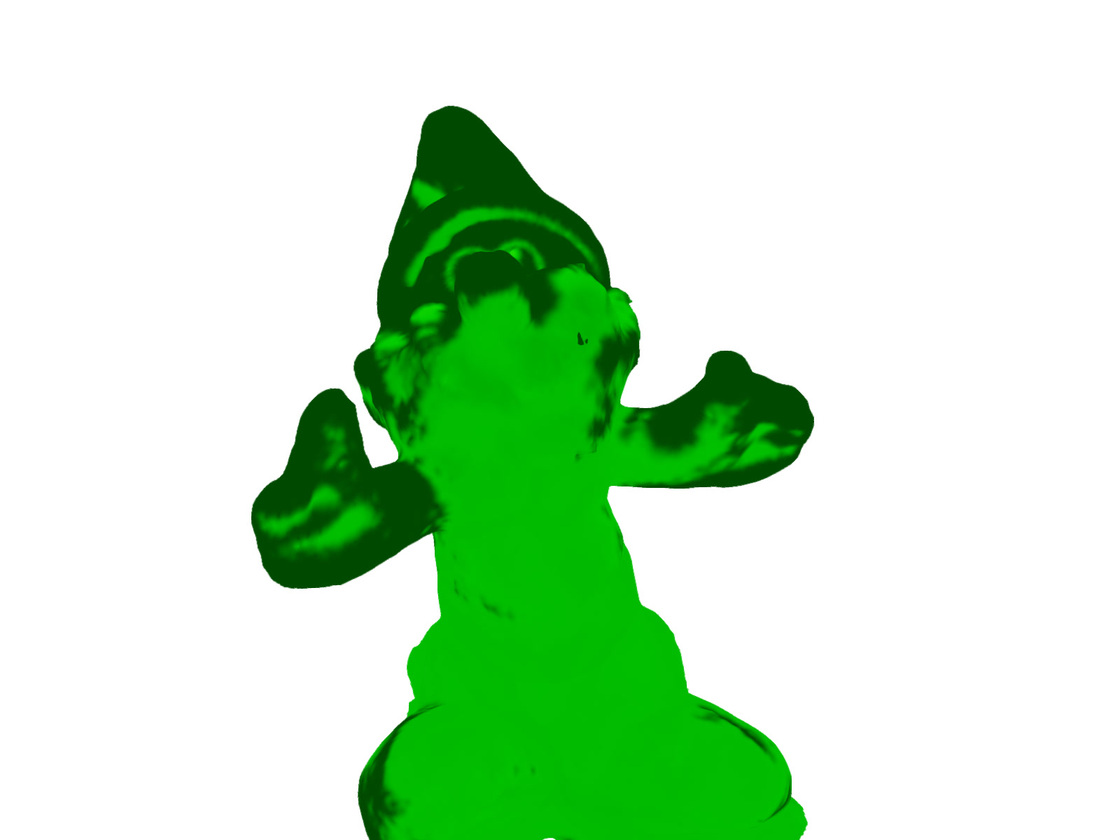}
    \end{subfigure}
    \hspace{1pt}
    \begin{subfigure}[h]{0.14\paperwidth}
        \includegraphics[width=\textwidth]{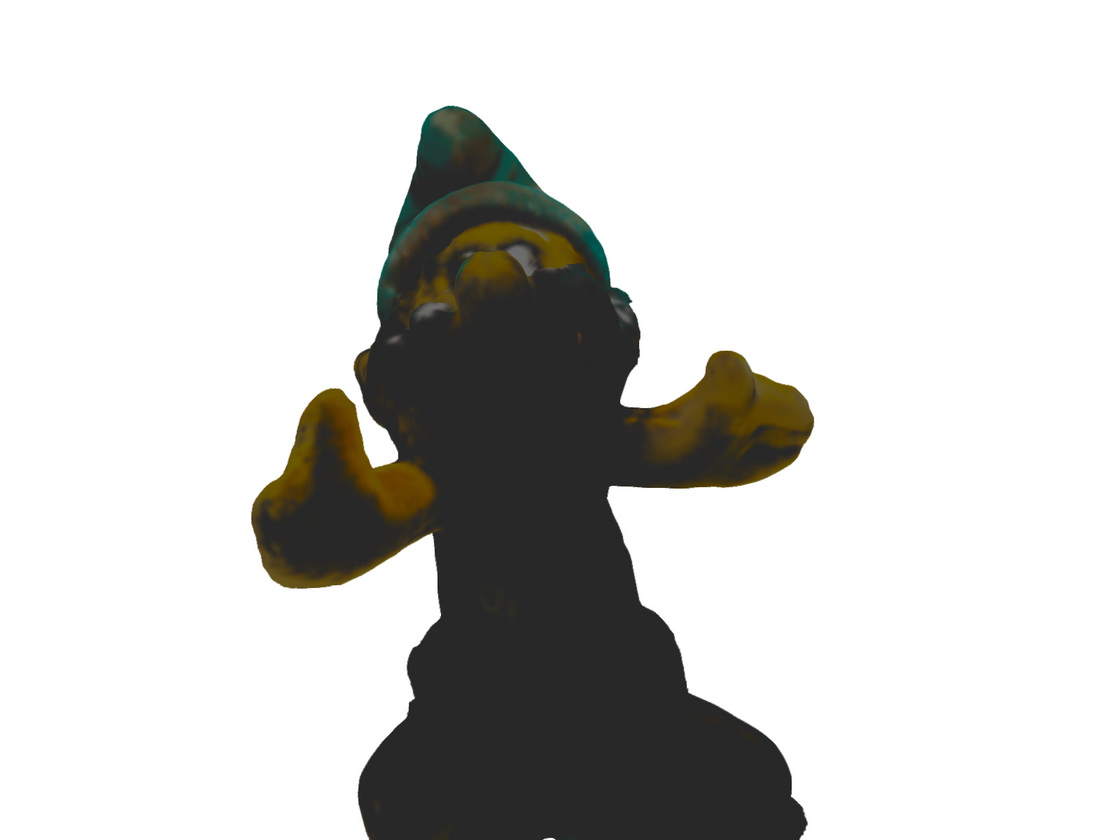}
    \end{subfigure}
    \hspace{1pt}
    \begin{subfigure}[h]{0.14\paperwidth}
        \includegraphics[width=\textwidth]{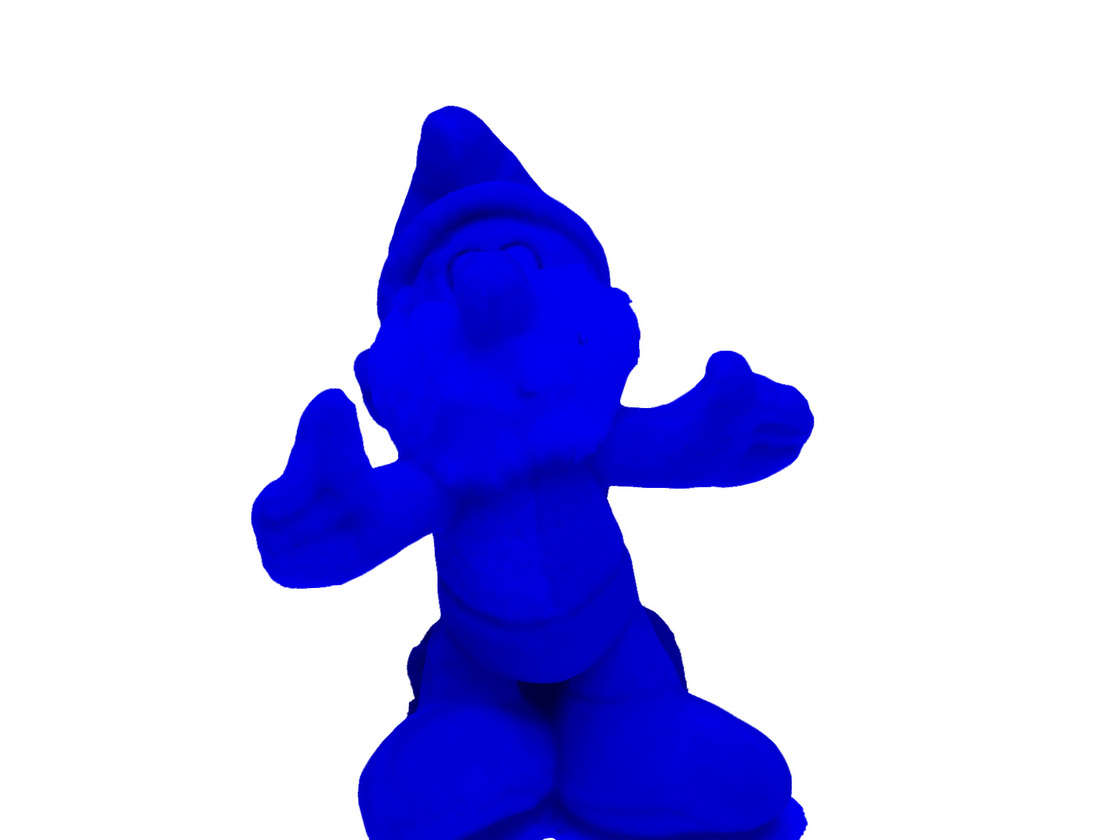}
    \end{subfigure}

    \smallskip
    \rotatebox[origin=b]{90}{scan63}\quad
    \begin{subfigure}[h]{0.17\paperwidth}
        \caption{PBR (default)}
        \includegraphics[width=\textwidth]{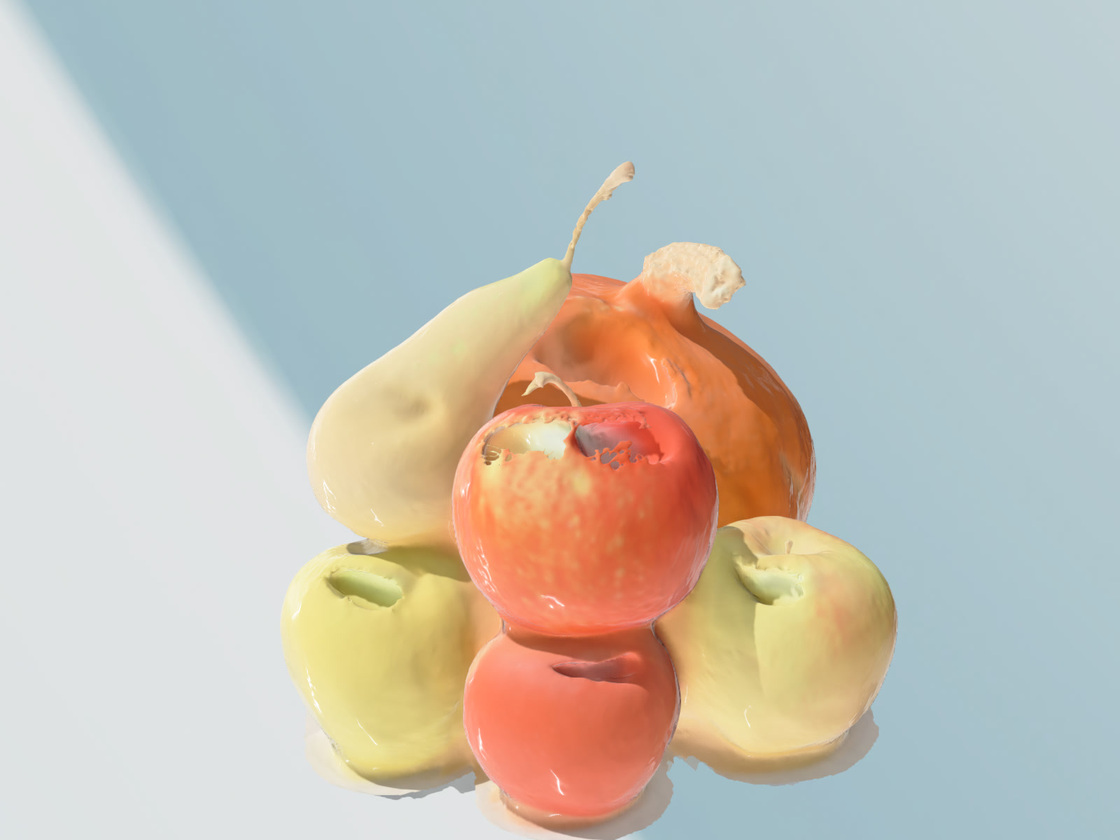}
    \end{subfigure}
    \hspace{1pt}
    \begin{subfigure}[h]{0.17\paperwidth}
        \caption{PBR (pillars)}
        \includegraphics[width=\textwidth]{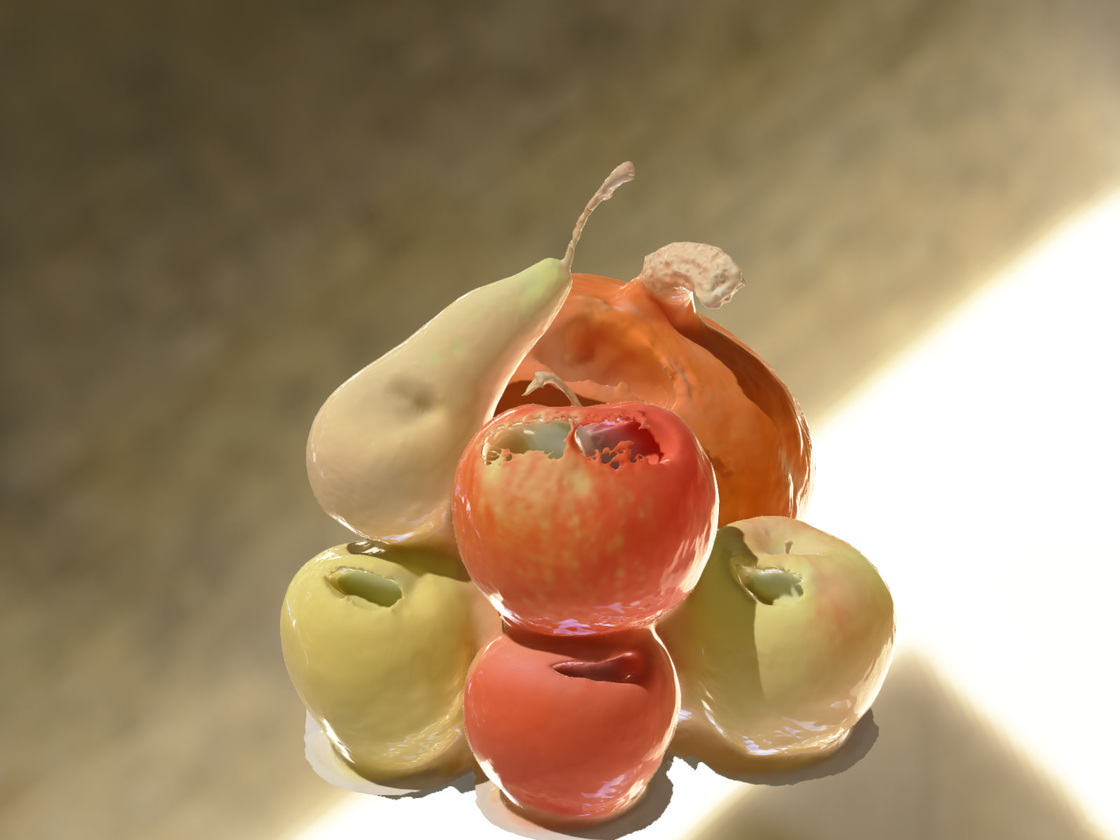}
    \end{subfigure}
    \hspace{1pt}
    \begin{subfigure}[h]{0.17\paperwidth}
        \caption{Neural rendering}
        \includegraphics[width=\textwidth]{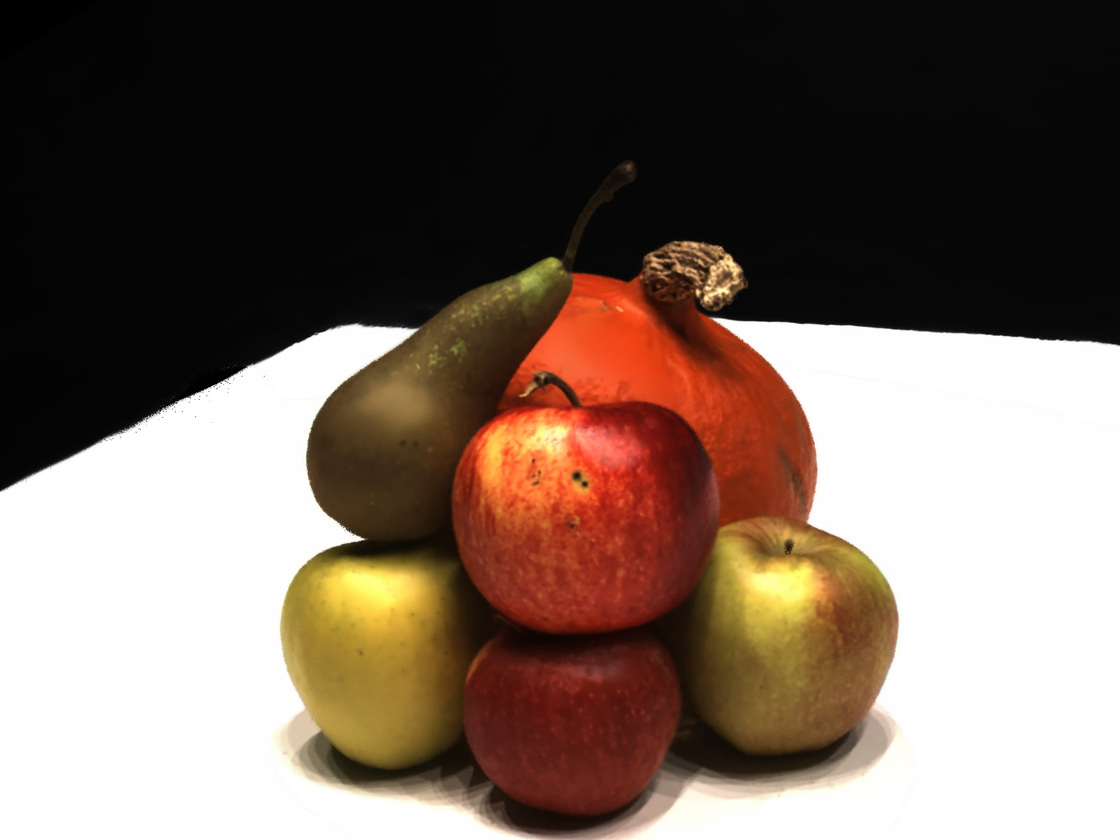}
    \end{subfigure}
    \hspace{1pt}
    \begin{subfigure}[h]{0.17\paperwidth}
        \caption{Groundtruth}
        \includegraphics[width=\textwidth]{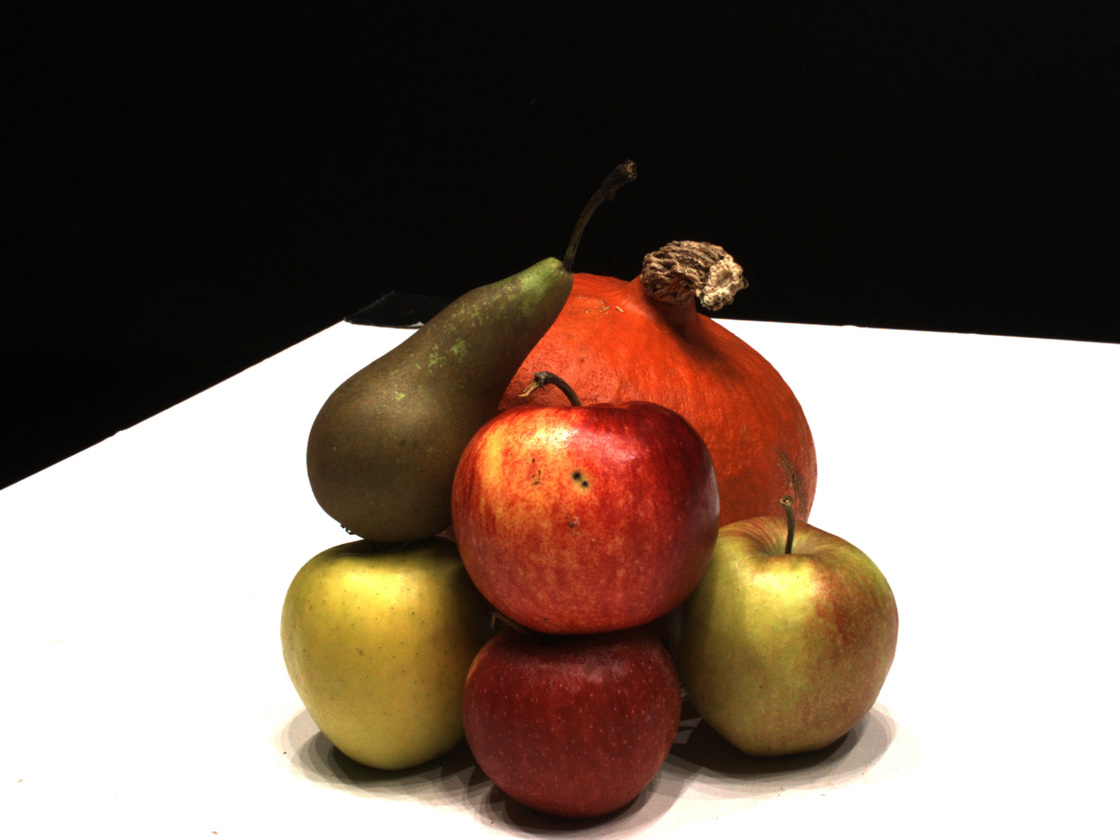}
    \end{subfigure}

    \smallskip
    \rotatebox[origin=b]{90}{scan65}\quad
    \begin{subfigure}[h]{0.17\paperwidth}
        \includegraphics[width=\textwidth]{assets/raw_images/NDJIR/default_epoch1500__groups_gcb50379_dataset_DTU_scan65/model_01499_512grid_trimmed_base_color_mesh00_filtered02_defaultLit_default_32.png.jpg}
    \end{subfigure}
    \hspace{1pt}
    \begin{subfigure}[h]{0.17\paperwidth}
        \includegraphics[width=\textwidth]{assets/raw_images/NDJIR/default_epoch1500__groups_gcb50379_dataset_DTU_scan65/model_01499_512grid_trimmed_base_color_mesh00_filtered02_defaultLit_pillars_32.png.jpg}
    \end{subfigure}
    \hspace{1pt}
    \begin{subfigure}[h]{0.17\paperwidth}
        \includegraphics[width=\textwidth]{assets/raw_images/NDJIR/default_epoch1500__groups_gcb50379_dataset_DTU_scan65/Eval-rendered-image-1600x1200/000032-000.png.jpg}
    \end{subfigure}
    \hspace{1pt}
    \begin{subfigure}[h]{0.17\paperwidth}
        \includegraphics[width=\textwidth]{assets/DTU/scan65/image/000032.png.jpg}
    \end{subfigure}

    \smallskip
    \rotatebox[origin=b]{90}{scan69}\quad
    \begin{subfigure}[h]{0.17\paperwidth}
        \includegraphics[width=\textwidth]{assets/raw_images/NDJIR/default_epoch1500__groups_gcb50379_dataset_DTU_scan69/model_01499_512grid_trimmed_base_color_mesh00_filtered02_defaultLit_default_32.png.jpg}
    \end{subfigure}
    \hspace{1pt}
    \begin{subfigure}[h]{0.17\paperwidth}
        \includegraphics[width=\textwidth]{assets/raw_images/NDJIR/default_epoch1500__groups_gcb50379_dataset_DTU_scan69/model_01499_512grid_trimmed_base_color_mesh00_filtered02_defaultLit_pillars_32.png.jpg}
    \end{subfigure}
    \hspace{1pt}
    \begin{subfigure}[h]{0.17\paperwidth}
        \includegraphics[width=\textwidth]{assets/raw_images/NDJIR/default_epoch1500__groups_gcb50379_dataset_DTU_scan69/Eval-rendered-image-1600x1200/000032-000.png.jpg}
    \end{subfigure}
    \hspace{1pt}
    \begin{subfigure}[h]{0.17\paperwidth}
        \includegraphics[width=\textwidth]{assets/DTU/scan69/image/000032.png.jpg}
    \end{subfigure}

    \smallskip
    \rotatebox[origin=b]{90}{scan83}\quad
    \begin{subfigure}[h]{0.17\paperwidth}
        \includegraphics[width=\textwidth]{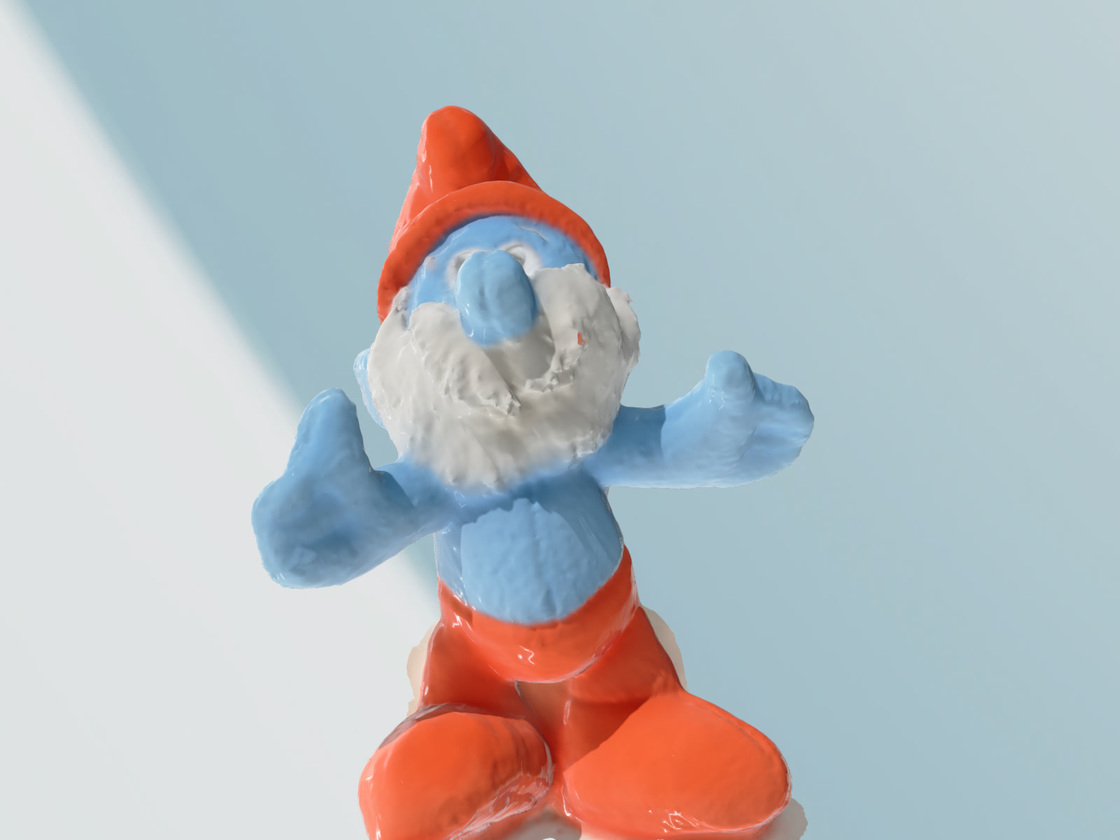}
    \end{subfigure}
    \hspace{1pt}
    \begin{subfigure}[h]{0.17\paperwidth}
        \includegraphics[width=\textwidth]{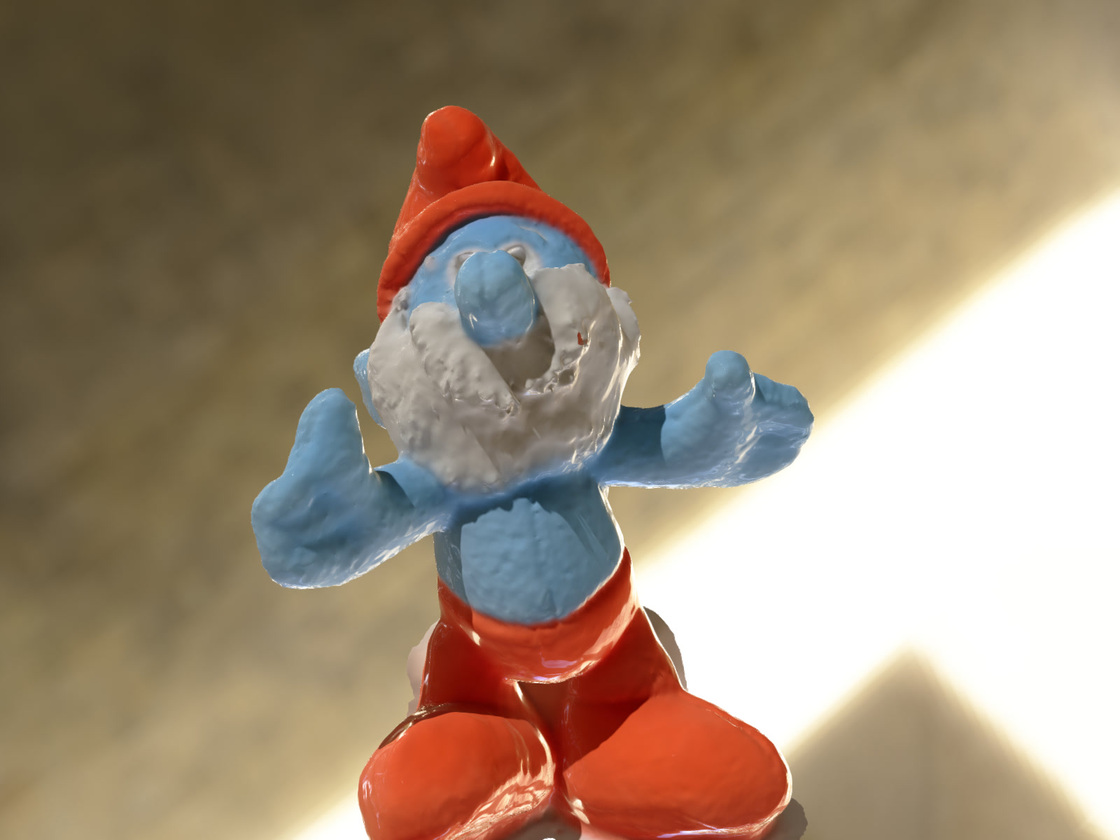}
    \end{subfigure}
    \hspace{1pt}
    \begin{subfigure}[h]{0.17\paperwidth}
        \includegraphics[width=\textwidth]{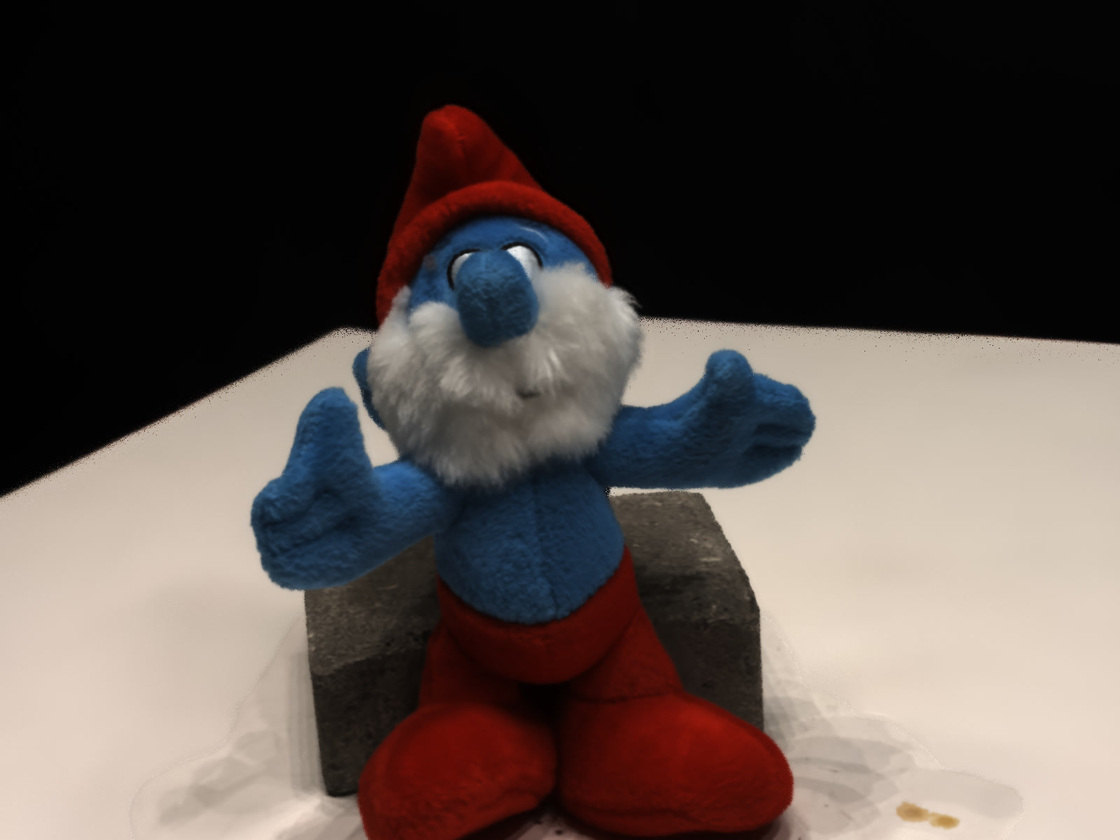}
    \end{subfigure}
    \hspace{1pt}
    \begin{subfigure}[h]{0.17\paperwidth}
        \includegraphics[width=\textwidth]{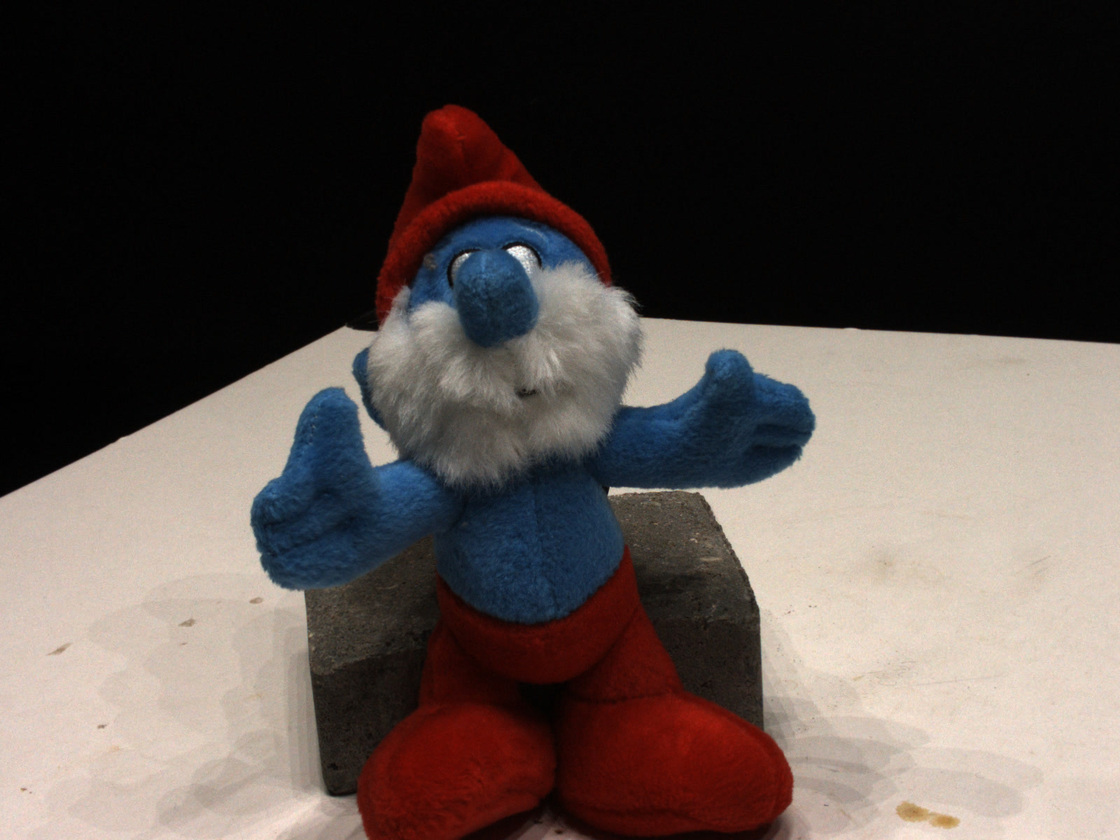}
    \end{subfigure}

    \caption{\textbf{Decomposed materials and rendered images.} ($\cdot$) is a given environment map in Open3D.}
    \label{fig:main_results_02}
\end{figure*}
\begin{figure*}[tbp]
    \centering
    \rotatebox[origin=b]{90}{scan97}\quad
    \begin{subfigure}[h]{0.14\paperwidth}
        \caption{normals}
        \includegraphics[width=\textwidth]{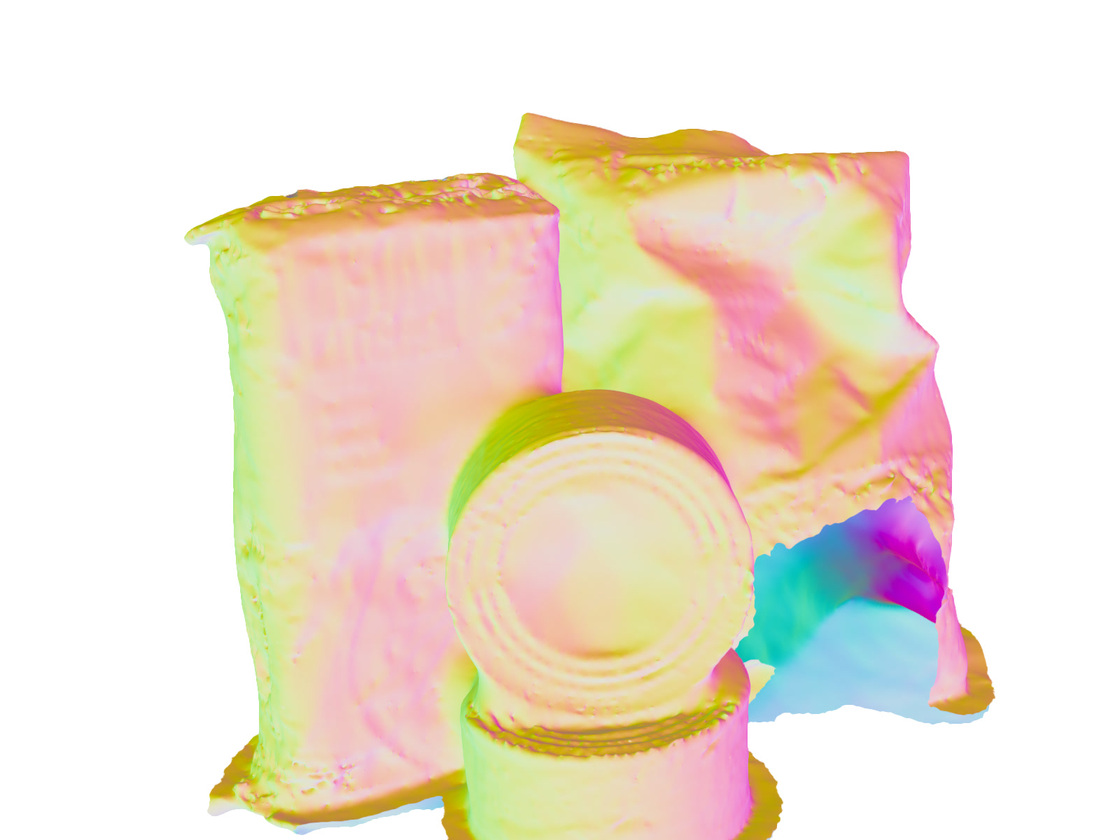}
    \end{subfigure}
    \hspace{1pt}
    \begin{subfigure}[h]{0.14\paperwidth}
        \caption{base color}
        \includegraphics[width=\textwidth]{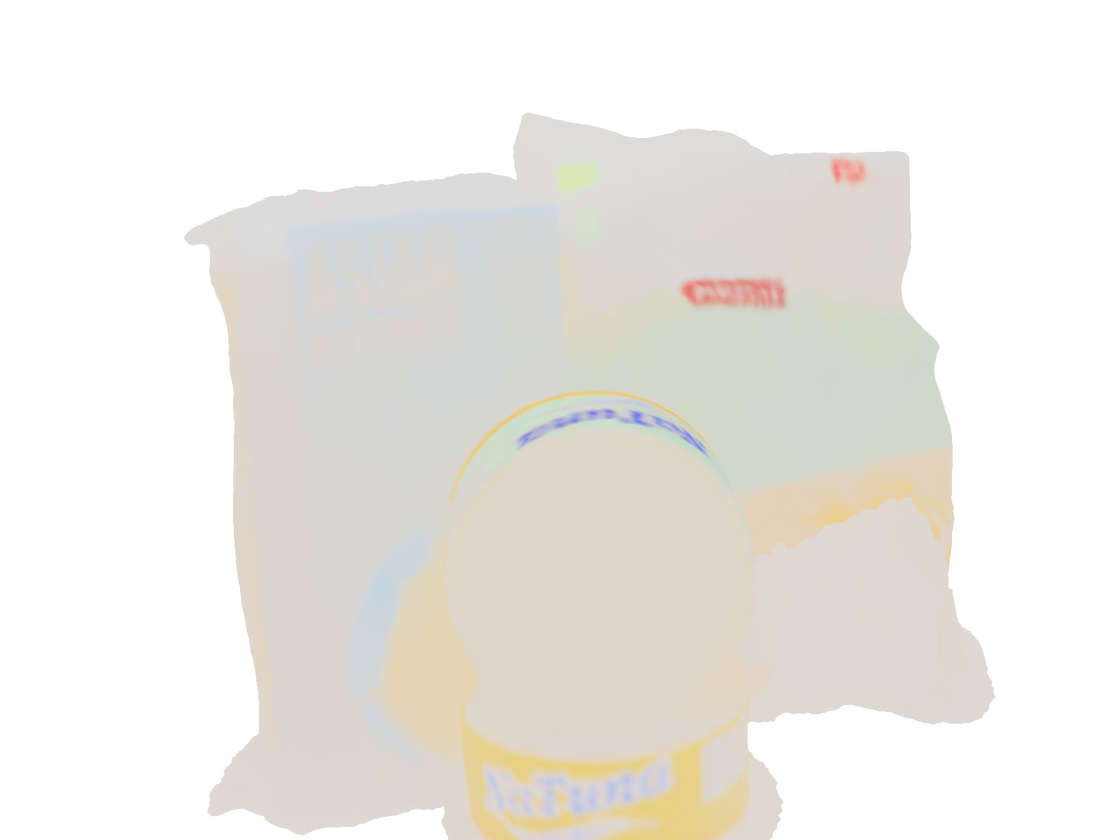}
    \end{subfigure}
    \hspace{1pt}
    \begin{subfigure}[h]{0.14\paperwidth}
        \caption{roughness}
        \includegraphics[width=\textwidth]{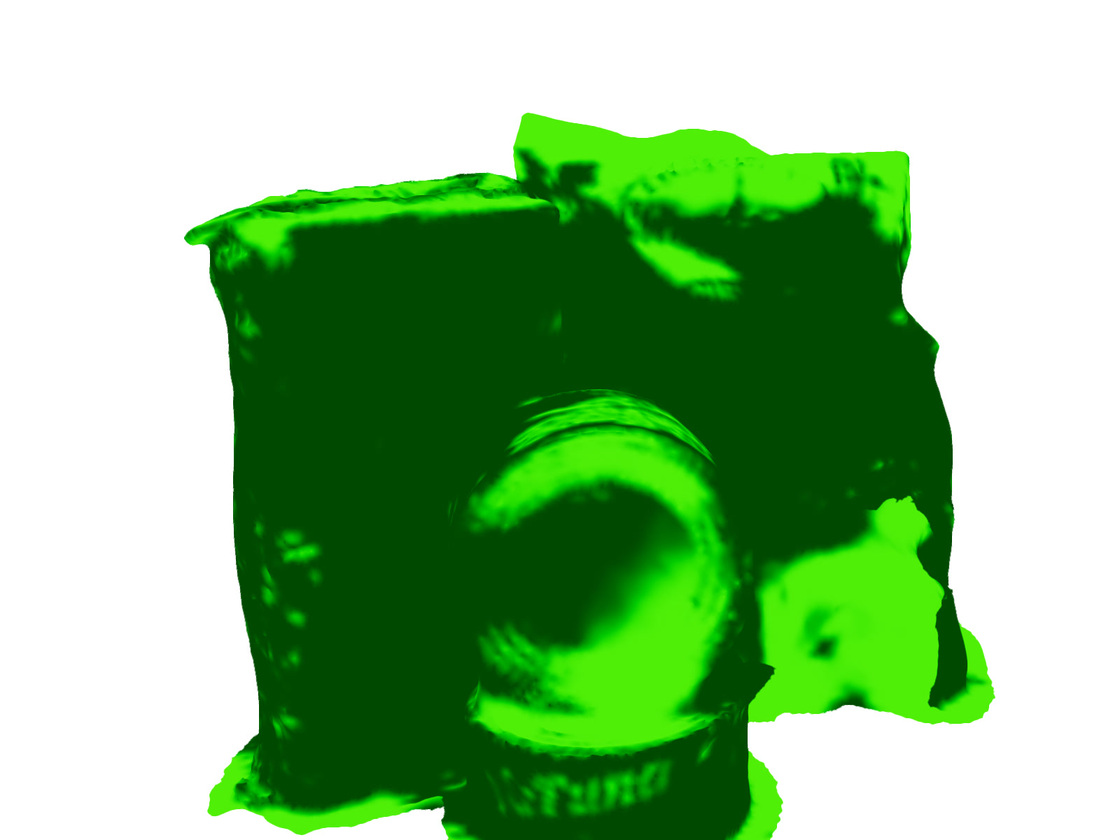}
    \end{subfigure}
    \hspace{1pt}
    \begin{subfigure}[h]{0.14\paperwidth}
        \caption{specular reflectance}
        \includegraphics[width=\textwidth]{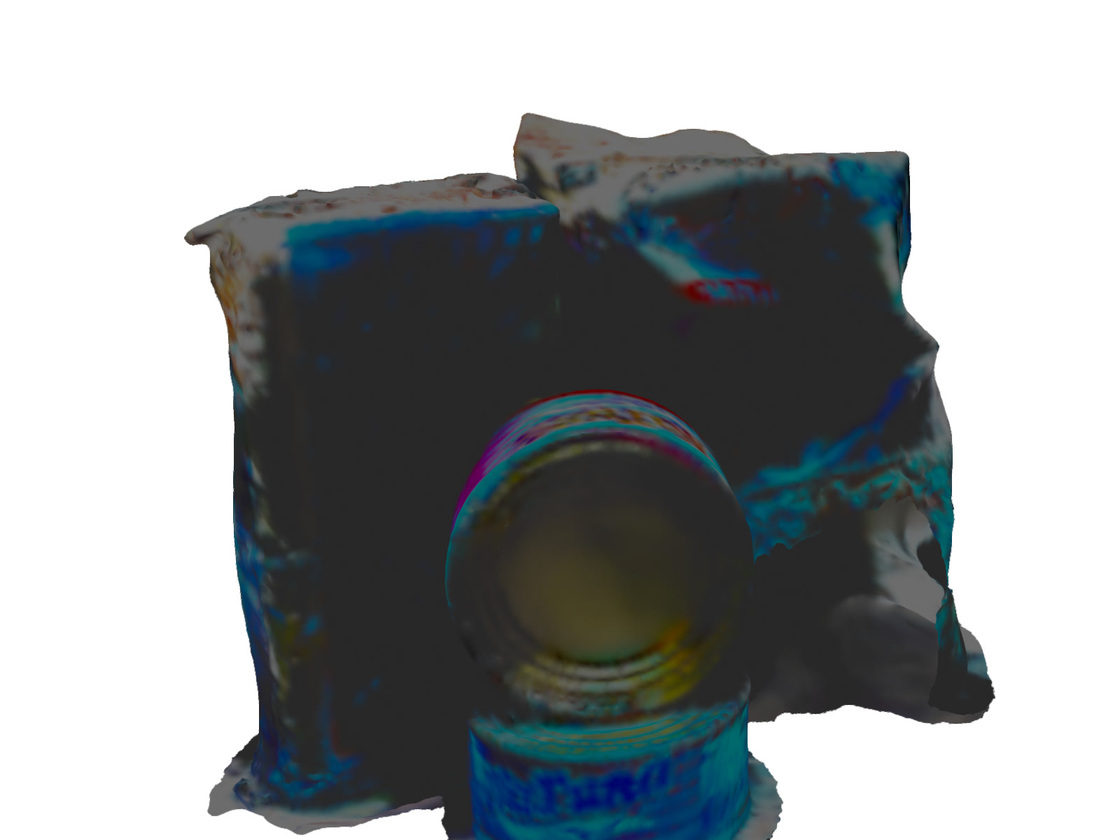}
    \end{subfigure}
    \hspace{1pt}
    \begin{subfigure}[h]{0.14\paperwidth}
        \caption{implicit illumination}
        \includegraphics[width=\textwidth]{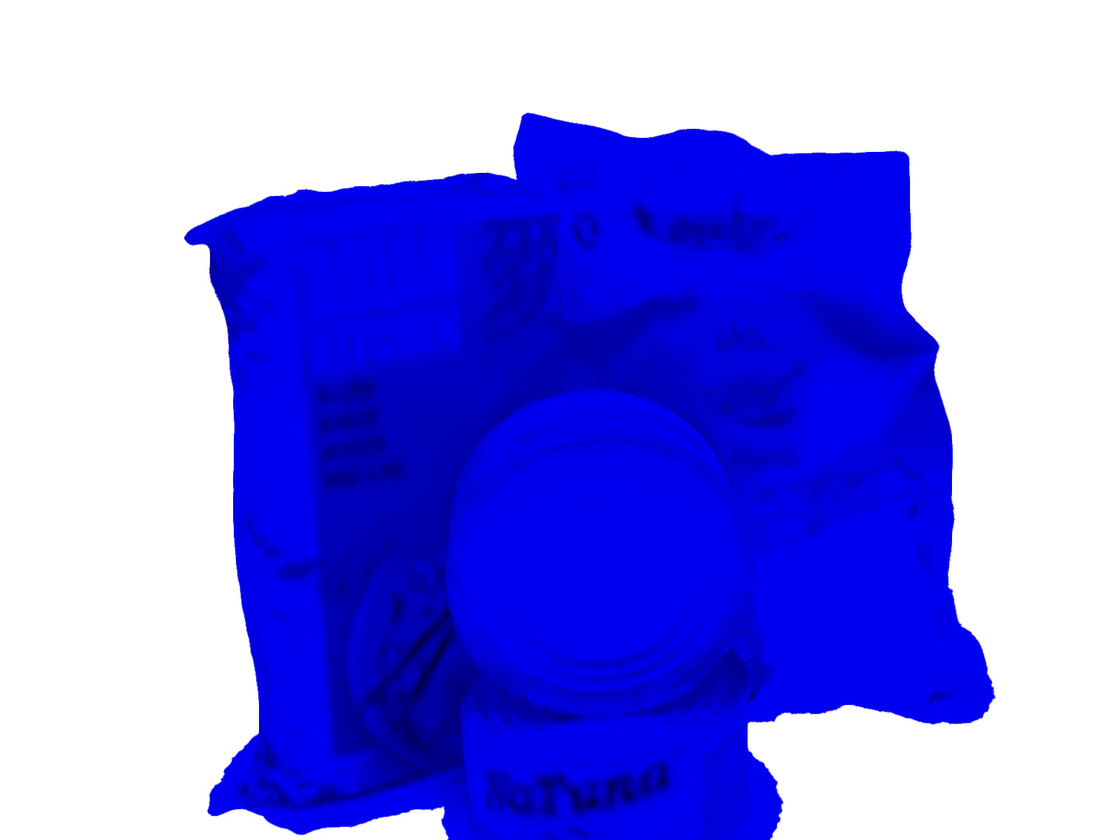}
    \end{subfigure}

    \smallskip
    \rotatebox[origin=b]{90}{scan105}\quad
    \begin{subfigure}[h]{0.14\paperwidth}
        \includegraphics[width=\textwidth]{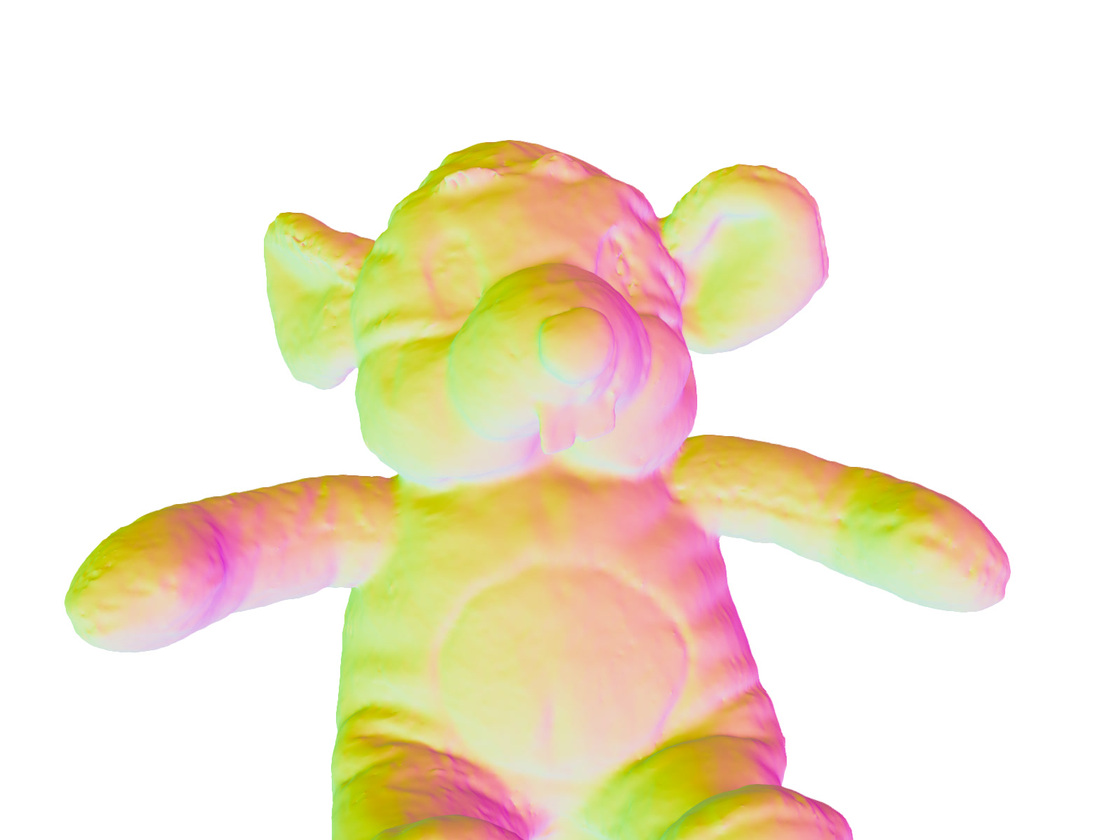}
    \end{subfigure}
    \hspace{1pt}
    \begin{subfigure}[h]{0.14\paperwidth}
        \includegraphics[width=\textwidth]{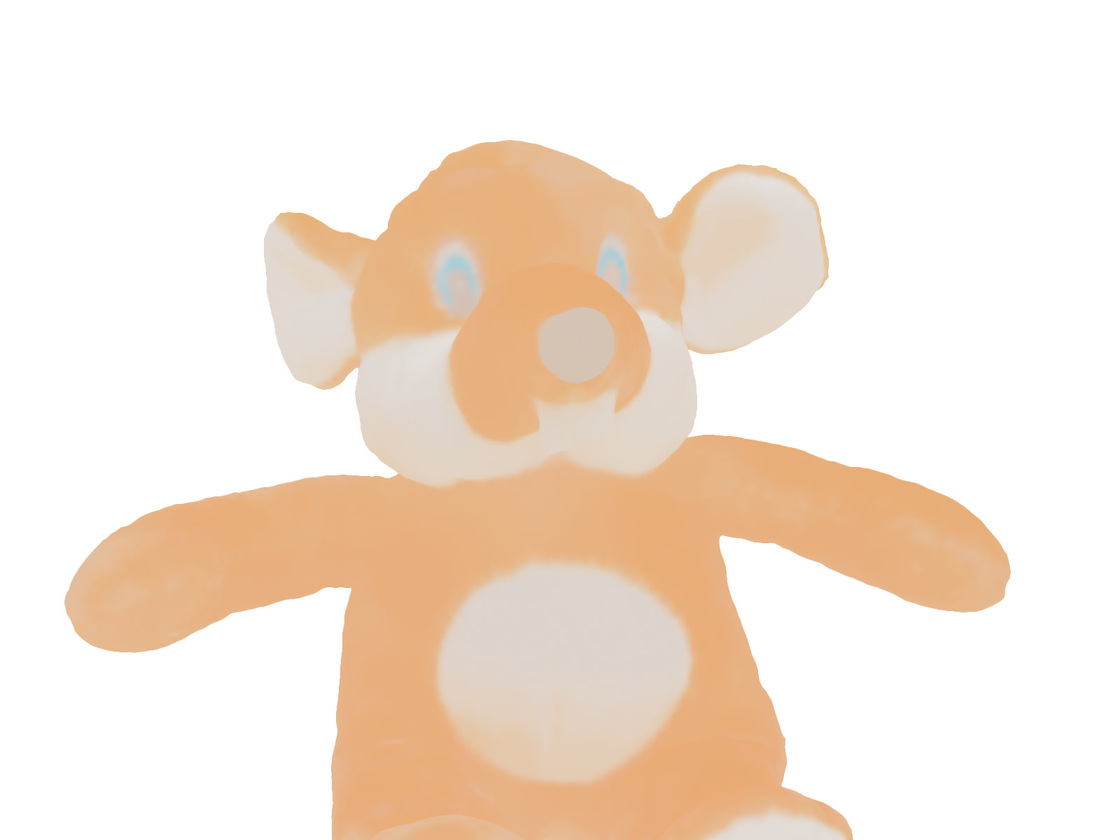}
    \end{subfigure}
    \hspace{1pt}
    \begin{subfigure}[h]{0.14\paperwidth}
        \includegraphics[width=\textwidth]{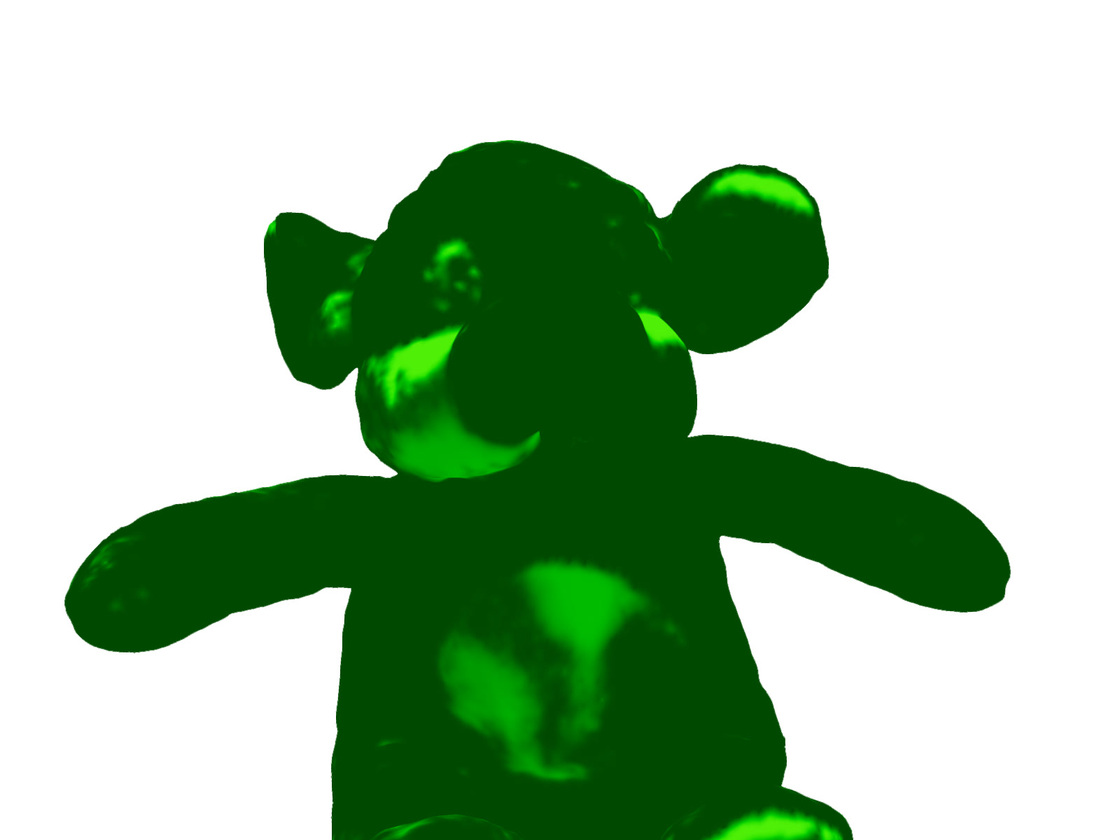}
    \end{subfigure}
    \hspace{1pt}
    \begin{subfigure}[h]{0.14\paperwidth}
        \includegraphics[width=\textwidth]{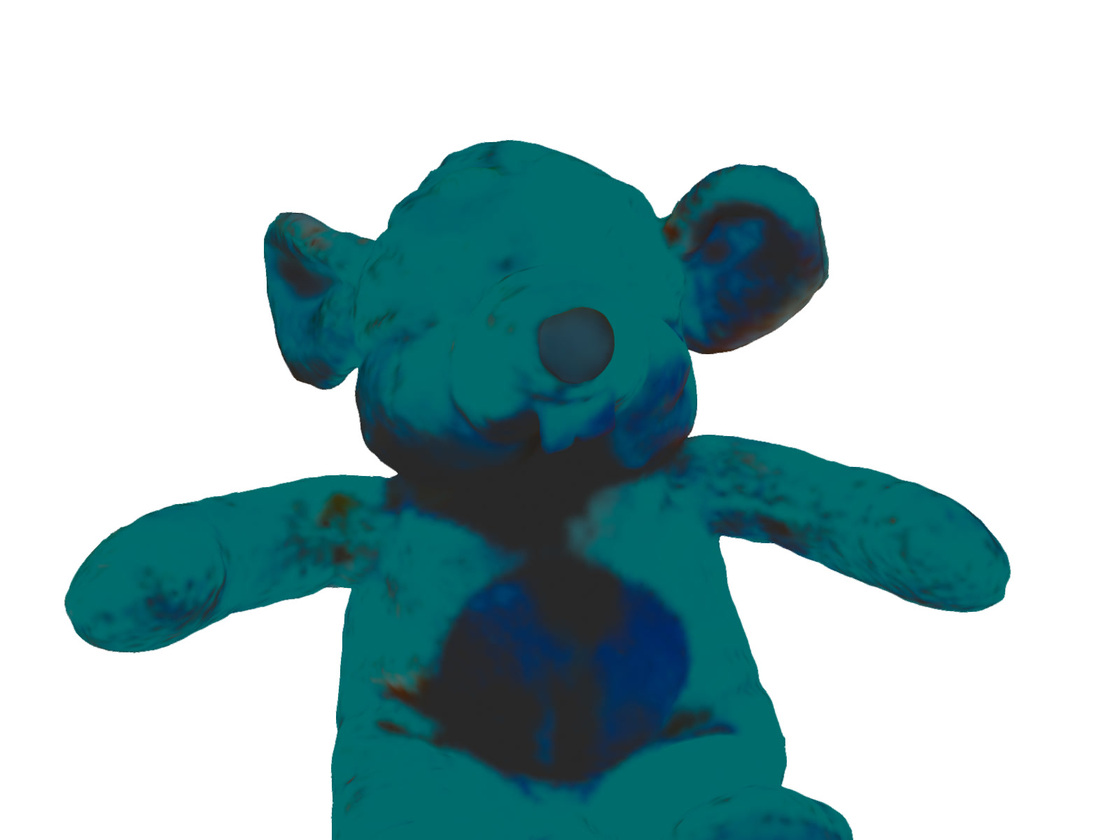}
    \end{subfigure}
    \hspace{1pt}
    \begin{subfigure}[h]{0.14\paperwidth}
        \includegraphics[width=\textwidth]{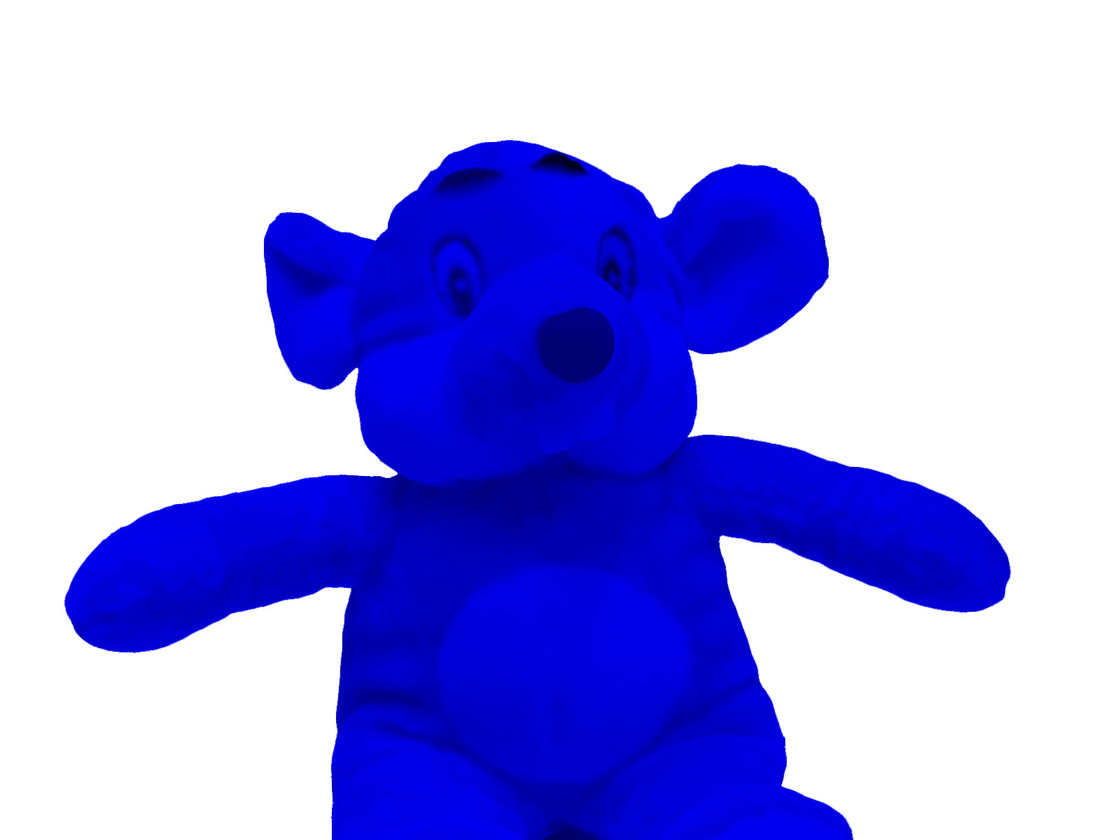}
    \end{subfigure}

    \smallskip
    \rotatebox[origin=b]{90}{scan106}\quad
    \begin{subfigure}[h]{0.14\paperwidth}
        \includegraphics[width=\textwidth]{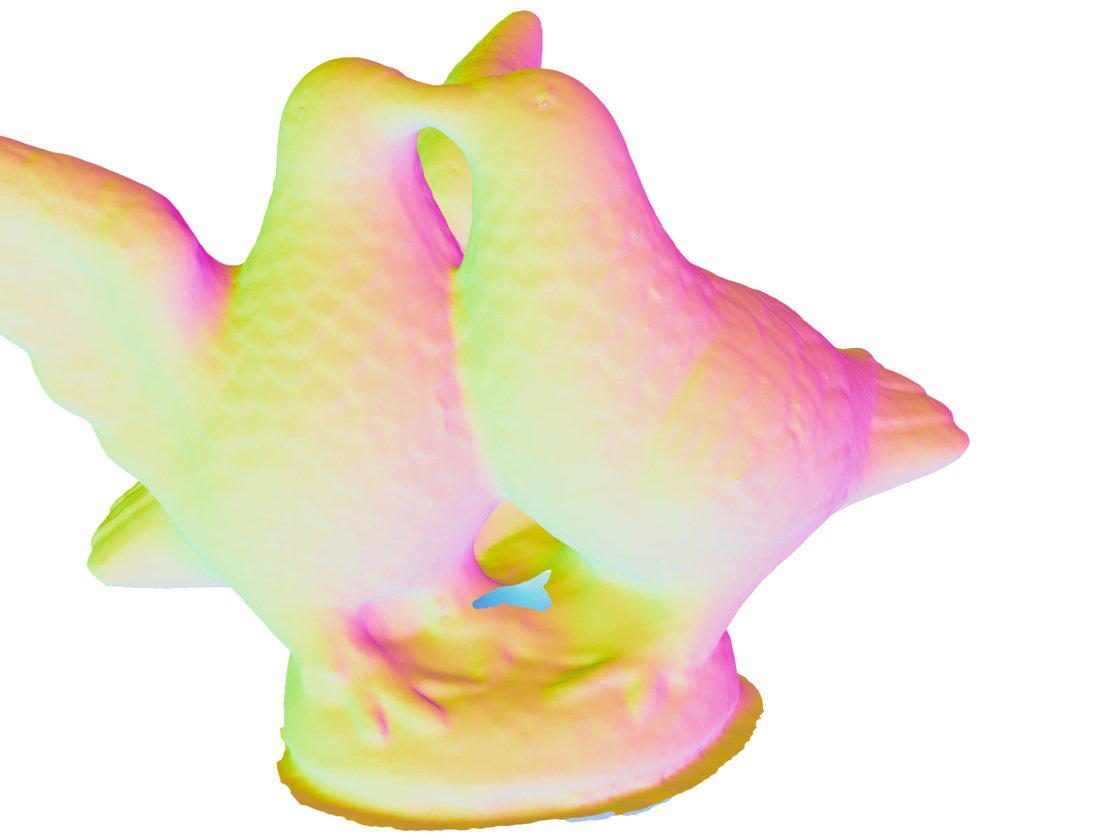}
    \end{subfigure}
    \hspace{1pt}
    \begin{subfigure}[h]{0.14\paperwidth}
        \includegraphics[width=\textwidth]{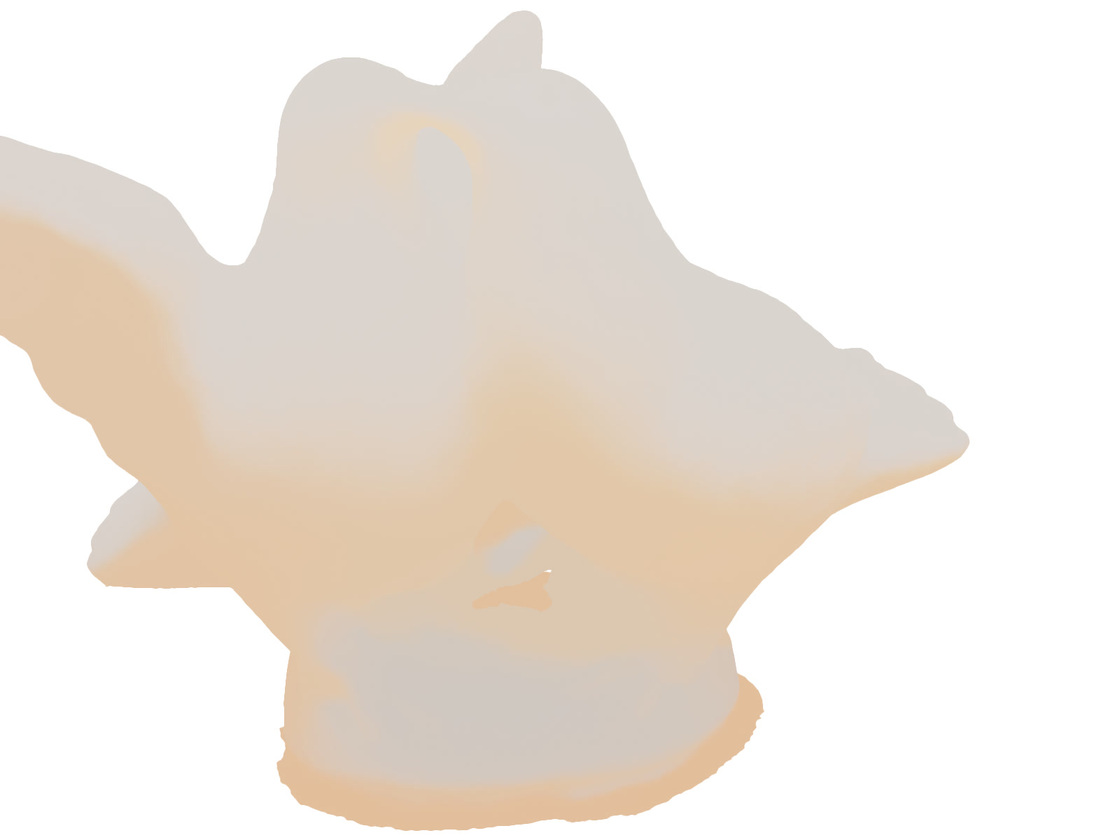}
    \end{subfigure}
    \hspace{1pt}
    \begin{subfigure}[h]{0.14\paperwidth}
        \includegraphics[width=\textwidth]{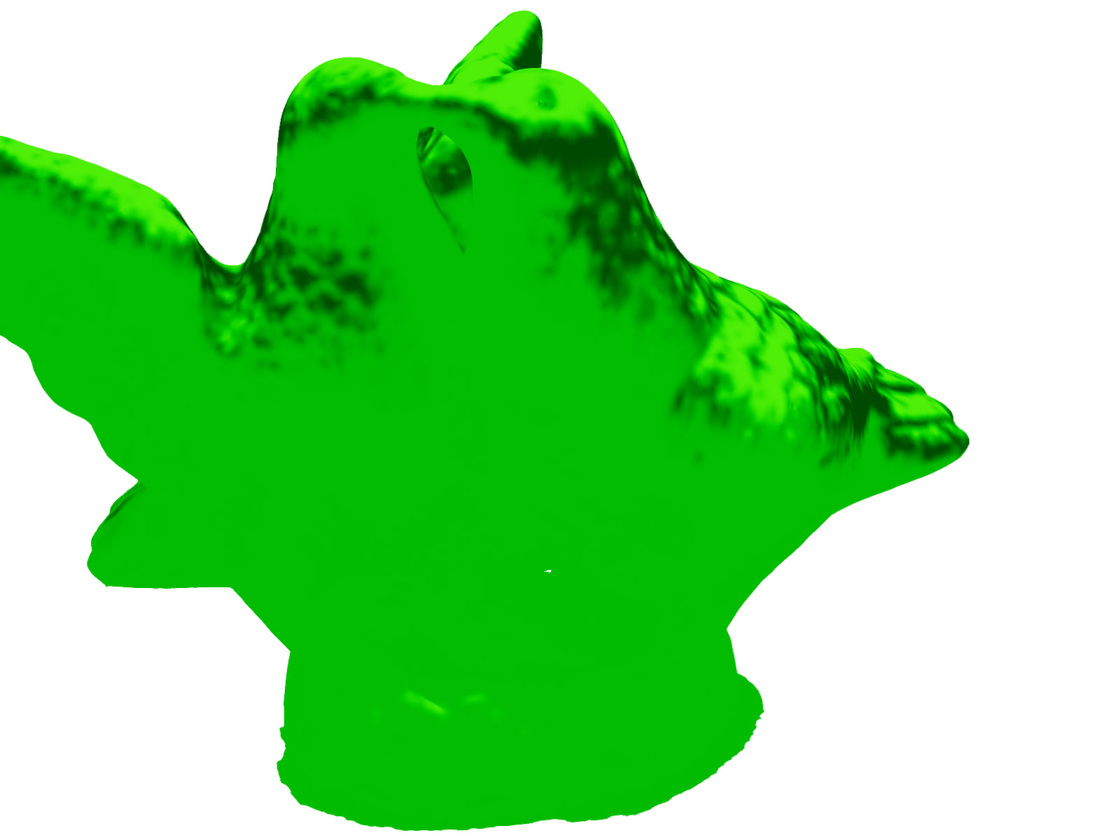}
    \end{subfigure}
    \hspace{1pt}
    \begin{subfigure}[h]{0.14\paperwidth}
        \includegraphics[width=\textwidth]{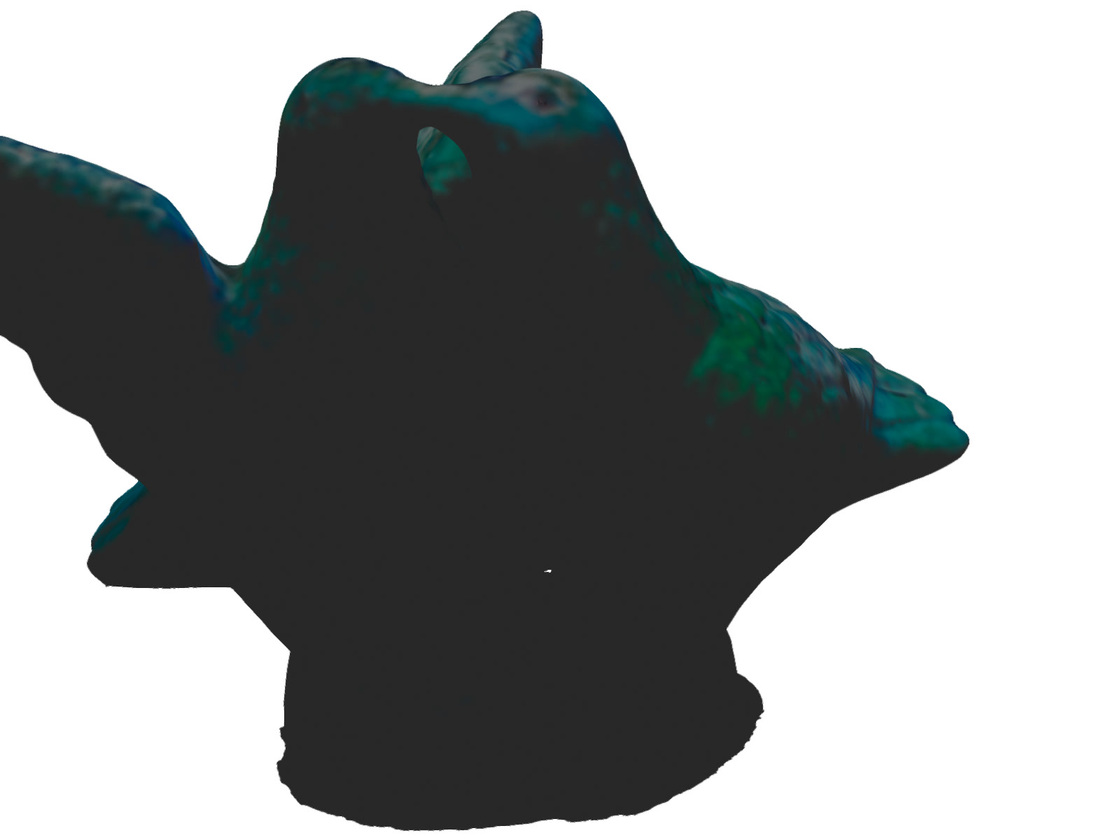}
    \end{subfigure}
    \hspace{1pt}
    \begin{subfigure}[h]{0.14\paperwidth}
        \includegraphics[width=\textwidth]{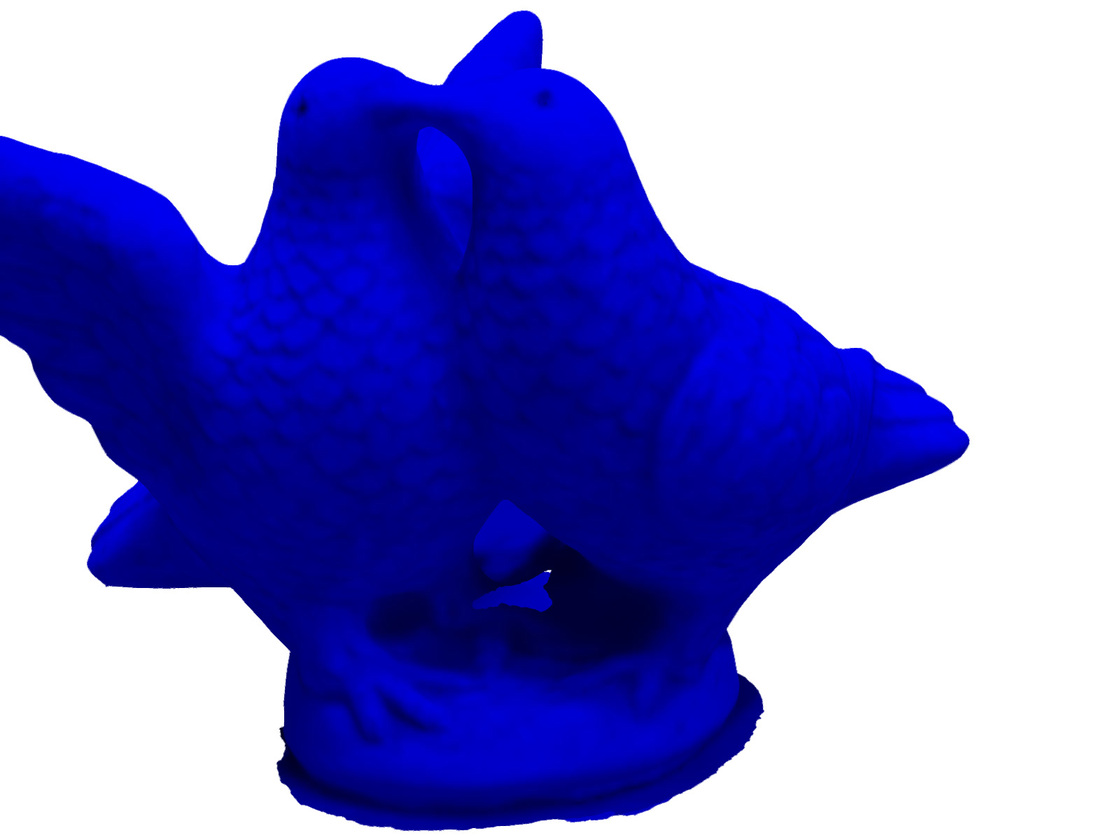}
    \end{subfigure}

    \smallskip
    \rotatebox[origin=b]{90}{scan110}\quad
    \begin{subfigure}[h]{0.14\paperwidth}
        \includegraphics[width=\textwidth]{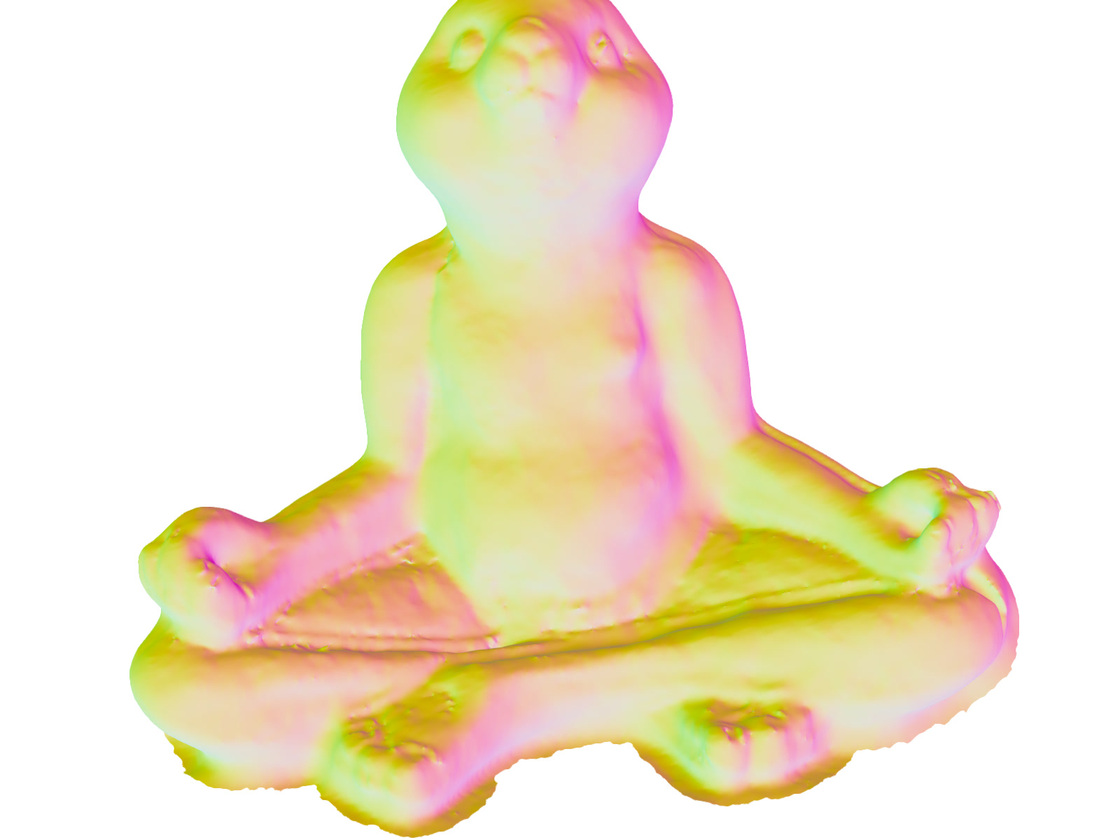}
    \end{subfigure}
    \hspace{1pt}
    \begin{subfigure}[h]{0.14\paperwidth}
        \includegraphics[width=\textwidth]{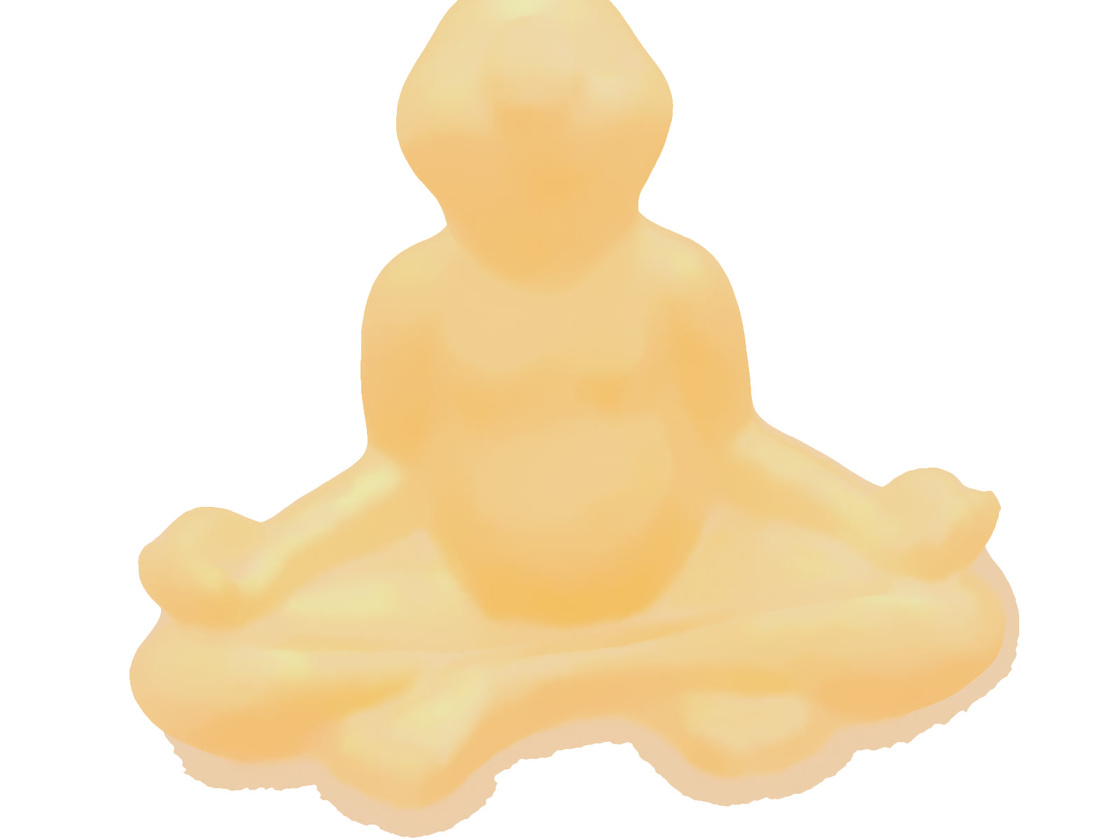}
    \end{subfigure}
    \hspace{1pt}
    \begin{subfigure}[h]{0.14\paperwidth}
        \includegraphics[width=\textwidth]{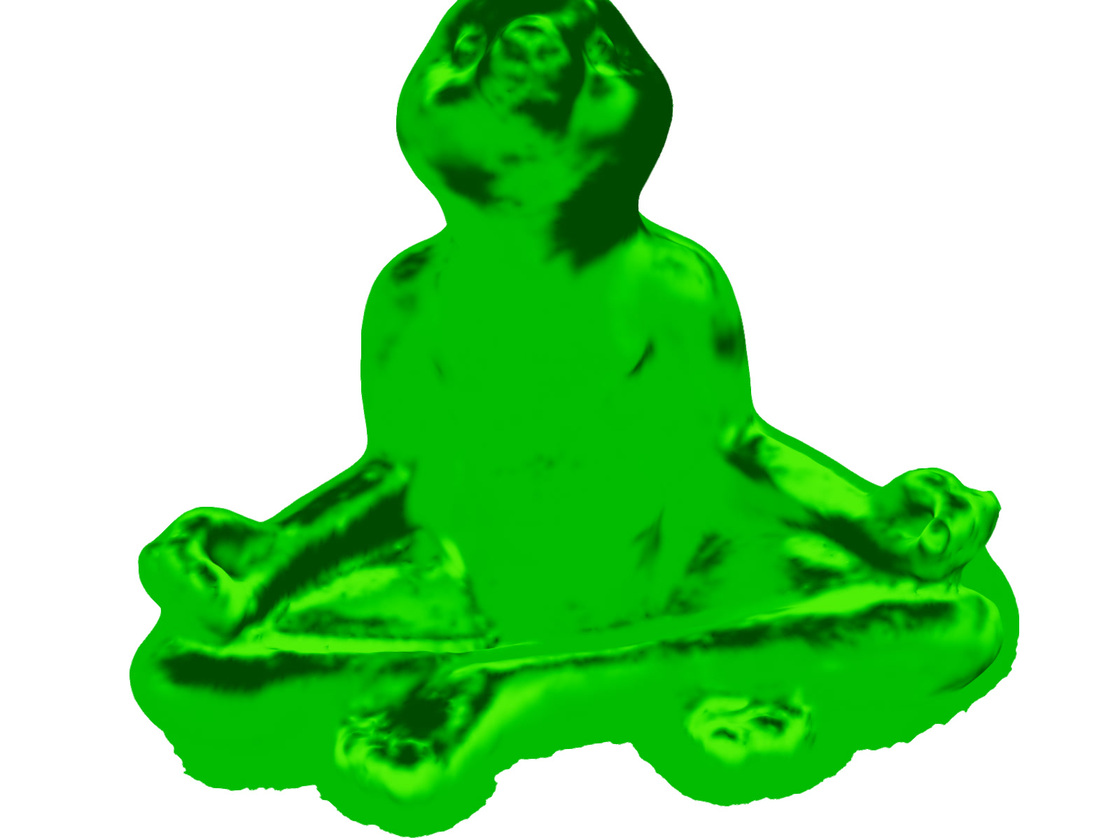}
    \end{subfigure}
    \hspace{1pt}
    \begin{subfigure}[h]{0.14\paperwidth}
        \includegraphics[width=\textwidth]{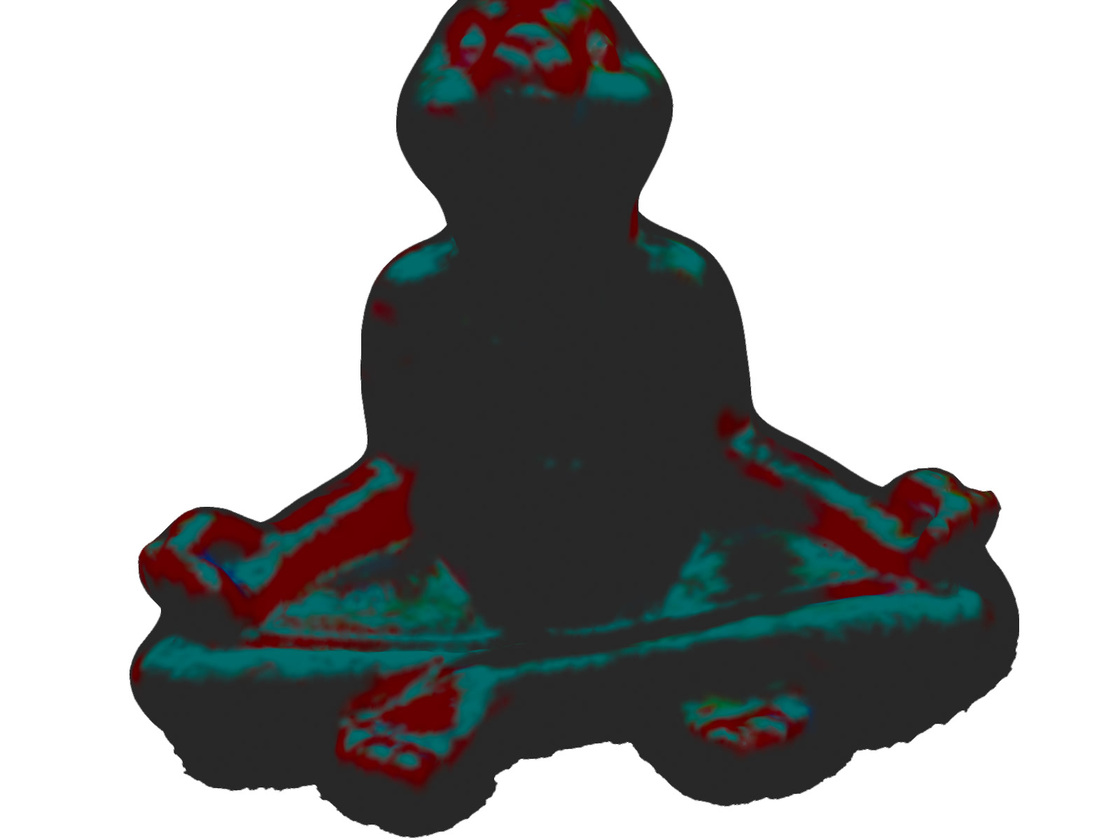}
    \end{subfigure}
    \hspace{1pt}
    \begin{subfigure}[h]{0.14\paperwidth}
        \includegraphics[width=\textwidth]{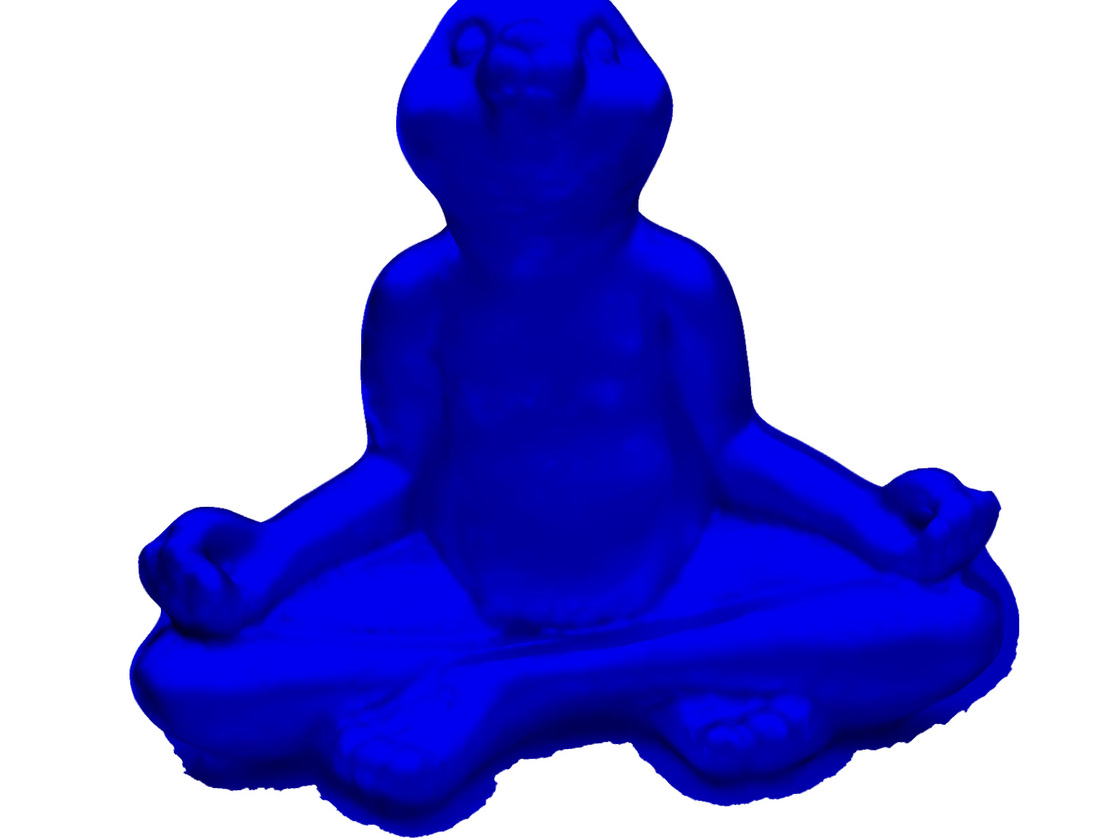}
    \end{subfigure}

    \smallskip
    \rotatebox[origin=b]{90}{scan97}\quad
    \begin{subfigure}[h]{0.17\paperwidth}
        \caption{PBR (default)}
        \includegraphics[width=\textwidth]{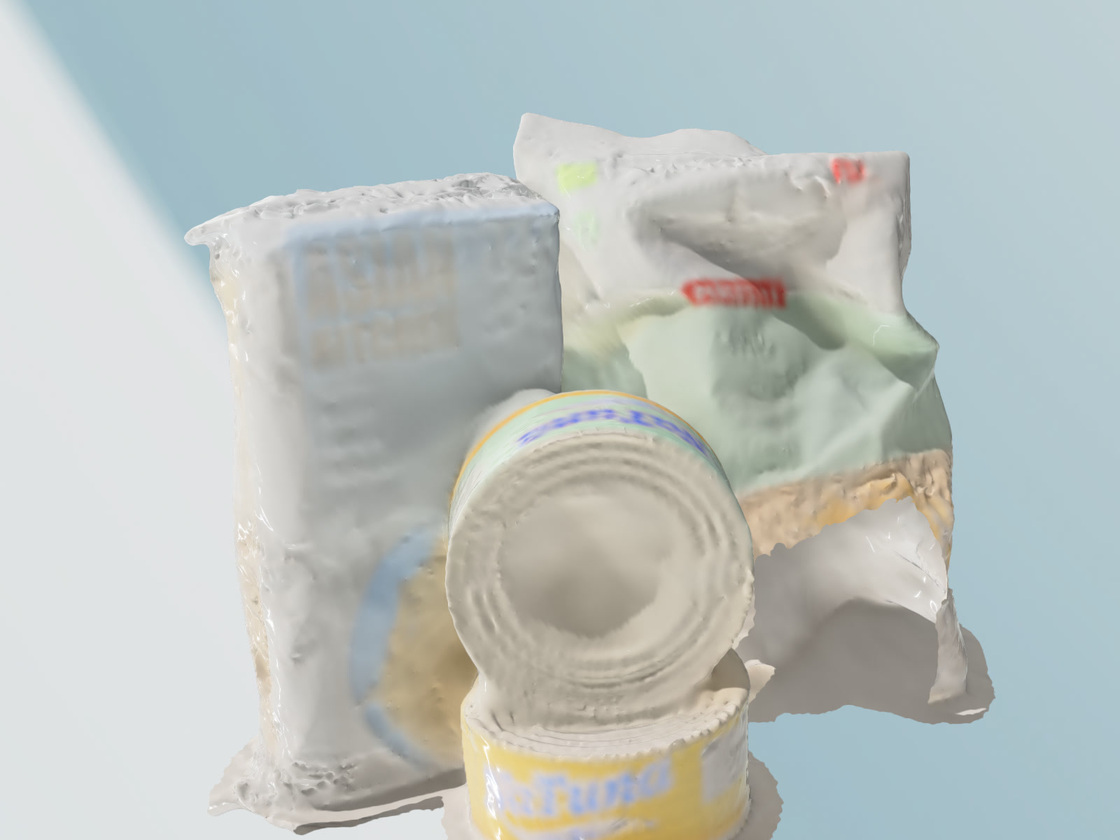}
    \end{subfigure}
    \hspace{1pt}
    \begin{subfigure}[h]{0.17\paperwidth}
        \caption{PBR (pillars)}
        \includegraphics[width=\textwidth]{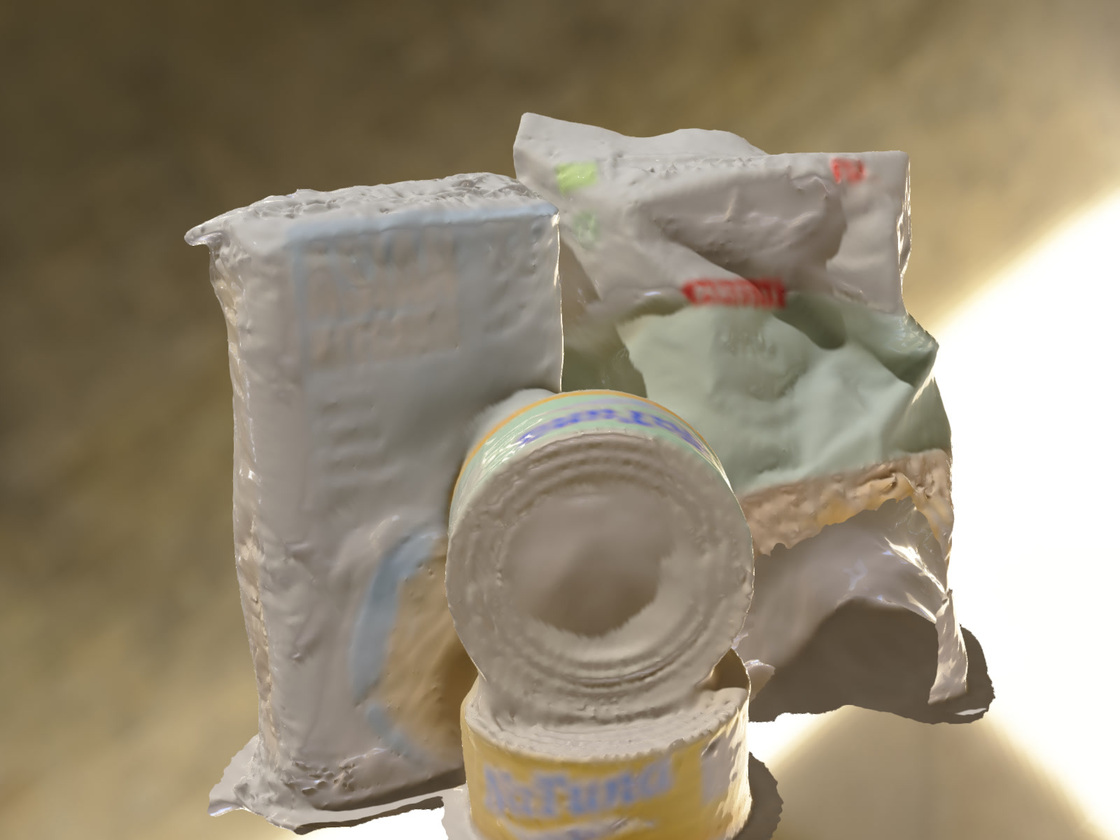}
    \end{subfigure}
    \hspace{1pt}
    \begin{subfigure}[h]{0.17\paperwidth}
        \caption{Neural rendering}
        \includegraphics[width=\textwidth]{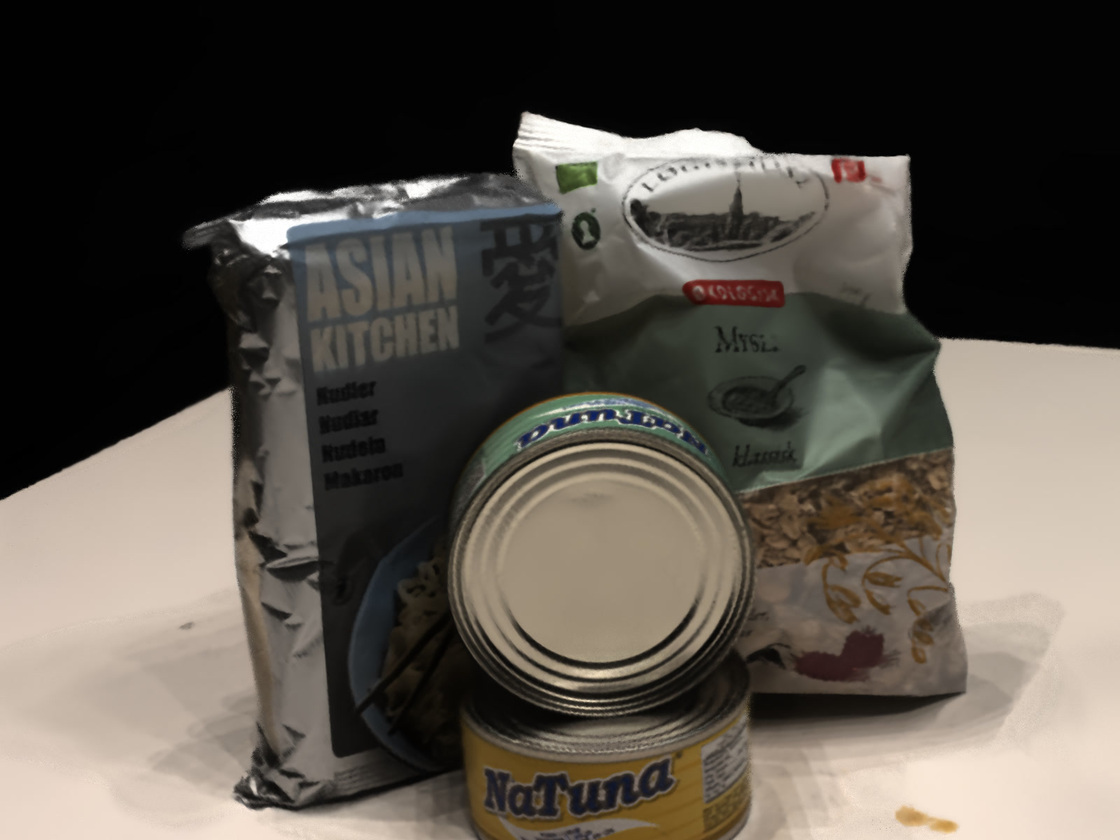}
    \end{subfigure}
    \hspace{1pt}
    \begin{subfigure}[h]{0.17\paperwidth}
        \caption{Groundtruth}
        \includegraphics[width=\textwidth]{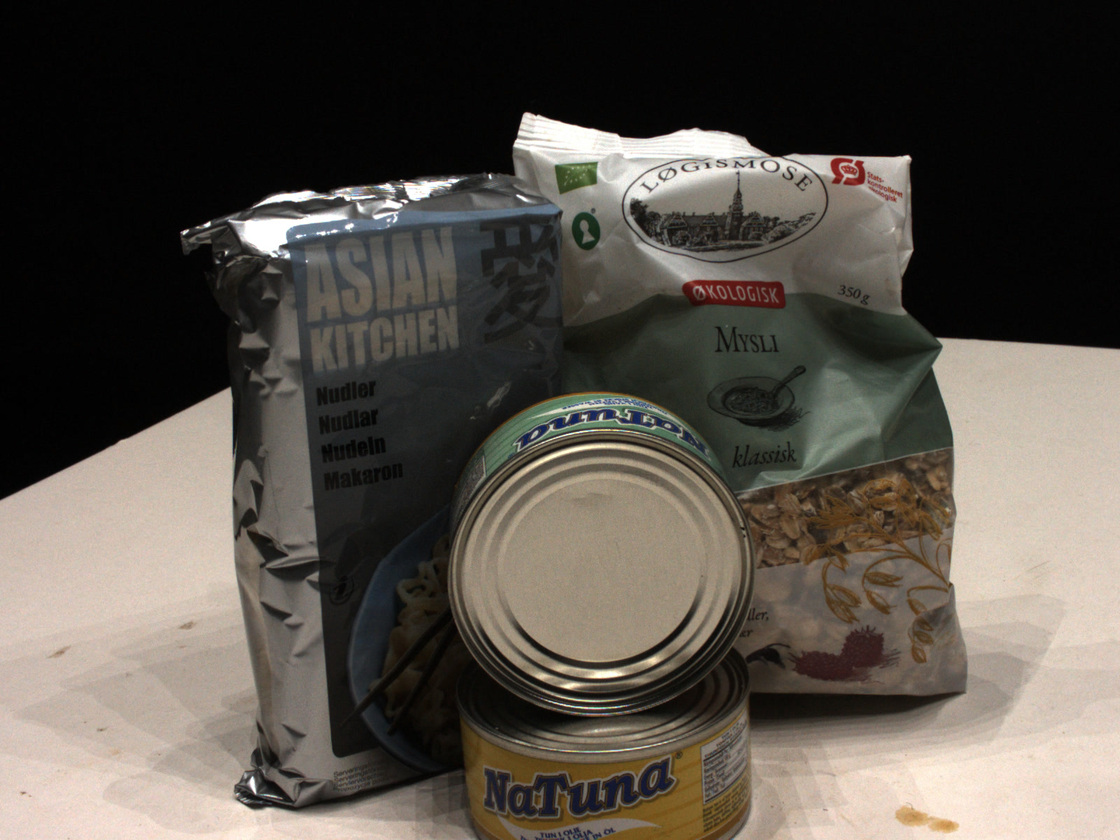}
    \end{subfigure}

    \smallskip
    \rotatebox[origin=b]{90}{scan105}\quad
    \begin{subfigure}[h]{0.17\paperwidth}
        \includegraphics[width=\textwidth]{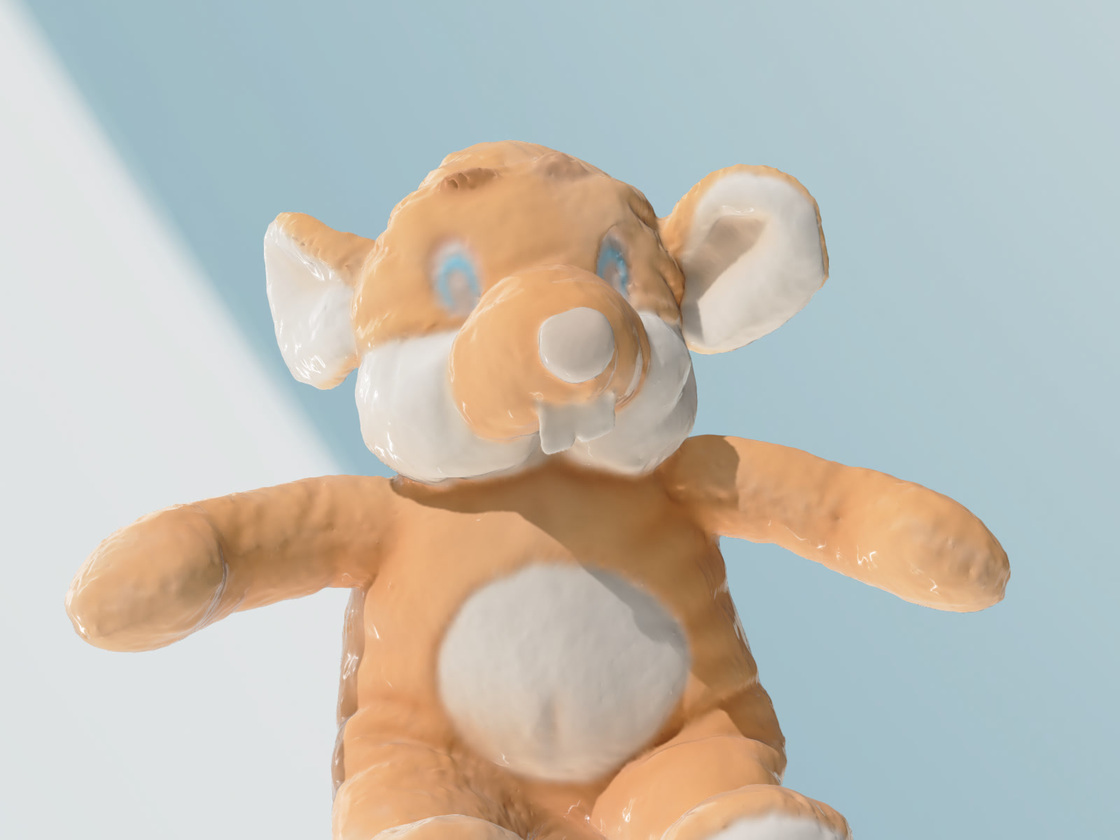}
    \end{subfigure}
    \hspace{1pt}
    \begin{subfigure}[h]{0.17\paperwidth}
        \includegraphics[width=\textwidth]{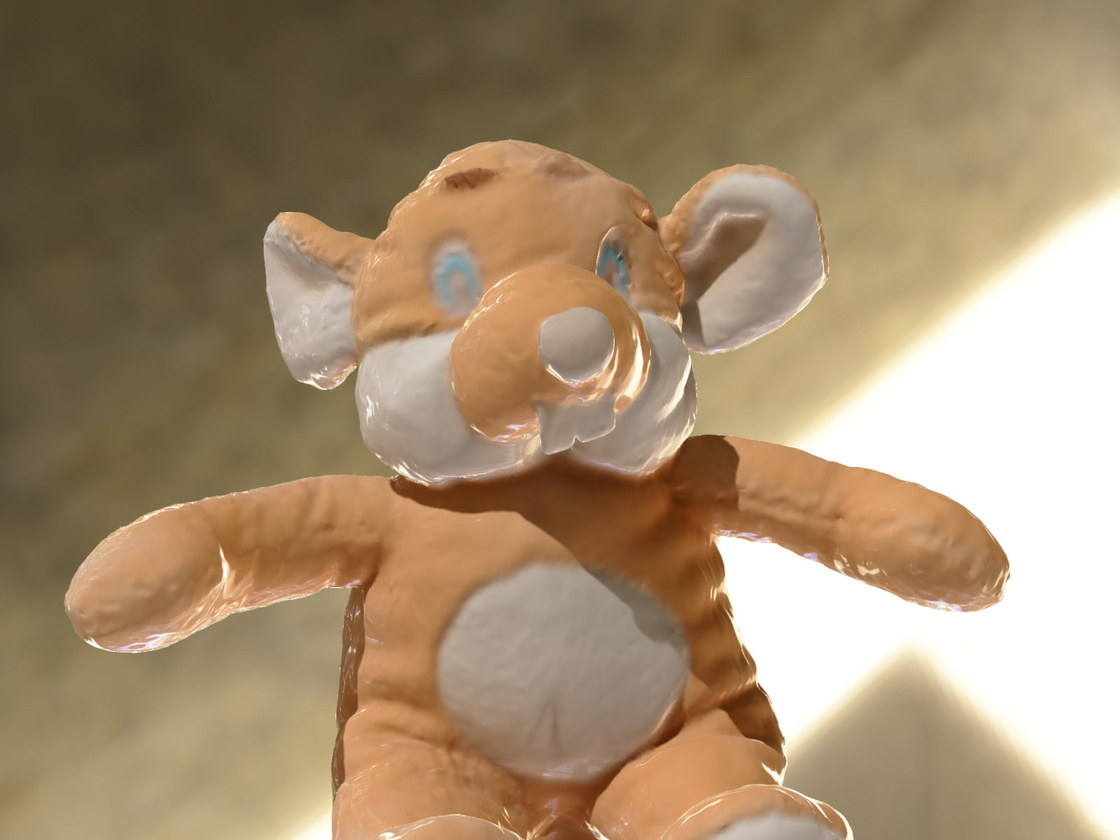}
    \end{subfigure}
    \hspace{1pt}
    \begin{subfigure}[h]{0.17\paperwidth}
        \includegraphics[width=\textwidth]{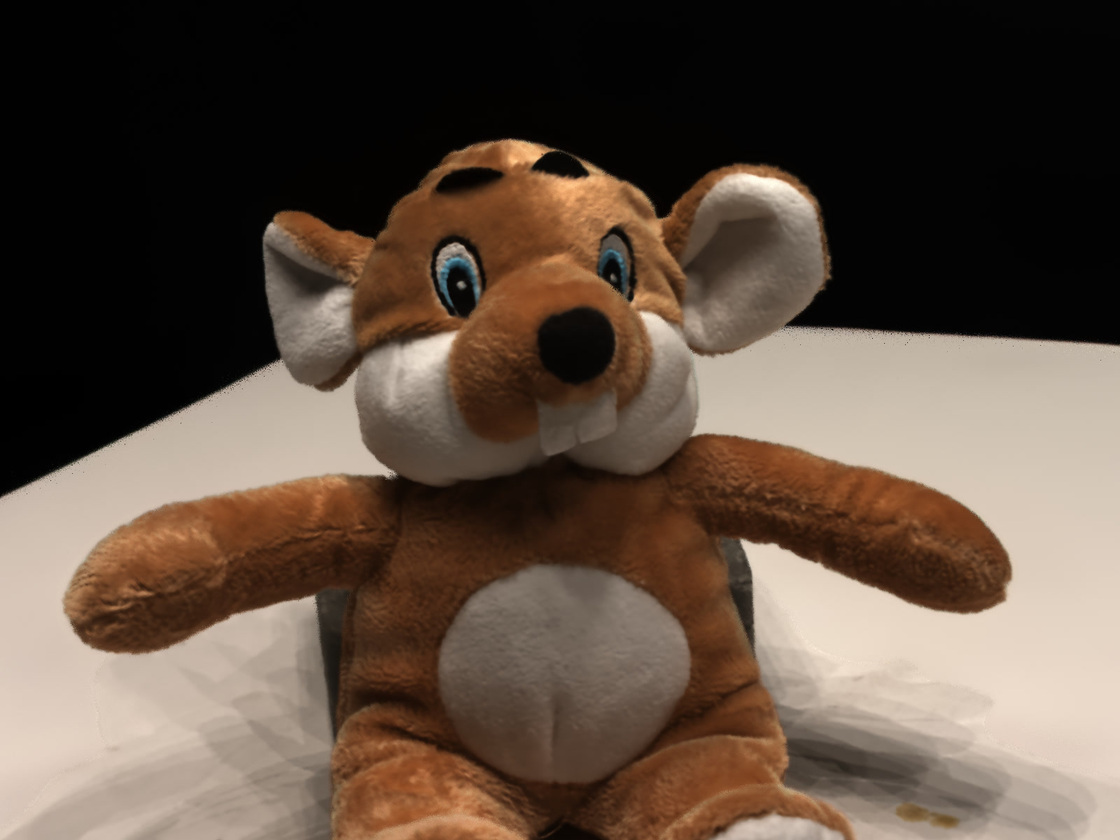}
    \end{subfigure}
    \hspace{1pt}
    \begin{subfigure}[h]{0.17\paperwidth}
        \includegraphics[width=\textwidth]{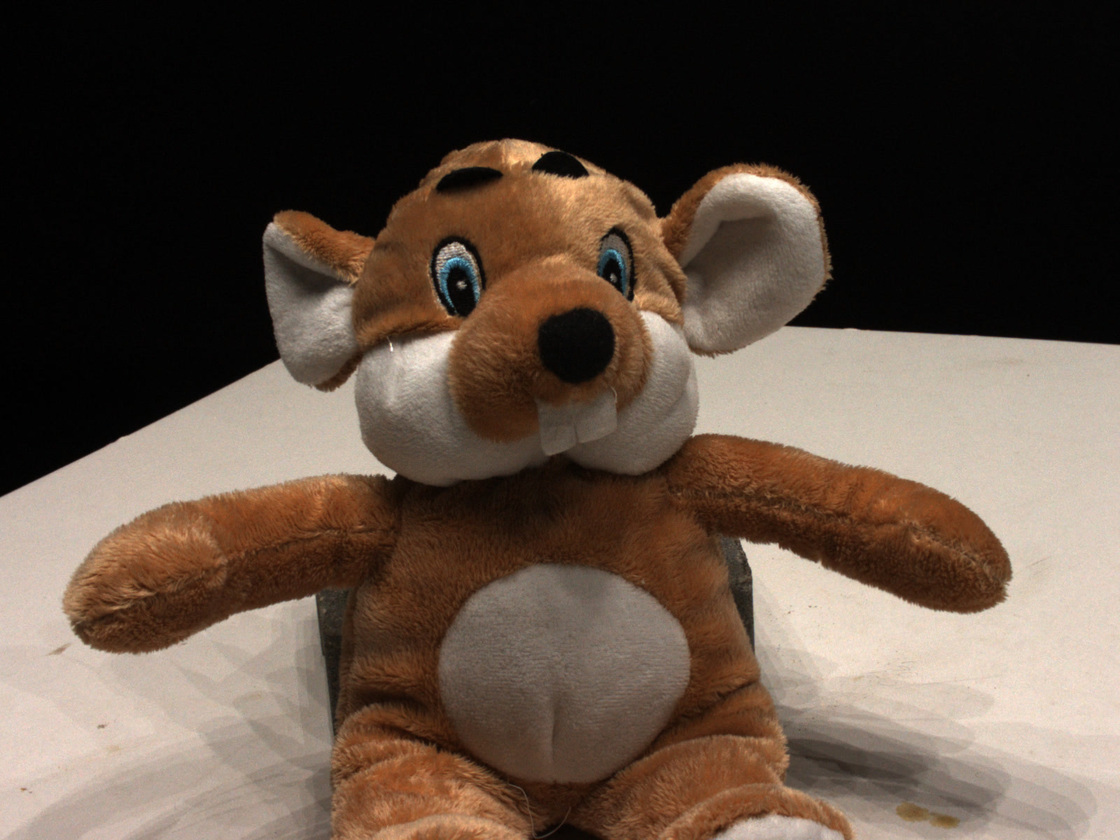}
    \end{subfigure}

    \smallskip
    \rotatebox[origin=b]{90}{scan106}\quad
    \begin{subfigure}[h]{0.17\paperwidth}
        \includegraphics[width=\textwidth]{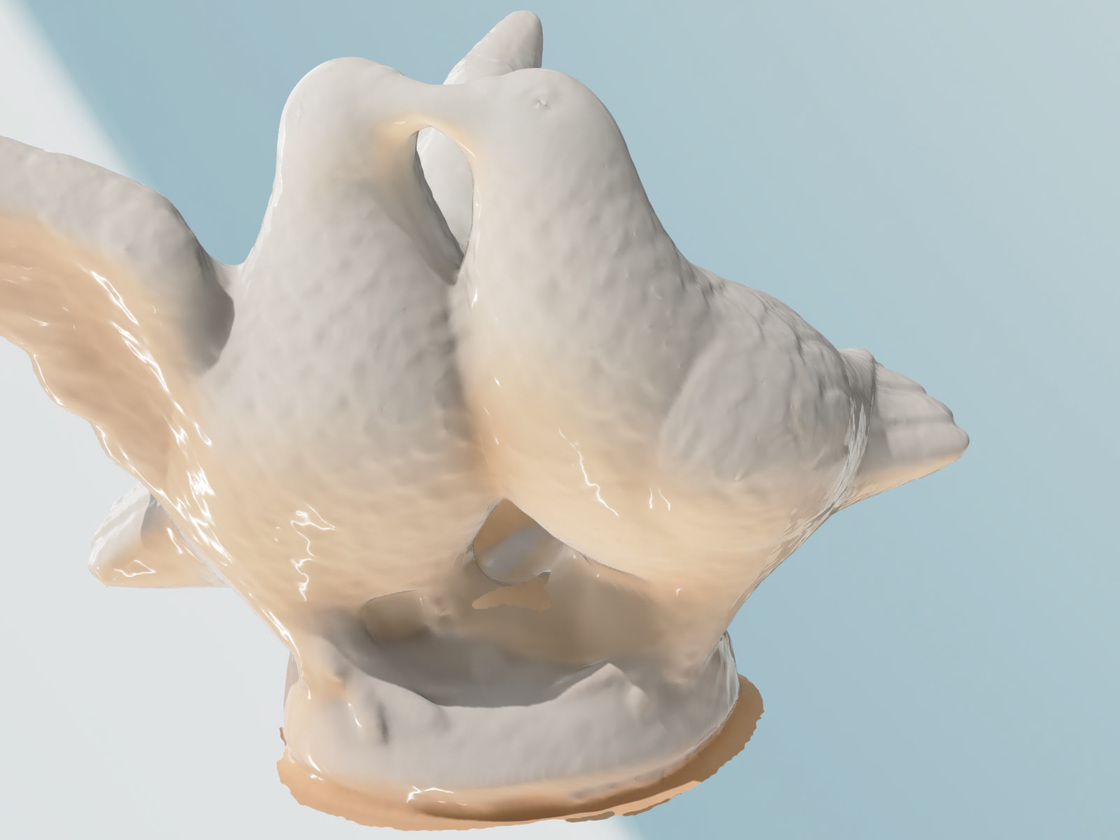}
    \end{subfigure}
    \hspace{1pt}
    \begin{subfigure}[h]{0.17\paperwidth}
        \includegraphics[width=\textwidth]{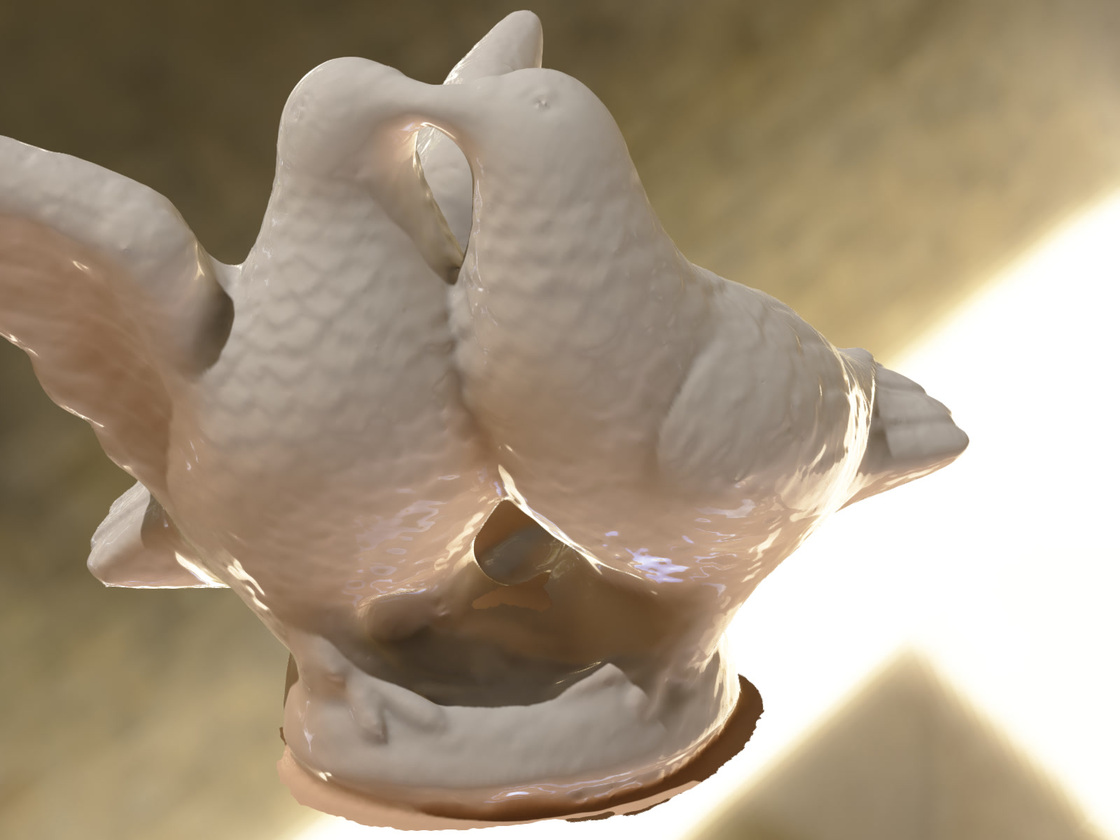}
    \end{subfigure}
    \hspace{1pt}
    \begin{subfigure}[h]{0.17\paperwidth}
        \includegraphics[width=\textwidth]{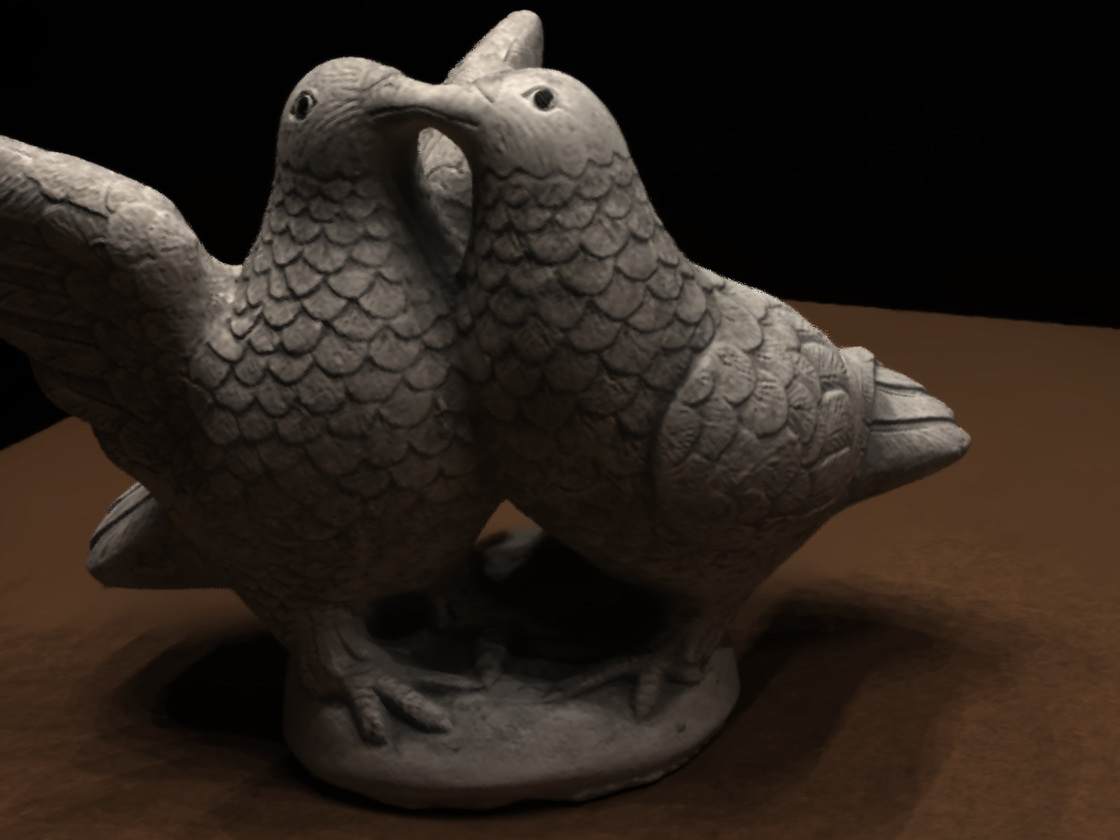}
    \end{subfigure}
    \hspace{1pt}
    \begin{subfigure}[h]{0.17\paperwidth}
        \includegraphics[width=\textwidth]{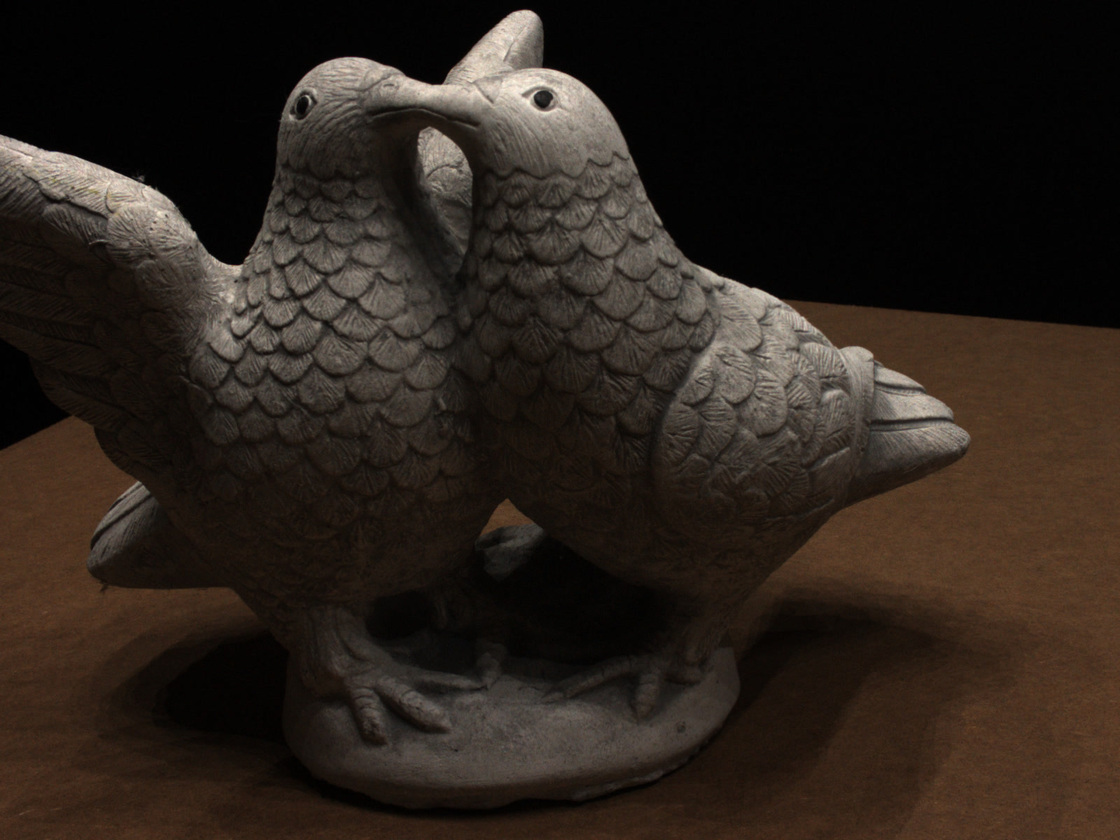}
    \end{subfigure}

    \smallskip
    \rotatebox[origin=b]{90}{scan110}\quad
    \begin{subfigure}[h]{0.17\paperwidth}
        \includegraphics[width=\textwidth]{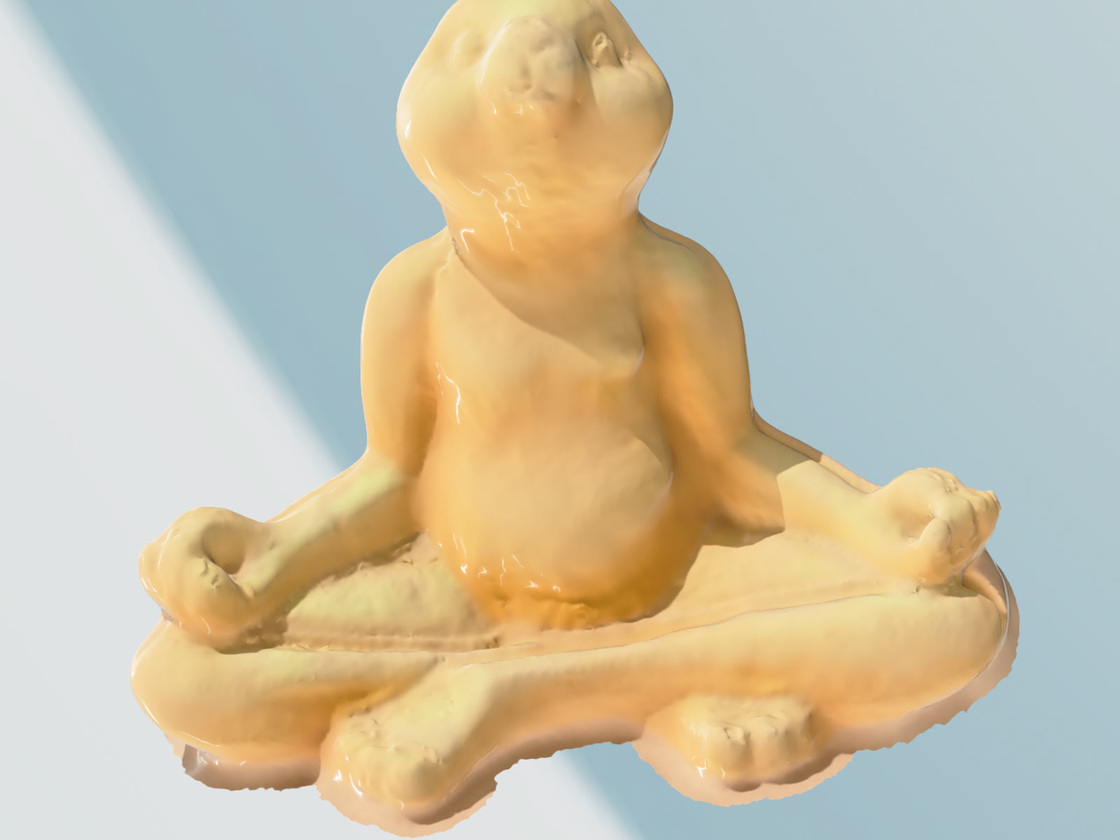}
    \end{subfigure}
    \hspace{1pt}
    \begin{subfigure}[h]{0.17\paperwidth}
        \includegraphics[width=\textwidth]{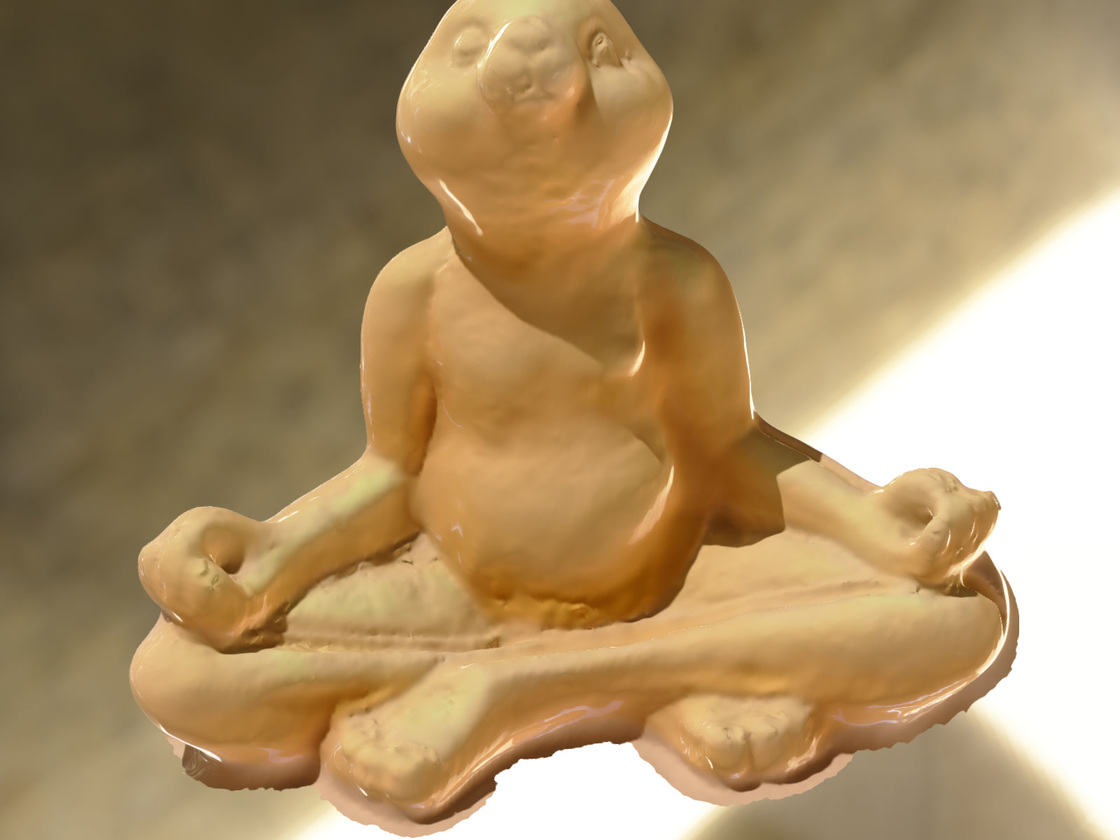}
    \end{subfigure}
    \hspace{1pt}
    \begin{subfigure}[h]{0.17\paperwidth}
        \includegraphics[width=\textwidth]{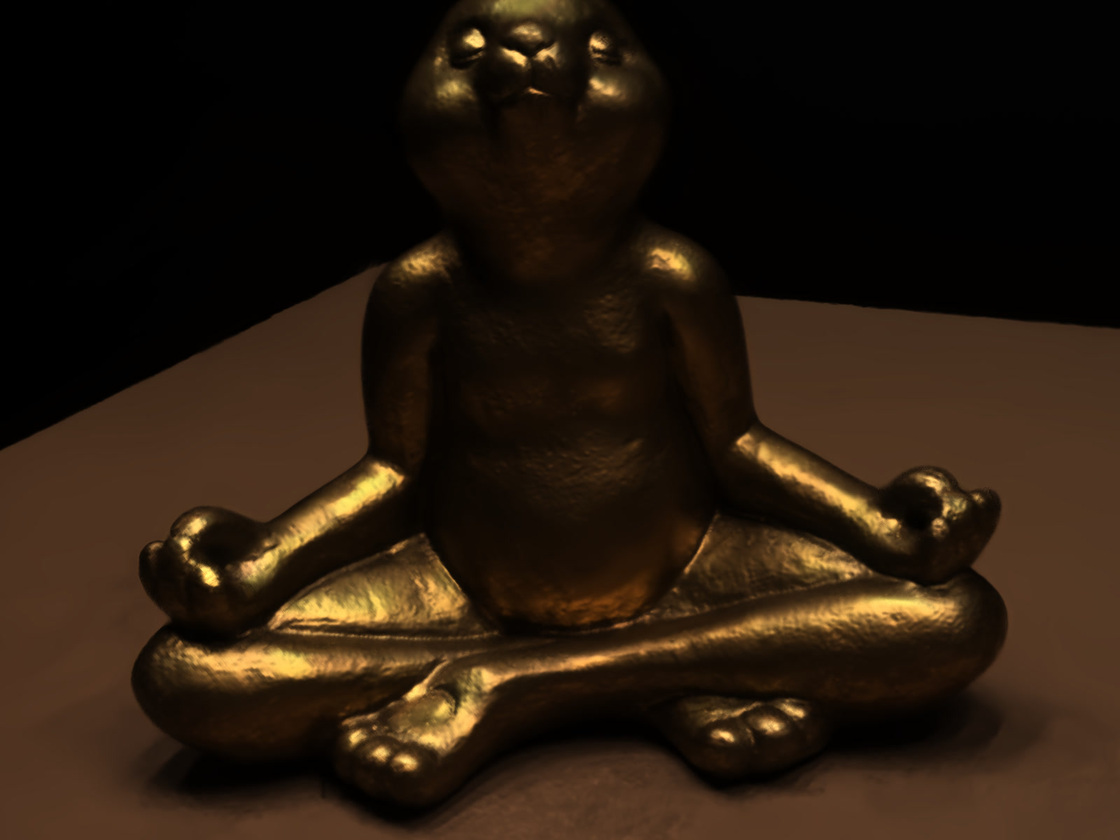}
    \end{subfigure}
    \hspace{1pt}
    \begin{subfigure}[h]{0.17\paperwidth}
        \includegraphics[width=\textwidth]{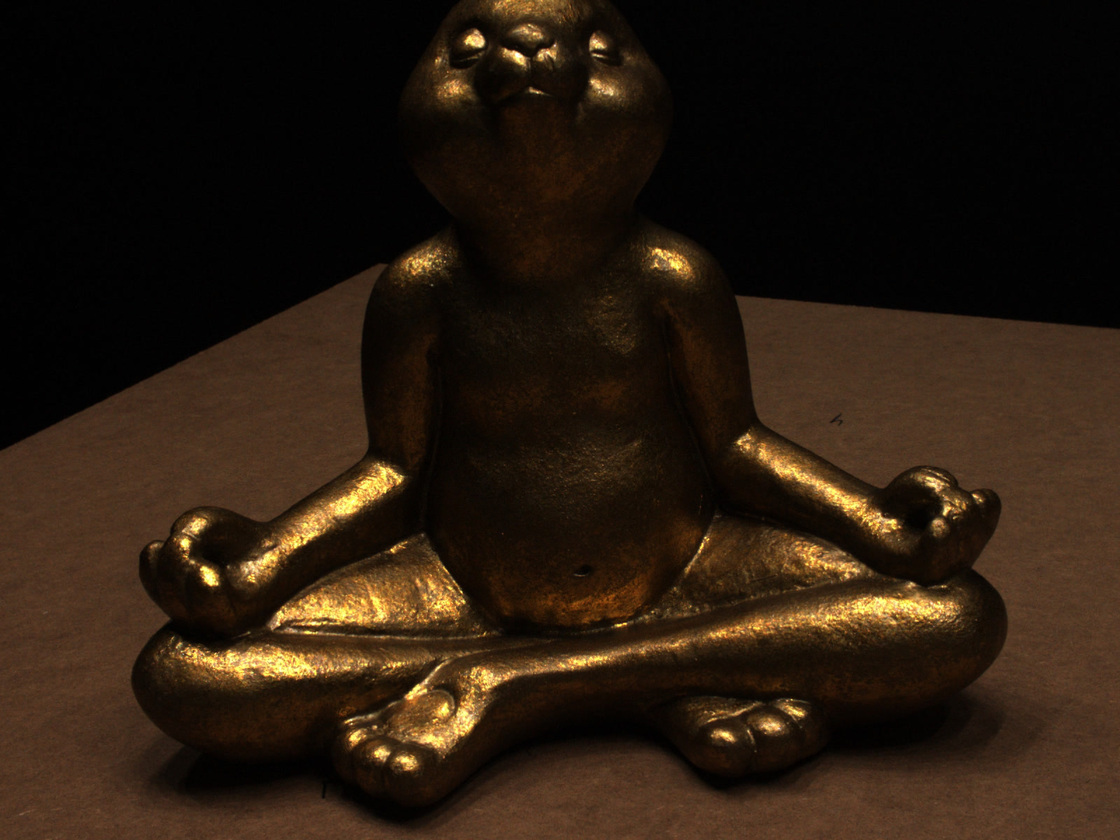}
    \end{subfigure}

    \caption{\textbf{Decomposed materials and rendered images.} ($\cdot$) is a given environment map in Open3D.}
    \label{fig:main_results_03}
\end{figure*}
\begin{figure*}[tbp]
    \centering
    \rotatebox[origin=b]{90}{scan114}\quad
    \begin{subfigure}[h]{0.14\paperwidth}
        \caption{normals}
        \includegraphics[width=\textwidth]{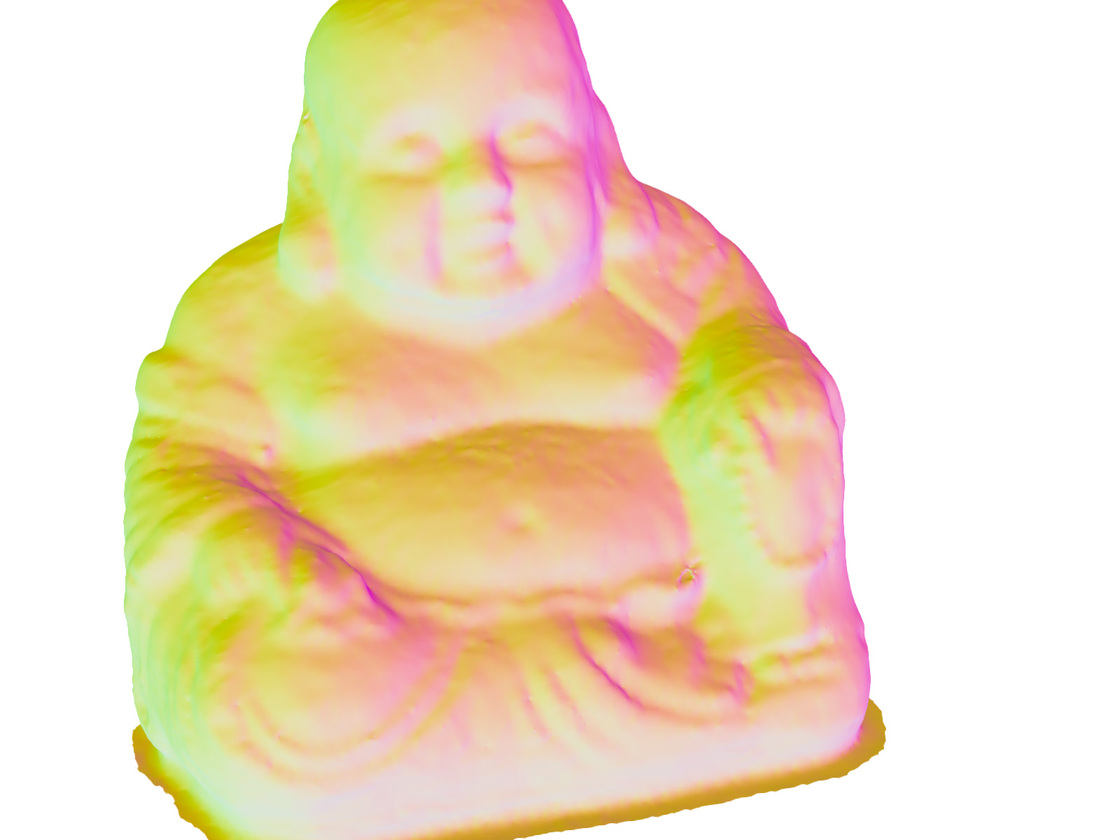}
    \end{subfigure}
    \hspace{1pt}
    \begin{subfigure}[h]{0.14\paperwidth}
        \caption{base color}
        \includegraphics[width=\textwidth]{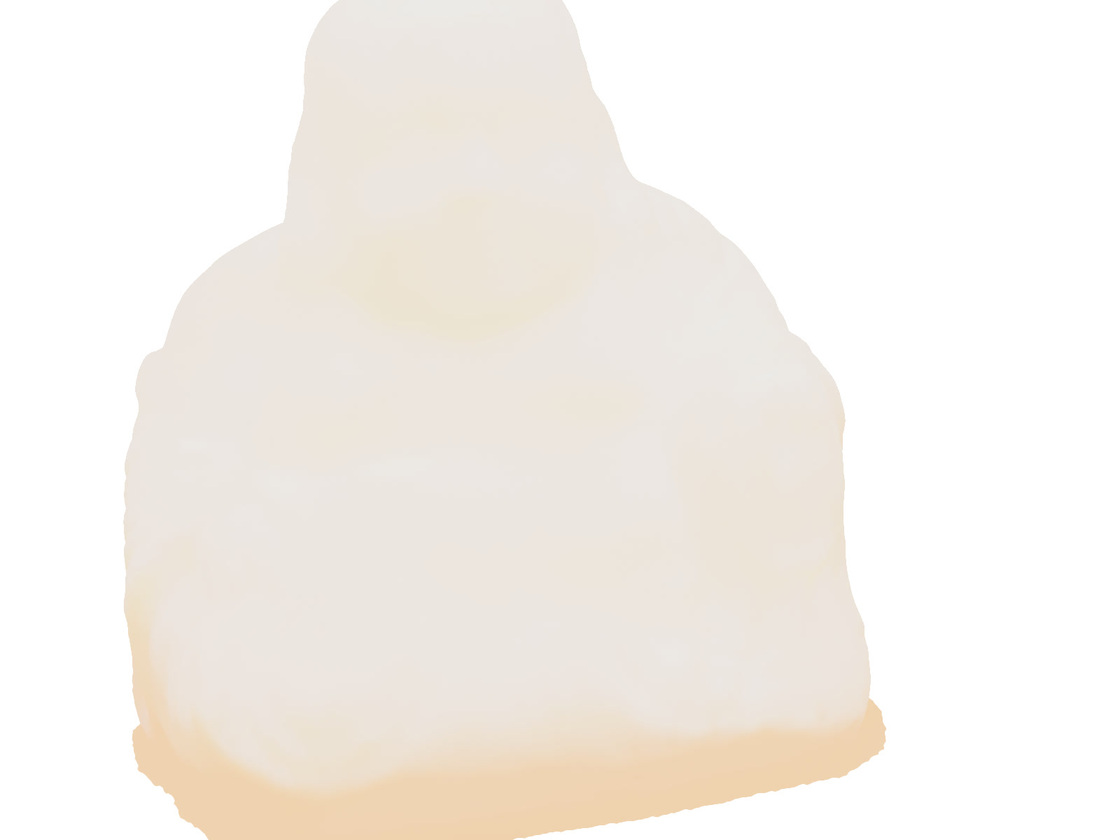}
    \end{subfigure}
    \hspace{1pt}
    \begin{subfigure}[h]{0.14\paperwidth}
        \caption{roughness}
        \includegraphics[width=\textwidth]{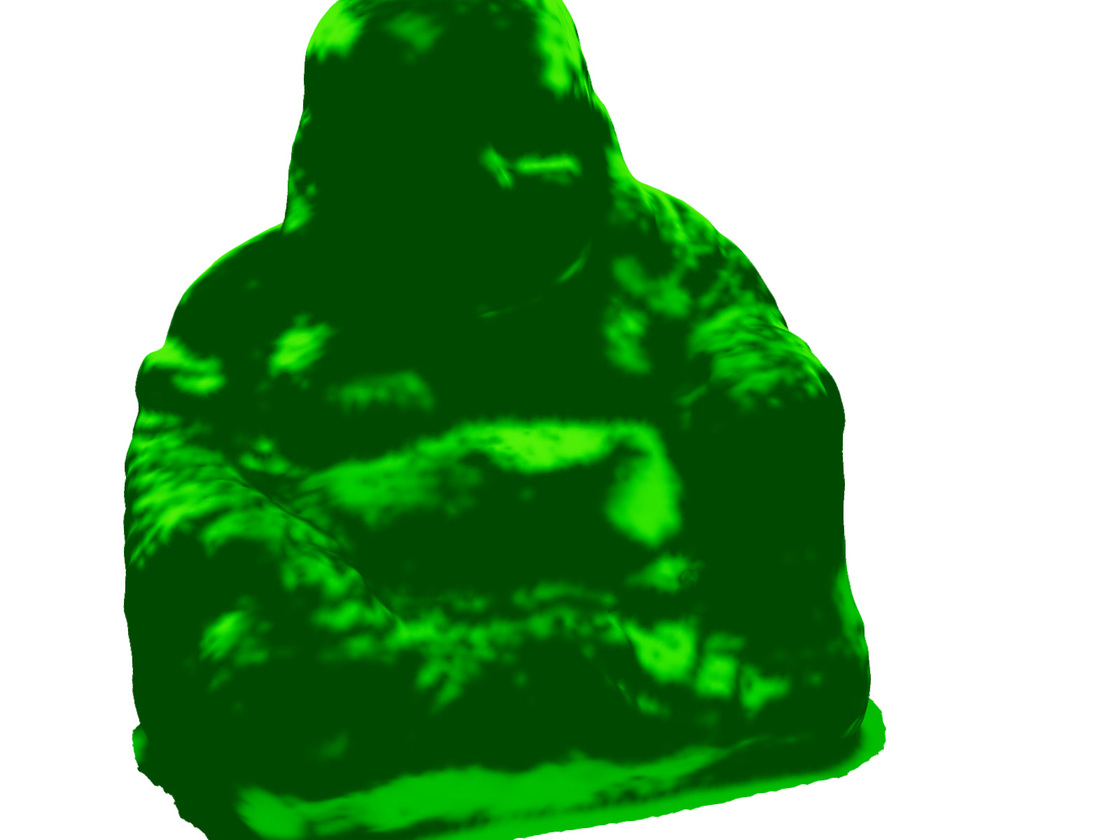}
    \end{subfigure}
    \hspace{1pt}
    \begin{subfigure}[h]{0.14\paperwidth}
        \caption{specular reflectance}
        \includegraphics[width=\textwidth]{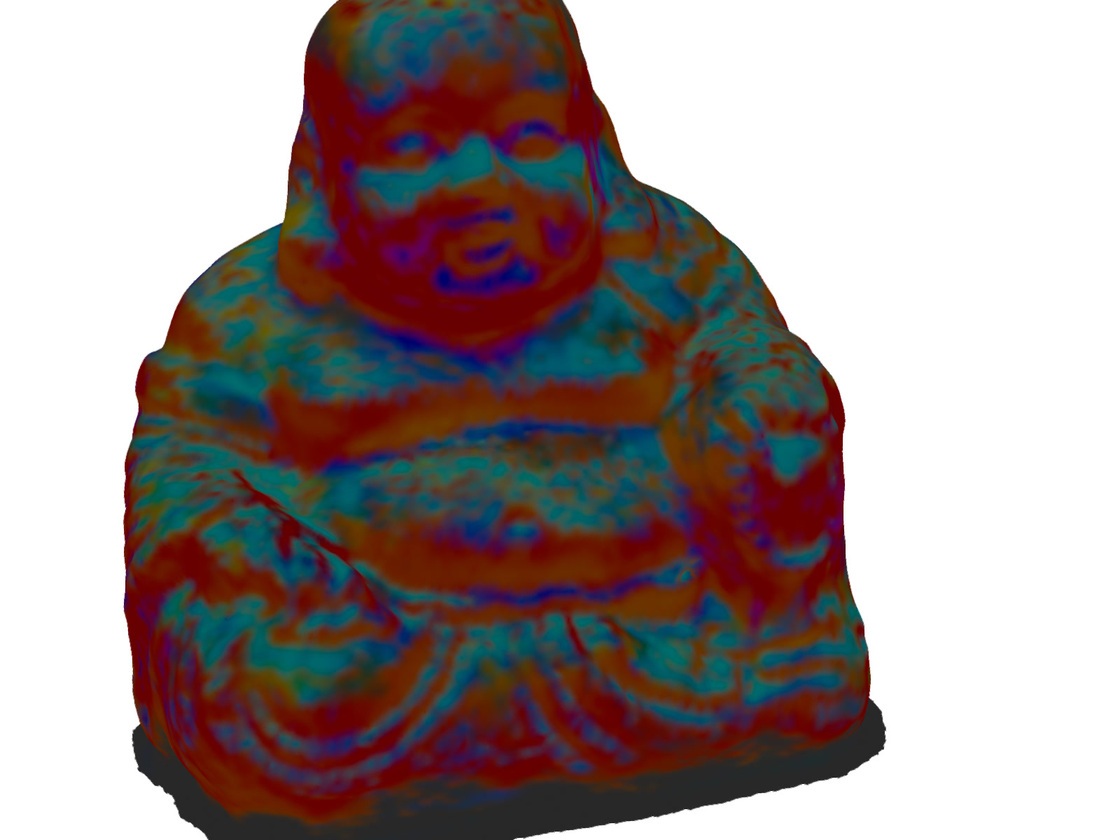}
    \end{subfigure}
    \hspace{1pt}
    \begin{subfigure}[h]{0.14\paperwidth}
        \caption{implicit illumination}
        \includegraphics[width=\textwidth]{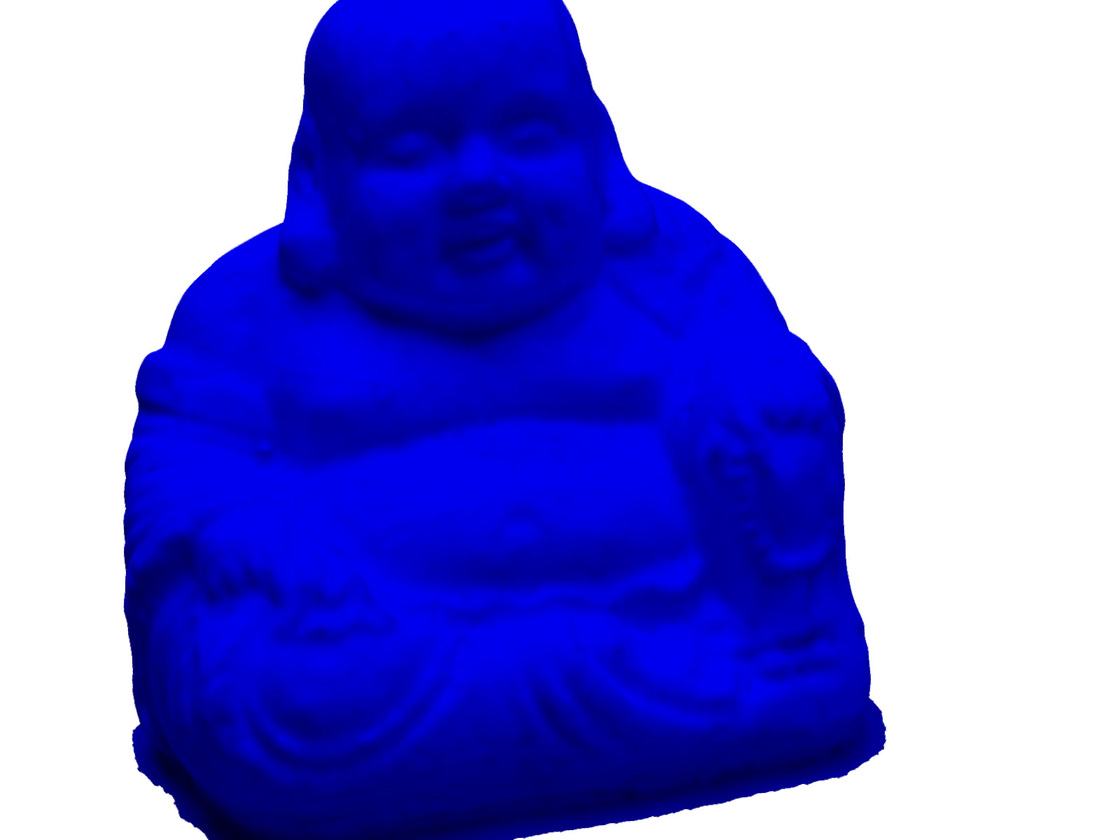}
    \end{subfigure}

    \smallskip
    \rotatebox[origin=b]{90}{scan118}\quad
    \begin{subfigure}[h]{0.14\paperwidth}
        \includegraphics[width=\textwidth]{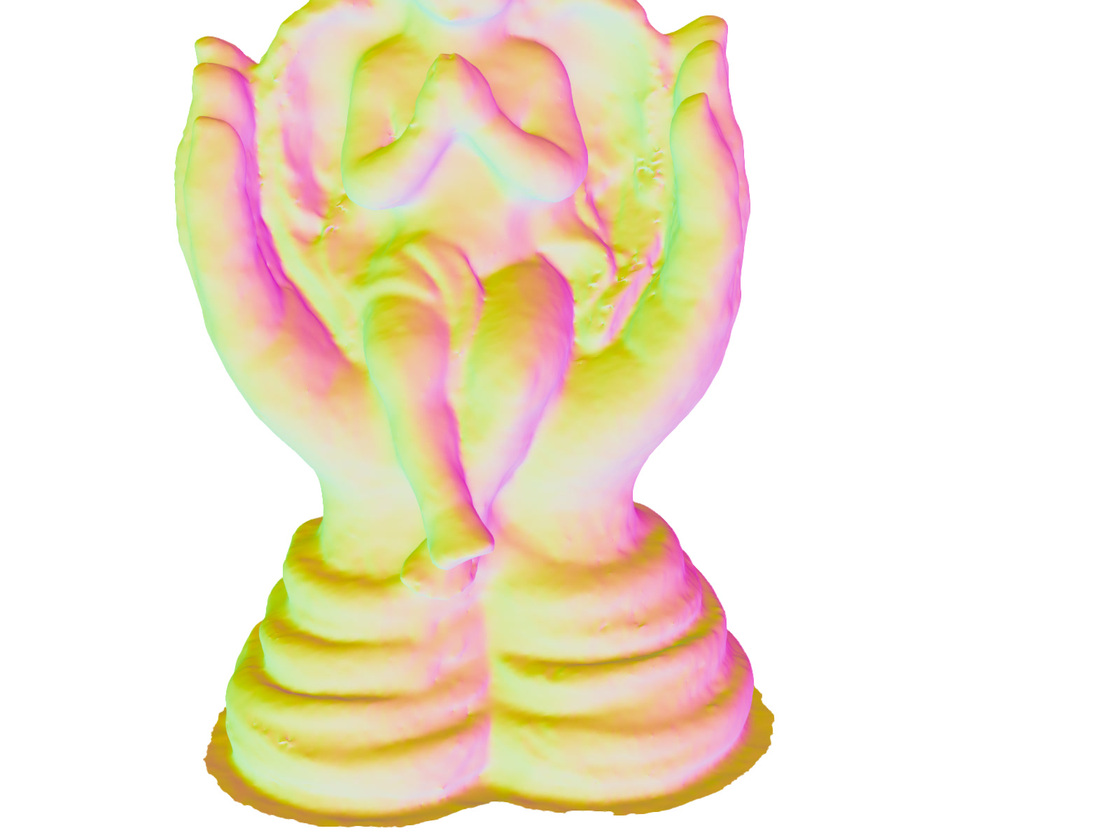}
    \end{subfigure}
    \hspace{1pt}
    \begin{subfigure}[h]{0.14\paperwidth}
        \includegraphics[width=\textwidth]{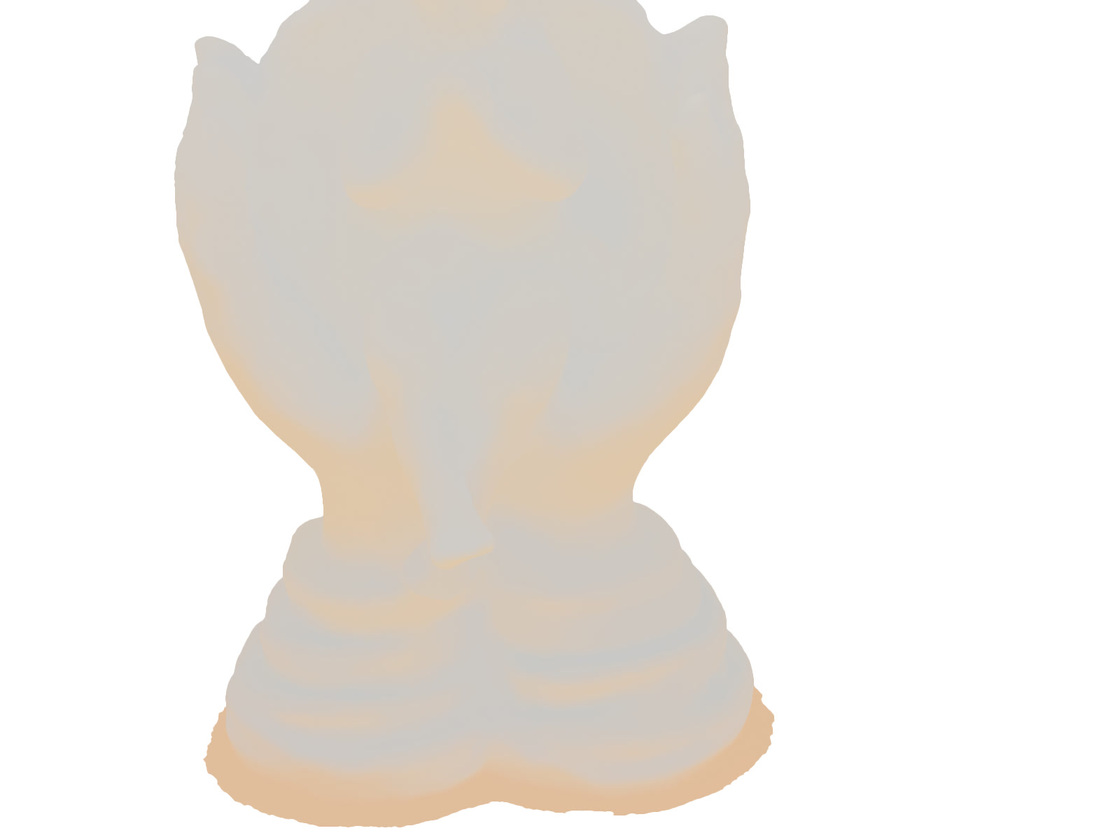}
    \end{subfigure}
    \hspace{1pt}
    \begin{subfigure}[h]{0.14\paperwidth}
        \includegraphics[width=\textwidth]{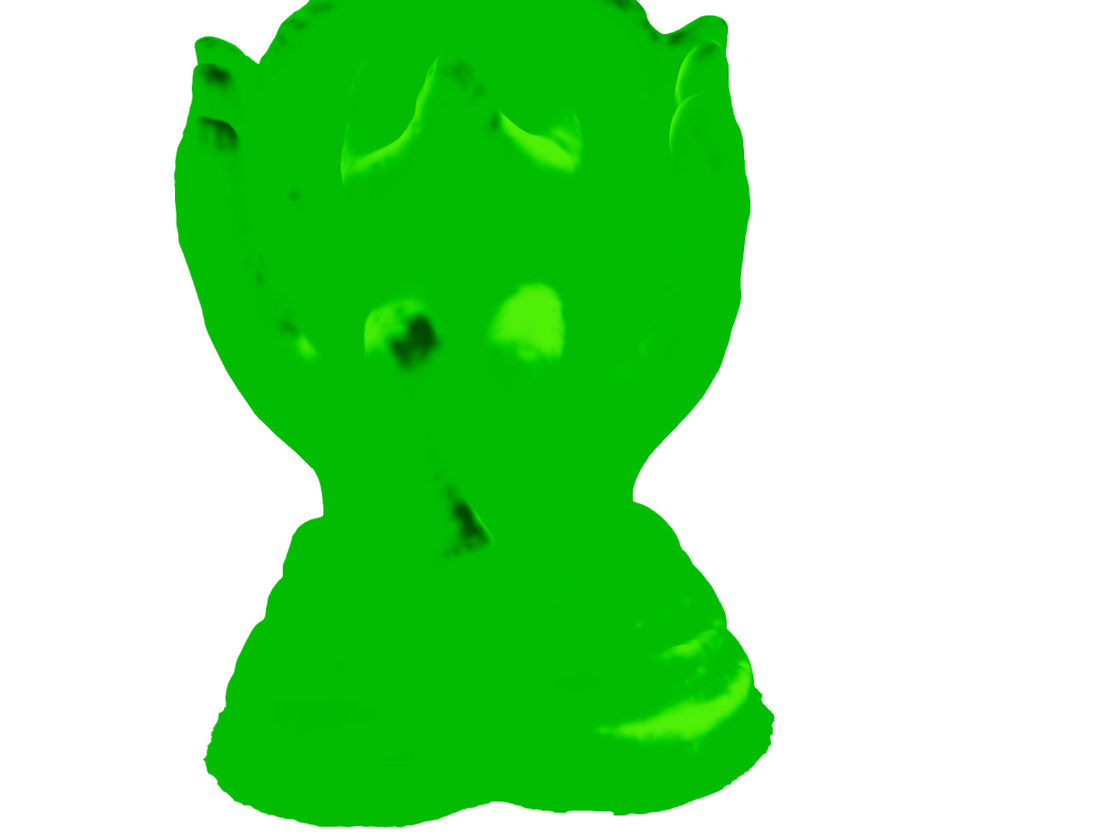}
    \end{subfigure}
    \hspace{1pt}
    \begin{subfigure}[h]{0.14\paperwidth}
        \includegraphics[width=\textwidth]{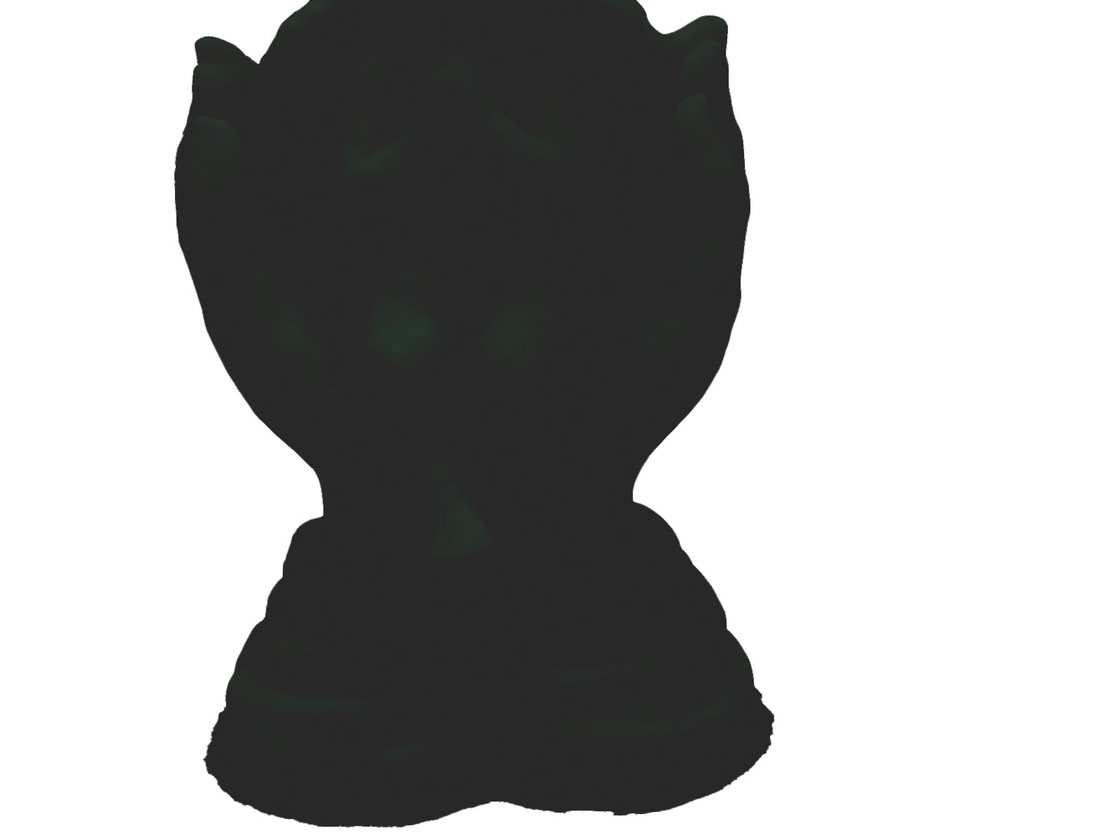}
    \end{subfigure}
    \hspace{1pt}
    \begin{subfigure}[h]{0.14\paperwidth}
        \includegraphics[width=\textwidth]{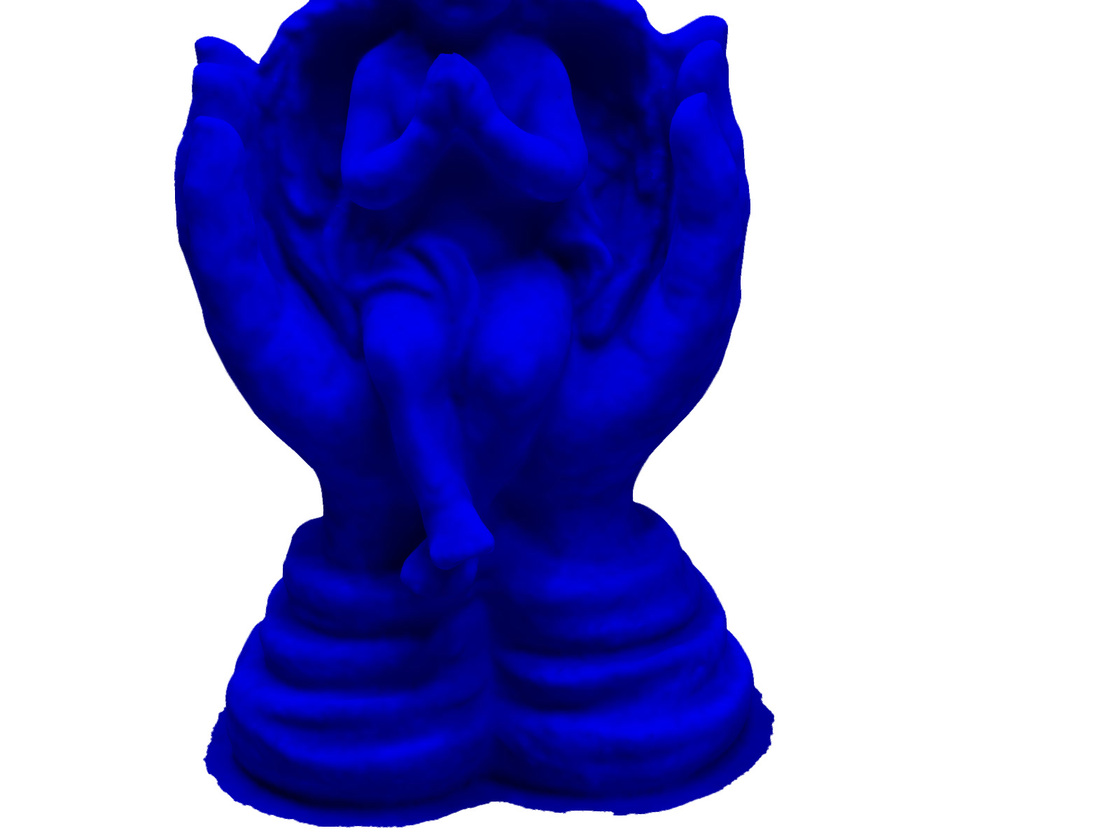}
    \end{subfigure}

    \smallskip
    \rotatebox[origin=b]{90}{scan122}\quad
    \begin{subfigure}[h]{0.14\paperwidth}
        \includegraphics[width=\textwidth]{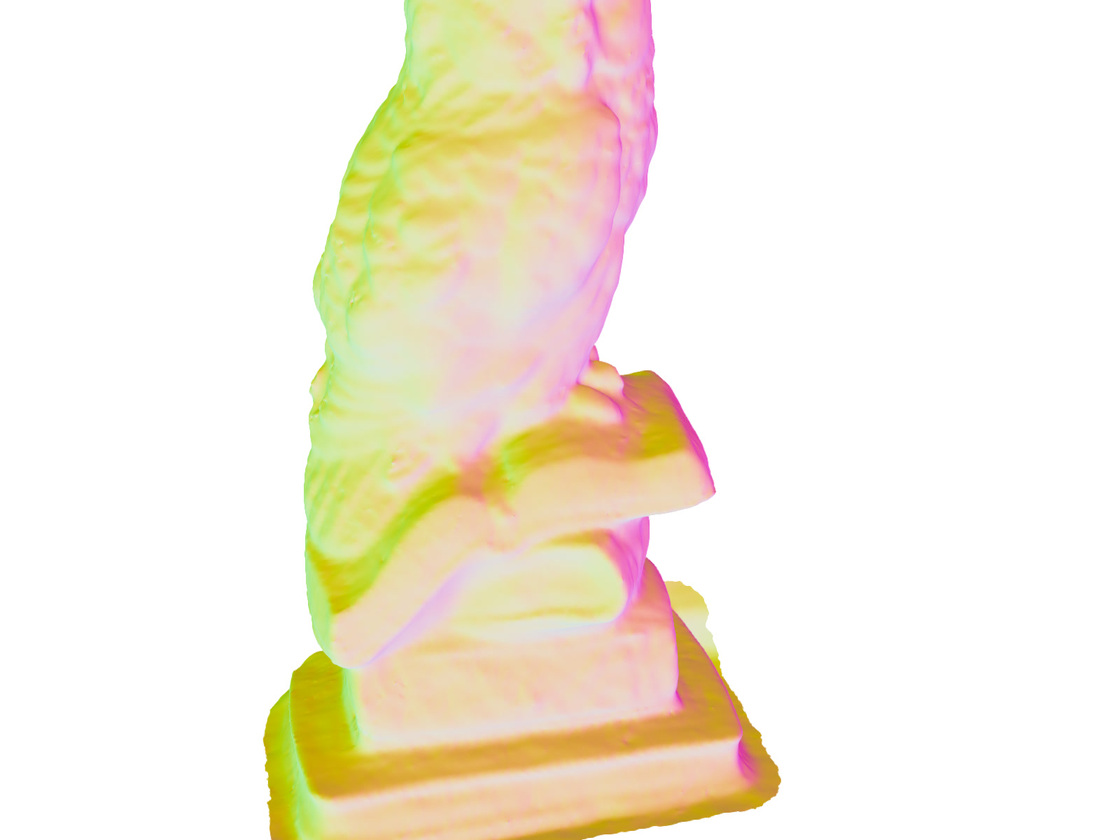}
    \end{subfigure}
    \hspace{1pt}
    \begin{subfigure}[h]{0.14\paperwidth}
        \includegraphics[width=\textwidth]{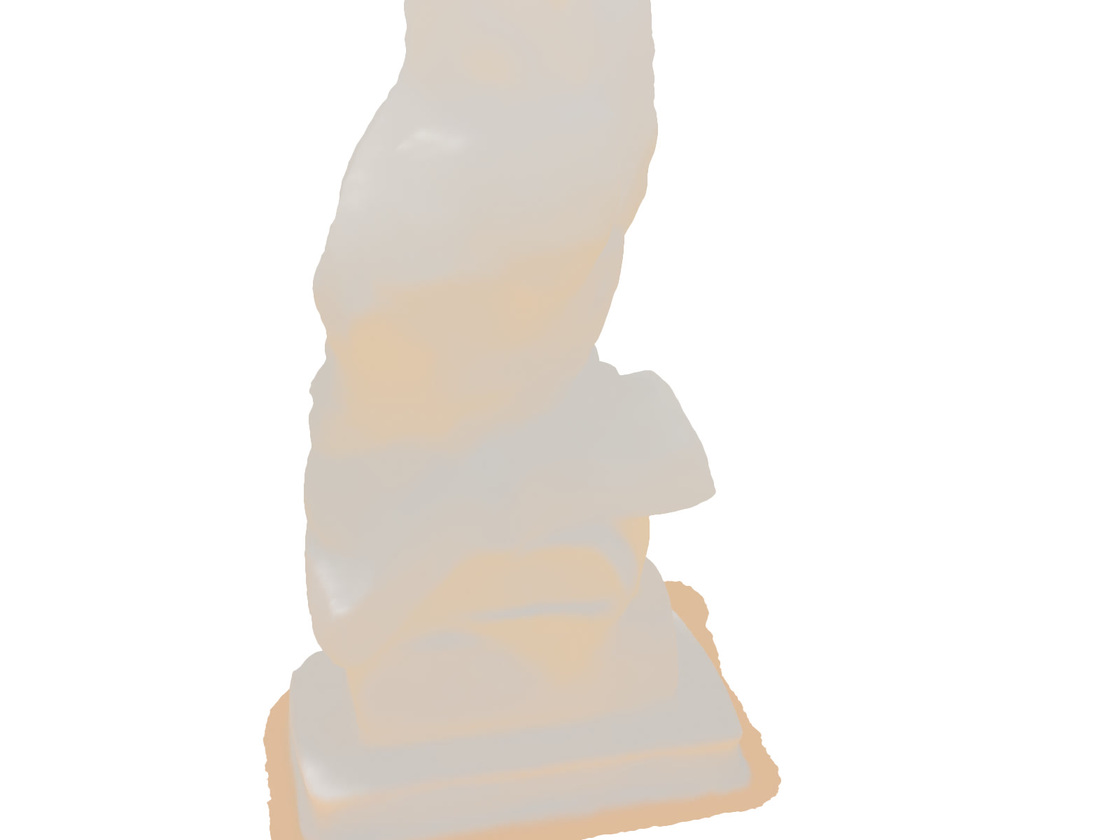}
    \end{subfigure}
    \hspace{1pt}
    \begin{subfigure}[h]{0.14\paperwidth}
        \includegraphics[width=\textwidth]{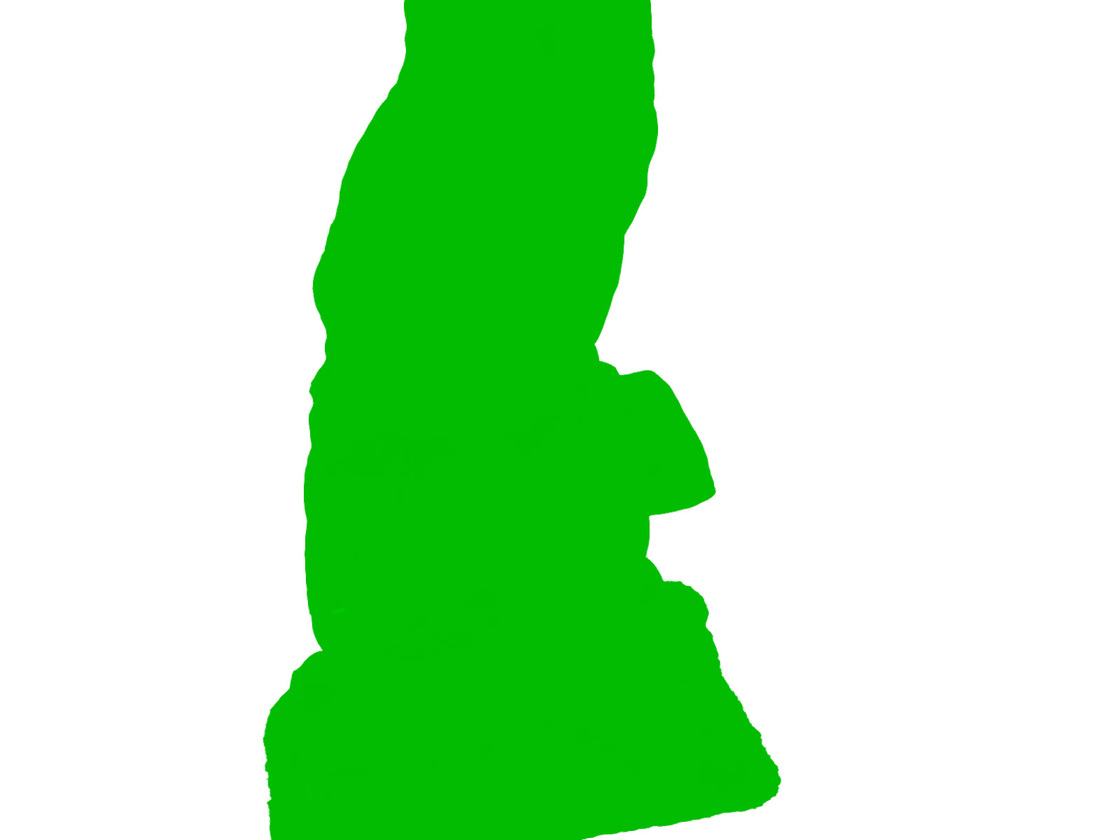}
    \end{subfigure}
    \hspace{1pt}
    \begin{subfigure}[h]{0.14\paperwidth}
        \includegraphics[width=\textwidth]{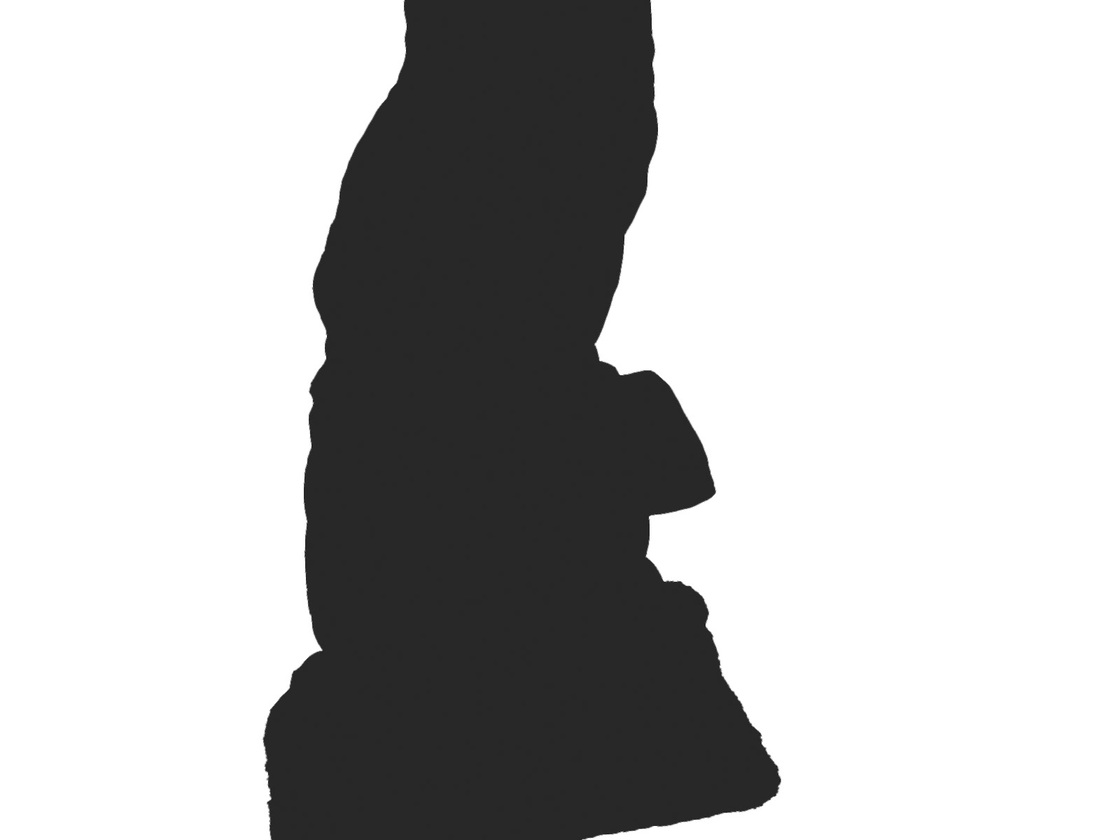}
    \end{subfigure}
    \hspace{1pt}
    \begin{subfigure}[h]{0.14\paperwidth}
        \includegraphics[width=\textwidth]{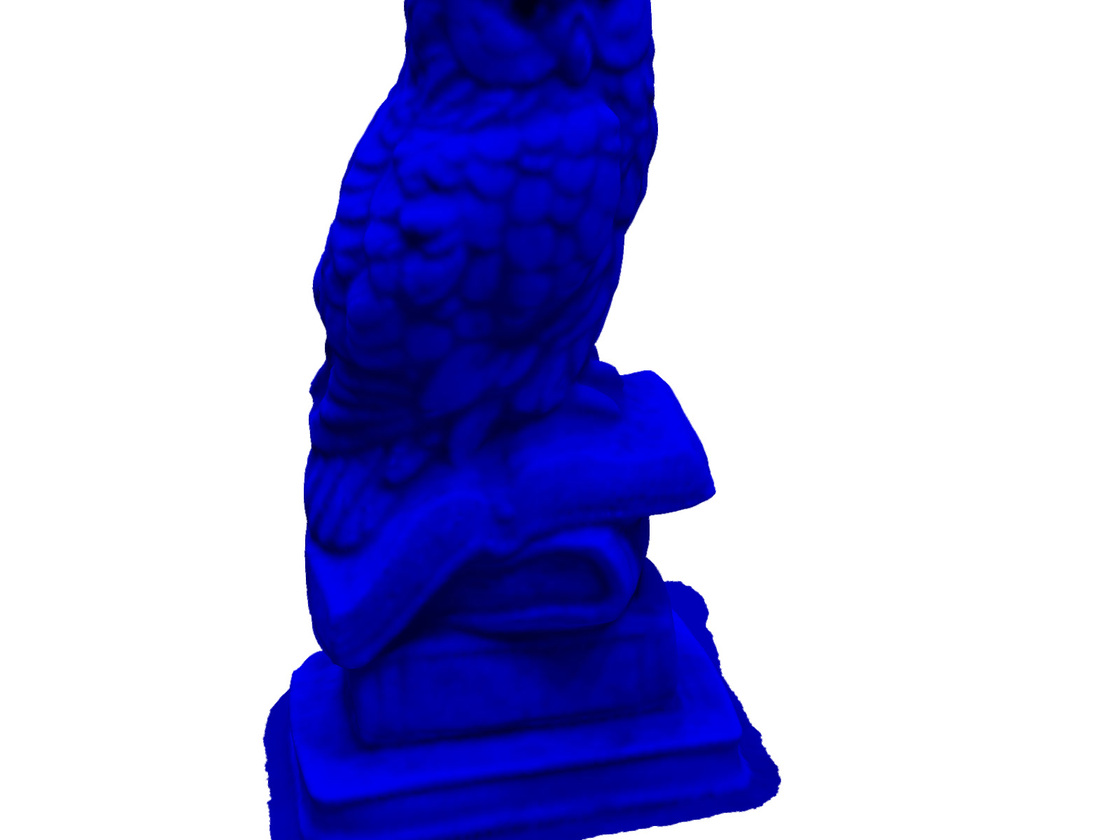}
    \end{subfigure}

    \smallskip
    \rotatebox[origin=b]{90}{scan114}\quad
    \begin{subfigure}[h]{0.17\paperwidth}
        \caption{PBR (default)}
        \includegraphics[width=\textwidth]{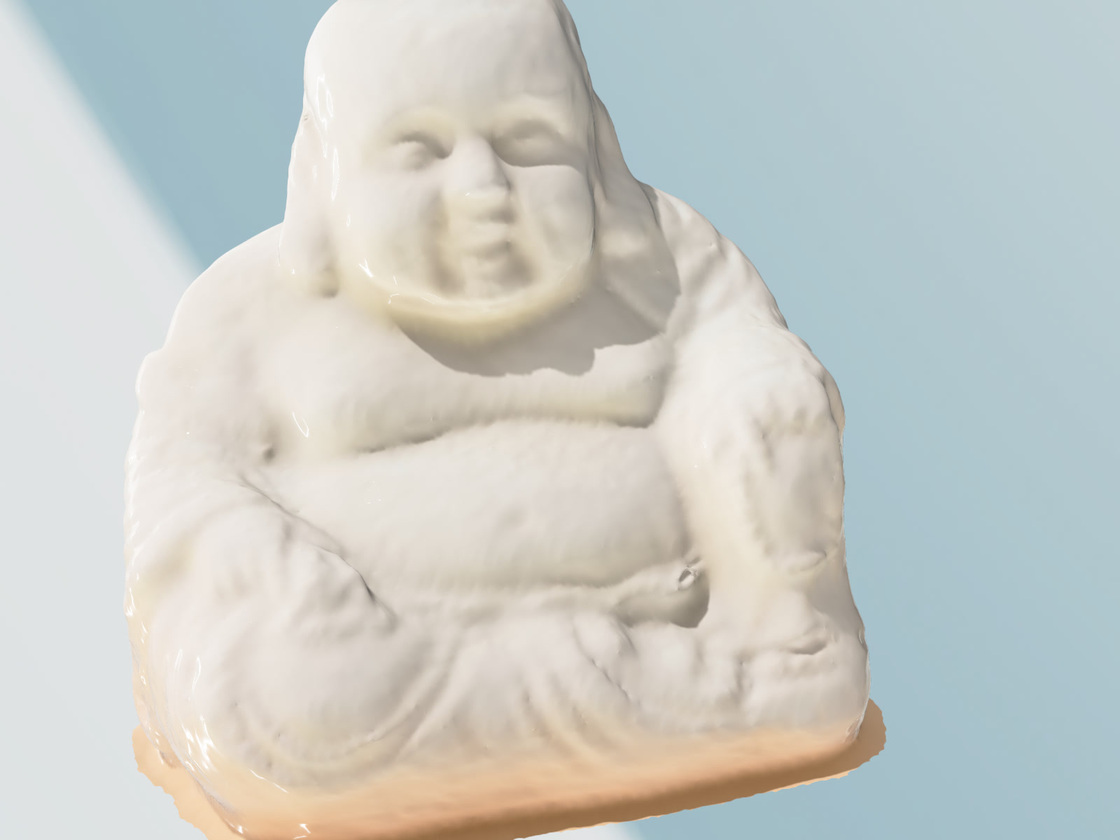}
    \end{subfigure}
    \hspace{1pt}
    \begin{subfigure}[h]{0.17\paperwidth}
        \caption{PBR (pillars)}
        \includegraphics[width=\textwidth]{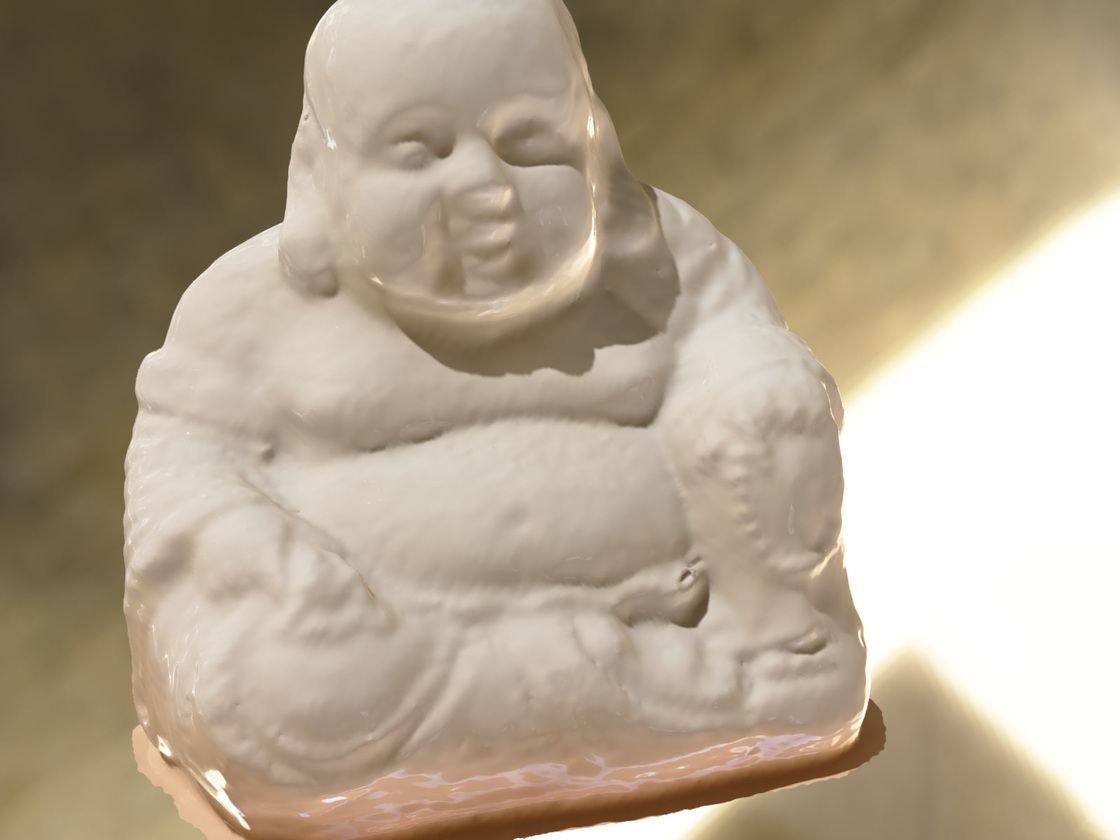}
    \end{subfigure}
    \hspace{1pt}
    \begin{subfigure}[h]{0.17\paperwidth}
        \caption{Neural rendering}
        \includegraphics[width=\textwidth]{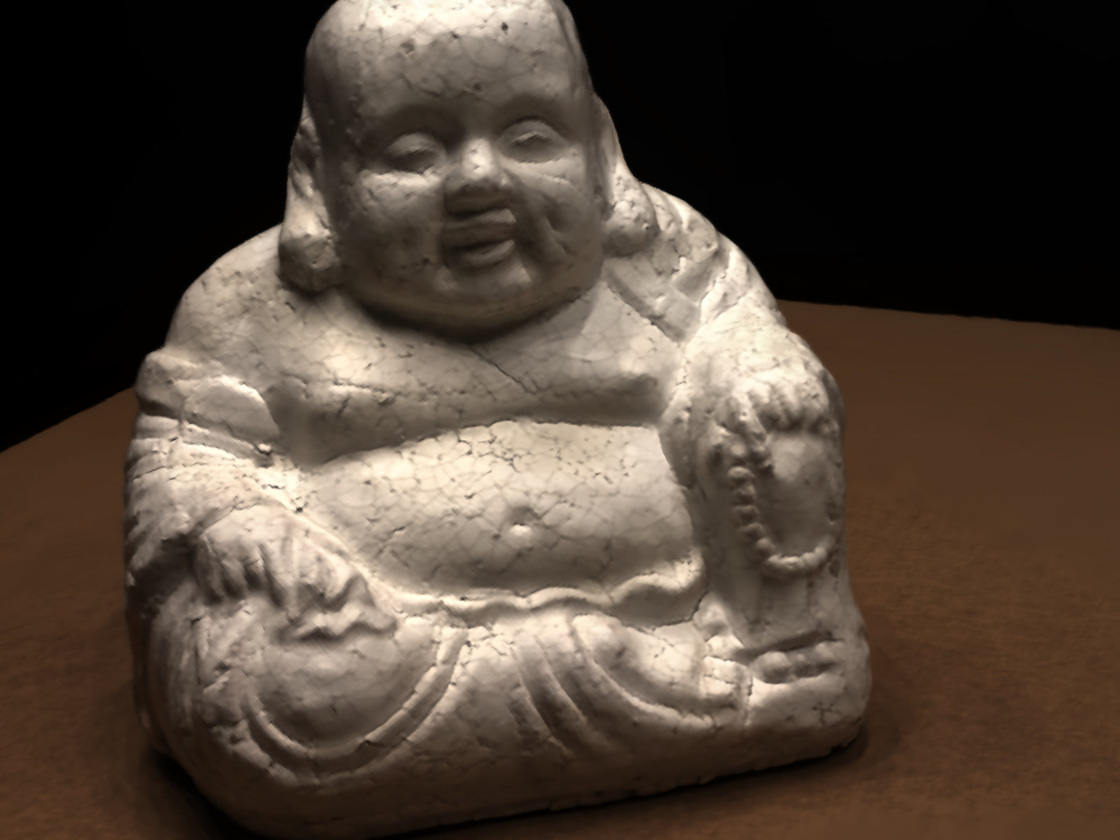}
    \end{subfigure}
    \hspace{1pt}
    \begin{subfigure}[h]{0.17\paperwidth}
        \caption{Groundtruth}
        \includegraphics[width=\textwidth]{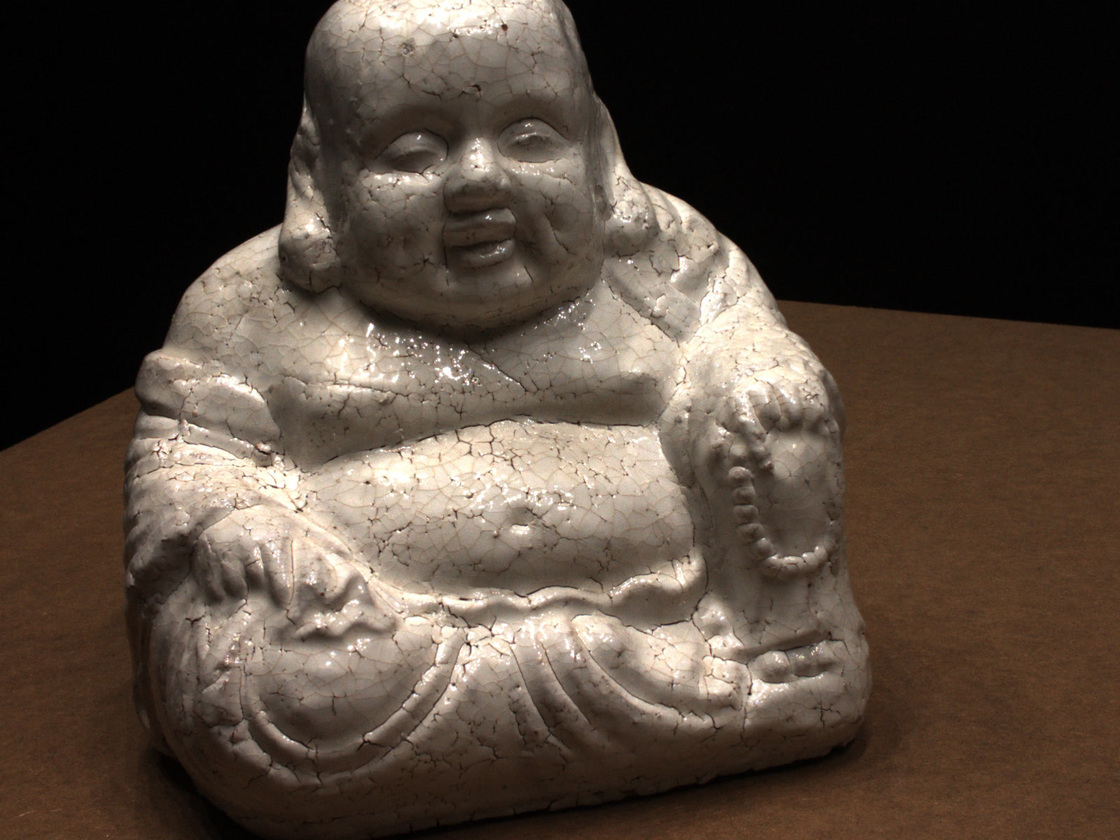}
    \end{subfigure}

    \smallskip
    \rotatebox[origin=b]{90}{scan118}\quad
    \begin{subfigure}[h]{0.17\paperwidth}
        \includegraphics[width=\textwidth]{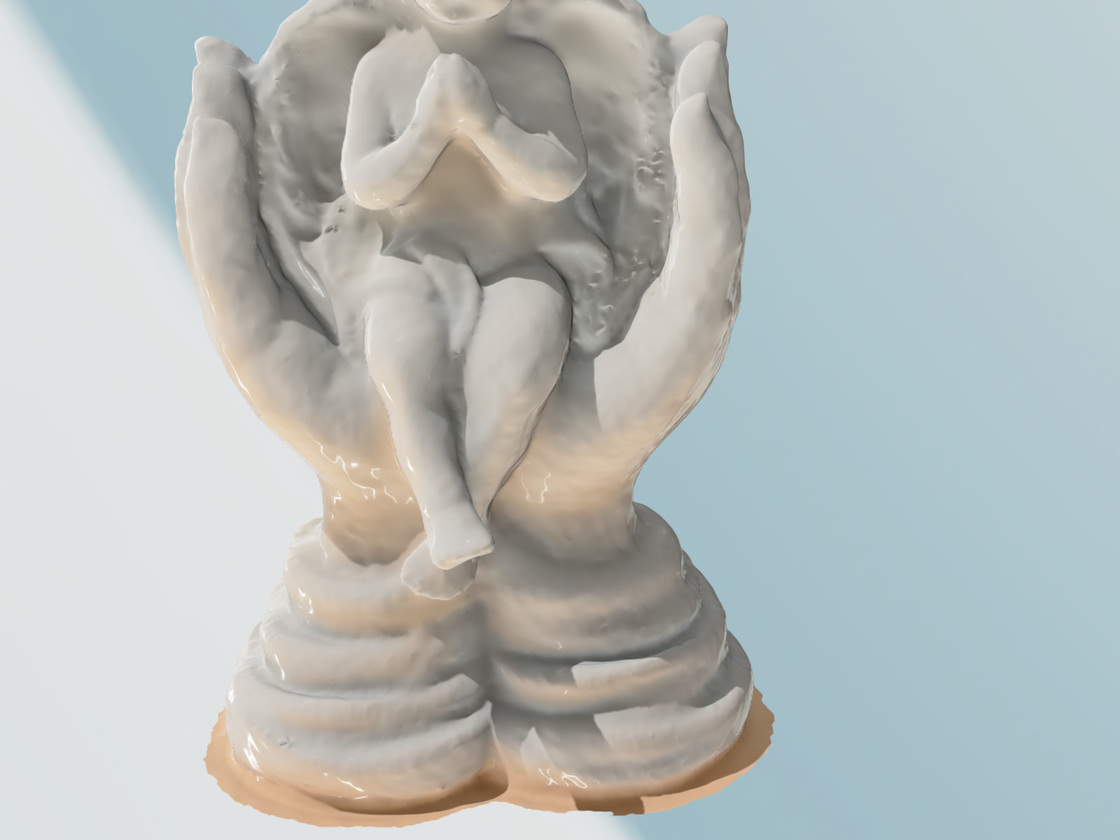}
    \end{subfigure}
    \hspace{1pt}
    \begin{subfigure}[h]{0.17\paperwidth}
        \includegraphics[width=\textwidth]{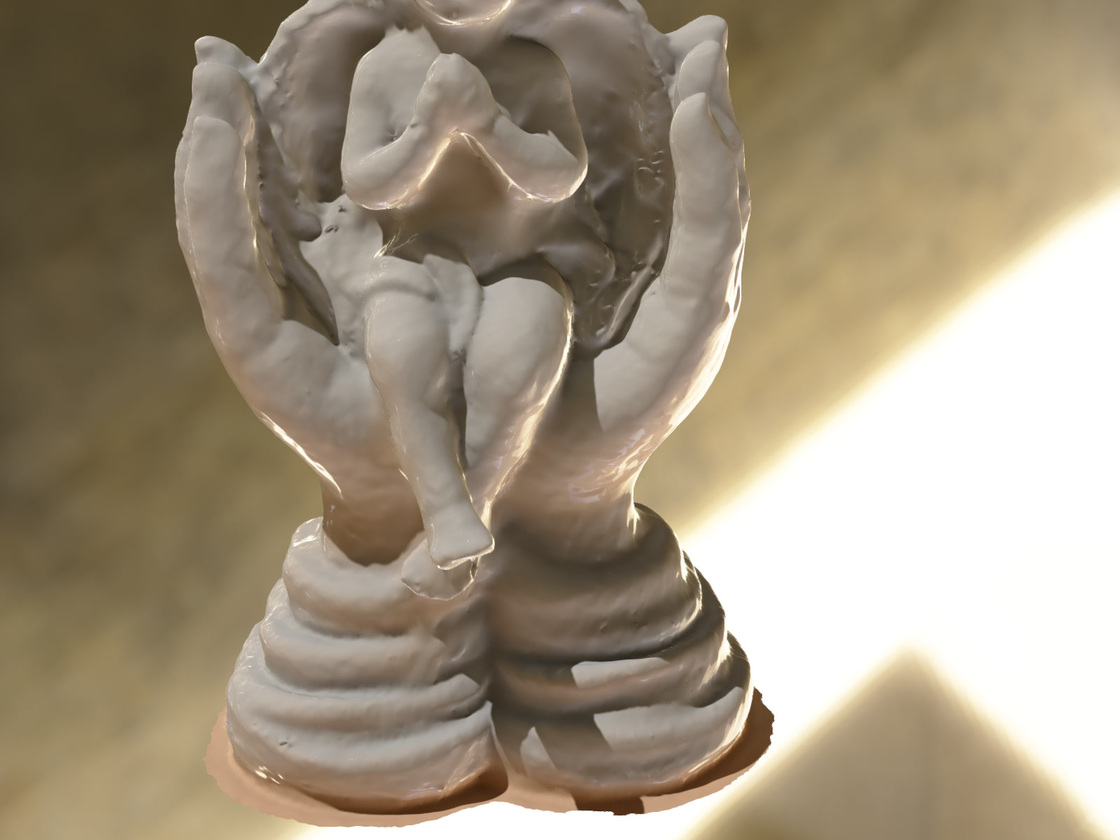}
    \end{subfigure}
    \hspace{1pt}
    \begin{subfigure}[h]{0.17\paperwidth}
        \includegraphics[width=\textwidth]{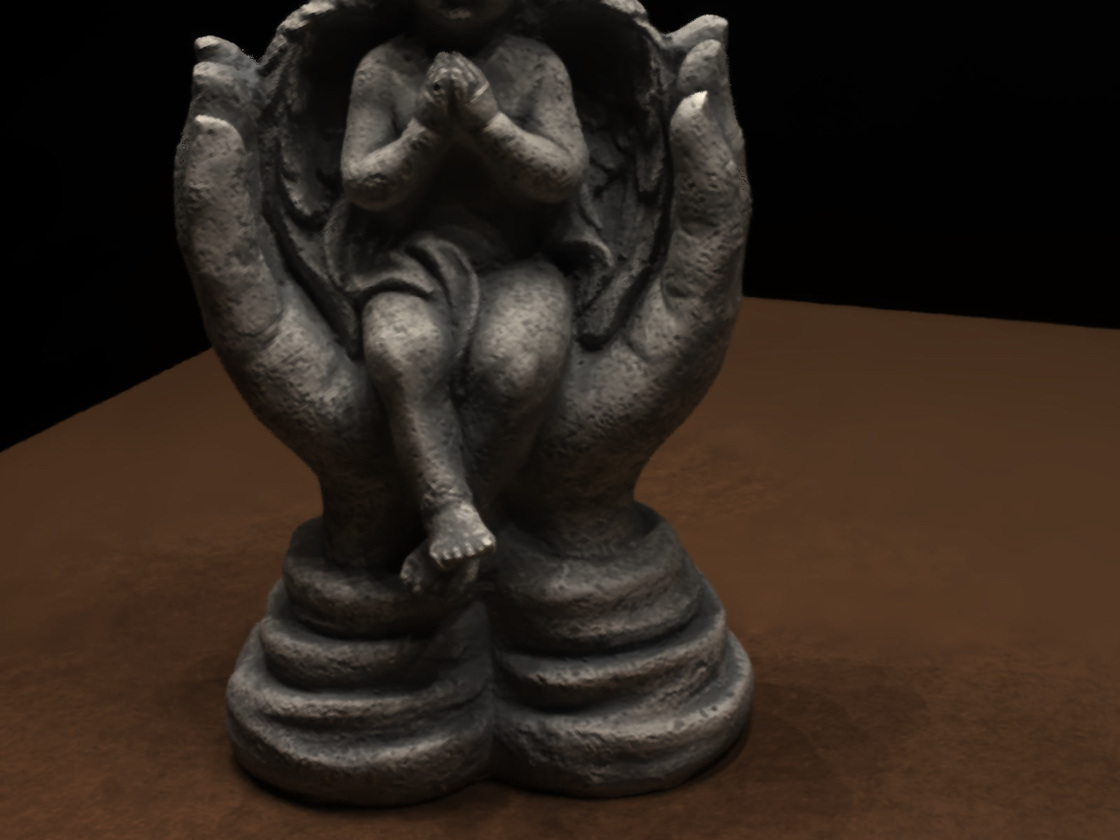}
    \end{subfigure}
    \hspace{1pt}
    \begin{subfigure}[h]{0.17\paperwidth}
        \includegraphics[width=\textwidth]{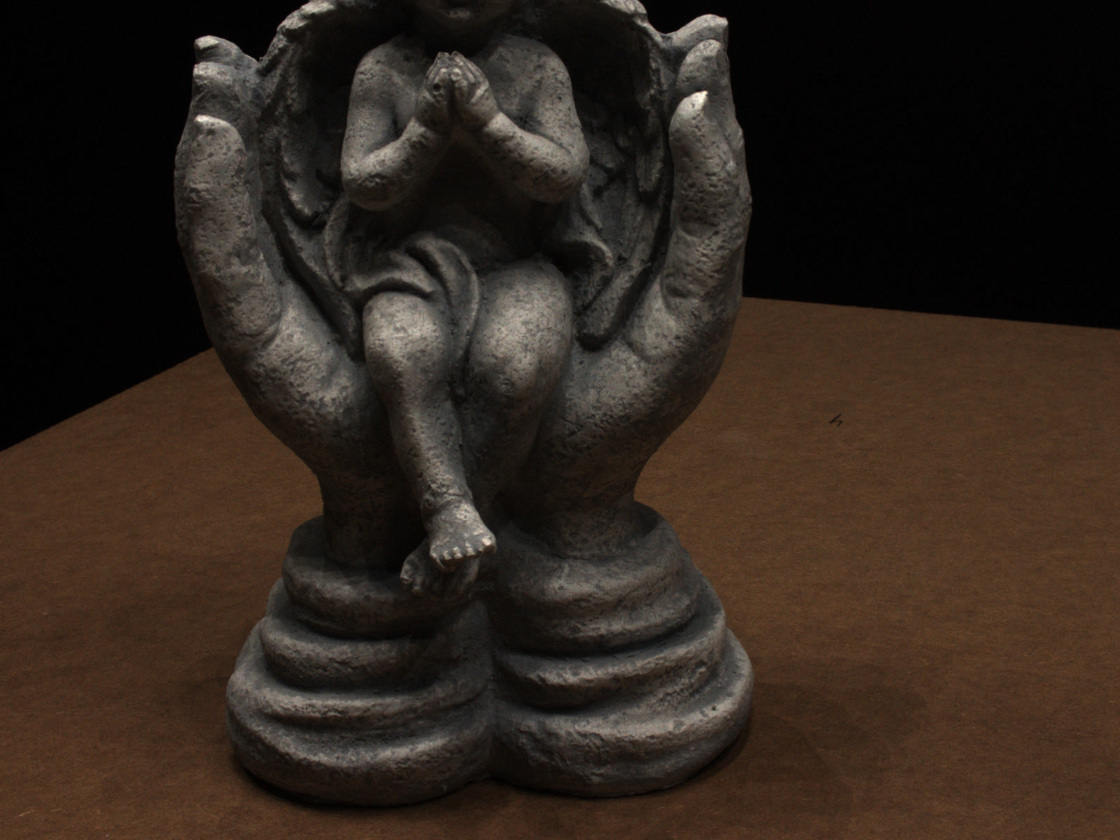}
    \end{subfigure}

    \smallskip
    \rotatebox[origin=b]{90}{scan122}\quad
    \begin{subfigure}[h]{0.17\paperwidth}
        \includegraphics[width=\textwidth]{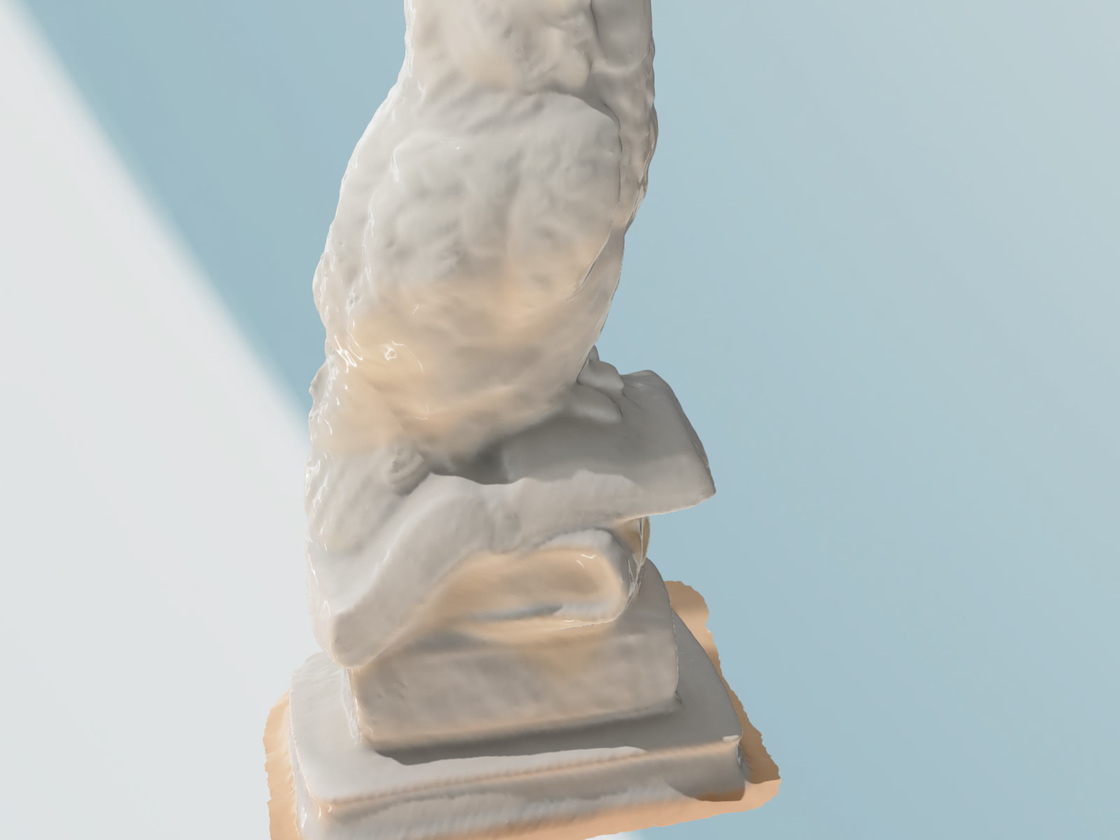}
    \end{subfigure}
    \hspace{1pt}
    \begin{subfigure}[h]{0.17\paperwidth}
        \includegraphics[width=\textwidth]{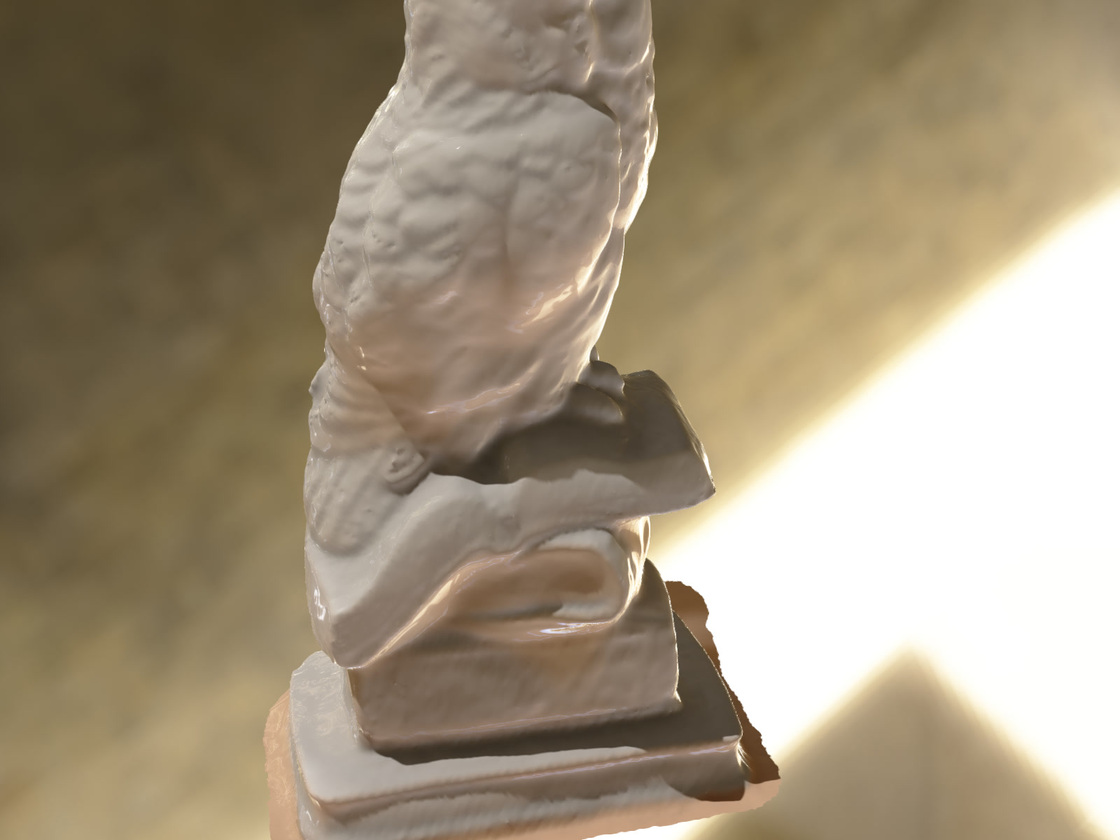}
    \end{subfigure}
    \hspace{1pt}
    \begin{subfigure}[h]{0.17\paperwidth}
        \includegraphics[width=\textwidth]{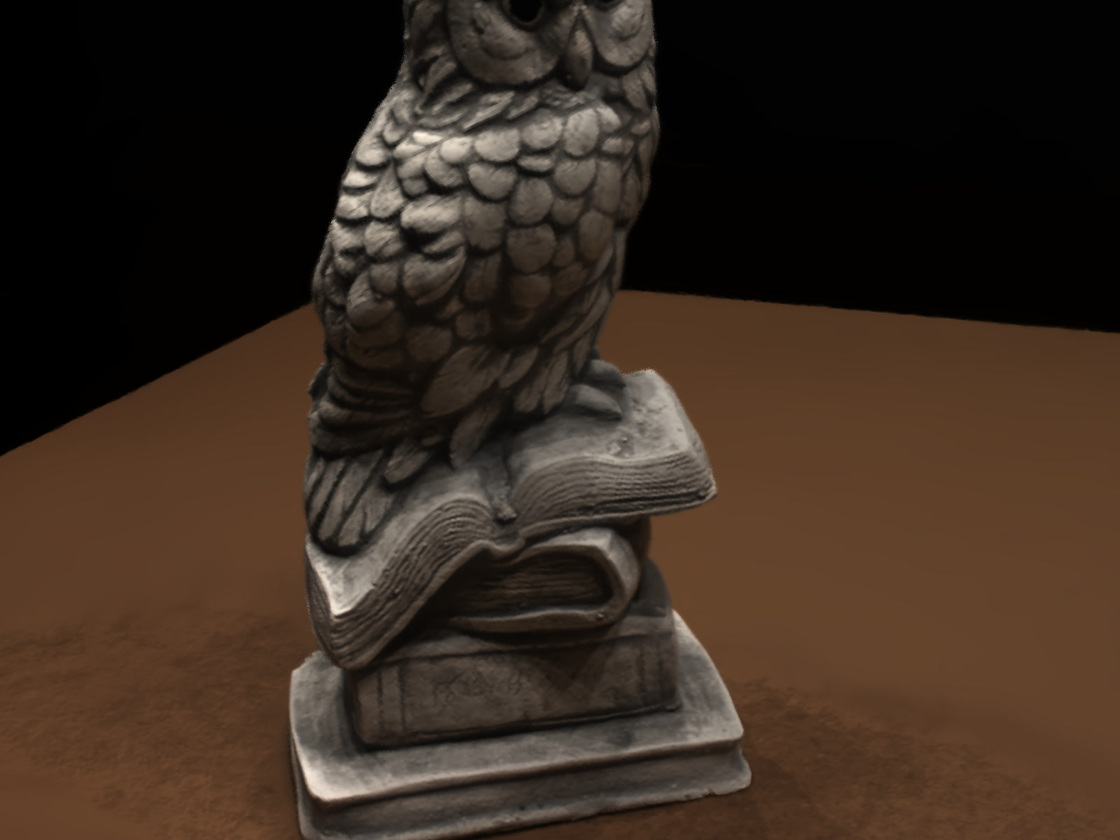}
    \end{subfigure}
    \hspace{1pt}
    \begin{subfigure}[h]{0.17\paperwidth}
        \includegraphics[width=\textwidth]{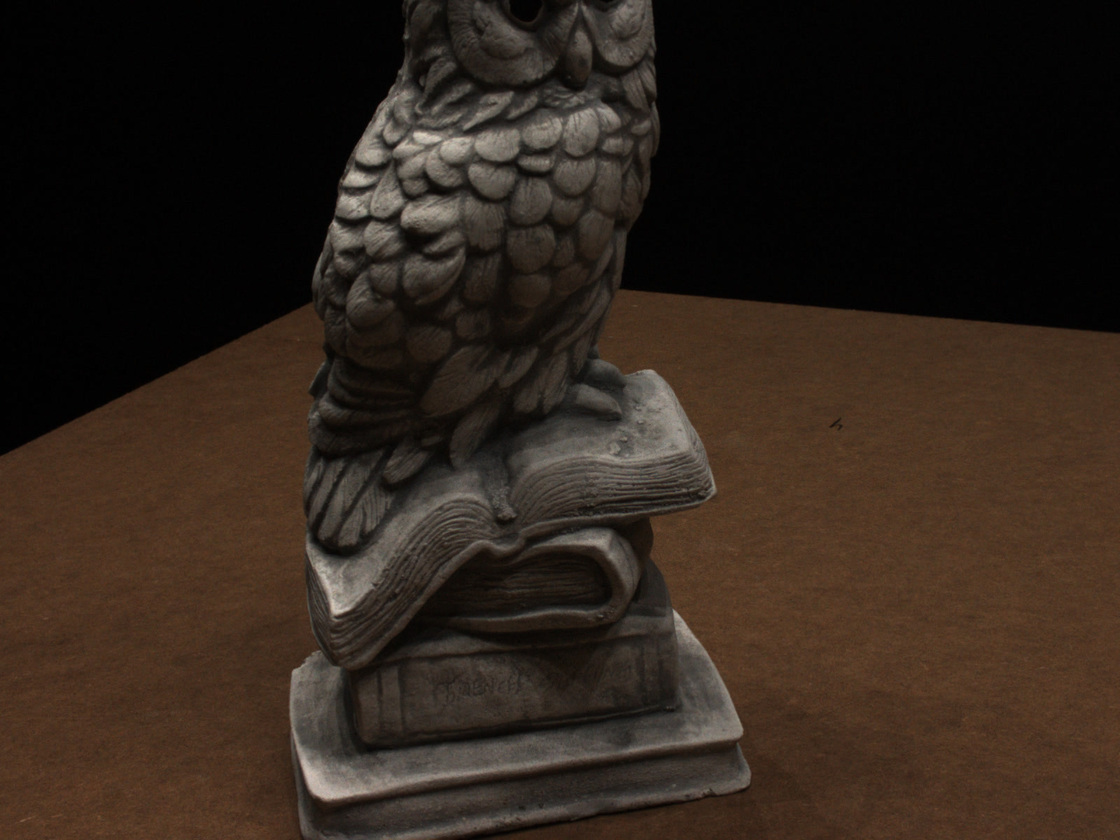}
    \end{subfigure}

    \caption{\textbf{Decomposed materials and rendered images.} ($\cdot$) is a given environment map in Open3D.}
    \label{fig:main_results_04}
\end{figure*}

\textbf{Decomposed geometry, light, and materials:} \cref{fig:main_results_01,fig:main_results_02,fig:main_results_03,fig:main_results_04} show decomposed components of primary results. Geometry is well extracted, base color is flatter, and comprehensively, combinations of roughness, specular reflectance, and implicit illumination exhibits a photogrammetric settings: high intensity and/or specularity corresponds to lower roughness, low specular reflectance shows low specularity, and implicit illumination captures light distribution spatially.

However, there are several failure cases. 1) Black color of base color are removed, e.g., eyes of scan37, scan55, scan69, scan83, and scan109. This might be mainly due to the implicit illumination and photogrammetric networks. Since either network can be able to output $0$, base color can be any values in that case. With strong base color prior, such values are towards colors in neighbors. Nonetheless, this might be trade-off to remove shadows from base color, we can observe clear removals of shadow in base color, e.g, the block shadow in scan40 and hat shadow in scan69. To some extent, we can add black color on base color as seen in the rebaking examples of the main script at the cost of non-flat base color and baked shadow. Further study of better light model is needed in order to mitigate such drawbacks. 2) Quality of geometric reconstruction is highly affected by hight specularity, especially on yellow colors. On some yellow colored regions with high specularity, e.g., in scan63 and scan110, geometric reconstructions clearly worse. Correspondingly, the roughness is intuitively flipped on such regions in scan63. For better geometric reconstruction on such regions, we may need more view-consistent constraints as modeled in \cite{DBLP:conf/cvpr/DarmonBDMA22}. 3) As noted in the main script, we do not model metalness, so decomposing metallic object is not optimal.

\textbf{Rendered images:} \cref{fig:main_results_01,fig:main_results_02,fig:main_results_03,fig:main_results_04} also show rendered images of primary results. Neural rendered images look real, and PBR images are towards photorealism.

\begin{figure*}[tbp]
    \centering
    \begin{subfigure}[h]{0.18\paperwidth}
        \includegraphics[width=\textwidth]{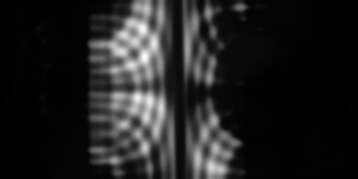}
        \caption{scan24 (0.22, 43.24)}
    \end{subfigure}
    \hspace{1pt}
    \begin{subfigure}[h]{0.18\paperwidth}
        \includegraphics[width=\textwidth]{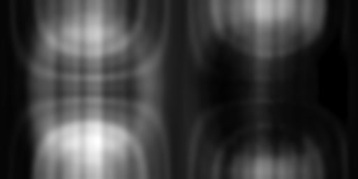}
        \caption{scan37 (0.21, 3641.38)}
    \end{subfigure}
    \hspace{1pt}
    \begin{subfigure}[h]{0.18\paperwidth}
        \includegraphics[width=\textwidth]{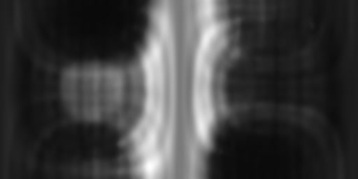}
        \caption{scan40 (1.05, 26.24)}
    \end{subfigure}
    \hspace{1pt}
    \begin{subfigure}[h]{0.18\paperwidth}
        \includegraphics[width=\textwidth]{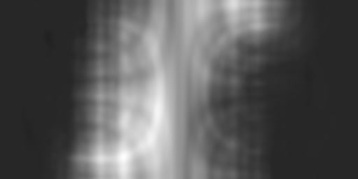}
        \caption{scan55 (0.54, 4.23)}
    \end{subfigure}

    \smallskip
    \begin{subfigure}[h]{0.18\paperwidth}
        \includegraphics[width=\textwidth]{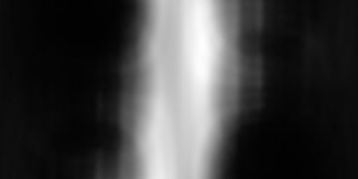}
        \caption{scan63 (2.50, 277.83)}
    \end{subfigure}
    \hspace{1pt}
    \begin{subfigure}[h]{0.18\paperwidth}
        \includegraphics[width=\textwidth]{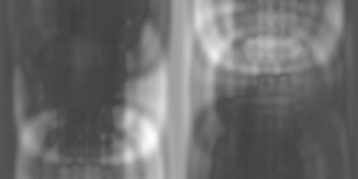}
        \caption{scan65 (0.64, 2.63)}
    \end{subfigure}
    \hspace{1pt}
    \begin{subfigure}[h]{0.18\paperwidth}
        \includegraphics[width=\textwidth]{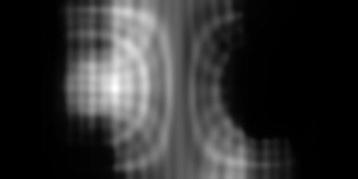}
        \caption{scan69 (0.75, 286.02)}
    \end{subfigure}
    \hspace{1pt}
    \begin{subfigure}[h]{0.18\paperwidth}
        \includegraphics[width=\textwidth]{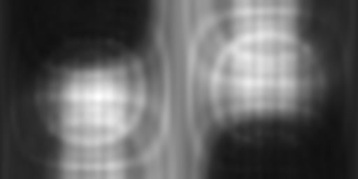}
        \caption{scan83 (146.73, 2263.60)}
    \end{subfigure}

    \smallskip
    \begin{subfigure}[h]{0.18\paperwidth}
        \includegraphics[width=\textwidth]{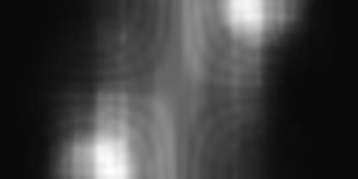}
        \caption{scan97 (1.01, 25.89)}
    \end{subfigure}
    \hspace{1pt}
    \begin{subfigure}[h]{0.18\paperwidth}
        \includegraphics[width=\textwidth]{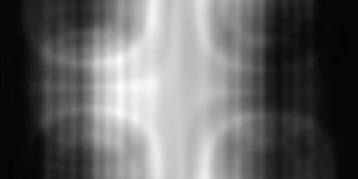}
        \caption{scan105 (35.49, 1650.91)}
    \end{subfigure}
    \hspace{1pt}
    \begin{subfigure}[h]{0.18\paperwidth}
        \includegraphics[width=\textwidth]{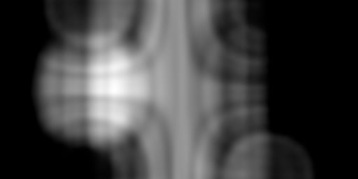}
        \caption{scan106 (0.66, 1138.29)}
    \end{subfigure}
    \hspace{1pt}
    \begin{subfigure}[h]{0.18\paperwidth}
        \includegraphics[width=\textwidth]{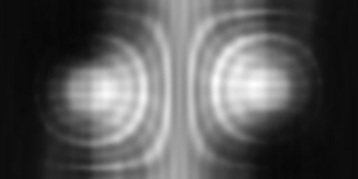}
        \caption{scan110 (45.94, 1494.81)}
    \end{subfigure}

    \smallskip
    \begin{subfigure}[h]{0.18\paperwidth}
        \includegraphics[width=\textwidth]{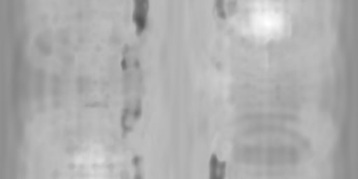}
        \caption{scan114 (0.26, 0.80)}
    \end{subfigure}
    \hspace{1pt}
    \begin{subfigure}[h]{0.18\paperwidth}
        \includegraphics[width=\textwidth]{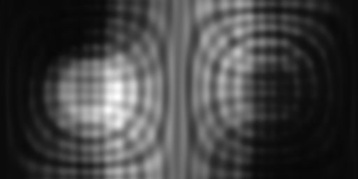}
        \caption{scan118 (35.85, 1324.66)}
    \end{subfigure}
    \hspace{1pt}
    \begin{subfigure}[h]{0.18\paperwidth}
        \includegraphics[width=\textwidth]{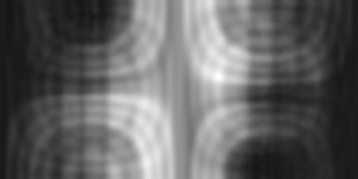}
        \caption{scan122 (64.61, 740.62)}
    \end{subfigure}

    \caption{\textbf{Decomposed environment lights} of all scenes. Values are the actual (min, max) values. For visibility, values are normalized such that the maximum is $255$}
    \label{fig:decomposed_environment_lights}
\end{figure*}

\textbf{Decomposed environment lights:} \cref{fig:decomposed_environment_lights} illustrates decomposed environment lights of all scenes. All environment lights exhibit non-uniformity and capture the DTU MVS light setting except for the one of scan114 which clearly fails to decompose environment lights and shows strong uniformity.

\subsection{More analysis and ablation study}
\label{subsec:more_analysis_and_ablation_study}

\begin{figure}[tbp]
    \centering
    \captionsetup[subfigure]{font=scriptsize}
    
    \begin{subfigure}[h]{0.18\paperwidth}
        \includegraphics[width=\textwidth]{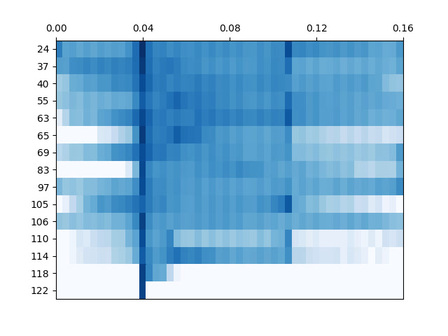}
        \caption{NDJIR}
    \end{subfigure}
    \hspace{1pt}
    \begin{subfigure}[h]{0.18\paperwidth}
        \includegraphics[width=\textwidth]{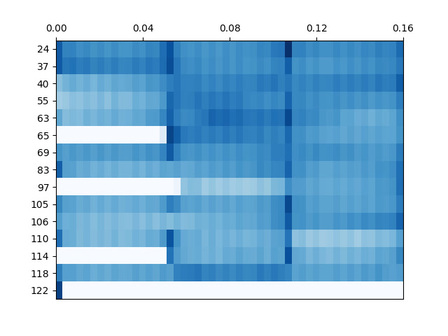}
        \caption{w/o priors ($128$)}
    \end{subfigure}

    \smallskip
    \begin{subfigure}[h]{0.18\paperwidth}
        \includegraphics[width=\textwidth]{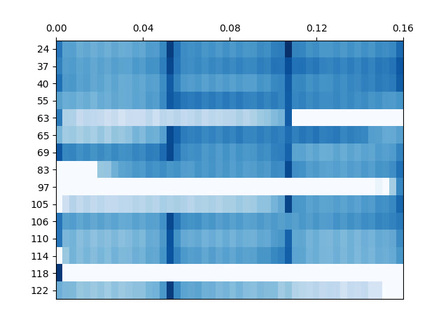}
        \caption{w/o priors ($32$)}
    \end{subfigure}
    \hspace{1pt}
    \begin{subfigure}[h]{0.18\paperwidth}
        \includegraphics[width=\textwidth]{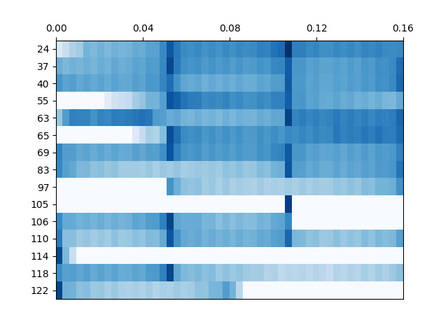}
        \caption{w/o priors ($8$)}
    \end{subfigure}

    \caption{\textbf{Specular reflectance distributions} with varying number of light samples per pixel ($N$) over all scenes of DTU MVS dataset. Specular reflectance of $3$ channels are averaged.}
    \label{fig:specular_reflectance_distributions}
\end{figure}

\textbf{Distribution of specular reflectance:} In \cref{fig:specular_reflectance_distributions}, similar to the tends of the roughness distribution as in the main script, when we increase the number of light samples, especially to $128$, degeneration of the network mitigates. Using Bayesian prior \cite{DBLP:conf/nips/KendallG17}, we can clearly observe the peak of the distribution around $0.04$.

\begin{figure}[tbp]
    \centering
    \footnotesize
    \captionsetup[subfigure]{font=scriptsize}
    
    \rotatebox[origin=b]{90}{scan55}\quad
    \begin{subfigure}[h]{0.16\paperwidth}
        \caption{importance}
        \includegraphics[width=\textwidth]{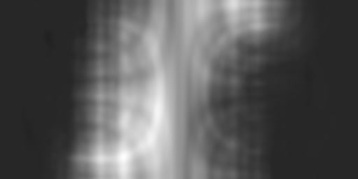}
        \caption{(0.54, 4.23)}        
    \end{subfigure}
    \hspace{1pt}
    \begin{subfigure}[h]{0.16\paperwidth}
        \caption{uniform}
        \includegraphics[width=\textwidth]{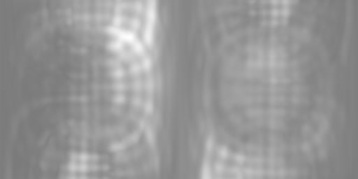}
        \caption{(0.74, 1.53)}
    \end{subfigure}

    \smallskip
    \rotatebox[origin=b]{90}{scan69}\quad
    \begin{subfigure}[h]{0.16\paperwidth}
        \includegraphics[width=\textwidth]{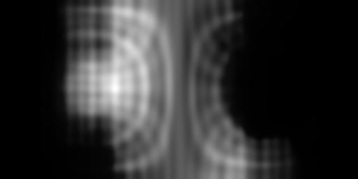}
        \caption{(0.75, 286.02)}
    \end{subfigure}
    \hspace{1pt}
    \begin{subfigure}[h]{0.16\paperwidth}
        \includegraphics[width=\textwidth]{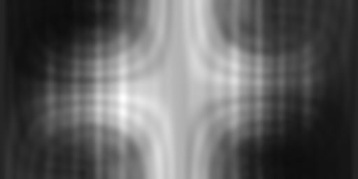}
        \caption{(109.57, 2168.52)}
    \end{subfigure}
    
    \caption{\textbf{Comparison of decomposed environment lights} between uniform sampling and importance sampling. Values below are the actual (min, max) values. For visibility, values are normalized such that the maximum is $255$}
    \label{fig:environment_lights_with_samplings}
\end{figure}

\textbf{Decomposed environment lights with uniform sampling:} \cref{fig:environment_lights_with_samplings} shows difference of the decomposed environment maps between using importance and uniform samplings. When we use the importance sampling, clear separation of light intensity on the upper-hemisphere is observed. On the upper-hemisphere, the environment light intensity is strong; on the other hand, the environment light intensity is quite low on the lower-hemisphere, or on the floor. This indicates that modeling the specular term with importance sampling contributes to better decomposition of environment light.

\section{Custom dataset}
\label{sec:custom_dataset}

Scenes are captured in an author's room where an object is located on a small white desk, walls color is also white, the sun light through a window is completely blocked by a curtain, and a flat ceiling light is the only direct light source. 

\subsection{Preprocessing pipeline overview}
\label{subsec:preprocessing_pipeline_overview}

With custom dataset, the overall pipeline of preprocessing is as follows: 

\begin{enumerate}
    \item Take video and extract images
    \item Deblur images
    \item Create object masks
    \item Estimate camera parameters
    \item Normalize camera poses
\end{enumerate}

When we take videos, over the upper hemisphere of an object, one by her(him)self moves a smart-phone camera around $360 ^{\circ}$ over azimuthal angle for each $45 ^{\circ}$ and $90 ^{\circ}$ in polar angle. Some camera features, e.g., AI-enhancement and/or camera shake correction are not used. Approximately, $100$ images are uniformly extracted. For each image, we apply deblurring \cite{chen2022simple} and background matting \cite{rembg}, then use COLMAP \cite{schoenberger2016sfm,schoenberger2016mvs} with such images and masks to estimate camera poses and an intrinsic. Finally, we normalize camera poses such that their visual hulls are contained in the unit sphere \cite{DBLP:conf/nips/YarivKMGABL20}.

\subsection{Result of custom dataset}
\label{subsec:result_of_custom_dataset}

\begin{figure*}[tbp]
    \centering
    \captionsetup[subfigure]{font=scriptsize}
    \rotatebox[origin=b]{90}{Decomposed materials}\quad
    \begin{subfigure}[h]{0.14\paperwidth}
        \caption{normals}
        \includegraphics[width=\textwidth]{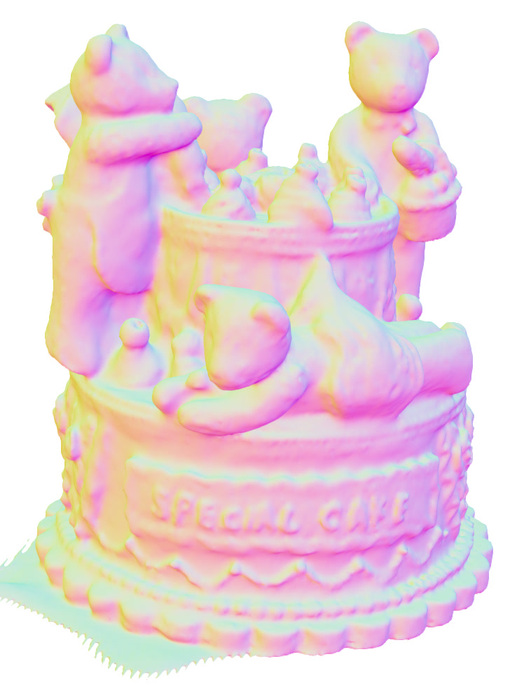}
    \end{subfigure}
    \hspace{1pt}
    \begin{subfigure}[h]{0.14\paperwidth}
        \caption{base color}
        \includegraphics[width=\textwidth]{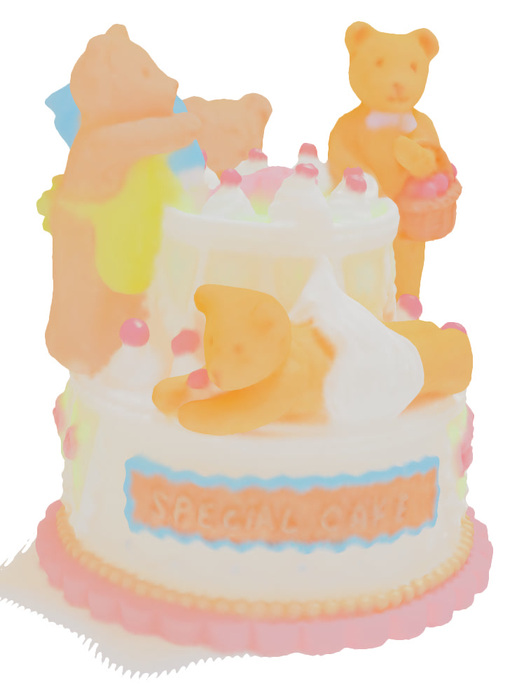}
    \end{subfigure}
    \hspace{1pt}
    \begin{subfigure}[h]{0.14\paperwidth}
        \caption{roughness}
        \includegraphics[width=\textwidth]{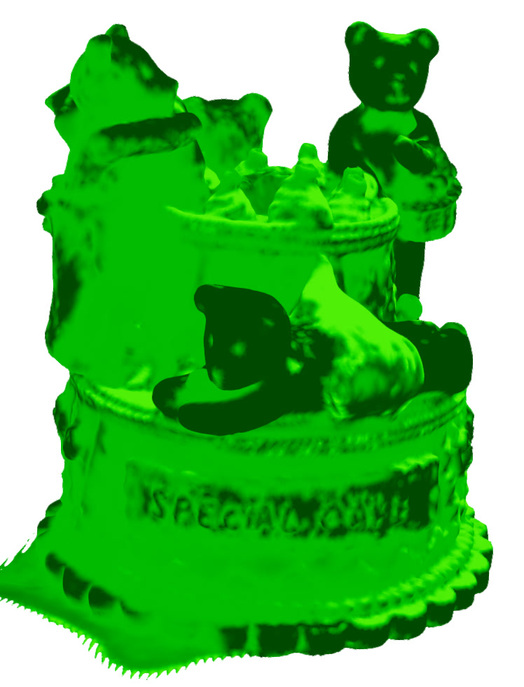}
    \end{subfigure}
    \hspace{1pt}
    \begin{subfigure}[h]{0.14\paperwidth}
        \caption{specular reflectance}
        \includegraphics[width=\textwidth]{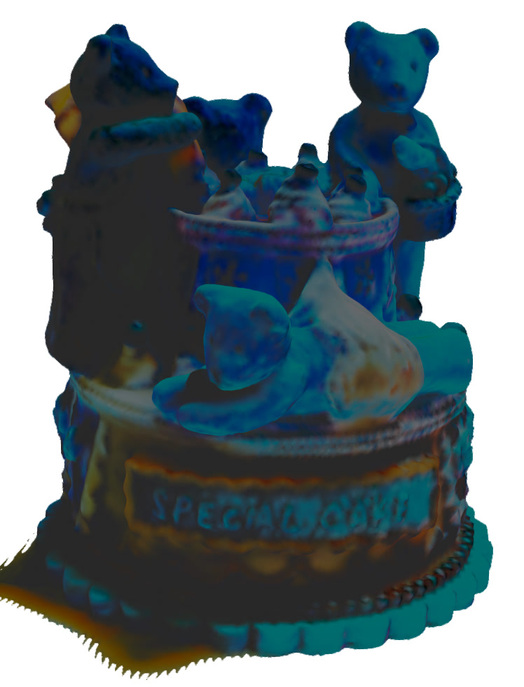}
    \end{subfigure}
    \hspace{1pt}
    \begin{subfigure}[h]{0.14\paperwidth}
        \caption{implicit illumination}
        \includegraphics[width=\textwidth]{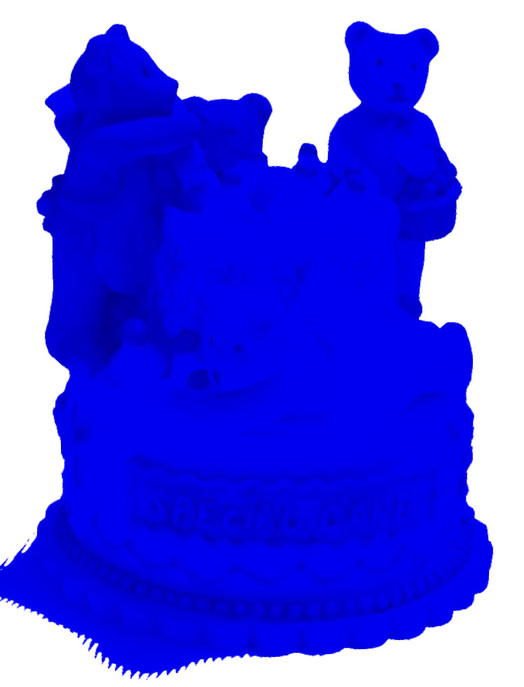}
    \end{subfigure}

    \smallskip
    \rotatebox[origin=b]{90}{PBR images}\quad
    \begin{subfigure}[h]{0.17\paperwidth}
        \caption{PBR (default)}
        \includegraphics[width=\textwidth]{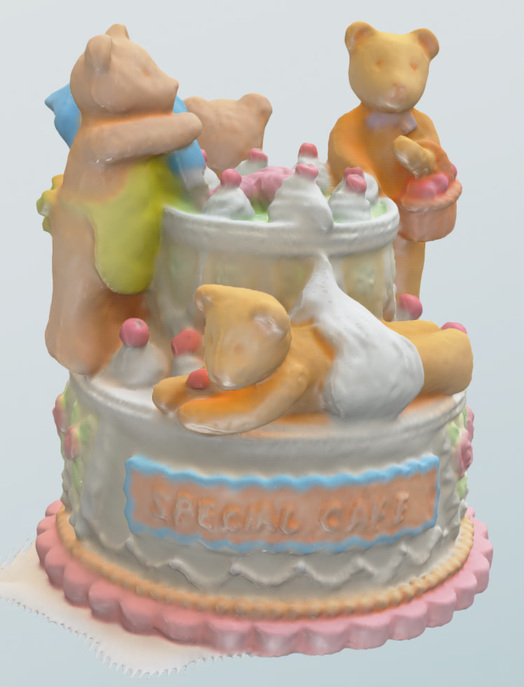}
    \end{subfigure}
    \hspace{1pt}
    \begin{subfigure}[h]{0.17\paperwidth}
        \caption{PBR (pillars)}
        \includegraphics[width=\textwidth]{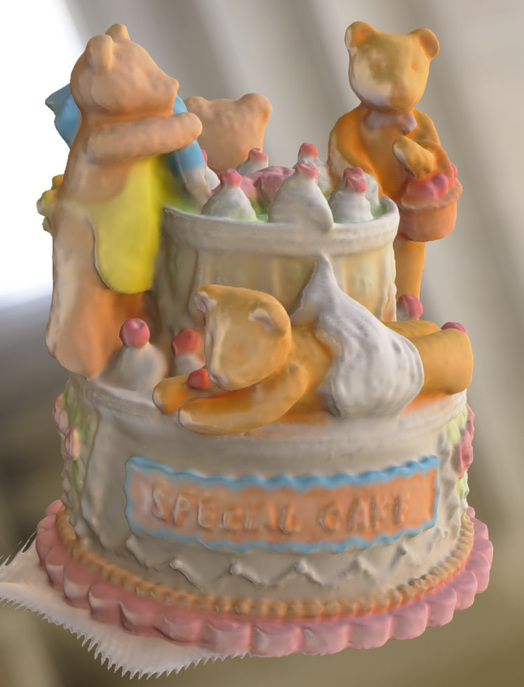}
    \end{subfigure}
    \hspace{1pt}
    \begin{subfigure}[h]{0.17\paperwidth}
        \caption{Neural rendering}
        \includegraphics[width=\textwidth]{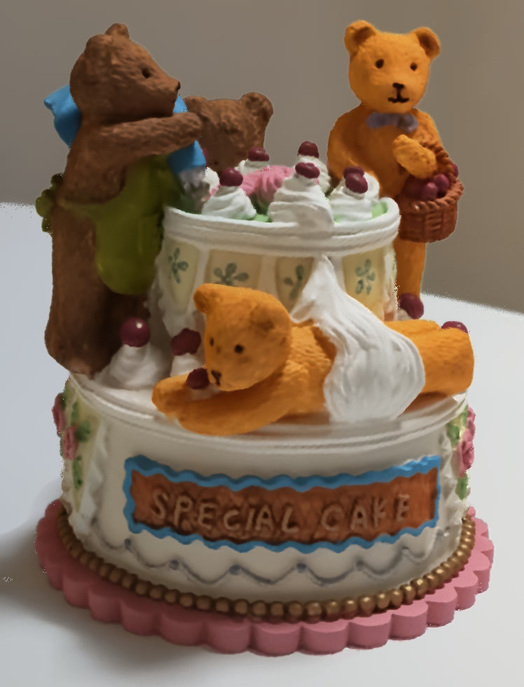}
    \end{subfigure}
    \hspace{1pt}
    \begin{subfigure}[h]{0.17\paperwidth}
        \caption{Groundtruth}
        \includegraphics[width=\textwidth]{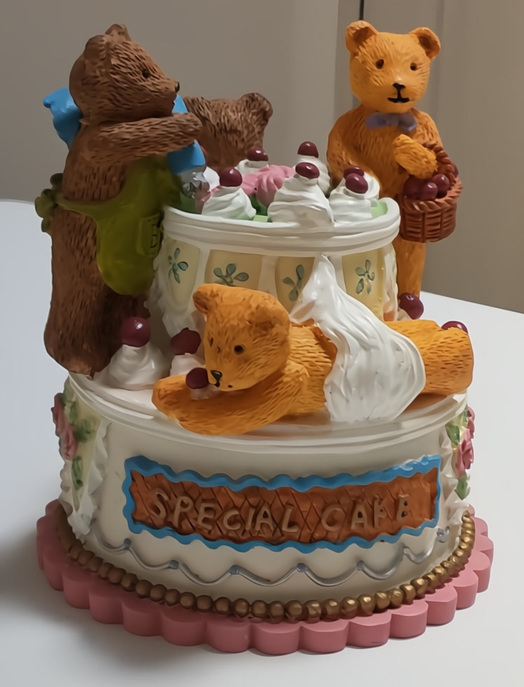}
    \end{subfigure}

    \smallskip
    \rotatebox[origin=b]{90}{Base color w/ illumination rebaked}\quad
    \begin{subfigure}[h]{0.14\paperwidth}
        \caption{$c=0$}
        \includegraphics[width=\textwidth]{assets/raw_images/NDJIR/results_custom_dataset/custom_rp1e4_bears_on_cake/resize/model_00999_512grid_raw_base_color_mesh00_defaultUnlit_default_57.png.jpg}
    \end{subfigure}
    \hspace{1pt}
    \begin{subfigure}[h]{0.14\paperwidth}
        \caption{$c=0.25$}
        \includegraphics[width=\textwidth]{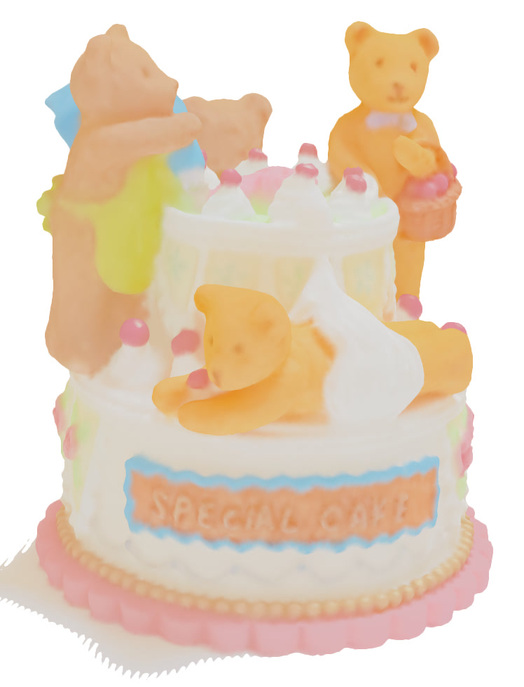}
    \end{subfigure}
    \hspace{1pt}
    \begin{subfigure}[h]{0.14\paperwidth}
        \caption{$c=0.5$}
        \includegraphics[width=\textwidth]{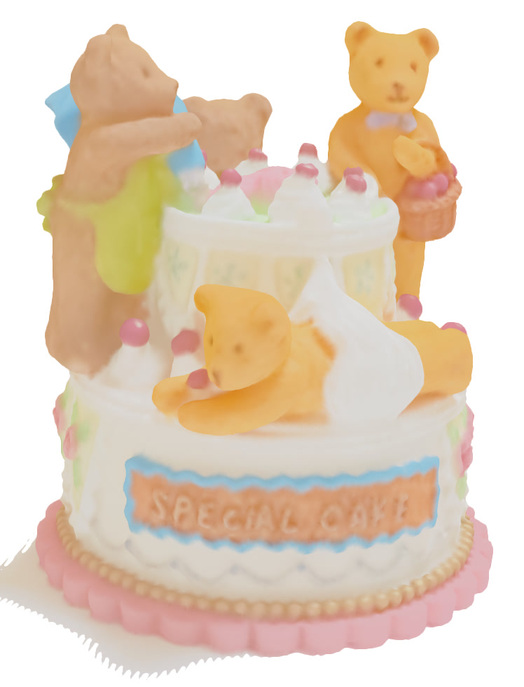}
    \end{subfigure}
    \hspace{1pt}
    \begin{subfigure}[h]{0.14\paperwidth}
        \caption{$c=0.75$}
        \includegraphics[width=\textwidth]{assets/raw_images/NDJIR/results_custom_dataset/custom_rp1e4_bears_on_cake/resize/model_00999_512grid_raw_base_color_mesh00_filtered01_ilbaked_0.5_defaultUnlit_default_57.png.jpg}
    \end{subfigure}
    \hspace{1pt}
    \begin{subfigure}[h]{0.14\paperwidth}
        \caption{$c=1$}
        \includegraphics[width=\textwidth]{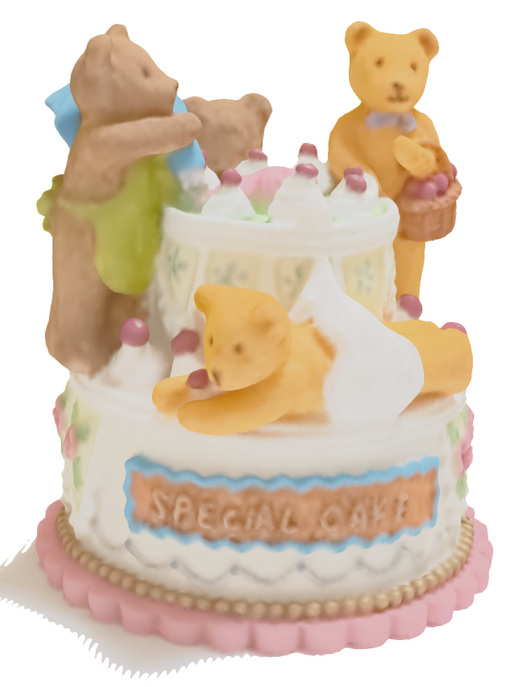}
    \end{subfigure}

    \smallskip
    \rotatebox[origin=b]{90}{PBR w/ illumination rebaked}\quad
    \begin{subfigure}[h]{0.14\paperwidth}
        \includegraphics[width=\textwidth]{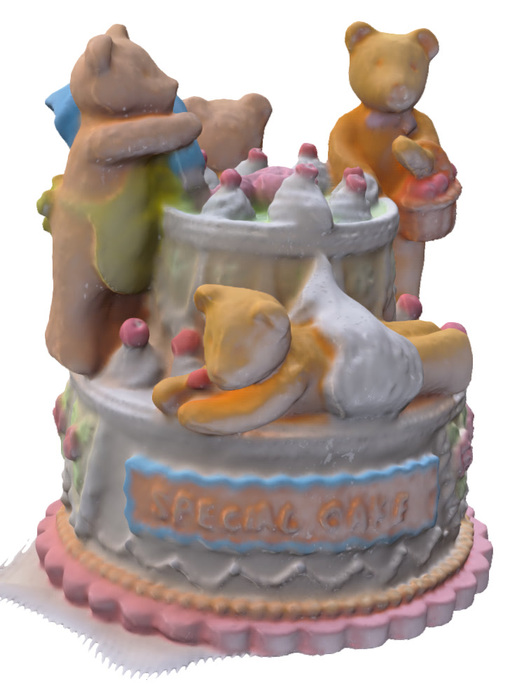}
    \end{subfigure}
    \hspace{1pt}
    \begin{subfigure}[h]{0.14\paperwidth}
        \includegraphics[width=\textwidth]{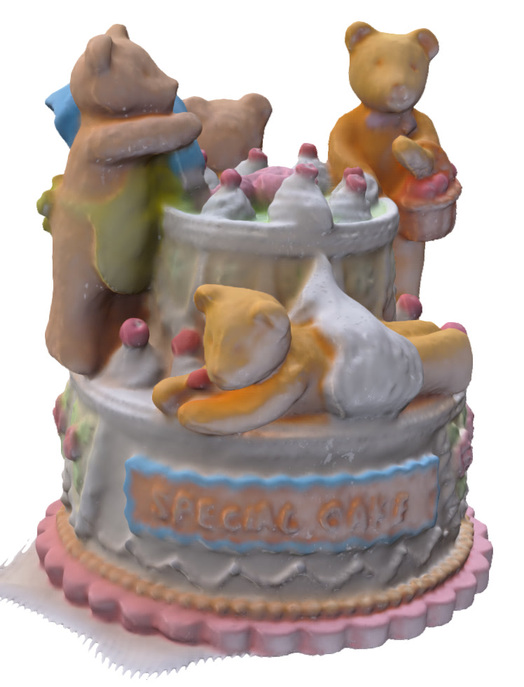}
    \end{subfigure}
    \hspace{1pt}
    \begin{subfigure}[h]{0.14\paperwidth}
        \includegraphics[width=\textwidth]{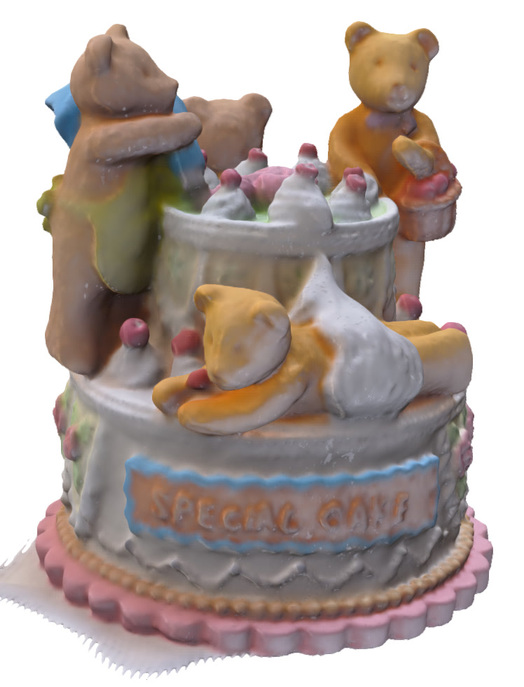}
    \end{subfigure}
    \hspace{1pt}
    \begin{subfigure}[h]{0.14\paperwidth}
        \includegraphics[width=\textwidth]{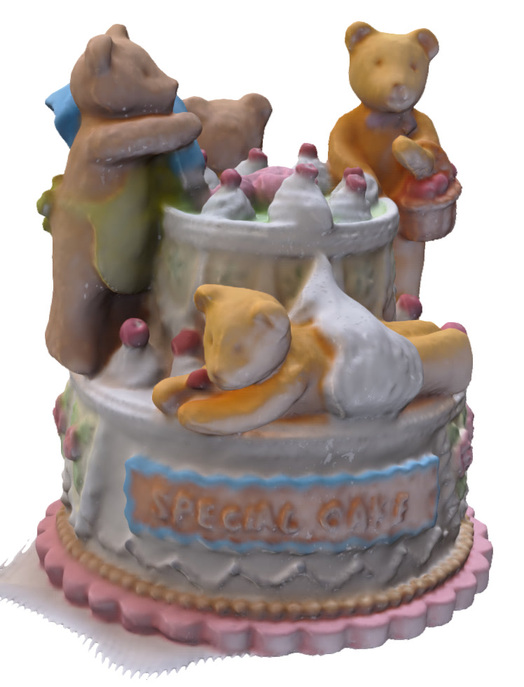}
    \end{subfigure}
    \hspace{1pt}
    \begin{subfigure}[h]{0.14\paperwidth}
        \includegraphics[width=\textwidth]{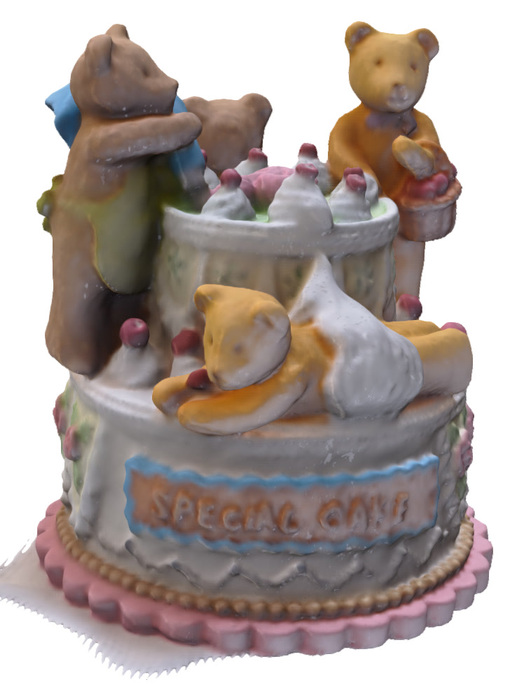}
    \end{subfigure}

    \caption{\textbf{Results of bears on cake scene.} ($\cdot$) is a given environment map in Open3D. $c$ is interpolation coefficient.}
    \label{fig:main_results_bears_on_cake}
\end{figure*}

\begin{figure*}[tbp]
    \centering
    \captionsetup[subfigure]{font=scriptsize}
    \rotatebox[origin=b]{90}{Decomposed materials}\quad
    \begin{subfigure}[h]{0.14\paperwidth}
        \caption{normals}
        \includegraphics[width=\textwidth]{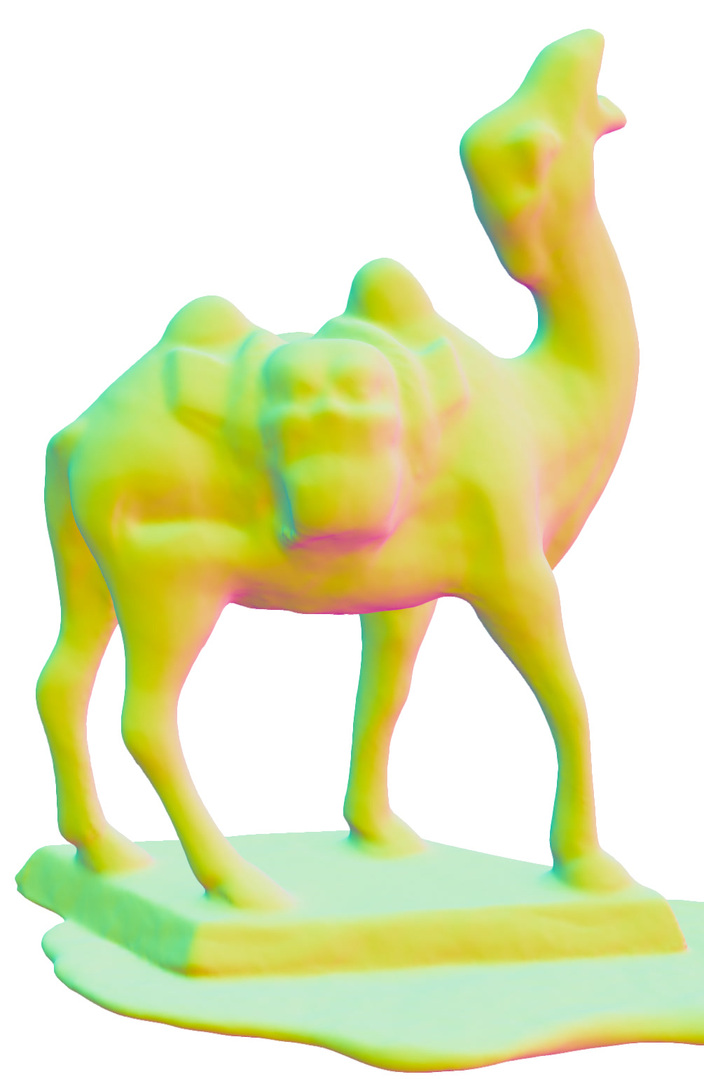}
    \end{subfigure}
    \hspace{1pt}
    \begin{subfigure}[h]{0.14\paperwidth}
        \caption{base color}
        \includegraphics[width=\textwidth]{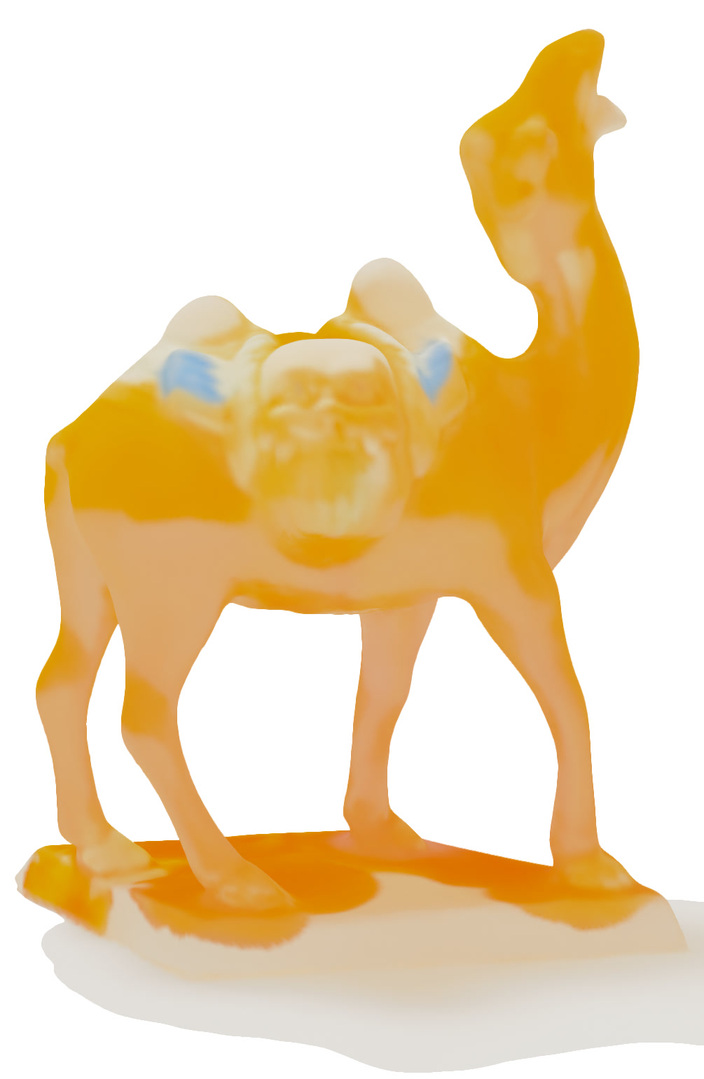}
    \end{subfigure}
    \hspace{1pt}
    \begin{subfigure}[h]{0.14\paperwidth}
        \caption{roughness}
        \includegraphics[width=\textwidth]{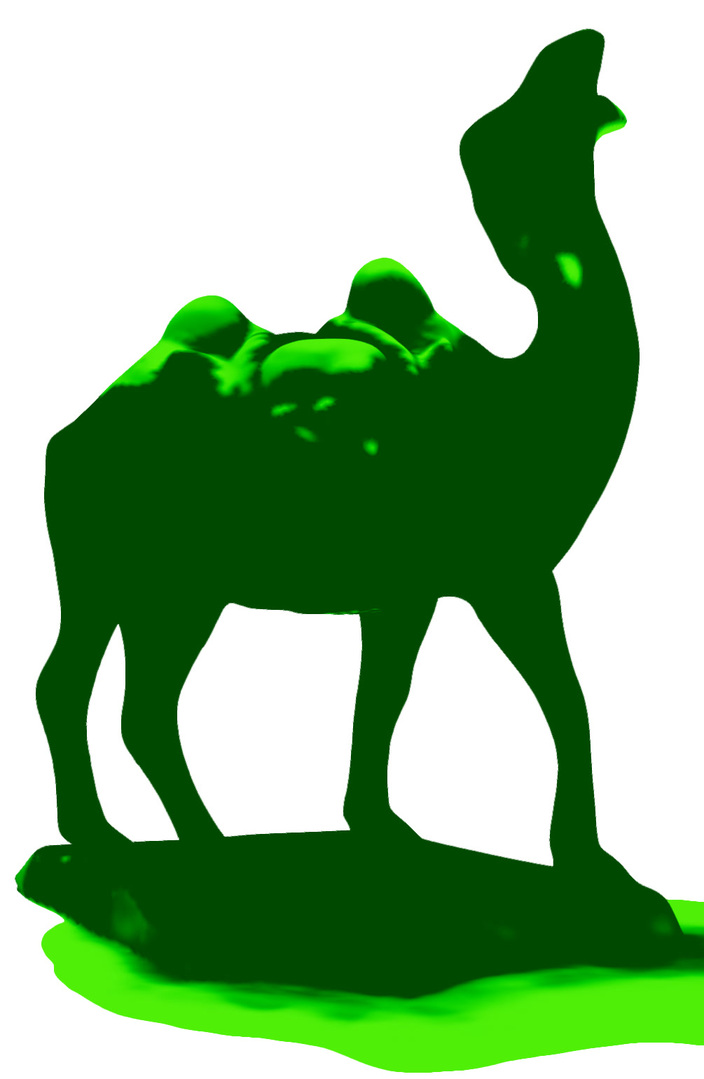}
    \end{subfigure}
    \hspace{1pt}
    \begin{subfigure}[h]{0.14\paperwidth}
        \caption{specular reflectance}
        \includegraphics[width=\textwidth]{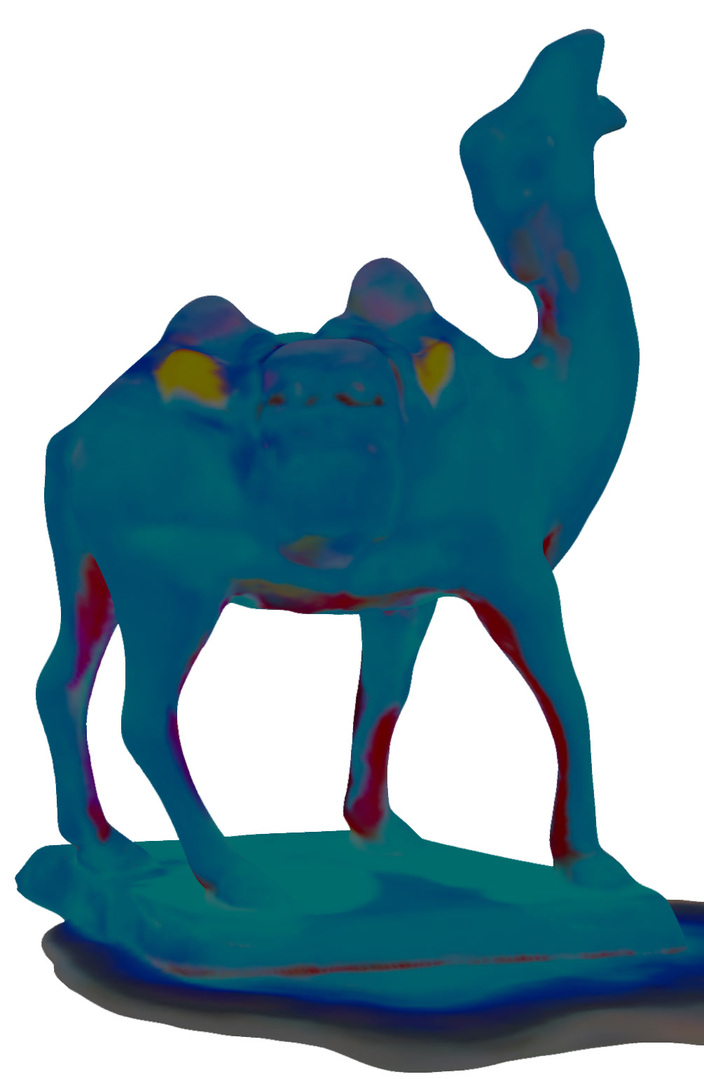}
    \end{subfigure}
    \hspace{1pt}
    \begin{subfigure}[h]{0.14\paperwidth}
        \caption{implicit illumination}
        \includegraphics[width=\textwidth]{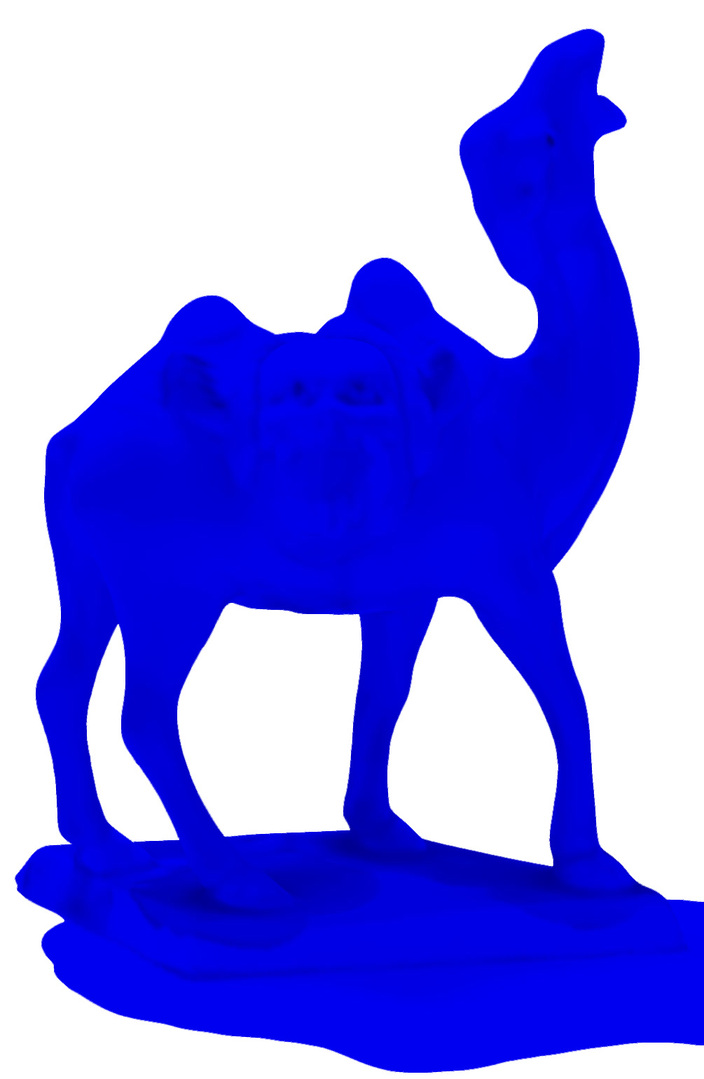}
    \end{subfigure}

    \smallskip
    \rotatebox[origin=b]{90}{PBR images}\quad
    \begin{subfigure}[h]{0.17\paperwidth}
        \caption{PBR (default)}
        \includegraphics[width=\textwidth]{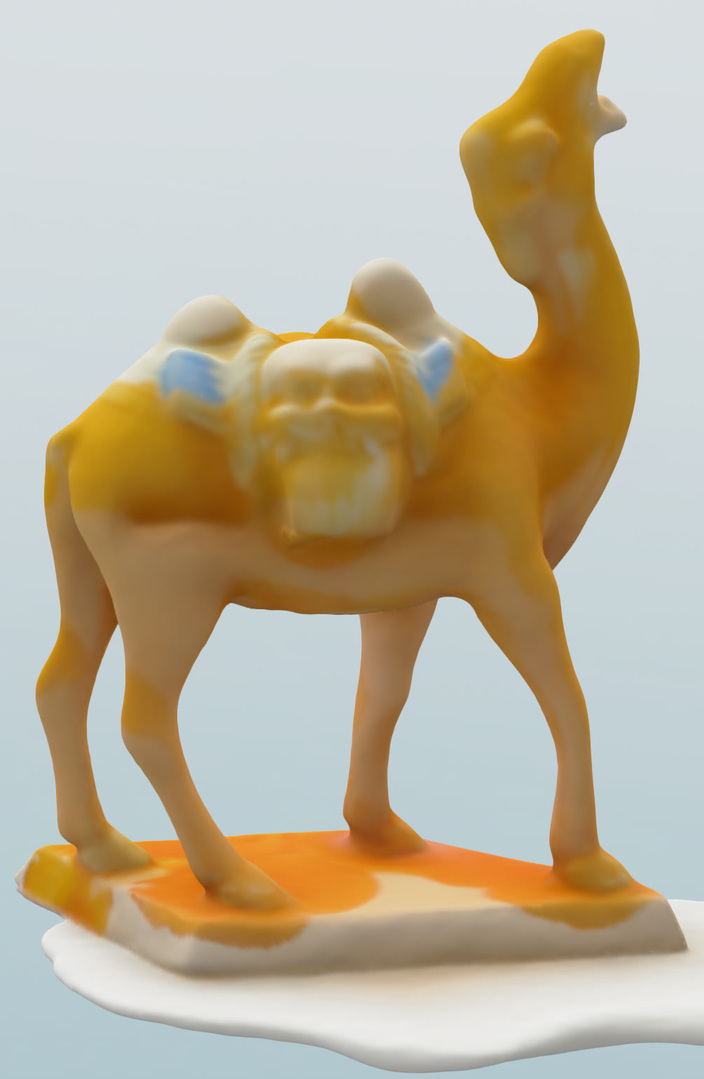}
    \end{subfigure}
    \hspace{1pt}
    \begin{subfigure}[h]{0.17\paperwidth}
        \caption{PBR (pillars)}
        \includegraphics[width=\textwidth]{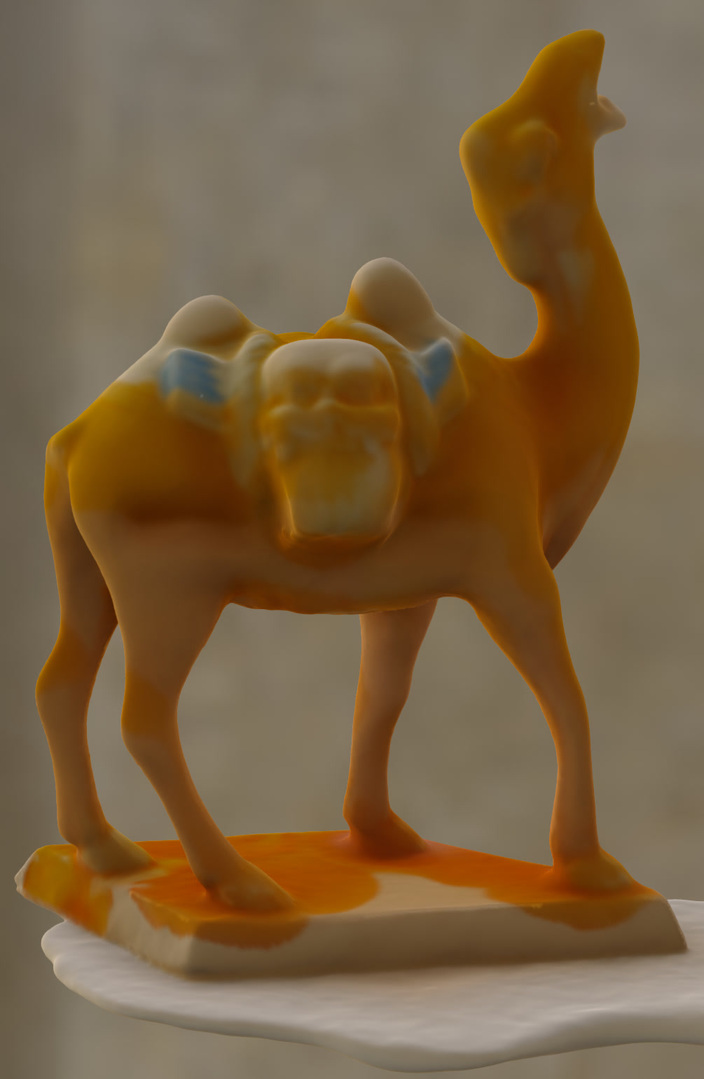}
    \end{subfigure}
    \hspace{1pt}
    \begin{subfigure}[h]{0.17\paperwidth}
        \caption{Neural rendering}
        \includegraphics[width=\textwidth]{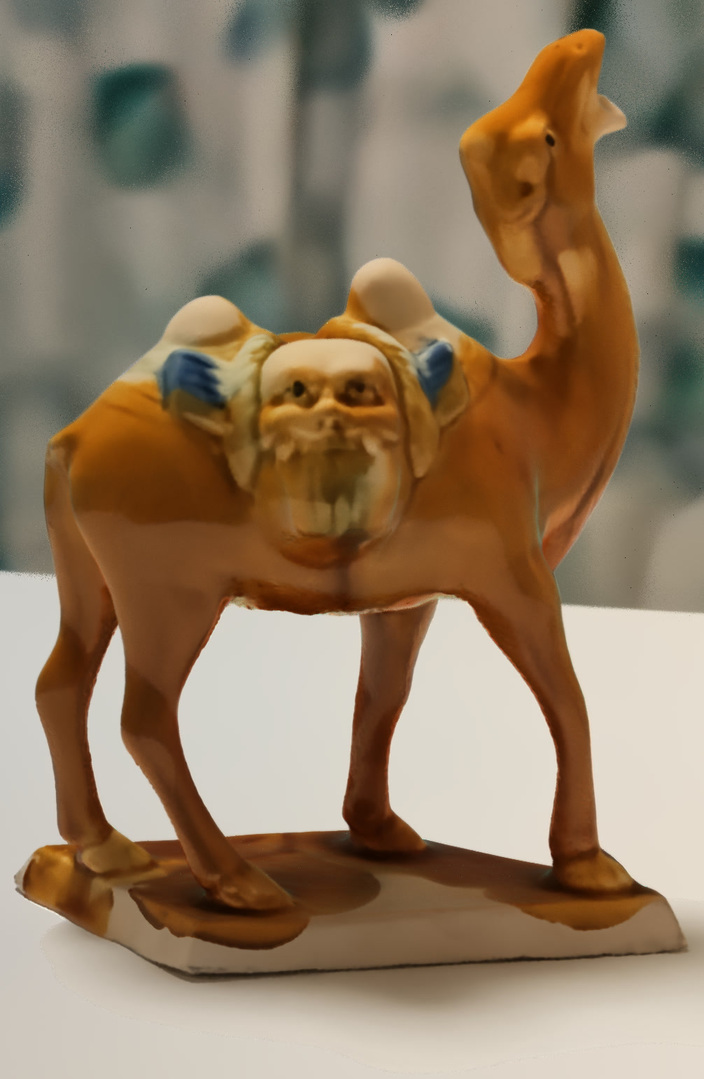}
    \end{subfigure}
    \hspace{1pt}
    \begin{subfigure}[h]{0.17\paperwidth}
        \caption{Groundtruth}
        \includegraphics[width=\textwidth]{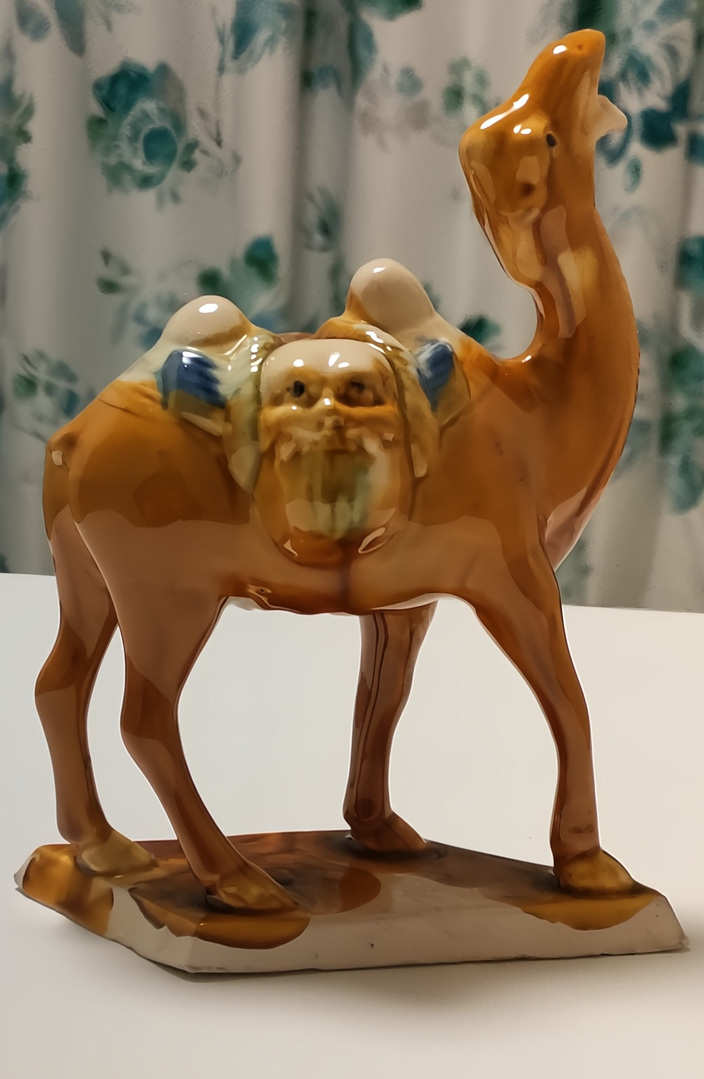}
    \end{subfigure}

    \smallskip
    \rotatebox[origin=b]{90}{Base color w/ illumination rebaked}\quad
    \begin{subfigure}[h]{0.14\paperwidth}
        \caption{$c=0$}
        \includegraphics[width=\textwidth]{assets/raw_images/NDJIR/results_custom_dataset/custom_rp1e4_cammel/resize/model_00999_512grid_raw_base_color_mesh00_defaultUnlit_default_75.png.jpg}
    \end{subfigure}
    \hspace{1pt}
    \begin{subfigure}[h]{0.14\paperwidth}
        \caption{$c=0.25$}
        \includegraphics[width=\textwidth]{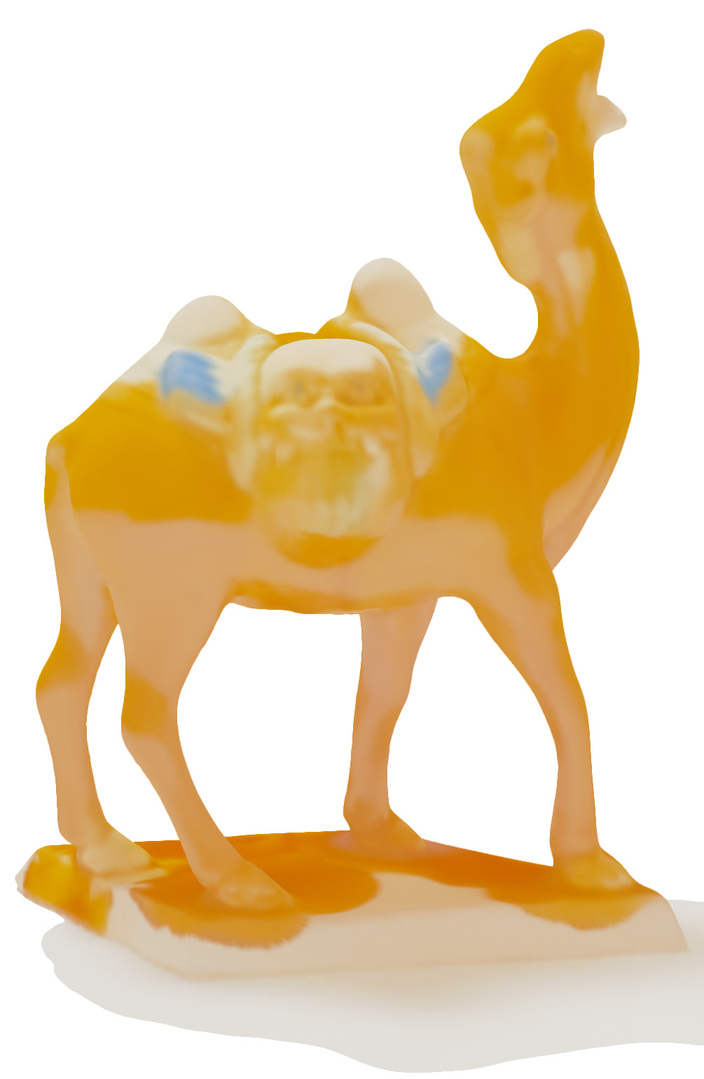}
    \end{subfigure}
    \hspace{1pt}
    \begin{subfigure}[h]{0.14\paperwidth}
        \caption{$c=0.5$}
        \includegraphics[width=\textwidth]{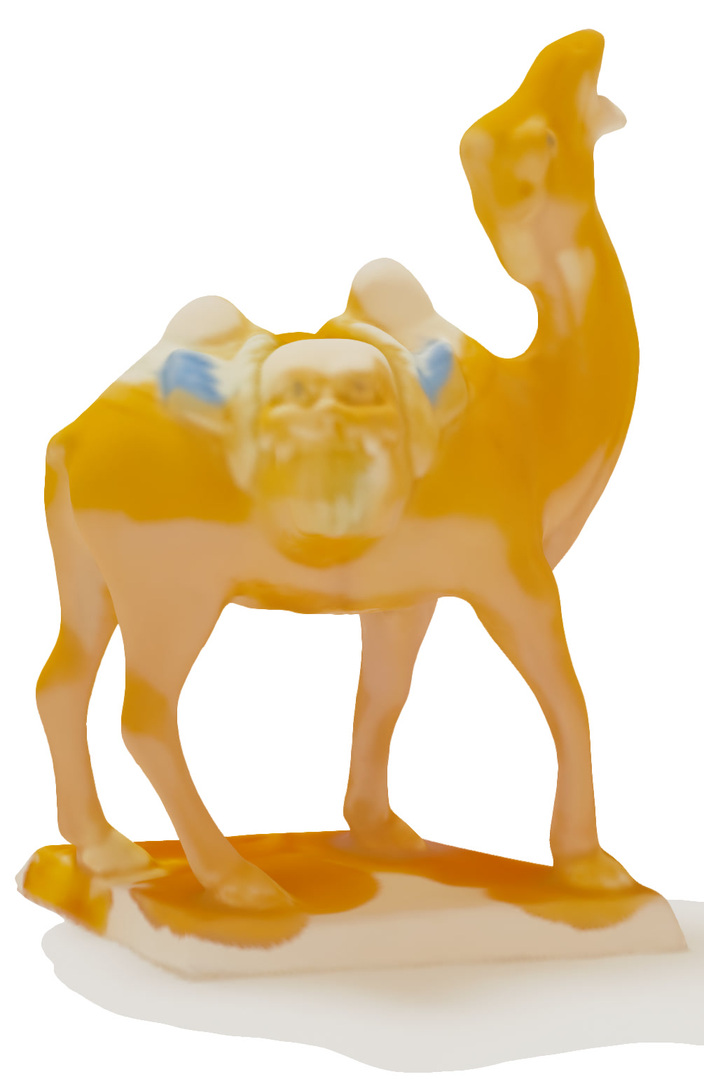}
    \end{subfigure}
    \hspace{1pt}
    \begin{subfigure}[h]{0.14\paperwidth}
        \caption{$c=0.75$}
        \includegraphics[width=\textwidth]{assets/raw_images/NDJIR/results_custom_dataset/custom_rp1e4_cammel/resize/model_00999_512grid_raw_base_color_mesh00_filtered01_ilbaked_0.5_defaultUnlit_default_75.png.jpg}
    \end{subfigure}
    \hspace{1pt}
    \begin{subfigure}[h]{0.14\paperwidth}
        \caption{$c=1$}
        \includegraphics[width=\textwidth]{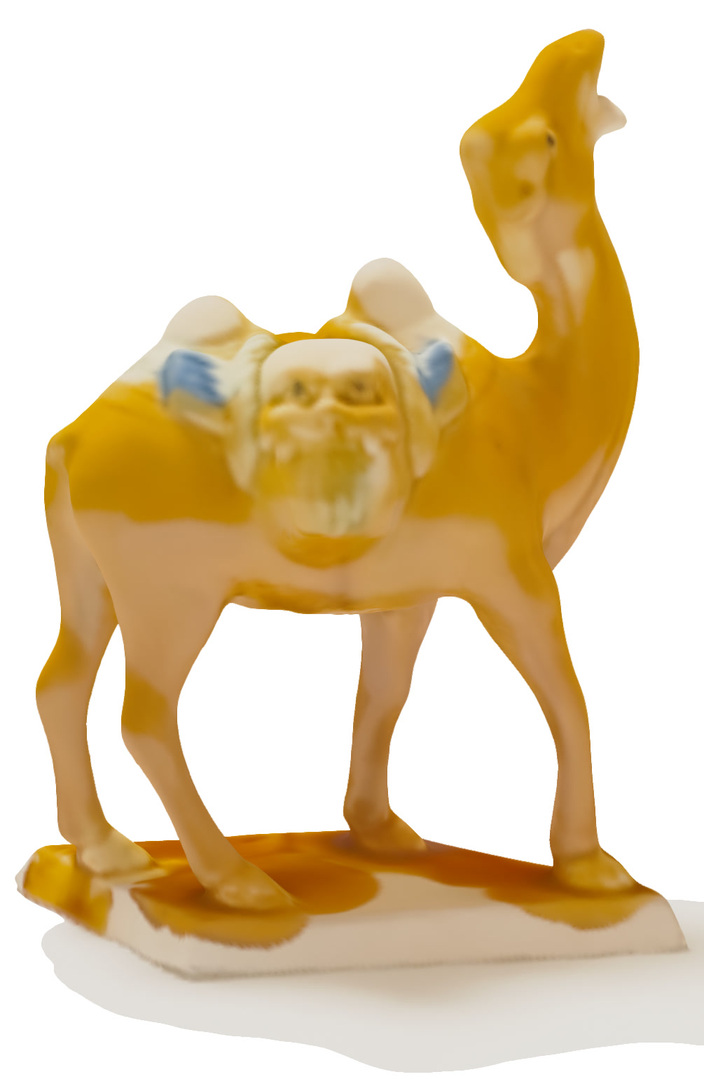}
    \end{subfigure}

    \smallskip
    \rotatebox[origin=b]{90}{PBR w/ illumination rebaked}\quad
    \begin{subfigure}[h]{0.14\paperwidth}
        \includegraphics[width=\textwidth]{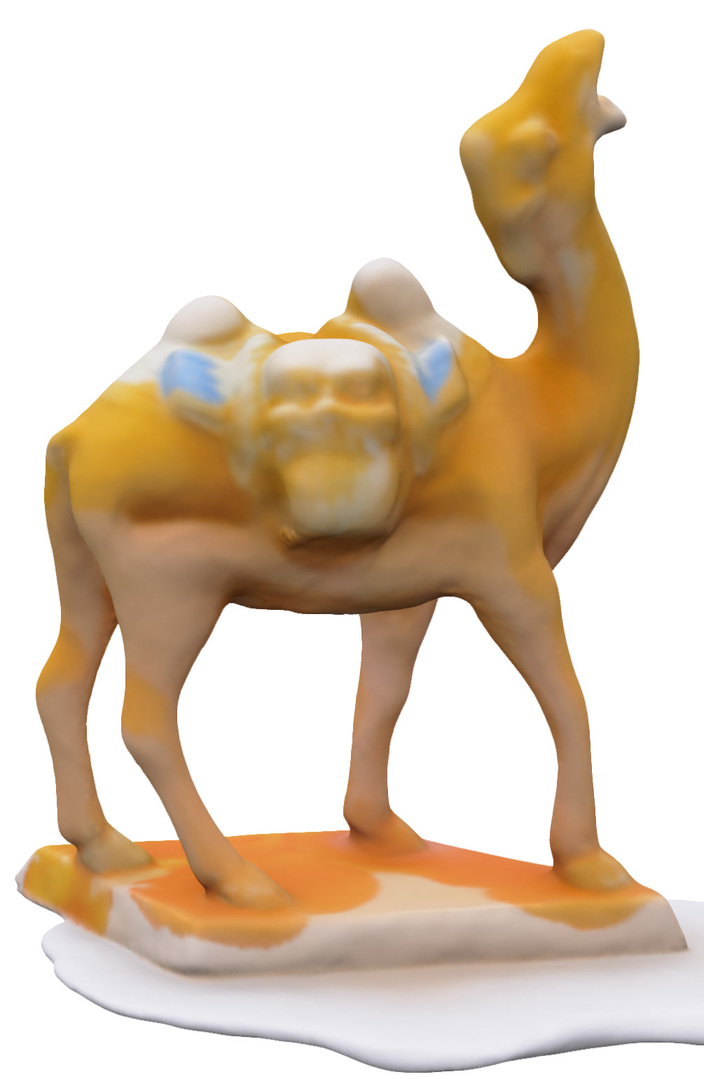}
    \end{subfigure}
    \hspace{1pt}
    \begin{subfigure}[h]{0.14\paperwidth}
        \includegraphics[width=\textwidth]{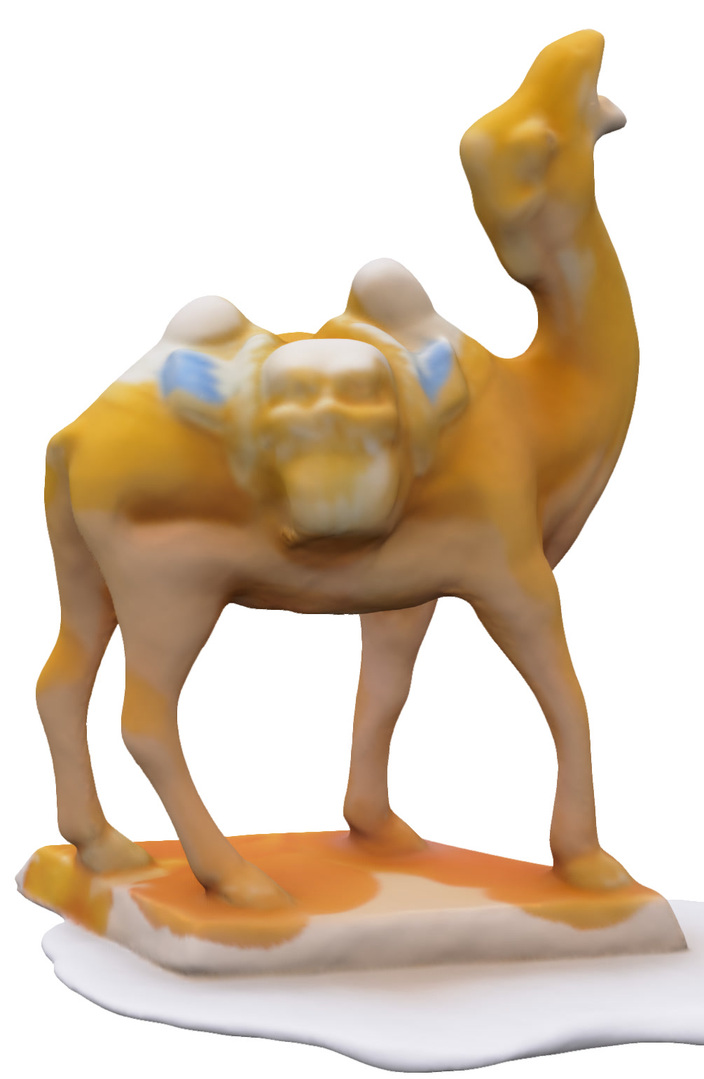}
    \end{subfigure}
    \hspace{1pt}
    \begin{subfigure}[h]{0.14\paperwidth}
        \includegraphics[width=\textwidth]{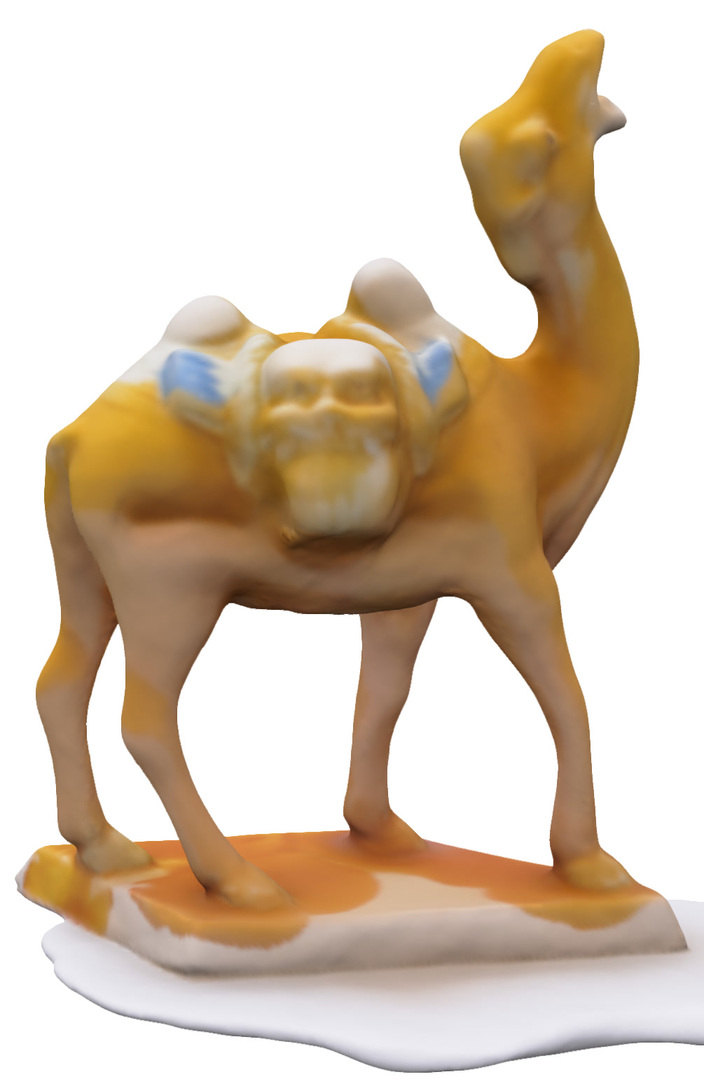}
    \end{subfigure}
    \hspace{1pt}
    \begin{subfigure}[h]{0.14\paperwidth}
        \includegraphics[width=\textwidth]{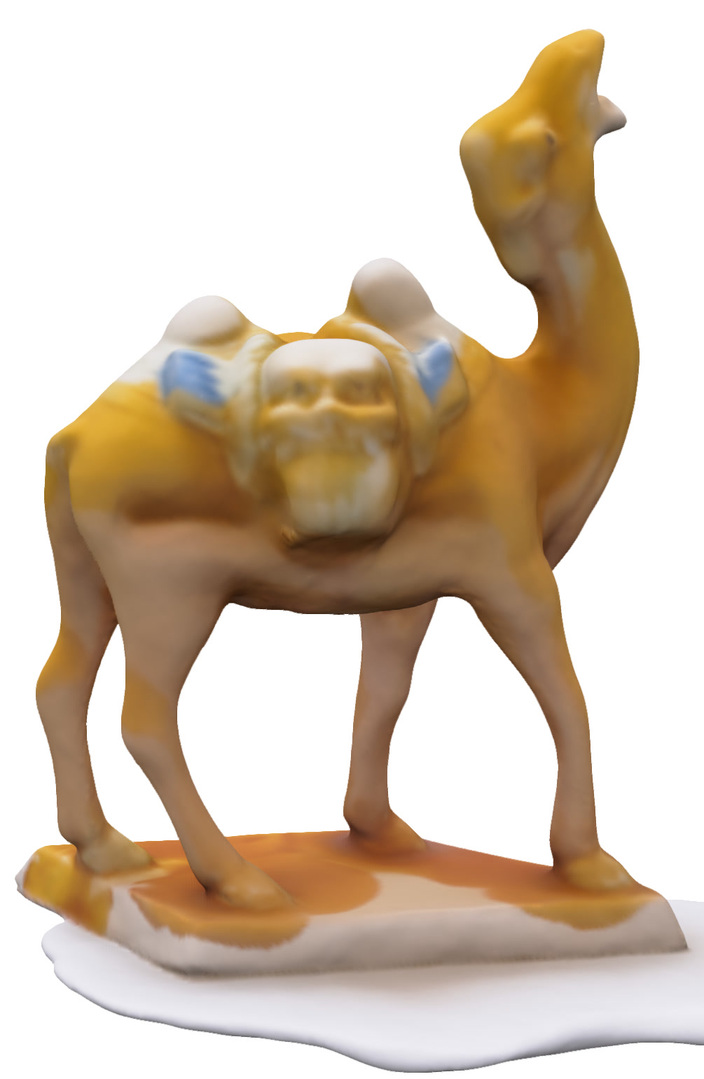}
    \end{subfigure}
    \hspace{1pt}
    \begin{subfigure}[h]{0.14\paperwidth}
        \includegraphics[width=\textwidth]{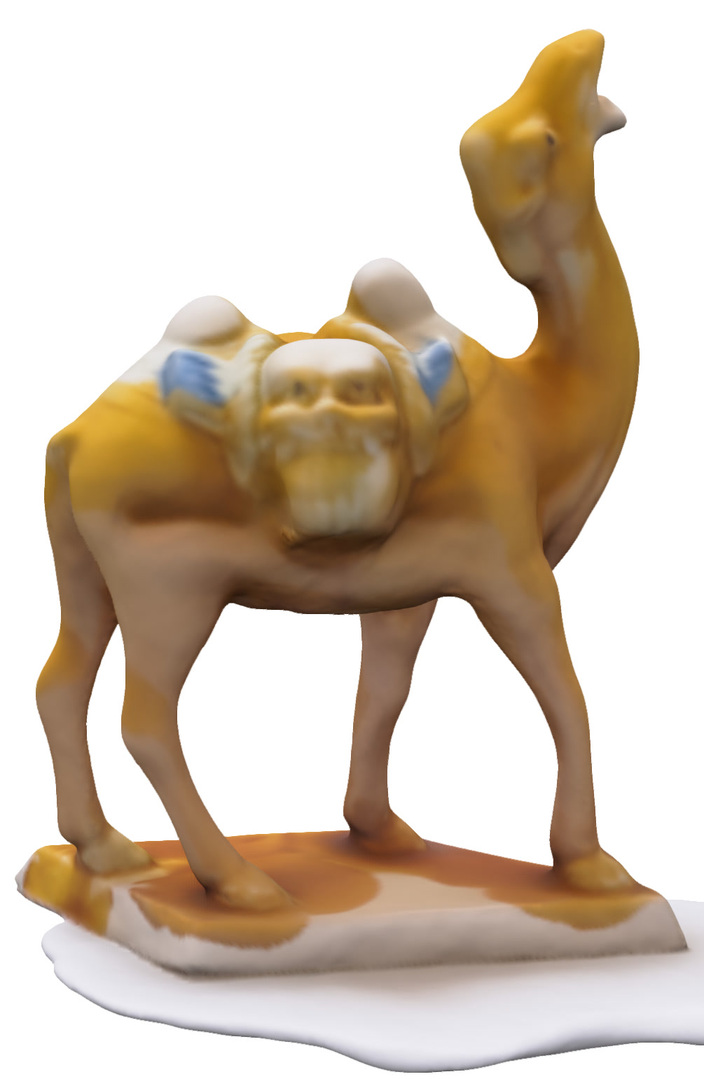}
    \end{subfigure}

    \caption{\textbf{Results of camel scene.} ($\cdot$) is a given environment map in Open3D. $c$ is interpolation coefficient.}
    \label{fig:main_results_camel}
\end{figure*}

\begin{figure*}[tbp]
    \centering
    \captionsetup[subfigure]{font=scriptsize}
    \rotatebox[origin=b]{90}{Decomposed materials}\quad
    \begin{subfigure}[h]{0.14\paperwidth}
        \centering
        \caption{normals}
        \includegraphics[width=0.6\textwidth]{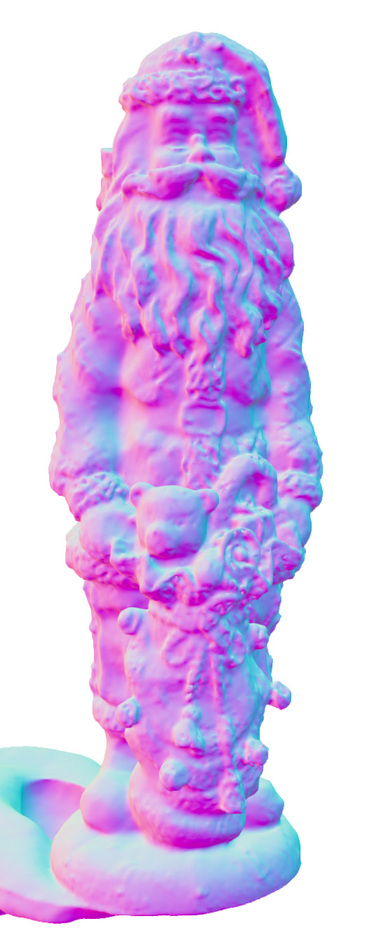}
    \end{subfigure}
    \hspace{1pt}
    \begin{subfigure}[h]{0.14\paperwidth}
        \centering
        \caption{base color}
        \includegraphics[width=0.6\textwidth]{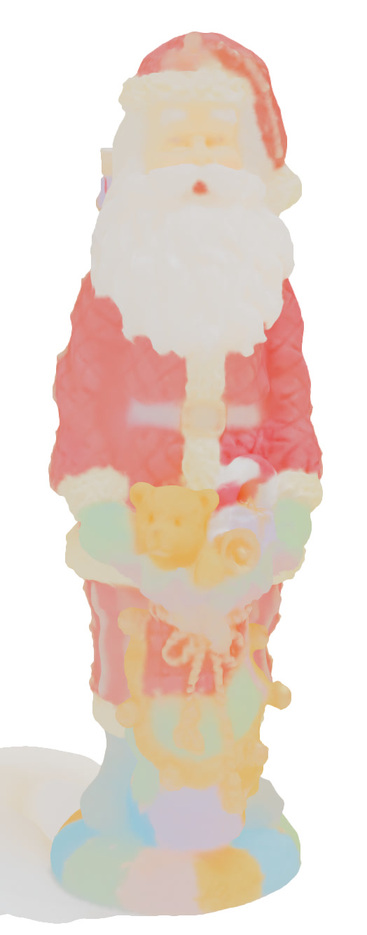}
    \end{subfigure}
    \hspace{1pt}
    \begin{subfigure}[h]{0.14\paperwidth}
        \centering
        \caption{roughness}
        \includegraphics[width=0.6\textwidth]{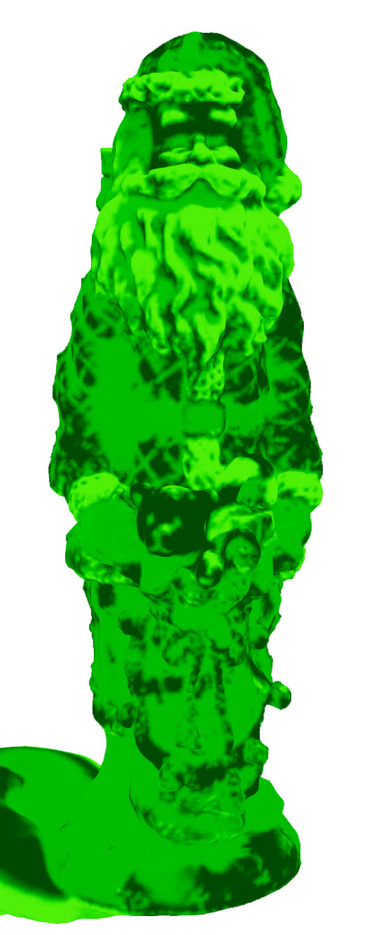}
    \end{subfigure}
    \hspace{1pt}
    \begin{subfigure}[h]{0.14\paperwidth}
        \centering
        \caption{specular reflectance}
        \includegraphics[width=0.6\textwidth]{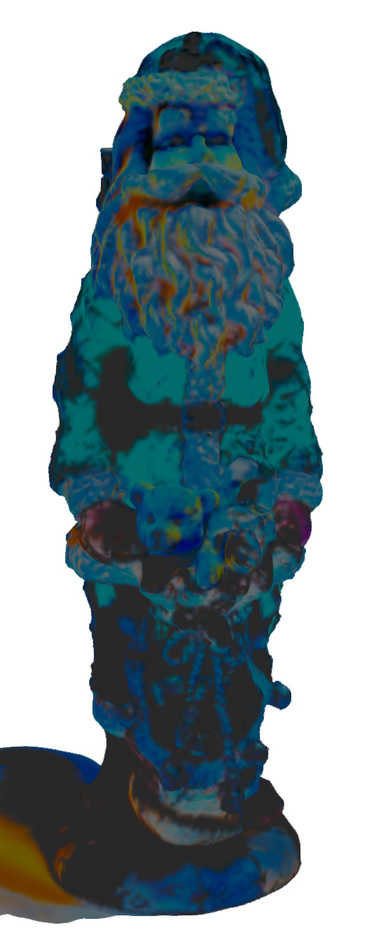}
    \end{subfigure}
    \hspace{1pt}
    \begin{subfigure}[h]{0.14\paperwidth}
        \centering
        \caption{implicit illumination}
        \includegraphics[width=0.6\textwidth]{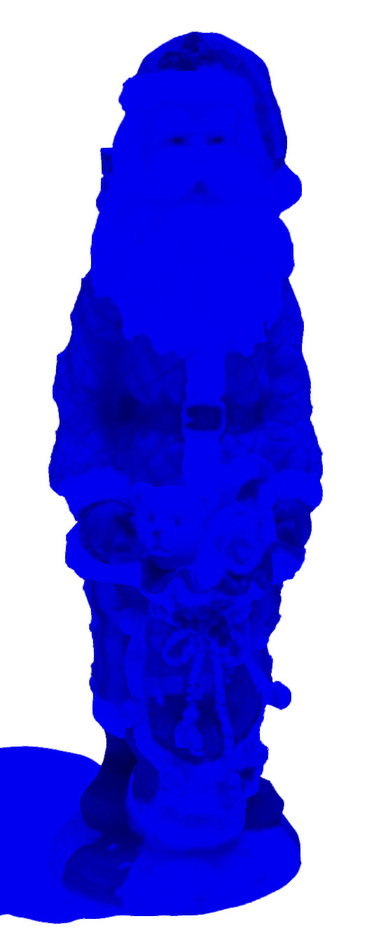}
    \end{subfigure}

    \smallskip
    \rotatebox[origin=b]{90}{PBR images}\quad
    \begin{subfigure}[h]{0.17\paperwidth}
        \centering
        \caption{PBR (default)}
        \includegraphics[width=0.6\textwidth]{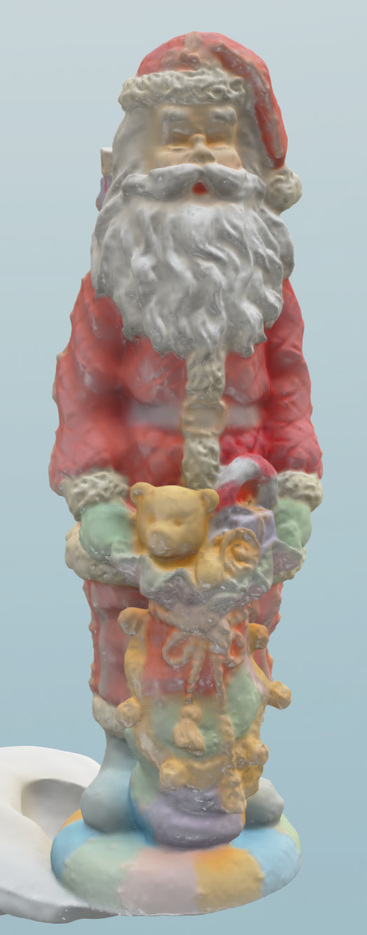}
    \end{subfigure}
    \hspace{1pt}
    \begin{subfigure}[h]{0.17\paperwidth}
        \centering
        \caption{PBR (pillars)}
        \includegraphics[width=0.6\textwidth]{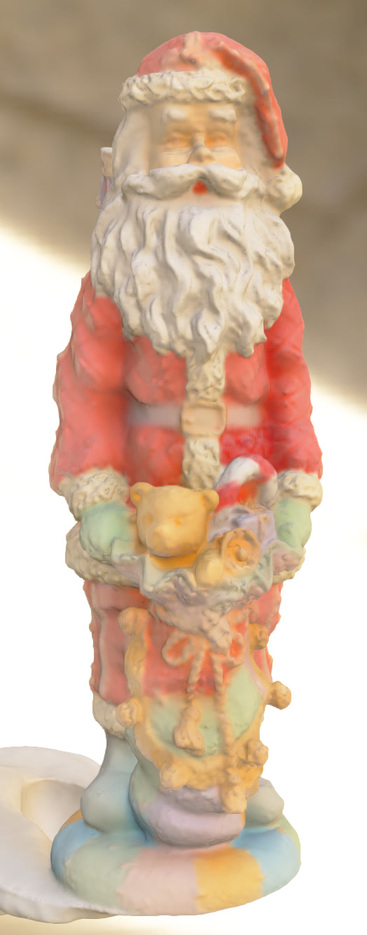}
    \end{subfigure}
    \hspace{1pt}
    \begin{subfigure}[h]{0.17\paperwidth}
        \centering
        \caption{Neural rendering}
        \includegraphics[width=0.6\textwidth]{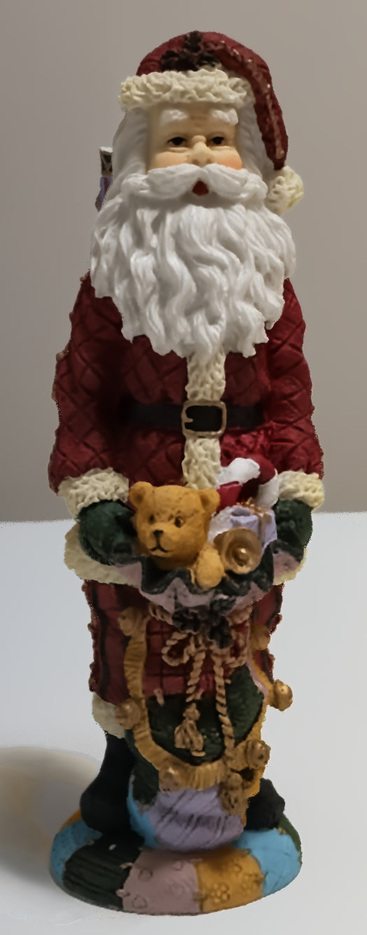}
    \end{subfigure}
    \hspace{1pt}
    \begin{subfigure}[h]{0.17\paperwidth}
        \centering
        \caption{Groundtruth}
        \includegraphics[width=0.6\textwidth]{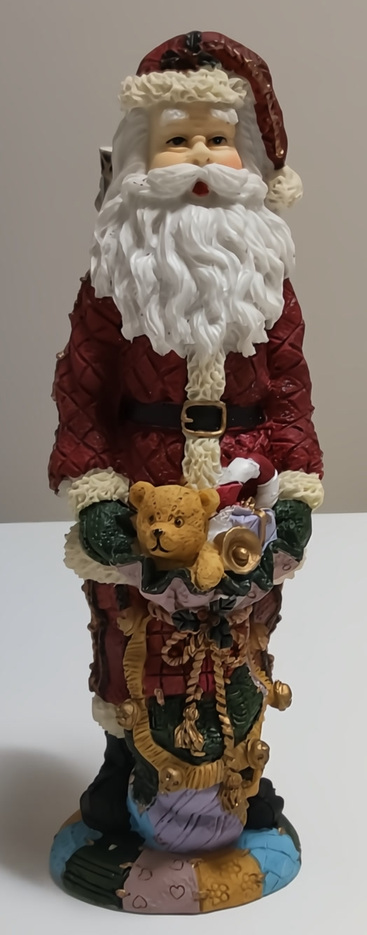}
    \end{subfigure}

    \smallskip
    \rotatebox[origin=b]{90}{Base color w/ illumination rebaked}\quad
    \begin{subfigure}[h]{0.14\paperwidth}
        \centering
        \caption{$c=0$}
        \includegraphics[width=0.6\textwidth]{assets/raw_images/NDJIR/results_custom_dataset/custom_rp1e4_santa/resize/model_00999_512grid_raw_base_color_mesh00_defaultUnlit_default_51.png.jpg}
    \end{subfigure}
    \hspace{1pt}    
    \begin{subfigure}[h]{0.14\paperwidth}
        \centering
        \caption{$c=0.25$}
        \includegraphics[width=0.6\textwidth]{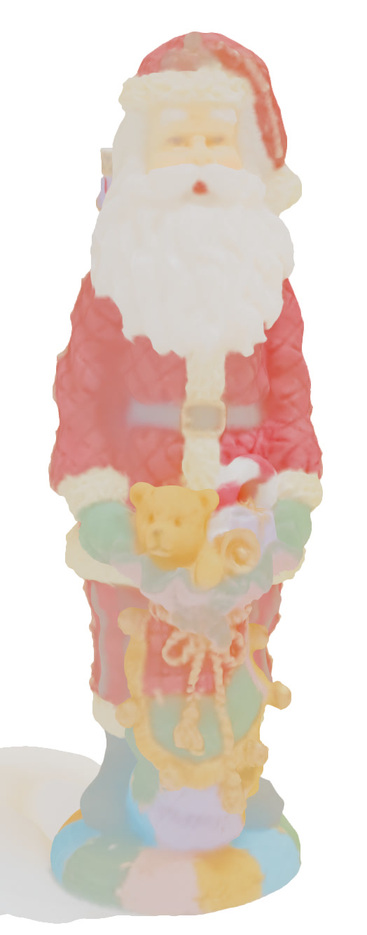}
    \end{subfigure}
    \hspace{1pt}
    \begin{subfigure}[h]{0.14\paperwidth}
        \centering
        \caption{$c=0.5$}
        \includegraphics[width=0.6\textwidth]{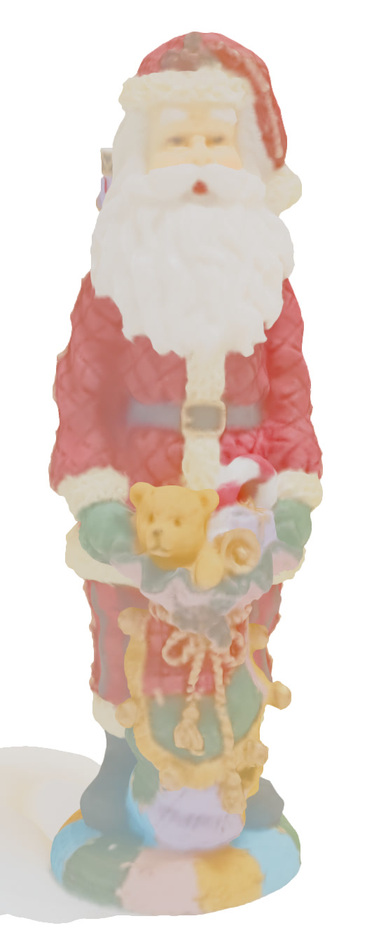}
    \end{subfigure}
    \hspace{1pt}
    \begin{subfigure}[h]{0.14\paperwidth}
        \centering
        \caption{$c=0.75$}
        \includegraphics[width=0.6\textwidth]{assets/raw_images/NDJIR/results_custom_dataset/custom_rp1e4_santa/resize/model_00999_512grid_raw_base_color_mesh00_filtered01_ilbaked_0.5_defaultUnlit_default_51.png.jpg}
    \end{subfigure}
    \hspace{1pt}
    \begin{subfigure}[h]{0.14\paperwidth}
        \centering
        \caption{$c=1$}
        \includegraphics[width=0.6\textwidth]{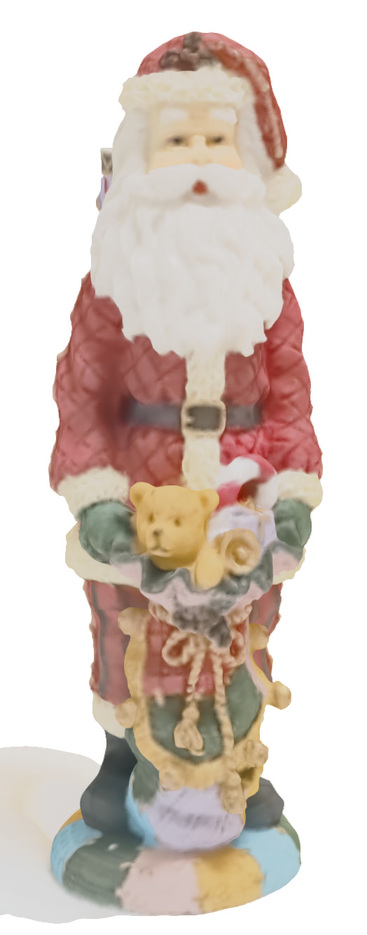}
    \end{subfigure}

    \smallskip
    \rotatebox[origin=b]{90}{PBR w/ illumination rebaked}\quad
    \begin{subfigure}[h]{0.14\paperwidth}
        \centering
        \includegraphics[width=0.6\textwidth]{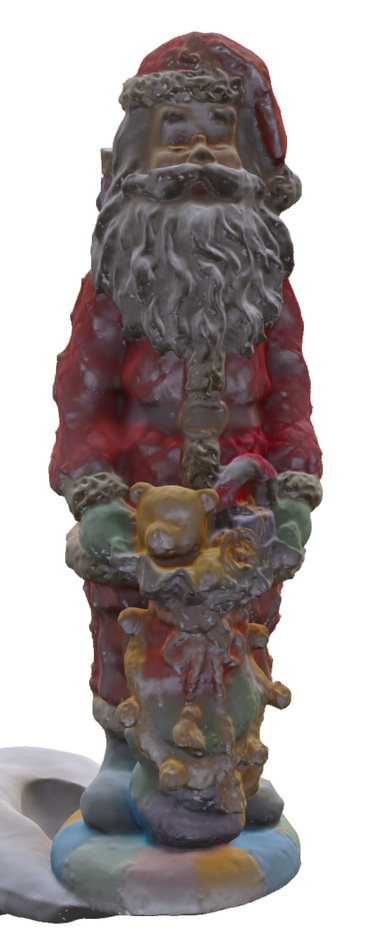}
    \end{subfigure}
    \hspace{1pt}
    \begin{subfigure}[h]{0.14\paperwidth}
        \centering
        \includegraphics[width=0.6\textwidth]{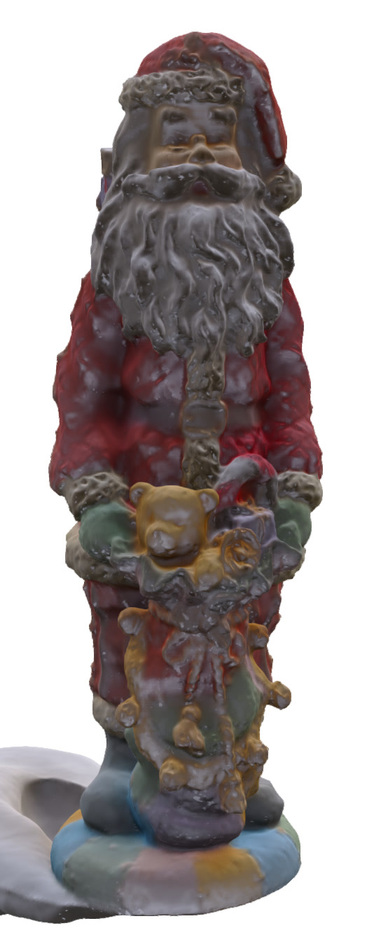}
    \end{subfigure}
    \hspace{1pt}
    \begin{subfigure}[h]{0.14\paperwidth}
        \centering
        \includegraphics[width=0.6\textwidth]{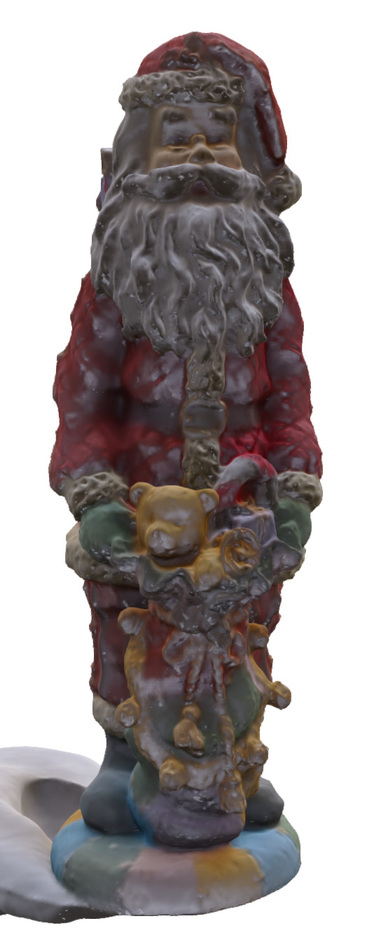}
    \end{subfigure}
    \hspace{1pt}
    \begin{subfigure}[h]{0.14\paperwidth}
        \centering
        \includegraphics[width=0.6\textwidth]{assets/raw_images/NDJIR/results_custom_dataset/custom_rp1e4_santa/resize/model_00999_512grid_raw_base_color_mesh00_filtered01_ilbaked_0.5_defaultLit_konzerthaus_51.png.jpg}
    \end{subfigure}
    \hspace{1pt}
    \begin{subfigure}[h]{0.14\paperwidth}
        \centering
        \includegraphics[width=0.6\textwidth]{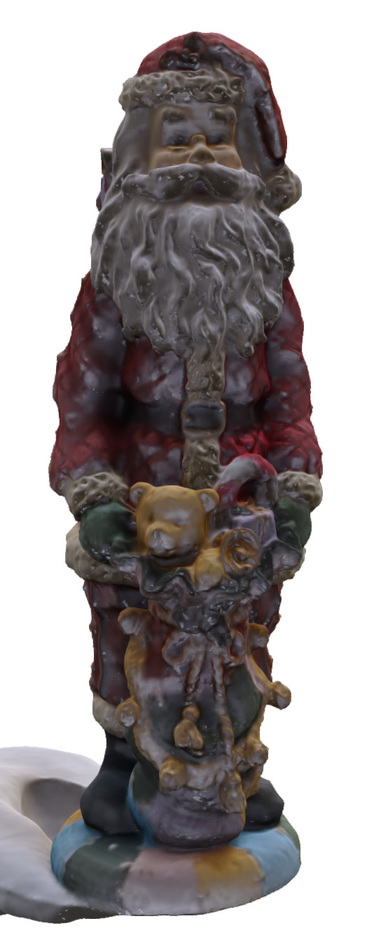}
    \end{subfigure}

    \caption{\textbf{Results of Santa Claus scene.} ($\cdot$) is a given environment map in Open3D. $c$ is interpolation coefficient.}
    \label{fig:main_results_santa}
\end{figure*}

\cref{fig:main_results_bears_on_cake,fig:main_results_camel,fig:main_results_santa} show decomposed materials, PBR images, base color with implicit illumination baked, and its PBR.

Note that different from DTU MVS dataset which uses the ground truth camera parameters, we reply on estimation for camera parameters such that we use Lanczos filter with the window size of $2$ in the voxel grid feature implementation, which we found is more stable training in custom dataset. Correspondingly, the average filter to meshes is applied one time, and the stronger total variation weight $\lambda_{TV} = 1.0$ and the roughness prior weight $\lambda_{r_{p}} = 10^{-4}$ are used. Training epoch is $1000$.

\clearpage

\clearpage

\clearpage

\clearpage



\begin{thebibliography}{10}\itemsep=-1pt

\bibitem{DBLP:conf/3dim/AsselinLL20}
Louis{-}Philippe Asselin, Denis Laurendeau, and Jean{-}Fran{\c{c}}ois Lalonde.
\newblock Deep {SVBRDF} estimation on real materials.
\newblock In Vitomir Struc and Francisco~G{\'{o}}mez Fern{\'{a}}ndez, editors,
  {\em 8th International Conference on 3D Vision, 3DV 2020, Virtual Event,
  Japan, November 25-28, 2020}, pages 1157--1166. {IEEE}, 2020.

\bibitem{DBLP:conf/cvpr/AtzmonL20}
Matan Atzmon and Yaron Lipman.
\newblock {SAL:} sign agnostic learning of shapes from raw data.
\newblock In {\em 2020 {IEEE/CVF} Conference on Computer Vision and Pattern
  Recognition, {CVPR} 2020, Seattle, WA, USA, June 13-19, 2020}, pages
  2562--2571. Computer Vision Foundation / {IEEE}, 2020.

\bibitem{DBLP:journals/corr/abs-2008-03824}
Sai Bi, Zexiang Xu, Pratul~P. Srinivasan, Ben Mildenhall, Kalyan Sunkavalli,
  Milos Hasan, Yannick Hold{-}Geoffroy, David~J. Kriegman, and Ravi
  Ramamoorthi.
\newblock Neural reflectance fields for appearance acquisition.
\newblock {\em CoRR}, abs/2008.03824, 2020.

\bibitem{DBLP:conf/iccv/BossBJBLL21}
Mark Boss, Raphael Braun, Varun Jampani, Jonathan~T. Barron, Ce Liu, and
  Hendrik P.~A. Lensch.
\newblock Nerd: Neural reflectance decomposition from image collections.
\newblock In {\em 2021 {IEEE/CVF} International Conference on Computer Vision,
  {ICCV} 2021, Montreal, QC, Canada, October 10-17, 2021}, pages 12664--12674.
  {IEEE}, 2021.

\bibitem{DBLP:conf/nips/BossJBLBL21}
Mark Boss, Varun Jampani, Raphael Braun, Ce Liu, Jonathan~T. Barron, and
  Hendrik P.~A. Lensch.
\newblock Neural-pil: Neural pre-integrated lighting for reflectance
  decomposition.
\newblock In Marc'Aurelio Ranzato, Alina Beygelzimer, Yann~N. Dauphin, Percy
  Liang, and Jennifer~Wortman Vaughan, editors, {\em Advances in Neural
  Information Processing Systems 34: Annual Conference on Neural Information
  Processing Systems 2021, NeurIPS 2021, December 6-14, 2021, virtual}, pages
  10691--10704, 2021.

\bibitem{DBLP:conf/cvpr/BossJKLK20}
Mark Boss, Varun Jampani, Kihwan Kim, Hendrik P.~A. Lensch, and Jan Kautz.
\newblock Two-shot spatially-varying {BRDF} and shape estimation.
\newblock In {\em 2020 {IEEE/CVF} Conference on Computer Vision and Pattern
  Recognition, {CVPR} 2020, Seattle, WA, USA, June 13-19, 2020}, pages
  3981--3990. Computer Vision Foundation / {IEEE}, 2020.

\bibitem{jax2018github}
James Bradbury, Roy Frostig, Peter Hawkins, Matthew~James Johnson, Chris Leary,
  Dougal Maclaurin, George Necula, Adam Paszke, Jake Vander{P}las, Skye
  Wanderman-{M}ilne, and Qiao Zhang.
\newblock {JAX}: composable transformations of {P}ython+{N}um{P}y programs,
  2018.

\bibitem{DBLP:journals/corr/abs-2203-09517}
Anpei Chen, Zexiang Xu, Andreas Geiger, Jingyi Yu, and Hao Su.
\newblock Tensorf: Tensorial radiance fields.
\newblock {\em CoRR}, abs/2203.09517, 2022.

\bibitem{chen2022simple}
Liangyu Chen, Xiaojie Chu, Xiangyu Zhang, and Jian Sun.
\newblock Simple baselines for image restoration.
\newblock {\em arXiv preprint arXiv:2204.04676}, 2022.

\bibitem{DBLP:conf/nips/ChenLGSLJF19}
Wenzheng Chen, Huan Ling, Jun Gao, Edward~J. Smith, Jaakko Lehtinen, Alec
  Jacobson, and Sanja Fidler.
\newblock Learning to predict 3d objects with an interpolation-based
  differentiable renderer.
\newblock In Hanna~M. Wallach, Hugo Larochelle, Alina Beygelzimer, Florence
  d'Alch{\'{e}}{-}Buc, Emily~B. Fox, and Roman Garnett, editors, {\em Advances
  in Neural Information Processing Systems 32: Annual Conference on Neural
  Information Processing Systems 2019, NeurIPS 2019, December 8-14, 2019,
  Vancouver, BC, Canada}, pages 9605--9616, 2019.

\bibitem{Blender}
Blender~Online Community.
\newblock {\em Blender - a 3D modelling and rendering package}.
\newblock Blender Foundation, Stichting Blender Foundation, Amsterdam, 2018.

\bibitem{DBLP:journals/tog/CookT82}
Robert~L. Cook and Kenneth~E. Torrance.
\newblock A reflectance model for computer graphics.
\newblock {\em {ACM} Trans. Graph.}, 1(1):7--24, 1982.

\bibitem{DBLP:conf/nips/CourbariauxBD15}
Matthieu Courbariaux, Yoshua Bengio, and Jean{-}Pierre David.
\newblock Binaryconnect: Training deep neural networks with binary weights
  during propagations.
\newblock In Corinna Cortes, Neil~D. Lawrence, Daniel~D. Lee, Masashi Sugiyama,
  and Roman Garnett, editors, {\em Advances in Neural Information Processing
  Systems 28: Annual Conference on Neural Information Processing Systems 2015,
  December 7-12, 2015, Montreal, Quebec, Canada}, pages 3123--3131, 2015.

\bibitem{DBLP:conf/cvpr/DarmonBDMA22}
Fran{\c{c}}ois Darmon, B{\'{e}}n{\'{e}}dicte Bascle, Jean{-}Cl{\'{e}}ment
  Devaux, Pascal Monasse, and Mathieu Aubry.
\newblock Improving neural implicit surfaces geometry with patch warping.
\newblock In {\em {IEEE/CVF} Conference on Computer Vision and Pattern
  Recognition, {CVPR} 2022, New Orleans, LA, USA, June 18-24, 2022}, pages
  6250--6259. {IEEE}, 2022.

\bibitem{155328}
H. Drucker and Y. Le~Cun.
\newblock Double backpropagation increasing generalization performance.
\newblock In {\em IJCNN-91-Seattle International Joint Conference on Neural
  Networks}, volume~ii, pages 145--150 vol.2, 1991.

\bibitem{unrealengine}
{Epic Games}.
\newblock Unreal engine.

\bibitem{DBLP:conf/cvpr/Fridovich-KeilY22}
Sara Fridovich{-}Keil, Alex Yu, Matthew Tancik, Qinhong Chen, Benjamin Recht,
  and Angjoo Kanazawa.
\newblock Plenoxels: Radiance fields without neural networks.
\newblock In {\em {IEEE/CVF} Conference on Computer Vision and Pattern
  Recognition, {CVPR} 2022, New Orleans, LA, USA, June 18-24, 2022}, pages
  5491--5500. {IEEE}, 2022.

\bibitem{DBLP:journals/tog/GaoLDP0019}
Duan Gao, Xiao Li, Yue Dong, Pieter Peers, Kun Xu, and Xin Tong.
\newblock Deep inverse rendering for high-resolution {SVBRDF} estimation from
  an arbitrary number of images.
\newblock {\em {ACM} Trans. Graph.}, 38(4):134:1--134:15, 2019.

\bibitem{DBLP:conf/iccv/GarbinK0SV21}
Stephan~J. Garbin, Marek Kowalski, Matthew Johnson, Jamie Shotton, and Julien
  P.~C. Valentin.
\newblock Fastnerf: High-fidelity neural rendering at 200fps.
\newblock In {\em 2021 {IEEE/CVF} International Conference on Computer Vision,
  {ICCV} 2021, Montreal, QC, Canada, October 10-17, 2021}, pages 14326--14335.
  {IEEE}, 2021.

\bibitem{rembg}
Daniel Gatis.
\newblock Rembg, 2020.

\bibitem{DBLP:journals/jmlr/GlorotB10}
Xavier Glorot and Yoshua Bengio.
\newblock Understanding the difficulty of training deep feedforward neural
  networks.
\newblock In Yee~Whye Teh and D.~Mike Titterington, editors, {\em Proceedings
  of the Thirteenth International Conference on Artificial Intelligence and
  Statistics, {AISTATS} 2010, Chia Laguna Resort, Sardinia, Italy, May 13-15,
  2010}, volume~9 of {\em {JMLR} Proceedings}, pages 249--256. JMLR.org, 2010.

\bibitem{DBLP:journals/corr/GoyalDGNWKTJH17}
Priya Goyal, Piotr Doll{\'{a}}r, Ross~B. Girshick, Pieter Noordhuis, Lukasz
  Wesolowski, Aapo Kyrola, Andrew Tulloch, Yangqing Jia, and Kaiming He.
\newblock Accurate, large minibatch {SGD:} training imagenet in 1 hour.
\newblock {\em CoRR}, abs/1706.02677, 2017.

\bibitem{DBLP:conf/icml/GroppYHAL20}
Amos Gropp, Lior Yariv, Niv Haim, Matan Atzmon, and Yaron Lipman.
\newblock Implicit geometric regularization for learning shapes.
\newblock In {\em Proceedings of the 37th International Conference on Machine
  Learning, {ICML} 2020, 13-18 July 2020, Virtual Event}, volume 119 of {\em
  Proceedings of Machine Learning Research}, pages 3789--3799. {PMLR}, 2020.

\bibitem{WinNT}
Romain Guy and Mathias Agopian.
\newblock Physically based rendering in filament.

\bibitem{DBLP:journals/corr/abs-2206-03380}
Jon Hasselgren, Nikolai Hofmann, and Jacob Munkberg.
\newblock Shape, light {\&} material decomposition from images using monte
  carlo rendering and denoising.
\newblock {\em CoRR}, abs/2206.03380, 2022.

\bibitem{hayakawa2021neural}
Akio Hayakawa, Masato Ishii, Yoshiyuki Kobayashi, Akira Nakamura, Takuya
  Narihira, Yukio Obuchi, Andrew Shin, Takuya Yashima, and Kazuki Yoshiyama.
\newblock Neural network libraries: A deep learning framework designed from
  engineers' perspectives, 2021.

\bibitem{DBLP:conf/iccv/HedmanSMBD21}
Peter Hedman, Pratul~P. Srinivasan, Ben Mildenhall, Jonathan~T. Barron, and
  Paul~E. Debevec.
\newblock Baking neural radiance fields for real-time view synthesis.
\newblock In {\em 2021 {IEEE/CVF} International Conference on Computer Vision,
  {ICCV} 2021, Montreal, QC, Canada, October 10-17, 2021}, pages 5855--5864.
  {IEEE}, 2021.

\bibitem{DBLP:conf/rt/Jensen96}
Henrik~Wann Jensen.
\newblock Global illumination using photon maps.
\newblock In Xavier Pueyo and Peter Schr{\"{o}}der, editors, {\em Rendering
  Techniques '96, Proceedings of the Eurographics Workshop in Porto, Portugal,
  August 27-31, 1996}, Eurographics, pages 21--30. Springer, 1996.

\bibitem{DBLP:conf/cvpr/JensenDVTA14}
Rasmus~Ramsb{\o}l Jensen, Anders~Lindbjerg Dahl, George Vogiatzis, Engin Tola,
  and Henrik Aan{\ae}s.
\newblock Large scale multi-view stereopsis evaluation.
\newblock In {\em 2014 {IEEE} Conference on Computer Vision and Pattern
  Recognition, {CVPR} 2014, Columbus, OH, USA, June 23-28, 2014}, pages
  406--413. {IEEE} Computer Society, 2014.

\bibitem{DBLP:conf/siggraph/Kajiya86}
James~T. Kajiya.
\newblock The rendering equation.
\newblock In David~C. Evans and Russell~J. Athay, editors, {\em Proceedings of
  the 13th Annual Conference on Computer Graphics and Interactive Techniques,
  {SIGGRAPH} 1986, Dallas, Texas, USA, August 18-22, 1986}, pages 143--150.
  {ACM}, 1986.

\bibitem{DBLP:conf/siggraph/Karnewar0WM22}
Animesh Karnewar, Tobias Ritschel, Oliver Wang, and Niloy~J. Mitra.
\newblock Relu fields: The little non-linearity that could.
\newblock In Munkhtsetseg Nandigjav, Niloy~J. Mitra, and Aaron Hertzmann,
  editors, {\em {SIGGRAPH} '22: Special Interest Group on Computer Graphics and
  Interactive Techniques Conference, Vancouver, BC, Canada, August 7 - 11,
  2022}, pages 27:1--27:9. {ACM}, 2022.

\bibitem{DBLP:conf/cvpr/KatoUH18}
Hiroharu Kato, Yoshitaka Ushiku, and Tatsuya Harada.
\newblock Neural 3d mesh renderer.
\newblock In {\em 2018 {IEEE} Conference on Computer Vision and Pattern
  Recognition, {CVPR} 2018, Salt Lake City, UT, USA, June 18-22, 2018}, pages
  3907--3916. Computer Vision Foundation / {IEEE} Computer Society, 2018.

\bibitem{DBLP:conf/nips/KendallG17}
Alex Kendall and Yarin Gal.
\newblock What uncertainties do we need in bayesian deep learning for computer
  vision?
\newblock In Isabelle Guyon, Ulrike von Luxburg, Samy Bengio, Hanna~M. Wallach,
  Rob Fergus, S.~V.~N. Vishwanathan, and Roman Garnett, editors, {\em Advances
  in Neural Information Processing Systems 30: Annual Conference on Neural
  Information Processing Systems 2017, December 4-9, 2017, Long Beach, CA,
  {USA}}, pages 5574--5584, 2017.

\bibitem{DBLP:journals/corr/KingmaB14}
Diederik~P. Kingma and Jimmy Ba.
\newblock Adam: {A} method for stochastic optimization.
\newblock In Yoshua Bengio and Yann LeCun, editors, {\em 3rd International
  Conference on Learning Representations, {ICLR} 2015, San Diego, CA, USA, May
  7-9, 2015, Conference Track Proceedings}, 2015.

\bibitem{DBLP:journals/tog/KuangOCHAT22}
Zhengfei Kuang, Kyle Olszewski, Menglei Chai, Zeng Huang, Panos Achlioptas, and
  Sergey Tulyakov.
\newblock Neroic: neural rendering of objects from online image collections.
\newblock {\em {ACM} Trans. Graph.}, 41(4):56:1--56:12, 2022.

\bibitem{DBLP:journals/tog/LaineHKSLA20}
Samuli Laine, Janne Hellsten, Tero Karras, Yeongho Seol, Jaakko Lehtinen, and
  Timo Aila.
\newblock Modular primitives for high-performance differentiable rendering.
\newblock {\em {ACM} Trans. Graph.}, 39(6):194:1--194:14, 2020.

\bibitem{DBLP:conf/cvpr/LiSRSC20}
Zhengqin Li, Mohammad Shafiei, Ravi Ramamoorthi, Kalyan Sunkavalli, and
  Manmohan Chandraker.
\newblock Inverse rendering for complex indoor scenes: Shape, spatially-varying
  lighting and {SVBRDF} from a single image.
\newblock In {\em 2020 {IEEE/CVF} Conference on Computer Vision and Pattern
  Recognition, {CVPR} 2020, Seattle, WA, USA, June 13-19, 2020}, pages
  2472--2481. Computer Vision Foundation / {IEEE}, 2020.

\bibitem{DBLP:conf/nips/LinWL20}
Chen{-}Hsuan Lin, Chaoyang Wang, and Simon Lucey.
\newblock {SDF-SRN:} learning signed distance 3d object reconstruction from
  static images.
\newblock In Hugo Larochelle, Marc'Aurelio Ranzato, Raia Hadsell,
  Maria{-}Florina Balcan, and Hsuan{-}Tien Lin, editors, {\em Advances in
  Neural Information Processing Systems 33: Annual Conference on Neural
  Information Processing Systems 2020, NeurIPS 2020, December 6-12, 2020,
  virtual}, 2020.

\bibitem{DBLP:conf/nips/LiuGLCT20}
Lingjie Liu, Jiatao Gu, Kyaw~Zaw Lin, Tat{-}Seng Chua, and Christian Theobalt.
\newblock Neural sparse voxel fields.
\newblock In Hugo Larochelle, Marc'Aurelio Ranzato, Raia Hadsell,
  Maria{-}Florina Balcan, and Hsuan{-}Tien Lin, editors, {\em Advances in
  Neural Information Processing Systems 33: Annual Conference on Neural
  Information Processing Systems 2020, NeurIPS 2020, December 6-12, 2020,
  virtual}, 2020.

\bibitem{DBLP:conf/iccv/Liu0LL19}
Shichen Liu, Weikai Chen, Tianye Li, and Hao Li.
\newblock Soft rasterizer: {A} differentiable renderer for image-based 3d
  reasoning.
\newblock In {\em 2019 {IEEE/CVF} International Conference on Computer Vision,
  {ICCV} 2019, Seoul, Korea (South), October 27 - November 2, 2019}, pages
  7707--7716. {IEEE}, 2019.

\bibitem{DBLP:conf/cvpr/LiuZPSPC20}
Shaohui Liu, Yinda Zhang, Songyou Peng, Boxin Shi, Marc Pollefeys, and Zhaopeng
  Cui.
\newblock {DIST:} rendering deep implicit signed distance function with
  differentiable sphere tracing.
\newblock In {\em 2020 {IEEE/CVF} Conference on Computer Vision and Pattern
  Recognition, {CVPR} 2020, Seattle, WA, USA, June 13-19, 2020}, pages
  2016--2025. Computer Vision Foundation / {IEEE}, 2020.

\bibitem{DBLP:journals/tog/LombardiSSSLS19}
Stephen Lombardi, Tomas Simon, Jason~M. Saragih, Gabriel Schwartz, Andreas~M.
  Lehrmann, and Yaser Sheikh.
\newblock Neural volumes: learning dynamic renderable volumes from images.
\newblock {\em {ACM} Trans. Graph.}, 38(4):65:1--65:14, 2019.

\bibitem{DBLP:conf/eccv/LoperB14}
Matthew~M. Loper and Michael~J. Black.
\newblock Opendr: An approximate differentiable renderer.
\newblock In David~J. Fleet, Tom{\'{a}}s Pajdla, Bernt Schiele, and Tinne
  Tuytelaars, editors, {\em Computer Vision - {ECCV} 2014 - 13th European
  Conference, Zurich, Switzerland, September 6-12, 2014, Proceedings, Part
  {VII}}, volume 8695 of {\em Lecture Notes in Computer Science}, pages
  154--169. Springer, 2014.

\bibitem{DBLP:conf/cvpr/Martin-BruallaR21}
Ricardo Martin{-}Brualla, Noha Radwan, Mehdi S.~M. Sajjadi, Jonathan~T. Barron,
  Alexey Dosovitskiy, and Daniel Duckworth.
\newblock Nerf in the wild: Neural radiance fields for unconstrained photo
  collections.
\newblock In {\em {IEEE} Conference on Computer Vision and Pattern Recognition,
  {CVPR} 2021, virtual, June 19-25, 2021}, pages 7210--7219. Computer Vision
  Foundation / {IEEE}, 2021.

\bibitem{DBLP:conf/eccv/MildenhallSTBRN20}
Ben Mildenhall, Pratul~P. Srinivasan, Matthew Tancik, Jonathan~T. Barron, Ravi
  Ramamoorthi, and Ren Ng.
\newblock Nerf: Representing scenes as neural radiance fields for view
  synthesis.
\newblock In Andrea Vedaldi, Horst Bischof, Thomas Brox, and Jan{-}Michael
  Frahm, editors, {\em Computer Vision - {ECCV} 2020 - 16th European
  Conference, Glasgow, UK, August 23-28, 2020, Proceedings, Part {I}}, volume
  12346 of {\em Lecture Notes in Computer Science}, pages 405--421. Springer,
  2020.

\bibitem{DBLP:journals/tog/MullerESK22}
Thomas M{\"{u}}ller, Alex Evans, Christoph Schied, and Alexander Keller.
\newblock Instant neural graphics primitives with a multiresolution hash
  encoding.
\newblock {\em {ACM} Trans. Graph.}, 41(4):102:1--102:15, 2022.

\bibitem{DBLP:conf/cvpr/MunkbergCHES0GF22}
Jacob Munkberg, Wenzheng Chen, Jon Hasselgren, Alex Evans, Tianchang Shen,
  Thomas M{\"{u}}ller, Jun Gao, and Sanja Fidler.
\newblock Extracting triangular 3d models, materials, and lighting from images.
\newblock In {\em {IEEE/CVF} Conference on Computer Vision and Pattern
  Recognition, {CVPR} 2022, New Orleans, LA, USA, June 18-24, 2022}, pages
  8270--8280. {IEEE}, 2022.

\bibitem{DBLP:conf/cvpr/NiemeyerMOG20}
Michael Niemeyer, Lars~M. Mescheder, Michael Oechsle, and Andreas Geiger.
\newblock Differentiable volumetric rendering: Learning implicit 3d
  representations without 3d supervision.
\newblock In {\em 2020 {IEEE/CVF} Conference on Computer Vision and Pattern
  Recognition, {CVPR} 2020, Seattle, WA, USA, June 13-19, 2020}, pages
  3501--3512. Computer Vision Foundation / {IEEE}, 2020.

\bibitem{cuda}
{NVIDIA Corporation}.
\newblock {NVIDIA CUDA C} programming guide, 2010.
\newblock Version 3.2.

\bibitem{paszke2017automatic}
Adam Paszke, Sam Gross, Soumith Chintala, Gregory Chanan, Edward Yang, Zachary
  DeVito, Zeming Lin, Alban Desmaison, Luca Antiga, and Adam Lerer.
\newblock Automatic differentiation in pytorch.
\newblock 2017.

\bibitem{DBLP:conf/eccv/SangC20}
Shen Sang and Manmohan Chandraker.
\newblock Single-shot neural relighting and {SVBRDF} estimation.
\newblock In Andrea Vedaldi, Horst Bischof, Thomas Brox, and Jan{-}Michael
  Frahm, editors, {\em Computer Vision - {ECCV} 2020 - 16th European
  Conference, Glasgow, UK, August 23-28, 2020, Proceedings, Part {XIX}}, volume
  12364 of {\em Lecture Notes in Computer Science}, pages 85--101. Springer,
  2020.

\bibitem{schoenberger2016sfm}
Johannes~Lutz Sch\"{o}nberger and Jan-Michael Frahm.
\newblock Structure-from-motion revisited.
\newblock In {\em Conference on Computer Vision and Pattern Recognition
  (CVPR)}, 2016.

\bibitem{schoenberger2016mvs}
Johannes~Lutz Sch\"{o}nberger, Enliang Zheng, Marc Pollefeys, and Jan-Michael
  Frahm.
\newblock Pixelwise view selection for unstructured multi-view stereo.
\newblock In {\em European Conference on Computer Vision (ECCV)}, 2016.

\bibitem{DBLP:conf/nips/ShenGYLF21}
Tianchang Shen, Jun Gao, Kangxue Yin, Ming{-}Yu Liu, and Sanja Fidler.
\newblock Deep marching tetrahedra: a hybrid representation for high-resolution
  3d shape synthesis.
\newblock In Marc'Aurelio Ranzato, Alina Beygelzimer, Yann~N. Dauphin, Percy
  Liang, and Jennifer~Wortman Vaughan, editors, {\em Advances in Neural
  Information Processing Systems 34: Annual Conference on Neural Information
  Processing Systems 2021, NeurIPS 2021, December 6-14, 2021, virtual}, pages
  6087--6101, 2021.

\bibitem{DBLP:conf/cvpr/0004SC22}
Cheng Sun, Min Sun, and Hwann{-}Tzong Chen.
\newblock Direct voxel grid optimization: Super-fast convergence for radiance
  fields reconstruction.
\newblock In {\em {IEEE/CVF} Conference on Computer Vision and Pattern
  Recognition, {CVPR} 2022, New Orleans, LA, USA, June 18-24, 2022}, pages
  5449--5459. {IEEE}, 2022.

\bibitem{DBLP:conf/siggraph/TakikawaET0MJF22}
Towaki Takikawa, Alex Evans, Jonathan Tremblay, Thomas M{\"{u}}ller, Morgan
  McGuire, Alec Jacobson, and Sanja Fidler.
\newblock Variable bitrate neural fields.
\newblock In Munkhtsetseg Nandigjav, Niloy~J. Mitra, and Aaron Hertzmann,
  editors, {\em {SIGGRAPH} '22: Special Interest Group on Computer Graphics and
  Interactive Techniques Conference, Vancouver, BC, Canada, August 7 - 11,
  2022}, pages 41:1--41:9. {ACM}, 2022.

\bibitem{DBLP:conf/nips/WangLLTKW21}
Peng Wang, Lingjie Liu, Yuan Liu, Christian Theobalt, Taku Komura, and Wenping
  Wang.
\newblock Neus: Learning neural implicit surfaces by volume rendering for
  multi-view reconstruction.
\newblock In Marc'Aurelio Ranzato, Alina Beygelzimer, Yann~N. Dauphin, Percy
  Liang, and Jennifer~Wortman Vaughan, editors, {\em Advances in Neural
  Information Processing Systems 34: Annual Conference on Neural Information
  Processing Systems 2021, NeurIPS 2021, December 6-14, 2021, virtual}, pages
  27171--27183, 2021.

\bibitem{DBLP:conf/cvpr/WuLBWF22}
Liwen Wu, Jae~Yong Lee, Anand Bhattad, Yu{-}Xiong Wang, and David~A. Forsyth.
\newblock Diver: Real-time and accurate neural radiance fields with
  deterministic integration for volume rendering.
\newblock In {\em {IEEE/CVF} Conference on Computer Vision and Pattern
  Recognition, {CVPR} 2022, New Orleans, LA, USA, June 18-24, 2022}, pages
  16179--16188. {IEEE}, 2022.

\bibitem{DBLP:conf/nips/YarivGKL21}
Lior Yariv, Jiatao Gu, Yoni Kasten, and Yaron Lipman.
\newblock Volume rendering of neural implicit surfaces.
\newblock In Marc'Aurelio Ranzato, Alina Beygelzimer, Yann~N. Dauphin, Percy
  Liang, and Jennifer~Wortman Vaughan, editors, {\em Advances in Neural
  Information Processing Systems 34: Annual Conference on Neural Information
  Processing Systems 2021, NeurIPS 2021, December 6-14, 2021, virtual}, pages
  4805--4815, 2021.

\bibitem{DBLP:conf/nips/YarivKMGABL20}
Lior Yariv, Yoni Kasten, Dror Moran, Meirav Galun, Matan Atzmon, Ronen Basri,
  and Yaron Lipman.
\newblock Multiview neural surface reconstruction by disentangling geometry and
  appearance.
\newblock In Hugo Larochelle, Marc'Aurelio Ranzato, Raia Hadsell,
  Maria{-}Florina Balcan, and Hsuan{-}Tien Lin, editors, {\em Advances in
  Neural Information Processing Systems 33: Annual Conference on Neural
  Information Processing Systems 2020, NeurIPS 2020, December 6-12, 2020,
  virtual}, 2020.

\bibitem{DBLP:conf/iccv/YuLT0NK21}
Alex Yu, Ruilong Li, Matthew Tancik, Hao Li, Ren Ng, and Angjoo Kanazawa.
\newblock Plenoctrees for real-time rendering of neural radiance fields.
\newblock In {\em 2021 {IEEE/CVF} International Conference on Computer Vision,
  {ICCV} 2021, Montreal, QC, Canada, October 10-17, 2021}, pages 5732--5741.
  {IEEE}, 2021.

\bibitem{DBLP:journals/tog/ZeltnerSGJ21}
Tizian Zeltner, S{\'{e}}bastien Speierer, Iliyan Georgiev, and Wenzel Jakob.
\newblock Monte carlo estimators for differential light transport.
\newblock {\em {ACM} Trans. Graph.}, 40(4):78:1--78:16, 2021.

\bibitem{DBLP:conf/nips/ZhangYTR21}
Jason Zhang, Gengshan Yang, Shubham Tulsiani, and Deva Ramanan.
\newblock Ners: Neural reflectance surfaces for sparse-view 3d reconstruction
  in the wild.
\newblock In Marc'Aurelio Ranzato, Alina Beygelzimer, Yann~N. Dauphin, Percy
  Liang, and Jennifer~Wortman Vaughan, editors, {\em Advances in Neural
  Information Processing Systems 34: Annual Conference on Neural Information
  Processing Systems 2021, NeurIPS 2021, December 6-14, 2021, virtual}, pages
  29835--29847, 2021.

\bibitem{DBLP:conf/cvpr/ZhangLLS22}
Kai Zhang, Fujun Luan, Zhengqi Li, and Noah Snavely.
\newblock {IRON:} inverse rendering by optimizing neural sdfs and materials
  from photometric images.
\newblock In {\em {IEEE/CVF} Conference on Computer Vision and Pattern
  Recognition, {CVPR} 2022, New Orleans, LA, USA, June 18-24, 2022}, pages
  5555--5564. {IEEE}, 2022.

\bibitem{DBLP:conf/cvpr/ZhangLWBS21}
Kai Zhang, Fujun Luan, Qianqian Wang, Kavita Bala, and Noah Snavely.
\newblock Physg: Inverse rendering with spherical gaussians for physics-based
  material editing and relighting.
\newblock In {\em {IEEE} Conference on Computer Vision and Pattern Recognition,
  {CVPR} 2021, virtual, June 19-25, 2021}, pages 5453--5462. Computer Vision
  Foundation / {IEEE}, 2021.

\bibitem{DBLP:journals/corr/abs-2010-07492}
Kai Zhang, Gernot Riegler, Noah Snavely, and Vladlen Koltun.
\newblock Nerf++: Analyzing and improving neural radiance fields.
\newblock {\em CoRR}, abs/2010.07492, 2020.

\bibitem{DBLP:journals/tog/ZhangSDDFB21}
Xiuming Zhang, Pratul~P. Srinivasan, Boyang Deng, Paul~E. Debevec, William~T.
  Freeman, and Jonathan~T. Barron.
\newblock Nerfactor: neural factorization of shape and reflectance under an
  unknown illumination.
\newblock {\em {ACM} Trans. Graph.}, 40(6):237:1--237:18, 2021.

\bibitem{DBLP:conf/cvpr/ZhangSHFJZ22}
Yuanqing Zhang, Jiaming Sun, Xingyi He, Huan Fu, Rongfei Jia, and Xiaowei Zhou.
\newblock Modeling indirect illumination for inverse rendering.
\newblock In {\em {IEEE/CVF} Conference on Computer Vision and Pattern
  Recognition, {CVPR} 2022, New Orleans, LA, USA, June 18-24, 2022}, pages
  18622--18631. {IEEE}, 2022.

\bibitem{DBLP:conf/ijcnn/ZhengYLLL15}
Hao Zheng, Zhanlei Yang, Wenju Liu, Jizhong Liang, and Yanpeng Li.
\newblock Improving deep neural networks using softplus units.
\newblock In {\em 2015 International Joint Conference on Neural Networks,
  {IJCNN} 2015, Killarney, Ireland, July 12-17, 2015}, pages 1--4. {IEEE},
  2015.

\bibitem{Zhou2018}
Qian-Yi Zhou, Jaesik Park, and Vladlen Koltun.
\newblock {Open3D}: {A} modern library for {3D} data processing.
\newblock {\em arXiv:1801.09847}, 2018.

\end{thebibliography}
\end{document}